\documentclass{article}

\usepackage[final]{neurips_2023}

\usepackage[utf8]{inputenc} 
\usepackage[T1]{fontenc}    
\usepackage{hyperref}       
\usepackage{url}            
\usepackage{booktabs}       
\usepackage{amsfonts}       
\usepackage{nicefrac}       
\usepackage{microtype}      
\usepackage{xcolor}         
\usepackage{amsmath,amsthm}
\usepackage{graphicx,psfrag,epsf}
\usepackage{enumerate}
\usepackage{natbib}
\usepackage{amssymb,color}
\usepackage{algorithm} 
\usepackage{algorithmic} 
\usepackage{appendix}
\usepackage{subfigure}
\usepackage{multirow}
\usepackage{rotating}
\usepackage{natbib}
\usepackage{threeparttable}

\def\argmin{\mathop{\rm argmin}}

\def\Var{\mathop{\rm Var}}
\def\sign{\mathop{\rm sign}}

\def\ba{\mathop{\bf a}}

\def\bc{\mathop{\bf c}}

\def\bx{\mathop{\bf x}}

\def\by{\mathop{\bf y}}

\def\bW{\mathop{\bf W}}

\newcommand {\bfalpha} {\mbox{\boldmath $\alpha$}}

\newtheorem{thm}{Theorem}
\newtheorem{lem}{Lemma}

\newtheorem{rem}{Remark}
\newtheorem{cor}{Corollary}
\graphicspath{{bounded KQR 1-d/}}

\title{Towards a Unified Analysis of Kernel-based Methods Under Covariate Shift}
\author{
  Xingdong Feng$^{1}$, Xin He$^{1}$, Caixing Wang$^{1}$\thanks{Caixing Wang is the corresponding author and all the authors contributed equally to this paper and their names are listed in alphabetical ordering.}, Chao Wang$^{1}$, Jingnan Zhang$^{2}$\\
  $^{1}$School of Statistics and Management, Shanghai University of Finance and Economics\\
  $^{2}$International Institute of Finance, School of Management, \\ University of Science and Technology of China \\
  \texttt{$\{$feng.xingdong, he.xin17$\}$@mail.shufe.edu.cn}\\
  \texttt{$\{$wang.caixing, wang.chao$\}$@stu.shufe.edu.cn}\\
  \texttt{jnzhang@ustc.edu.cn}
}

\begin{document}

\maketitle

\def \thefootnote{\dag} \makeatletter\def\Hy@Warning#1{}\makeatother 

\begin{abstract}
Covariate shift occurs prevalently in practice, where the input distributions of the source and target data are substantially different. Despite its practical importance in various learning problems, most of the existing methods only focus on some specific learning tasks and are not well validated theoretically and numerically. To tackle this problem, we propose a unified analysis of general nonparametric methods in a reproducing kernel Hilbert space (RKHS) under covariate shift.  Our theoretical results are established for a general loss belonging to a rich loss function family, which includes many commonly used methods as special cases, such as mean regression, quantile regression, likelihood-based classification, and margin-based classification. Two types of covariate shift problems are the focus of this paper and the sharp convergence rates are established for a general loss function to provide a unified theoretical analysis, which concurs with the optimal results in literature where the squared loss is used. Extensive numerical studies on synthetic and real examples confirm our theoretical findings and further illustrate the effectiveness of our proposed method.

\end{abstract}

\section{Introduction}\label{sec:intro}
\begin{figure}[!htbp]
    \centering
    \subfigure[]{
        \includegraphics[width=1.92in,  height=1.308in]{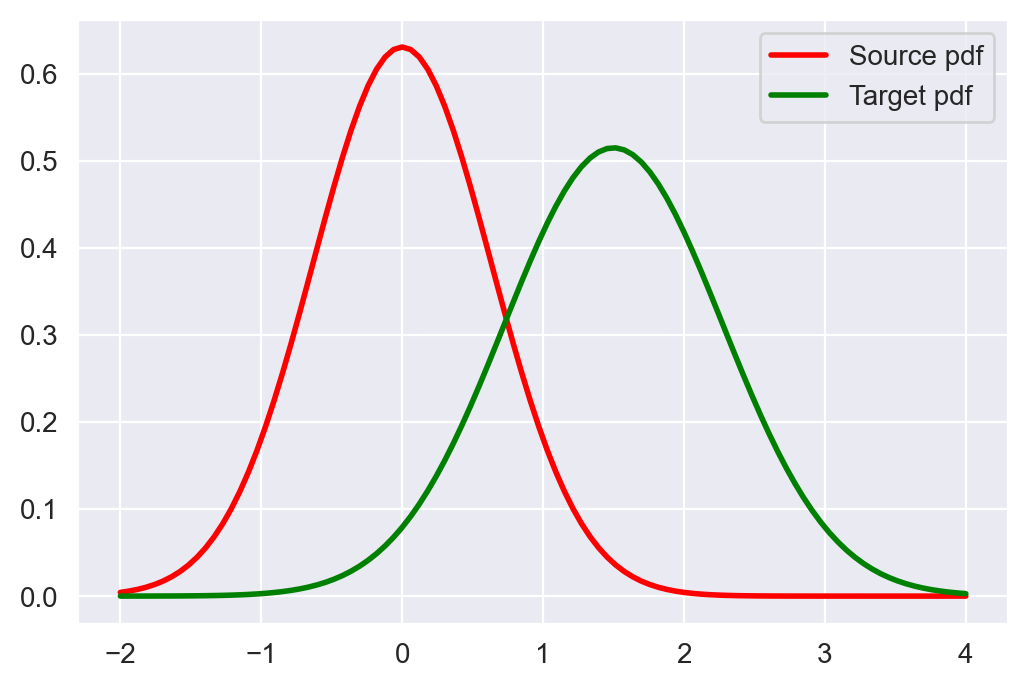}
        \label{pdf_cs}
    }
    \hspace{8mm}
    \subfigure[]{
 \includegraphics[width=1.92in,  height=1.308in]{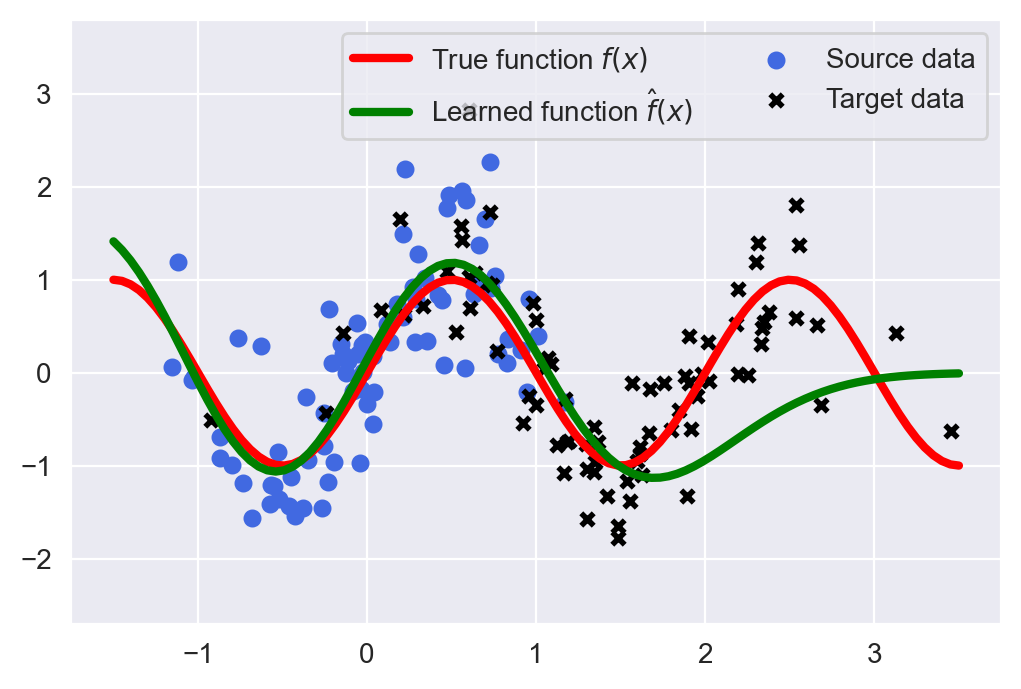}
        \label{learning_cs}
    }
  \caption{\footnotesize{(a) The probability density functions of normal distributions with $\mu_1=0, \sigma^2=0.4$ that the source data is driven from and $\mu_1=1.5, \sigma^2=0.6$ that the target data is driven from, respectively; (b) the learned function trained by using the source data and the true mean regression function. Note that the considered example serves as an illustration that satisfies case (ii) in Section \ref{sec:2.3}. }  }
  \label{fig1}
\end{figure}

Covariate shift is a phenomenon that commonly occurs in machine learning, where the distribution of input features (covariates) changes between the source (or training) and target (or test) data, while the conditional distribution of output values given covariates remains unchanged \citep{shimodaira2000improving,pan2010survey}.  Such a  phenomenon is illustrated in  Figure \ref{fig1} where the learned predictive function from the source data may significantly differ from the true function. Thus, the prediction performance can be largely degraded since the predictive function has not been trained on data that accurately represents the target environment. Covariate shift can arise in a variety of domains, including medicine and healthcare \citep{wei2015health,hajiramezanali2018bayesian}, remote sensing \citep{tuia2011using}, and natural language and speech processing \citep{yamada2009covariate, fei2015social}. Various factors contribute to covariate shift, such as data sampling bias (e.g., patient demographics in healthcare applications), changes in the studied population (e.g., vocabulary evolution in natural language processing), or measurement and observational errors (e.g., sensor noise or calibration errors in the sensor domain). Compared to the well-studied supervised learning without such a distribution mismatch \citep{gyorfi2002distribution}, there still exists some gap in both theoretically and numerically understanding the influence of the covariate shift under various kernel-based learning problems.

This paper provides a unified analysis of the kernel-based methods under covariate shift.  Moreover, a general loss function is considered, which is allowed to belong to a rich loss function family and thus includes many commonly used methods as special cases, such as mean regression, quantile regression, likelihood-based classification, and margin-based classification. This paper considers two types of covariate shift problems in which the importance ratio is assumed to be uniformly bounded or to have a bounded second  moment.  A unified theoretical analysis
 has been provided that the sharp convergence rates are established for a general loss function under two evaluation metrics, which concurs with the optimal results in \cite{ma2022optimally} where the squared loss is used.  Our theoretical findings are also validated by a variety of synthetic and real-life examples.

 
\textbf{Our contributions.}
The contributions of this paper are multi-fold.  We propose a unified analysis of the kernel-based methods under covariate shift, which provides an insightful understanding of the influences of covariate shift on the kernel-based methods both theoretically and numerically. By utilizing empirical process in learning theory, we investigate the theoretical behaviors of several estimators under various conditions, with particular emphasis on different importance ratio cases between the source and target distributions.  Specifically,  we show that the unweighted estimator achieves the optimal learning rates in terms of both $\mathcal{L}^2$-error and the excess risk with respect to the target distribution under the uniformly bounded case. Yet, the unweighted estimator is sub-optimal under the bounded second moment case. Then, we construct a weighted estimator by using an appropriate truncated ratio, which again attains a sharp convergence rate. Unlike kernel ridge regression (KRR), the estimator for the general loss does not have an explicit form, making many classical analysis techniques inapplicable \citep{lian2022distributed} and in our technical proofs, the theoretical derivation is much more involved due to the considered general loss function.  Numerous experiments on synthetic data and multi-source real data with various loss functions confirm our theoretical findings. To some extent, this paper provides a comprehensive study of the numerical performance of the kernel-based methods under various scenarios with covariate shift.

\textbf{Related work.} Some most related works including covariate shift adaptation and importance ratio estimation are presented below.

\textbf{Covariate shift adaptation.}
\cite{shimodaira2000improving} investigates the impact of covariate shift in parametric regression with maximum likelihood estimation and proposes an importance-weighting (IW) estimator, which has a significant improvement when the misspecification does exist. \cite{sugiyama2005model} further extend this work by analyzing an unbiased estimator for $\mathcal{L}^2$-error.  These fundamental works also motivate a variety of follow-up studies under the parametric setting
 \cite{sugiyama2006mixture,yamazaki2007asymptotic,wen2014robust,lei2021near}. Beyond the parametric setting,
\cite{kpotufe2021marginal} consider the nonparametric classification problem and provide a new minimax result that concisely captures the relative benefits of source and target labeled data under covariate shift. Focusing on the H{\"o}lder continuous regression function class, \cite{pathak2022new} propose a new measure of distribution mismatch between the source and target distributions. \cite{ma2022optimally} and \cite{gogolashvili2023importance} establish the optimal learning rates for KRR over RKHS. Recently, covariate shift in overparameterized models, such as high-dimensional models and neural networks, has also drawn tremendous attention from relevant researchers \citep{hendrycks2019benchmarking,byrd2019effect,hendrycks2021many,tripuraneni2021overparameterization}. 

\textbf{Importance ratio estimation.} A straightforward approach to estimating the importance ratio is separately estimating the source and target densities by kernel density estimation \citep{sugiyama2005input,baktashmotlagh2014domain} and then computing the ratio. In practice, it is more efficient to directly estimate the importance ratio by minimizing some   discrepancy measures between distributions, including kernel mean matching \citep{huang2006correcting, gretton2009covariate}, Kullback-Leibler divergence \citep{sugiyama2007covariate,sugiyama2007direct, sugiyama2012density} and non-negative Bregman divergence \citep{kato2021non}.

\textbf{Notation.} In this paper, we use $C$ to denote a universal constant that may vary from line to line. For two sequence $a_n$ and $b_n$, we say $a_n\asymp b_n$ if and only if $a_n=O(b_n)$ and $b_n=O(a_n)$ hold.

\section{Problem formulation}\label{sec:pro}

\subsection{Learning within a rich family of loss functions}\label{sec:2.1}
Suppose that  the random pair $(\bx, y)$ is drawn from some unknown distribution $P_{\bx,y}$ with ${\bx}=(x_1,\ldots,x_p)^{\top}\in {\mathcal{X}}$, where $\mathcal{X}\subset {\cal R}^p$ is a compact support  and $y \in \mathcal{Y}\subset {\cal R}$. In the literature of machine learning, the true target function $f^*$ is often defined as the minimizer of the expected risk with a pre-specified loss function  ${L}(\cdot,\cdot):{\cal Y}\times {\cal R}\rightarrow\mathcal{R}^{+}$ that
\begin{align}\label{expect_risk}
 f^* := \argmin\nolimits_{f} {\cal E}^{ L}(f)= \argmin\nolimits_{f  } E  \left[ {L}\big  (y, f(\bx) \big )  \right]. 
\end{align}

Throughout this paper, we consider a general loss function belonging to a rich loss function family that ${L}(y, \cdot)$ is assumed to be convex and locally $c_{{L}}$-Lipschitz continuous \citep{wainwright2019high,dasgupta2019feature}, that is for some $V \geq 0$, there exists a constant $c_{L}>0$ such that $|L(y, \omega)-L(y, \omega^{\prime})|\leq c_{L}|\omega-\omega^{\prime}|$ holds for all pairs $\omega, \omega^{\prime} \in [-V, V]$ and $y \in {\cal Y}$. 
A variety of popularly used loss functions satisfy these two conditions, including
\begin{itemize}
\item \textbf{Squared loss}: ${L}(y, f(\mathbf{x}))=(y-f(\mathbf{x}))^2$ with $c_L=2\left(M_y+V\right)$, for any $|y| \leq M_y$ with  constant $M_y>0$;
\item \textbf{Check loss}: ${L}(y, f(\mathbf{x}))=(y-f(\mathbf{x}))\left(\tau-I_{\{y\leq f(\mathbf{x})\}}\right)$ with $c_{{L}}=1$ and quantile level $\tau$;
\item \textbf{Huber loss}: ${L}(y, f(\mathbf{x}))=(y-f(\mathbf{x}))^2$, if $|y-f(\mathbf{x})| \leq \delta ; \delta|y-f(\mathbf{x})|-\frac{1}{2} \delta^2$, otherwise, with $c_{{L}}=\delta$;
\item \textbf{Logistic loss}: ${L}(y, f(\mathbf{x}))=\log (1+\exp (-y f(\mathbf{x})))/{\log 2}$ with $c_{{L}}= (\log 2)^{-1} e^V /(1+e^V)$;
\item \textbf{Hinge loss}: ${L}(y, f(\mathbf{x}))=(1-y f(\mathbf{x}))_{+}$with $c_{{L}}=1$ .
\end{itemize}
Note that the explicit form of $f^*$ may differ from one loss to another. For example, for the squared loss, $f^*(\bx)=E[y|\bx]$; for the check loss, $f^*(\bx)=Q_{\tau}(y|\bx)$ with $Q_\tau(y |\bx)=\inf \{y: P(Y \leq y |\bx) \geq \tau\}$; for the hinge loss, $f^*=$sign$(P(y=1|\bx)-1/2)$ with sign$(\cdot)$ denoting the sign function. Moreover, we require $f^* \in \mathcal{H}_K$, where ${\cal H}_K$ denotes the RKHS induced by a pre-specified kernel function $K(\cdot,\cdot): {\cal X}\times {\cal X} \rightarrow {\cal R}$.
In practice, one of the most important tasks in machine learning is to learn $f^*$ from the source data and use the estimated predictive function $\widehat{f}$ for prediction in the target data.

\subsection{Measurement under source and target distributions}
Classical learning problems often explicitly or implicitly
assume that the source and target data are drawn from the same distribution. Precisely, the source data comes from a joint distribution $P_S=P_{y|\bx}P^S_{\bx}$ where $P_{y|\bx}$ is the conditional distribution and $P^S_{\bx}$ is the input distribution with density function $\rho^S_{\bx}$. Then, the performance of the estimated predictive function $\widehat{f}$ is usually evaluated by the ${\cal L}^2( P_{\bx}^S)$-error or the excess risk error with respect to $P_S$ that
\begin{align*}
\|\widehat{f}-f^*\|_S^2=E_{\bx \sim S}\big[\big(\widehat{f}(\bx)-f^*(\bx)\big)^2\big], \quad  
\mathcal{E}^{{ L}}_S(\widehat{f})-\mathcal{E}^{{ L}}_S(f^*)=E_{S }\big[ { L}(y, \widehat{f}(\bx))- { L}(y, f^*(\bx))\big],
\end{align*}
where $E_{\bx \sim S}$ and $E_S$ are the expectation over $P_{\bx}^S$ and $P_S$ conditioning on the observed data, respectively. When covariate shift occurs, the target data may come from some totally different joint distribution  $P_T=P_{y|\bx}P^T_{\bx}$  in the sense that although $P_{y|\bx}$ representing the regression or classification rule remains invariant, $P^T_{\bx}$
 differs significantly from  $P^S_{\bx}$. We also assume that $P^T_{\bx}$  has  density function $\rho^T_{\bx}$. Obviously, it is reasonable to evaluate $\widehat{f}$ under $P_T$ instead of $P_S$.  Then, our primary interest is the $\mathcal{L}^2(P_{\bx}^T)$-error or the excess risk error with respect to $P_T$ that
\begin{align*}
 \|\widehat{f}-f^*\|_T^2=E_{\bx \sim T}\big[\big(\widehat{f}(\bx)-f^*(\bx)\big)^2\big], \quad  
{\cal E}_T^{ { L}}  (\widehat{f})-{\cal E}_T^{{ L}}(f^*)=E_{T }\big[ { L}(y, \widehat{f}(\bx))-{ L}(y, f^*(\bx))\big ],
\end{align*}
where $E_{\bx \sim T}$ and $E_T$ are the expectation over $P_{\bx}^T$ and $P_T$ conditioning on the observed data, respectively. In the rest of this paper,  $f^*$ is defined as the optimal function under the target distribution such that $f^*=\argmin\nolimits_{f  } E_T [ {L}\big  (y, f(\bx) \big ) ]$.

\subsection{Kernel-based estimation under covariate shift and importance ratio correction}\label{sec:2.3}
Suppose that random sample ${\cal Z}^S_n = \big \{\big(\bx_i^S, y_{i}^S\big) \big \}_{i=1}^{n}$ are  i.i.d.  drawn from the source distribution $P_S$. We consider the classical nonparametric  estimation problem in  RKHS \citep{vapnik1999nature} that 
\begin{align}\label{empirical_loss}
    \widehat{f} := \argmin_{f \in {\cal H}_K}
  \frac{1}{n} \sum_{i=1}^{n} {L}\big (y_{i}^S, f({\bx}_{i}^S) \big )+\lambda\|f\|_{K}^{2},
\end{align}
where $\lambda$ is the regularization parameter. Without covariate shift, the first term on the right side of \eqref{empirical_loss} is an unbiased estimator of $\mathcal{E}_T^{{L}}({f})$. However, it becomes biased when covariate shift occurs, and thus may lead to inaccurate predictive estimator. To tackle this issue, we consider the importance ratio  measuring the discrepancy between distributions that is
$
\phi(\bx)={\rho^T_{\bx}}(\bx)/{\rho^S_{\bx}}(\bx), 
$
for any $\bx \in {\cal X}$, and we notice that $\int\int { L}(y,f(\bx)) \rho^T_{\bx}(\bx)d P_{y|\bx}d\bx = \int\int\phi(\bx) { L}(y,f(\bx)) \rho^S_{\bx}(\bx)d P_{y|\bx}d\bx$. Inspired by this, the weighted version of \eqref{empirical_loss} can be used, which leads to an importance ratio weighted (IRW) estimator that
\begin{align}
   \widetilde{f}^{\phi}:= \argmin_{f \in {\cal H}_K}
  \frac{1}{n} \sum_{i=1}^{n} \phi({\bx}_i^S) {L}\big (y_{i}^S, f({\bx}_{i}^S) \big )+\lambda\|f\|_{K}^{2}. \label{weighted_estimation}
\end{align} 

Throughout  this paper, we focus on two types of the importance ratio that 
\begin{itemize}
	\item[(i)]  $\phi(\bx)$ is $\alpha$-uniformly bounded  that is
$\sup_{\bx \in {\cal X}}\phi(\bx) \leq \alpha$, for some positive constant $\alpha$;
	\item[(ii)] $\phi(\bx)$'s second moment   is bounded  that  is
$E_{\bx \sim S}[\phi^2(\bx)]\leq \beta^2$, for some constant $\beta^2\geq 1$.
\end{itemize}

Note that case (i) reduces to the classical case without covariate shift if $\alpha=1$. Yet, the bounded case is somewhat restrictive, and a much weaker condition is considered in case (ii). It is also clear that case (i) can be regarded as a special case of case (ii) by taking $\beta^2=\alpha$. To see this, we have $E_{\bx \sim S}[\phi(\bx)^2] =E_{\bx \sim T}[\phi(\bx)] \leq \alpha$.  It is interesting to notice that these two cases are related to   R{\'e}nyi divergence \citep{renyi1961measures} between $\rho^S_{\bx}$ and $\rho^T_{\bx}$. Specifically,  the conditions in cases (i) and (ii) are equivalent to requiring $D_{\infty}(\rho^T_{\bx}||\rho^S_{\bx})$ and $D_{2}(\rho^T_{\bx}||\rho^S_{\bx})$ being bounded \citep{cortes2010learning}, respectively. Moreover, as pointed out by \cite{ma2022optimally}, \eqref{weighted_estimation} may result in significant inflation of the variance, largely due to the unbounded importance ratio. Thus, it is natural to consider a truncated importance ratio with a pre-specified threshold value $\gamma_n >0$ \citep{ma2022optimally, gogolashvili2023importance}. Precisely, we consider the following truncation 
\begin{align*}
\phi_n(\bx):=\min\left\{\phi(\bx),\gamma_n\right\}.
\end{align*}
Accordingly, the following truncated important ratio weighted (TIRW) estimator is considered that
\begin{align}\label{IRRW_estimates}
   \widehat{f}^{\phi} := \argmin_{f \in {\cal H}_K}
  \frac{1}{n} \sum_{i=1}^{n} \phi_n({\bx}^S_i) {L}\big (y_{i}^S, f({\bx}_{i}^S) \big )+\lambda\|f\|_{K}^{2},
 \end{align}
where the theoretical suggestion of the choice of  $\gamma_n$ is provided in Section \ref{main_results}.

\section{Main results}\label{main_results}
In this section, we provide a unified theoretical analysis of the unweighted estimator in \eqref{empirical_loss} and the TIRW estimator in \eqref{IRRW_estimates} under two types of importance ratio. Let ${\cal H}_K$ denote the RKHS induced by a  symmetric, continuous, and positive semi-definite kernel function $K(\cdot,\cdot):{\cal X} \times {\cal X} \rightarrow \mathcal{R}$. Under some regularity conditions, Mercer’s theorem \citep{mercer1909xvi} guarantees that $K$ has an eigen-expansion of the form
\begin{align*}
K(\bx,{\bx}^{\prime})=\sum_{j=1}^{\infty}\mu_j \psi_j(\bx)\psi_j({\bx}^{\prime}),  
\end{align*}
where $\{\mu_j\}_{j \geq 1}$ are the eigenvalues and $\{\psi_j\}_{j \geq 1}$ are the orthonormal eigenfunctions of $\mathcal{L}^2(\mathcal{X},P_{\bx}^T)=\left\{f:\int_{\mathcal{X}}f^2(\bx)\rho_{\bx}^T(\bx)d\bx< \infty\right\}$. We expand the kernel in $\mathcal{L}^2(\mathcal{X}, P_{\bx}^T)$ for deriving bounds under the target distribution. For any $f \in \mathcal{H}_K$, we have $f=\sum_{j=1}^{\infty}a_j\psi_j$ with $a_j=\int_{\mathcal{X}}f(\bx)\psi_j(\bx)\rho^T_{\bx}(\bx)d \bx$ and the RKHS-norm is $\|f\|_K^2=\sum_{j=1}^{\infty}a_j^2/\mu_j < \infty$. The kernel complexity function of $\mathcal{H}_K$ is then given as $R(\delta)=\sqrt{\frac{1}{n}\sum_{j=1}^{\infty}\min(\delta^2,\mu_j\|f^*\|_K^2)}$.  In literature, $R(\delta)$ is important to quantify the localized Rademacher complexity of $\mathcal{H}_{K}$, which helps to build tight bounds in kernel-based methods \citep{mendelson2002geometric, Koltchinskii2010SPARSITYIM}. For theoretical simplicity, we assume $\|f^*\|_K=1$ in the rest of this paper, which is commonly considered in machine learning literature \citep{yang2017randomized,li2021towards}. The following technical assumptions are required for the theoretical analysis.

\noindent \textbf{Assumption 1}: There exists some constant $\kappa>0$ such that $\sup_{\bx \in \cal X}|K({\bx},{\bx})|\leq \kappa$, and  the eigenfunctions $\{\psi_j\}_{j \geq 1}$ are uniformly bounded such that  $\sup_{\bx \in \cal X}|\psi_j(\bx)|\leq 1$ for all $j \geq 1$.


\noindent \textbf{Assumption 2}: For some sufficiently small constant $u>0$, we assume $ E_{I}\left[L(y, f(\bx))\right]-E_{I}\left[L\left(y, f^*(\bx)\right)\right] \geqslant c_0\left\|f-f^*\right\|_I^2$ holds with some constant $c_0 >0$, for all $\left\|f-f^*\right\|_I \leq u$, where $I \in \{S,T\}$.

Assumption 1 imposes a boundedness condition on the kernel function and its corresponding eigenfunctions, which is commonly considered in the literature and satisfied by many popularly used kernel functions with the compact support condition \citep{Smale2007,steinwart2009optimal,mendelson2010regularization}. Assumption 2 is a local $c_0$-strong convexity condition of the expected loss function with respect to $\mathcal{L}^2(\mathcal{X},P^S_{\bx})$ and $\mathcal{L}^2(\mathcal{X},P^T_{\bx})$ at $f^*$. Similar assumptions are also considered by \cite{steinwart2008support,wainwright2019high,li2021towards,lian2022distributed}.
Note that many popularly used loss functions, including squared loss, check loss, Huber loss, logistic loss, and hinge loss, satisfy this assumption.  More detailed discussions on Assumption 2 are deferred to Section A.7 of the supplementary material.

\subsection{Convergence rate for the uniformly bounded case}
We begin to establish the convergence rate of the unweighted estimator $\widehat{f}$ under the uniformly bounded case with covariate shift.

\begin{thm}\label{thm1}
Under Assumptions 1-2, if the importance ratio is $\alpha$-uniformly bounded, let $\lambda > c_0\delta_n^2/4$ with $\delta_{n}$ being  the smallest  positive solution to   $C\sqrt{\log n} R(\sqrt{\alpha}\delta)\le c_0\delta^2/2$, then for some constant $c_1>0$, with probability at least $1-n^{-c_1}$, we have
\begin{align}\label{rw_l2_error1}
\|\widehat{f}-f^*\|_T^2 \leq \alpha\left(\delta_n^2+2c_0^{-1}\lambda\right).
\end{align}
Furthermore, based on \eqref{rw_l2_error1}, we have
\begin{align}\label{rw_risk1}
\mathcal{E}^L_T(\widehat{f})-\mathcal{E}^{L}_T(f^*) \leq c_L\alpha{\left(\delta_n^2+2c_0^{-1}\lambda\right)^{1/2}}. 
\end{align}
\end{thm}

Note that the existence and uniqueness of $\delta_n$ in Theorem \ref{thm1}  is guaranteed for any kernel class \citep{bartlett2005local}. To facilitate the comprehension of Theorem \ref{thm1}, we define an index as $d(\delta)=\min\{j \geq 1 | \mu_j \leq \delta^2\}$ for a given target error level $\delta>0$. It is known that the kernel with eigenvalues satisfying that $\sum_{j=d(\delta)+1}^{\infty}\mu_j \leq Cd(\delta)\delta^2$ is referred to as  the regular kernel class \citep{yang2017randomized}. From the definition of $d(\delta)$ and regular kernel, we can get that $\sum_{j=1}^{\infty}\min(\delta^2,\mu_j)\asymp d(\delta)\delta^2$, thus the inequality $C\sqrt{\log n} R(\sqrt{\alpha}\delta)\le c_0\delta^2/2$ in Theorem \ref{thm1} can be directly simplified as $\sqrt{\alpha\log n d(\sqrt{\alpha}\delta)/n}\leq C\delta$ under the assumption that $\|f^*\|_K=1$.  Consequently, we can rewrite \eqref{rw_l2_error1} as 
\begin{align}\label{uw_bound_1}
\|\widehat{f}-f^*\|_T^2  \le C\alpha\Big (\delta^2+\alpha \frac{\log n}{n} d(\sqrt{\alpha} \delta)\Big),
\end{align}
where $\delta$ satisfies $\alpha\frac{\log n}{n} d(\sqrt{\alpha}\delta)\le C\delta^2$ and $\lambda\asymp\delta^2$ (see Section C.4 of the supplementary material for the detailed proof). The bound in \eqref{uw_bound_1} controls a type of bias-variance tradeoff by the choice of $\delta$, and hence $\lambda$. 
Under the uniformly bounded case,  \cite{ma2022optimally} has established a  minimax rate of the order $\alpha\inf_{\delta>0}\{\delta^2+\frac{\sigma^2 d(\sqrt{\alpha}\delta)}{n}\}$ for the squared loss function. It is worthnoting that the result in \eqref{uw_bound_1} can also attain this lower bound up to a logarithmic factor under some weaker conditions.  Particularly, the assumption of \cite{ma2022optimally} that the noise terms are sub-Gaussian is no longer needed in \eqref{uw_bound_1}.  Although the convergence rate of \eqref{rw_risk1} is sub-optimal for  general loss function $L$ considered in Section \ref{sec:2.1}, it  becomes optimal at the rate of $c_0^{\prime}\alpha\left(\delta_n^2+2c_0^{-1}\lambda\right)$ if the following assumption holds.

\noindent \textbf{Assumption 3}: For some sufficiently small constant $u>0$, we assume $ E_{T}\left[L(y, f(\bx))\right]-E_{T}\left[L\left(y, f^*(\bx)\right)\right] \le c_0^{\prime}\left\|f-f^*\right\|_T^2$ holds with some constant $c_0^{\prime} >c_0>0$, for all $\left\|f-f^*\right\|_T \leq u$.

\vspace{0.5em}
In fact, Assumption 3 is a mild condition that can be satisfied by many commonly used losses. For instance, for the squared loss, the equality always holds with $c_0^{\prime}=1$; for the check loss, Assumption 3 is satisfied if the conditional density of the noise term is uniformly bounded \citep{zhang2021approximate,lian2022distributed}. Moreover, the regular kernel class includes kernels with either finite rank (i.e., linear 
or polynomial kernels), polynomially decayed eigenvalues (i.e., Sobolev kernels), or exponentially decayed eigenvalues (i.e., Gaussian kernels). Corollary \ref{cor1} provides the convergence rate of   $\widehat{f}$ over these three specific kernel classes. 

\begin{cor}\label{cor1}
Under Assumptions 1-3, if the kernel has a finite rank $D$, and let $\lambda=C\frac{\alpha D\log n}{n}$, then with probability at least $1-n^{-c_1}$, we have
\begin{align}\label{cor1_bound}
\|\widehat{f}-f^*\|_T^2  \asymp \mathcal{E}^{L}_T(\widehat{f})-\mathcal{E}^{L}_T(f^*)  \leq C\frac{\alpha^2D\log n}{n},   
\end{align}
and if the eigenvalues of the kernel decay polynomially such as $\mu_j \leq Cj^{-2r}$ with some constant $r>1/2$ for $j=1, 2, \ldots$,  and let $\lambda=C\alpha^{\frac{2r-1}{2r+1}}(\frac{\log n}{n})^{\frac{2r}{2r+1}}$, then with probability at least $1-n^{-c_1}$, we have 
\begin{align}\label{cor1_decay}
\|\widehat{f}-f^*\|_T^2  \asymp \mathcal{E}^{L}_T(\widehat{f})-\mathcal{E}^{L}_T(f^*)  \leq C\left(\frac{\alpha^2\log n}{n}\right)^{\frac{2r}{2r+1}},
\end{align}
and if the eigenvalues of the kernel decay exponentially such as $\mu_j \asymp e^{-Cj\log j}$, and let $\lambda=C\frac{\alpha\log^2 n}{n}$, then with probability at least $1-n^{-c_1}$, we have
\begin{align}\label{cor1_decay_2}
\|\widehat{f}-f^*\|_T^2  \asymp \mathcal{E}^{L}_T(\widehat{f})-\mathcal{E}^{L}_T(f^*)  \leq C\frac{\alpha^2\log^2 n}{n},
\end{align}
\end{cor}
Note that these bounds in \eqref{cor1_bound}-\eqref{cor1_decay_2} reduce to the known minimax lower bounds \citep{yang2017randomized} for the squared loss without covariate shift (i.e., $\alpha=1$). As the uniform boundedness implies the boundedness of the second moment, the convergence rate of the TIRW estimator $\widehat{f}^{\phi}$ for the uniformly bounded case is similar to Theorem \ref{thm2} in the next section by replacing $\alpha$ with  $\beta^2$. 

\subsection{Convergence rate for the second moment bounded case}

The optimality for the $\alpha$-uniformly bounded importance ratio condition relies on the inequality that $\|\widehat{f}-f^*\|_T^2 \leq \alpha \|\widehat{f}-f^*\|_S^2$. Yet, such a desired relation is not guaranteed for the second moment bounded case. Theorem \ref{thm3} shows that the unweighted estimator $\widehat{f}$ is still consistent, but not optimal.

\begin{thm}\label{thm3}
Under Assumptions 1-2, if the importance ratio satisfies that $E_{\bx \sim S}[\phi^2(\bx)]\leq \beta^2$, let $\lambda > c_0{\delta}_n^2/4$ with $\delta_{n}$ being  the smallest  positive solution to  $C\sqrt{\log n}R((c_0^{-1}c_{L}\sqrt{\beta^2}\delta)^{1/2})\leq c_0\delta^2/2$, then for some constant $c_2>0$, with probability at least $1-n^{-c_2}$, we have
\begin{align}\label{l_2}
\|\widehat{f}-f^*\|_T^2 \leq c_0^{-1}c_{L}\sqrt{\beta^2} \left(\delta_n^2+2c_0^{-1}\lambda\right)^{1/2}.
\end{align}
Furthermore, based on \eqref{l_2}, we have
\begin{align}\label{risk}
\mathcal{E}^{L}_T(\widehat{f})-\mathcal{E}^{L}_T(f^*) \leq c_{L}\sqrt{\beta^2} \left(\delta_n^{2}+2c_0^{-1}\lambda\right)^{1/2}.   
\end{align}
\end{thm}

To see the bound in \eqref{l_2} is sub-optimal, we  consider the kernel with finite rank $D$, we can show that $\delta_n\asymp (\frac{\sqrt{\beta^2}D\log n}{n} )^{1/3}$, and if we take $\lambda\asymp (\frac{\sqrt{\beta^2}D\log n}{n} )^{2/3}$, the unweighted estimator satisfies that $\|\widehat{f}-f^*\|_T^2= O_P((\frac{\beta^4D\log n}{n})^{1/3})$, which is far from the optimal rate (see Section C.4 of the supplementary material for examples of finite rank, polynomially and exponentially decay kernel classes). To deal with the sub-optimality, we consider the importance ratio correction ensuring that $E_S[\phi(\bx)L(y, f(\bx))]=E_T[L(y, f(\bx))]$. The following theorem shows that the TIRW estimator $\widehat{f}^{\phi}$ can again reach a sharp convergence rate up to logarithmic factors under some mild conditions. 

\begin{thm}\label{thm2}
Under Assumptions 1-2, if the importance  ratio satisfies that $E_{\bx \sim S}[\phi^2(\bx)]\leq \beta^2$, let $\lambda > c_0{\delta}_n^2/4$ with $\delta_{n}$ being  the smallest  positive solution to    $C\sqrt{\beta^2}\log nR(\delta) \leq c_0\delta^2/2$, and set the truncation level $\gamma_n=\sqrt{n\beta^2}$, then for some constant $c_3>0$, with probability at least $1-n^{-c_3}$, we have
\begin{align}\label{rw_l2_error2}
\|\widehat{f}^{\phi}-f^*\|_T^2 \leq \delta_n^{2}+2c_0^{-1}\lambda.    
\end{align}
Furthermore, based on \eqref{rw_l2_error2}, we have
\begin{align}\label{rw_risk2}
\mathcal{E}^{L}_T(\widehat{f}^{\phi})-\mathcal{E}^{L}_T(f^*) \leq \frac{1}{2}c_0\delta_n^{2}+2\lambda.   
\end{align}
\end{thm}

Similar to \eqref{uw_bound_1}, when dealing with regular kernel classes, the bounds in \eqref{rw_l2_error2} and \eqref{rw_risk2} become 
\begin{align}\label{IR_weighted_bound}
\|\widehat{f}^{\phi}-f^*\|_T^2  \asymp \mathcal{E}^{L}_T(\widehat{f}^{\phi})-\mathcal{E}^{L}_T(f^*)  \leq C \Big (\delta^2+\beta^2\frac{\log^2 n}{n}d(\delta)\Big ),
\end{align}
where $\delta$ is any solution of $\beta^2\frac{\log^2 n}{n}d(\delta) \leq C\delta^2$ and $\lambda\asymp\delta^2$. Since the second moment boundedness can be implied by the uniform boundedness, it can be concluded that $\widehat{f}^{\phi}$ also reaches the minimax lower bound for the square loss up to logarithmic factors (with $\alpha$ substituted by $\beta^2$). For three specific kernel classes, we also have the following Corollary for the TIRW estimator $\widehat{f}^{\phi}$.

\begin{cor}\label{cor2}
Under Assumptions 1-2, if the kernel has a finite rank $D$, and let $\lambda=C\frac{\beta^2 D\log^2 n}{n}$, then with probability at least $1-n^{-c_3}$, we have
\begin{align}\label{cor2_bound}
\|\widehat{f}^{\phi}-f^*\|_T^2  \asymp \mathcal{E}^{L}_T(\widehat{f}^{\phi})-\mathcal{E}^{L}_T(f^*)  \leq C\frac{\beta^2 D\log^2 n}{n},   
\end{align}
and if the eigenvalues of the kernel decay polynomially such as $\mu_j \leq Cj^{-2r}$ with a constant $r>1/2$ for $j=1, 2, \ldots$, and let $\lambda=C(\frac{\beta^2\log n}{n})^{\frac{2r}{2r+1}}$, then with probability at least $1-n^{-c_3}$, we have 
\begin{align}\label{cor2_decay}
\|\widehat{f}^{\phi}-f^*\|_T^2  \asymp \mathcal{E}^{L}_T(\widehat{f}^{\phi})-\mathcal{E}^{L}_T(f^*)  \leq C\left(\frac{\beta^2 \log^2 n}{n}\right)^{\frac{2r}{2r+1}},    
\end{align}
and if the eigenvalues of the kernel decay exponentially such as $\mu_j \asymp e^{-Cj\log j}$, and let $\lambda=C\frac{\beta^2\log^3 n}{n}$, then with probability at least $1-n^{-c_3}$, we have
\begin{align}\label{cor2_decay_2}
\|\widehat{f}-f^*\|_T^2  \asymp \mathcal{E}^{L}_T(\widehat{f})-\mathcal{E}^{L}_T(f^*)  \leq C\frac{\beta^2\log^3 n}{n},
\end{align}
\end{cor}

For easy reference, we summarize some important theoretical results that have been established by us in table \ref{table3}.

\begin{table}[H]
\centering
\caption{Established convergence rates under different cases}
\label{table3}
\begin{threeparttable}
\begin{footnotesize}
\scalebox{0.9}{
\begin{tabular}{ l c c c c} 
\hline\hline
 \multirow{3}*{Kernel class} & \multicolumn{2}{c}{Uniformly bounded case}&\multicolumn{2}{c}{Moment bounded case}\\
\cmidrule(r){2-3}  \cmidrule(r){4-5} &  Unweighted estimator  & TIRW estimator     &  Unweighted estimator  & TIRW estimator  \\
\hline
\multirow{3}*{Finite rank $D$}  & \multirow{3}*{$O_P(\frac{\alpha^2D\log n}{n})$} & \multirow{3}*{$O_P(\frac{\alpha D\log^2n}{n})$} & \multirow{3}*{$O_P((\frac{\beta^4 D\log n}{n} )^{1/3})$}    &\multirow{3}*{$O_P(\frac{\beta^2 D\log^2n}{n})$}\\
  &  &    &    &\\
   &  &    &    &\\
\hline
  \multirow{3}*{Polynomial decay}  & \multirow{3}*{$O_P((\frac{\alpha^2\log n}{n})^{\frac{2r}{2r+1}})$} & \multirow{3}*{$O_P((\frac{\alpha\log^2 n}{n})^{\frac{2r}{2r+1}})$} & \multirow{3}*{$O_P((\frac{\beta^4\log n}{n} )^{\frac{2r}{6r+1}})$}    &\multirow{3}*{$O_P((\frac{\beta^2 \log^2 n}{n})^{\frac{2r}{2r+1}})$}\\
  &  &    &    &\\
    &  &    &    &\\
  \hline 
  \multirow{3}*{Exponential decay }  & \multirow{3}*{$O_P(\frac{\alpha^2\log^2 n}{n})$} & \multirow{3}*{$O_P(\frac{\alpha \log^3n}{n})$} & \multirow{3}*{$O_P((\frac{\beta^4 \log^2 n}{n} )^{1/3})$}    &\multirow{3}*{$O_P(\frac{\beta^2 \log^3n}{n})$}\\
  &  &    &    &\\
   &  &    &    &\\
  \hline  \hline
\end{tabular}
}
\end{footnotesize}  
 \end{threeparttable}
\end{table}

\section{Numerical experiments}\label{sec:sim}
In this section, we validate our theoretical analyses by performing numerical experiments on both synthetic data and real applications. To estimate the importance ratios, we apply the Kullback-Leibler importance estimation procedure (KLIEP) \citep{sugiyama2007direct}, and the estimation details are provided in the supplementary material. 
For brevity, we only report the performance of kernel-based quantile regression (KQR) where the check loss is used in synthetic data analysis, and the performance of kernel support vector support machine (KSVM) with hinge loss in the real data analysis; completed results for other loss functions as well as the detailed settings and
results for the other examples, including for the multi-dimensional cases, can be found in Section A of the supplementary material. In our experiments, we consider the RKHS induced by  Gaussian kernel in all the examples, and all the experiments are replicated 100 times in the synthetic data analysis.

\subsection{Synthetic data analysis} 
\begin{figure}[H]
\graphicspath{{bounded_KQR_1_d/}}
    \centering
    \subfigure[]{
    \includegraphics[width=1.6in,  height=1.09in]{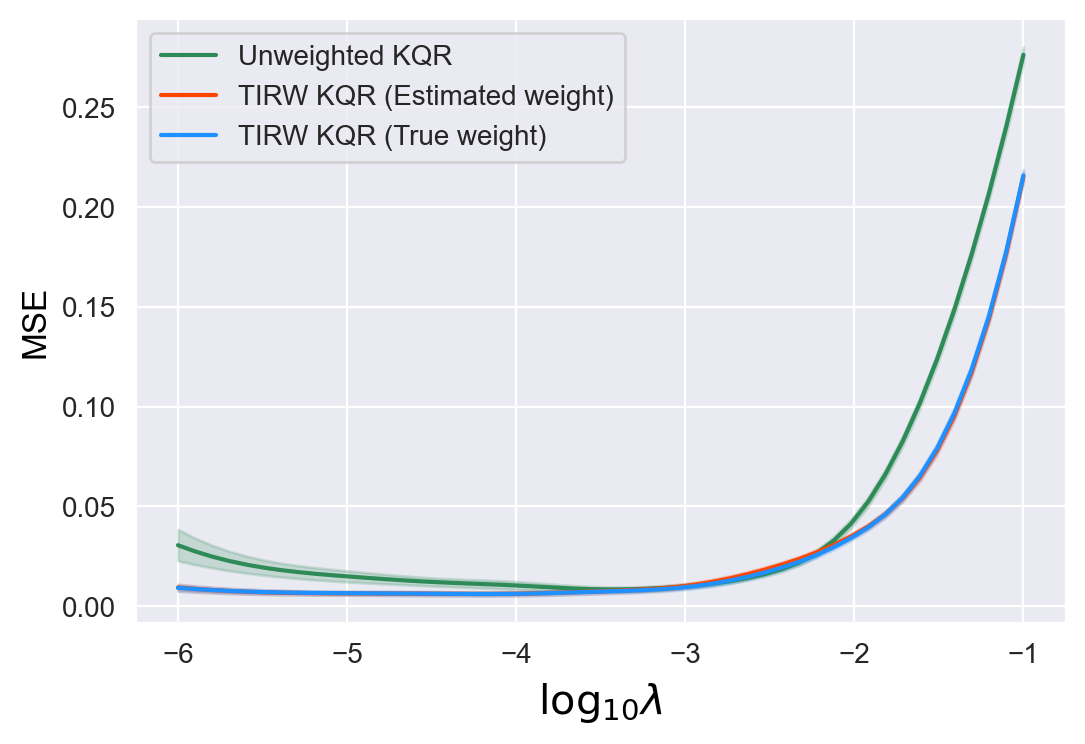}
\label{KQR_dimension1_uniformbounded_MSE_smooth_3}}
    \subfigure[]{
	\includegraphics[width=1.6in,  height=1.09in]{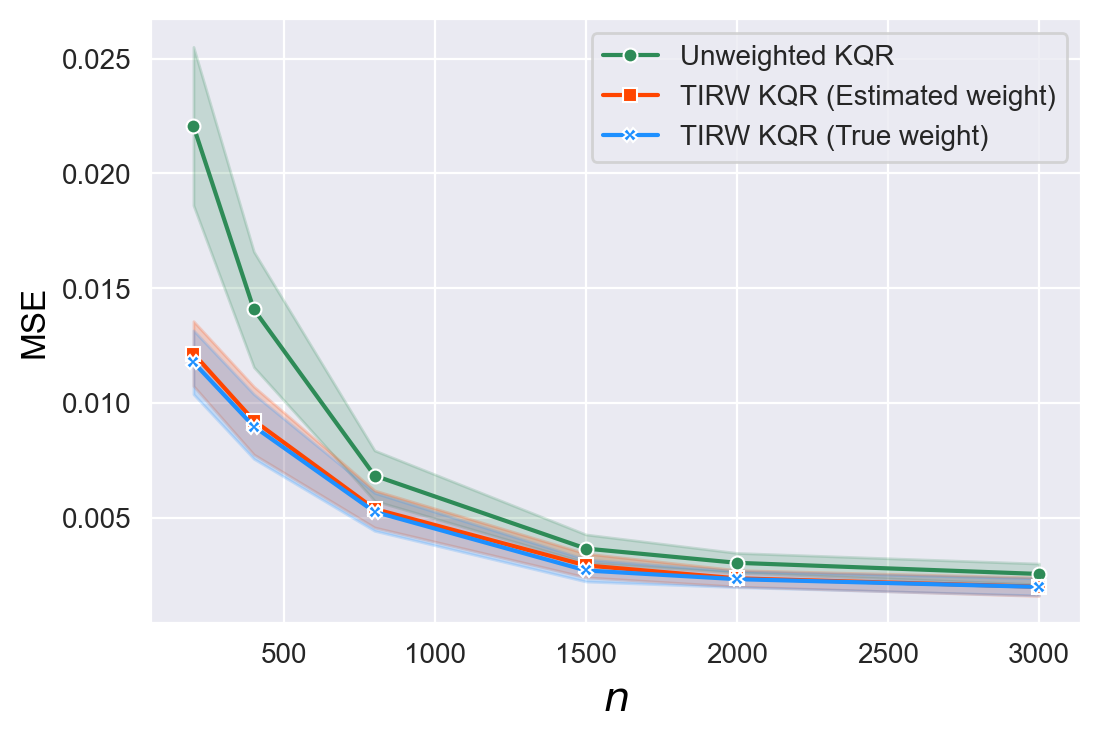}
\label{KQR_dimension1_uniformbounded_MSE_9}}
   \subfigure[]{
    \includegraphics[width=1.6in,  height=1.09in]{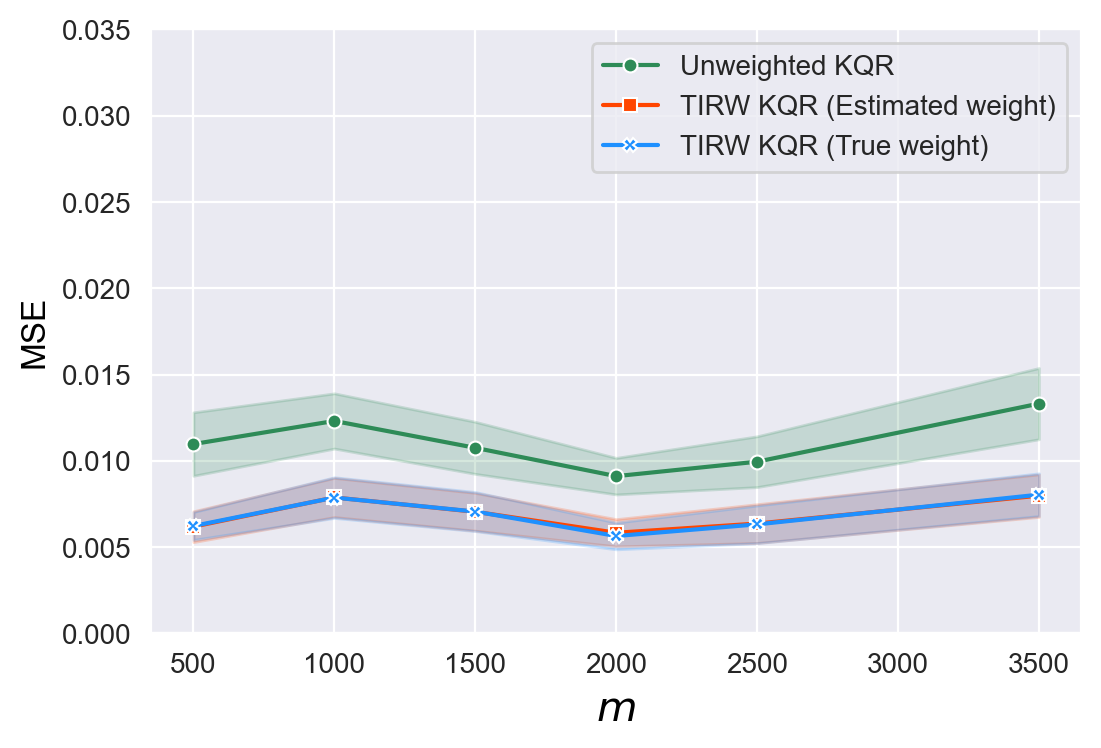}
\label{KQR_dimension1_uniformbounded_MSE_15}}
    \subfigure[]{
	\includegraphics[width=1.6in,  height=1.09in]{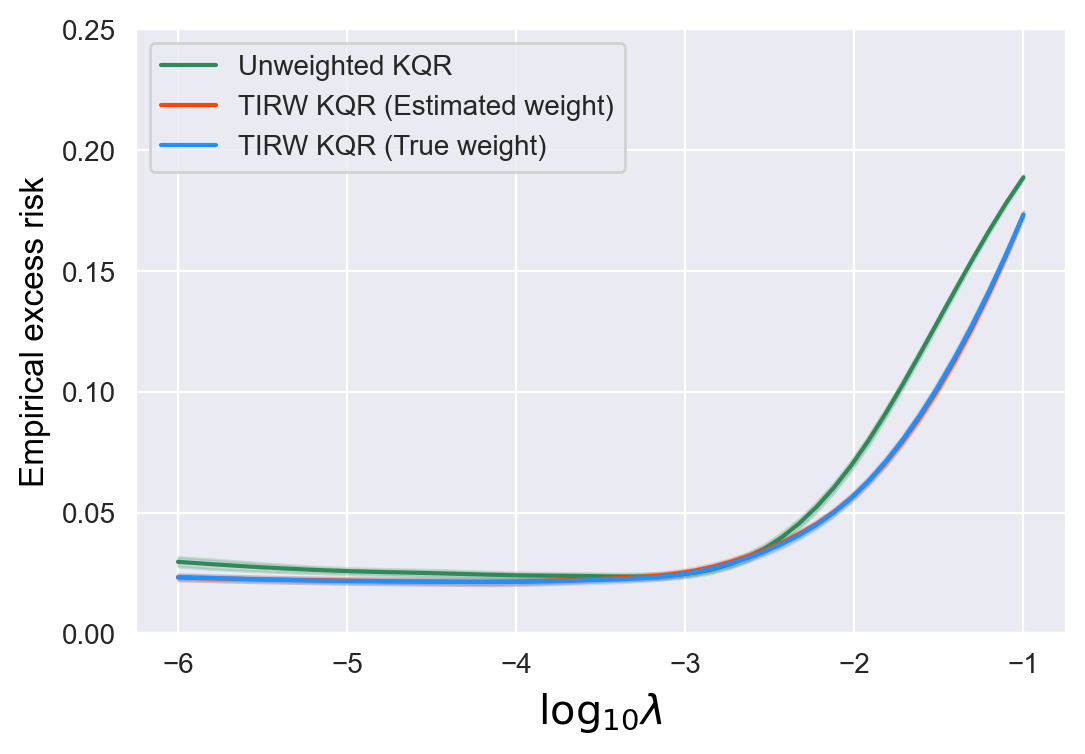}
\label{KQR_dimension1_uniformbounded_Empirical_smooth_3}}
 \subfigure[]{
    \includegraphics[width=1.6in,  height=1.09in]{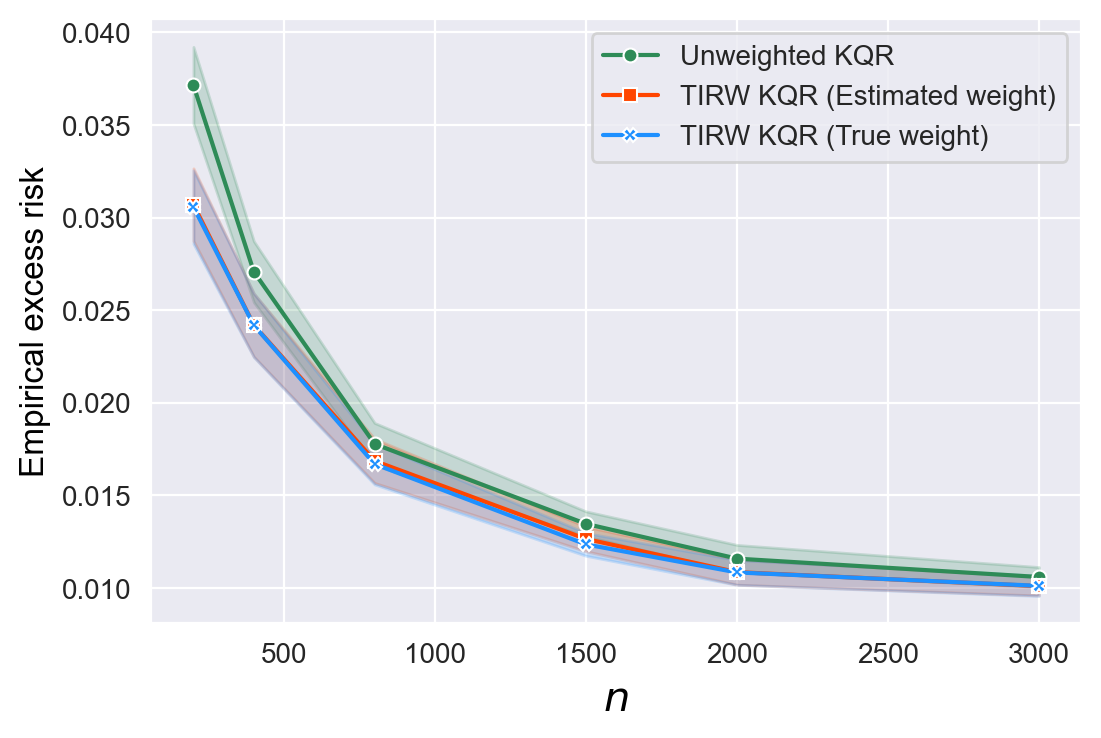}
\label{KQR_dimension1_uniformbounded_Empirical_9}}
    \subfigure[]{
	\includegraphics[width=1.6in,  height=1.09in]{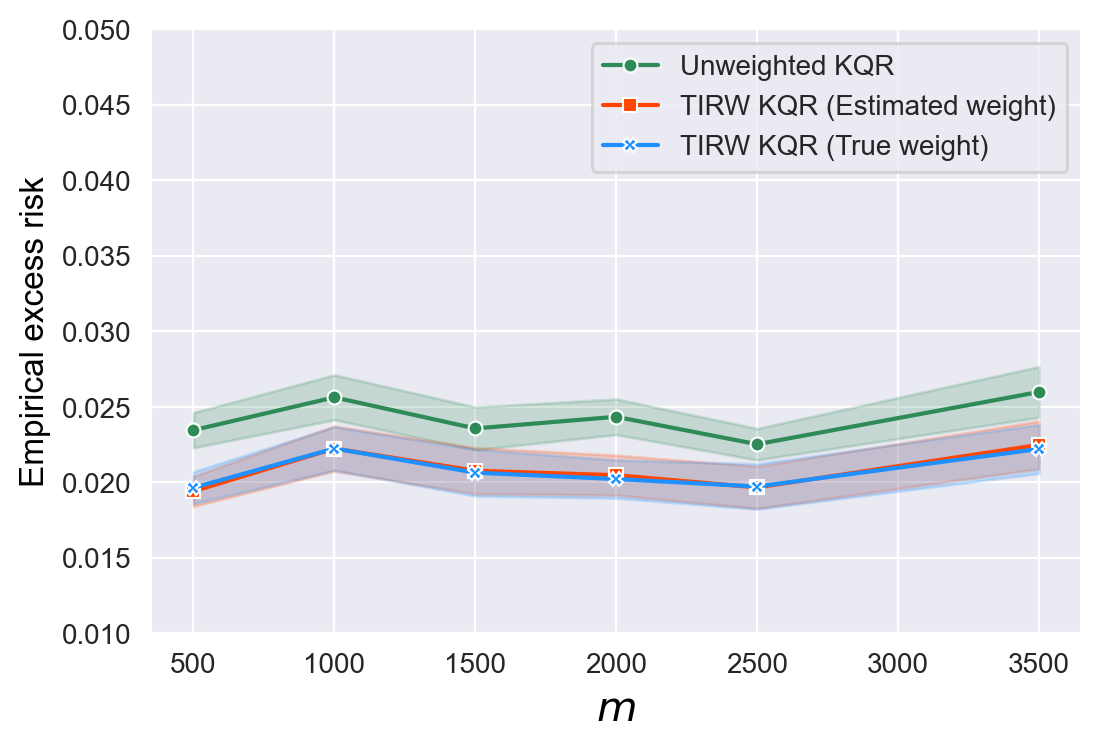}
\label{KQR_dimension1_uniformbounded_Empirical_15}}
\caption{\footnotesize{Averaged MSE and empirical excess risk for unweighted KQR, TIRW KQR with true weight and estimated weight, respectively. Note that in (a) and (d), the curves are plotted with respect to $\log_{10} \lambda$ with $n=500,m=1000$; in (b) and (e) the curves are plotted with respect to $n$ with fixed $m=1000,\lambda=10^{-4}$; in (c) and (f), the curves are plotted with respect to $m$ with fixed $n=500,\lambda=10^{-4}$.}}
\label{KQR_dimension1_bounded_main}
\end{figure}

\begin{figure}
\graphicspath{{unbounded_KQR_1_d/}}
    \centering
    \subfigure[]{
    \includegraphics[width=1.6in,  height=1.09in]{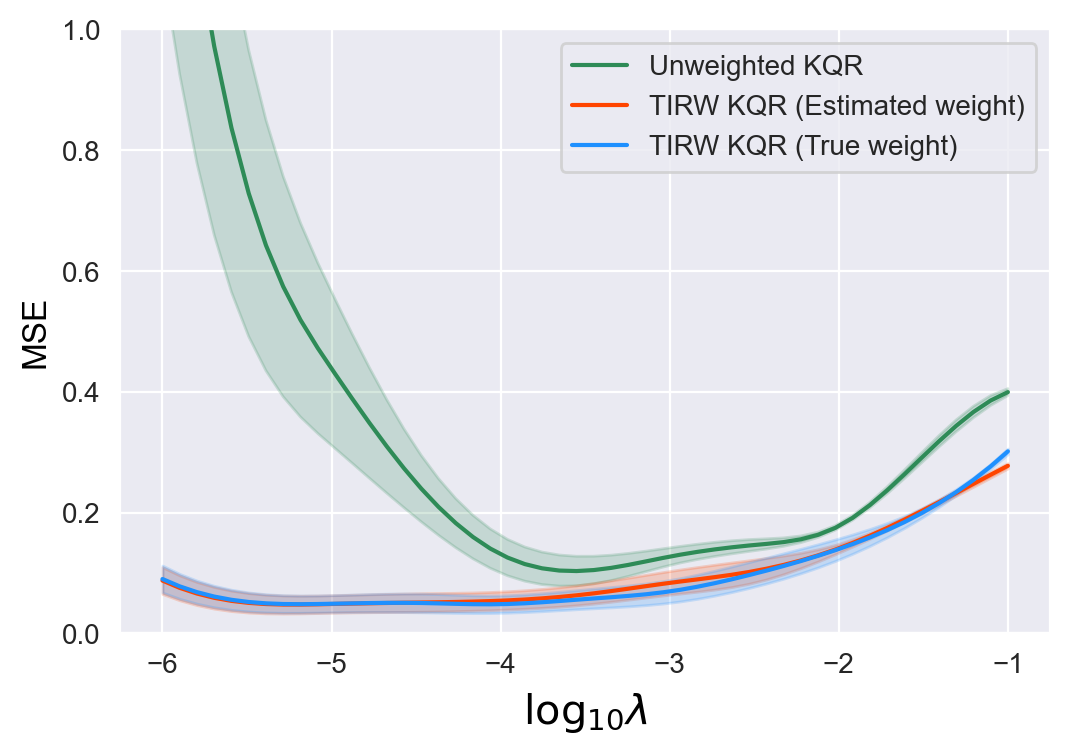}
    \label{KQR_dimension1_MSE_smooth_3}}
    \subfigure[]{
	\includegraphics[width=1.6in,  height=1.09in]{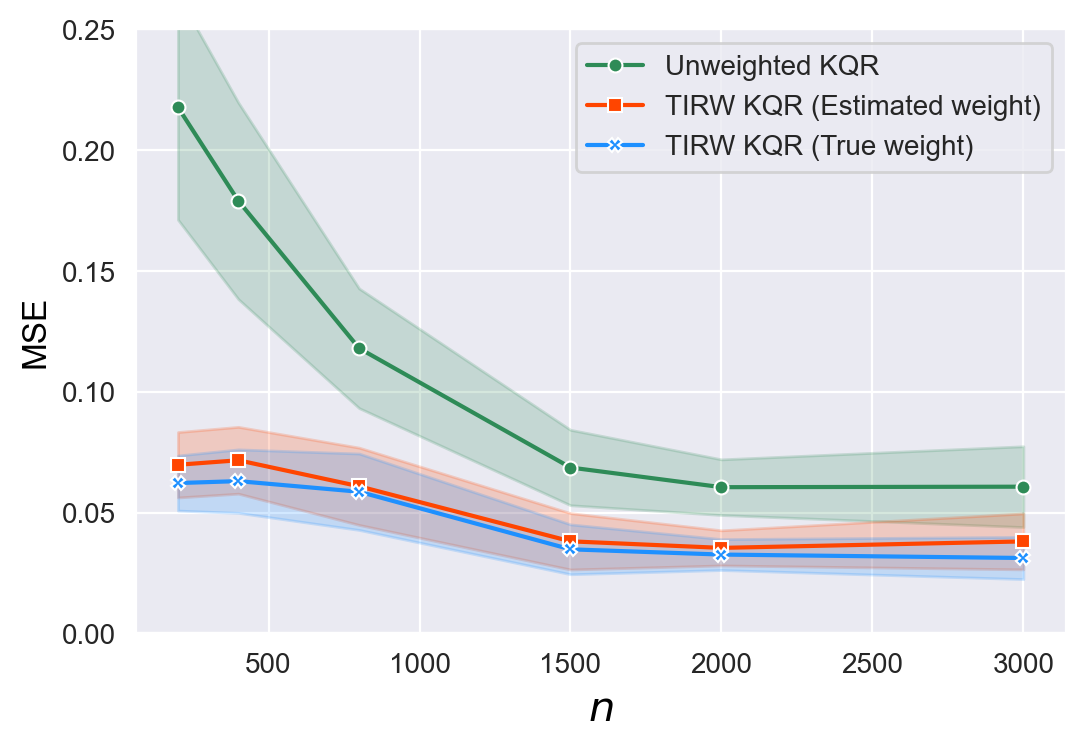}
\label{KQR_dimension1_MSE_9}}
   \subfigure[]{
    \includegraphics[width=1.6in,  height=1.09in]{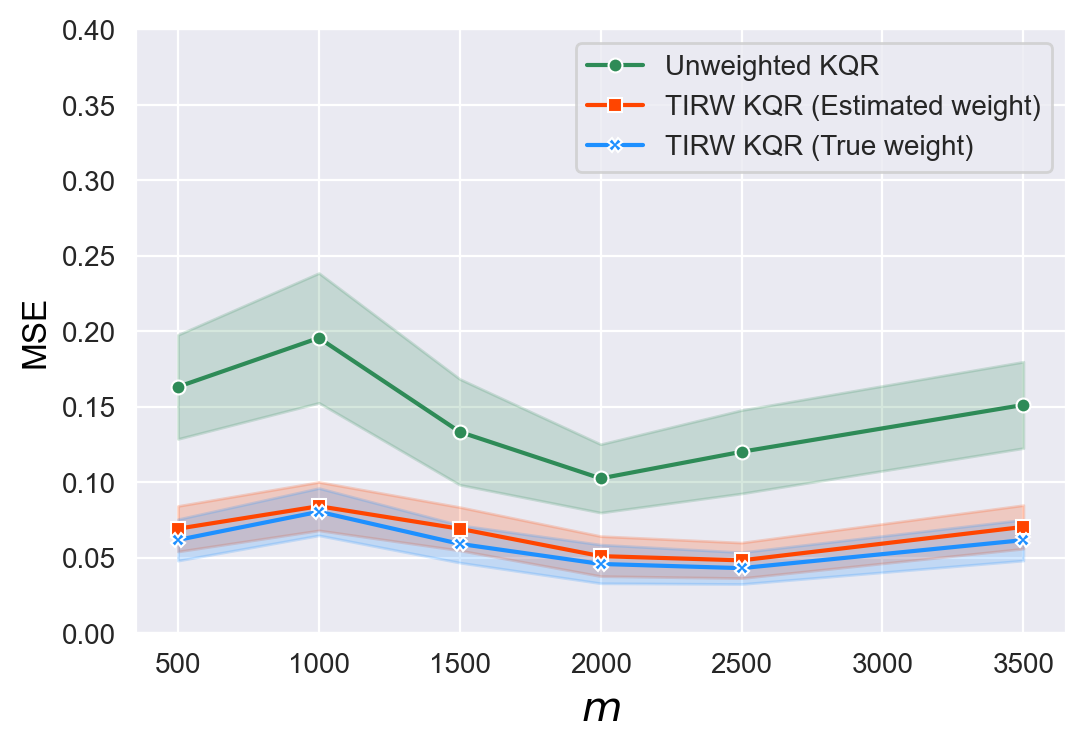}
        \label{KQR_dimension1_MSE_15}}
    \subfigure[]{
	\includegraphics[width=1.6in,  height=1.09in]{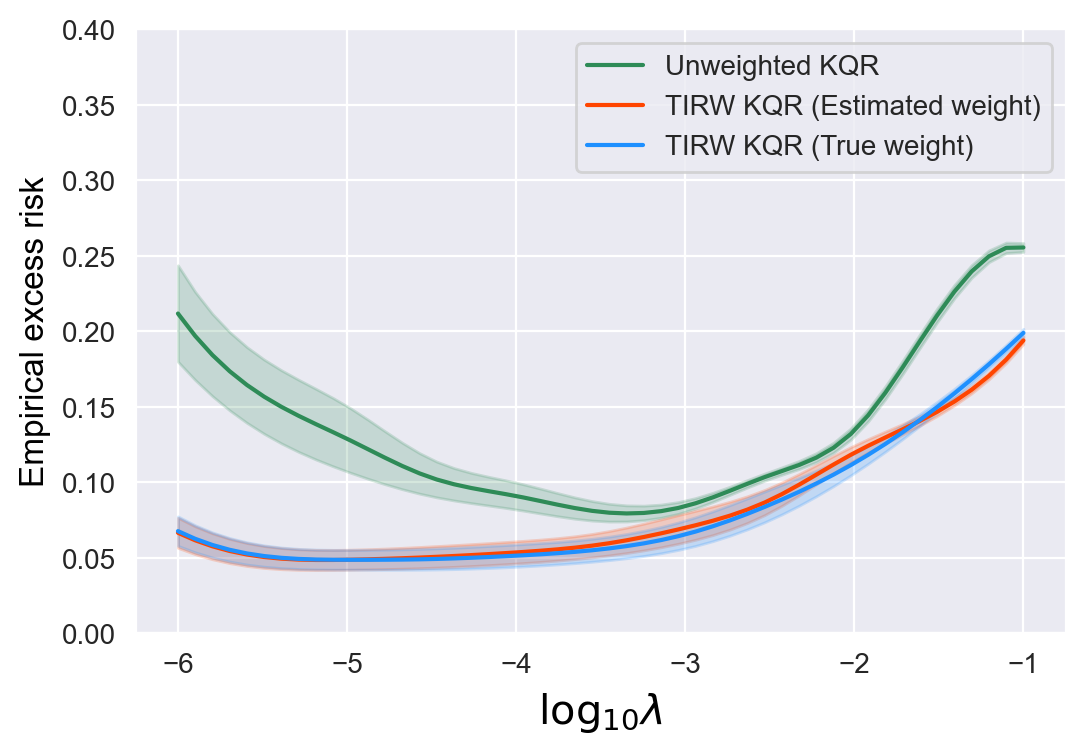}
\label{KQR_dimension1_Empirical_smooth_3}}
 \subfigure[]{
    \includegraphics[width=1.6in,  height=1.09in]{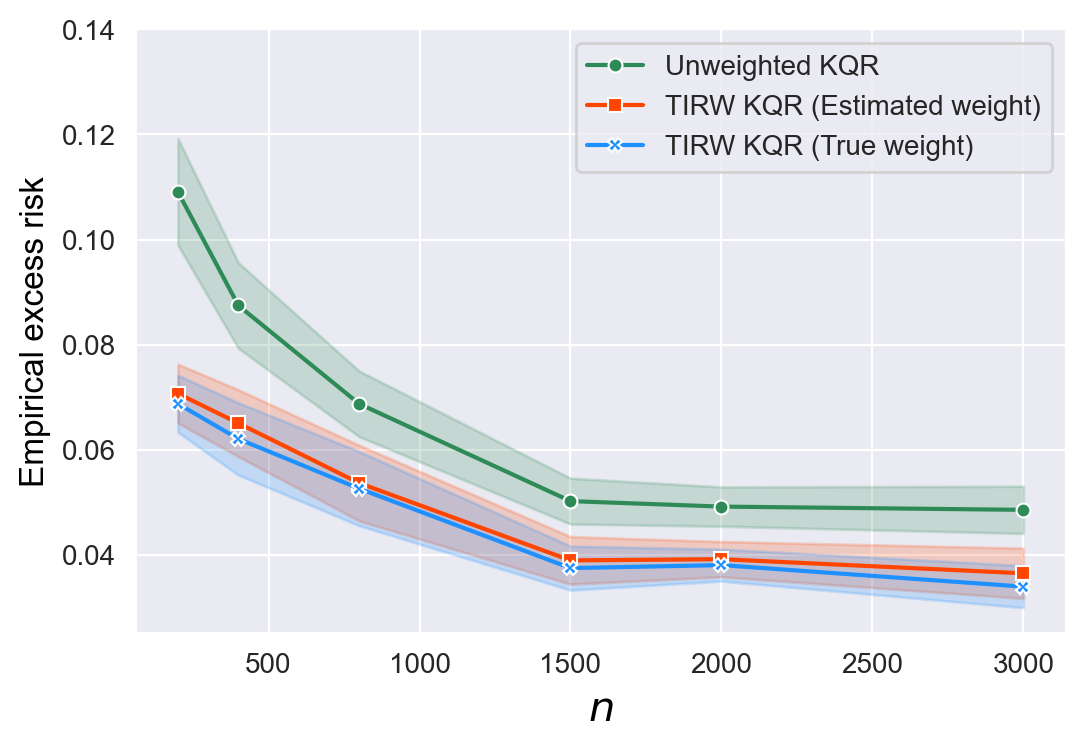}
        \label{KQR_dimension1_Empirical_9}}
    \subfigure[]{
	\includegraphics[width=1.6in,  height=1.09in]{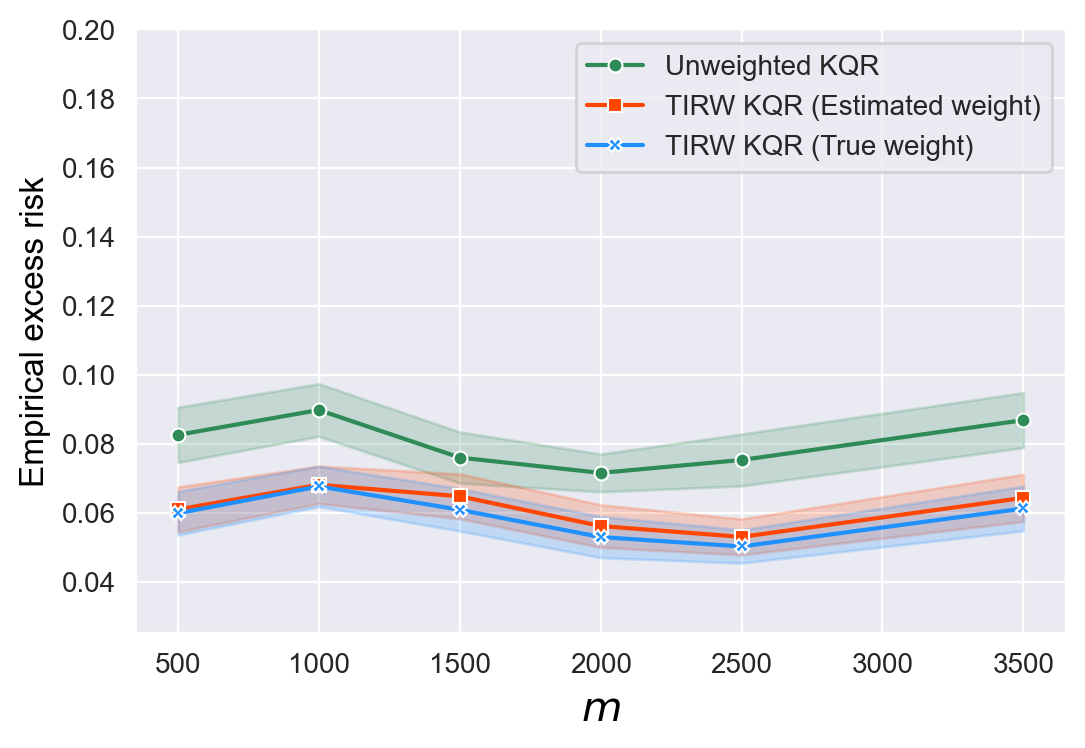}
     \label{KQR_dimension1_Empirical_15}}
 \caption{\footnotesize{Averaged MSE and empirical excess risk for unweighted KQR, TIRW KQR with true weight and estimated weight, respectively. Note that in (a) and (d), the curves are plotted with respect to $\log_{10} \lambda$ with $n=500,m=1000$; in (b) and (e) the curves are plotted with respect to $n$ with fixed $m=1000,\lambda=10^{-4}$; in (c) and (f), the curves are plotted with respect to $m$ with fixed $n=500,\lambda=10^{-4}$.}}
 \label{KQR_dimension1_unbounded_main}
\end{figure}

We investigate the performance of KQR under covariate shift with the following generating model 
\begin{align*}
y=f_0(x)+\left(1+r\left(x-0.5\right)^2\right)\sigma(\varepsilon-\Phi^{-1}(\tau)),  
\end{align*}
where $f_0(x)=\sin(\pi x)$ with $x \in {\cal R}$, $\Phi$ denotes the CDF function of the standard normal distribution and $\varepsilon\sim N(0,1)$. We consider $\rho_{\bx}^S \sim N(\mu_1,\sigma^2_1)$ and $\rho_{\bx}^T \sim N(\mu_2,\sigma^2_2)$ with  $\mu_1=0,\sigma_1^2=0.4, \mu_2=0.5,  \sigma_2^2=0.3$  for the uniformly bounded case and $\mu_1=0,\sigma_1^2=0.3,\mu_2=1,  \sigma_2^2=0.5$  for the moment bounded case, respectively. Moreover, we set $r=0$ and $\sigma=0.5$ for the homoscedastic case, and $r=1$ and $\sigma=0.3$ for the heteroscedastic case, respectively.  Note that the simulation results for the case that $\tau=0.3$ and $r=1$ are presented in Figure \ref{KQR_dimension1_bounded_main} under the uniformly bounded case and in Figure \ref{KQR_dimension1_unbounded_main} under the moment bounded case; completed results for other cases are provided in the supplementary material.  We compare the averaged mean square error (MSE) and empirical excess risk of the unweighted estimator and weighted estimator, either with true or estimated weights across different choices of regularization parameter $\lambda$, source sample size $n$, and target sample size $m$.

From (a) and (d) in Figure \ref{KQR_dimension1_bounded_main}, we can conclude that the error of the unweighted estimator is very close to that of the weighted estimator for the uniformly bounded case, which is consistent with our theoretical findings in Section \ref{main_results}. For the moment bounded case as demonstrated in (a) and (d) of Figure \ref{KQR_dimension1_unbounded_main}, the weighted estimator consistently outperforms its unweighted counterpart for all choices of $\lambda$. Even when $\lambda$ is far from its optimal choice, the weighted estimator still maintains a lower error level with significant improvement over the unweighted estimator whose error is on occasion extremely high for small choices of $\lambda$. Additionally, it is clear from Panels (b) and (e) of Figures \ref{KQR_dimension1_bounded_main} and  \ref{KQR_dimension1_unbounded_main} that in the moment bounded case, the error curves for the unweighted estimator have some significant gaps with those of the weighted estimator, while in the bounded case, these curves tend to coincide as $n$ grows. It is also clear from the table \ref{table3} that the trends in Figure 2 (b)-(e) and Figure 3 (b)-(e) almost agree with the explicit convergence rate where the Gaussian kernel with exponential decay is used. This phenomenon is consistent with our theoretical conclusion that the unweighted estimator can only achieve sub-optimal rates when the importance ratio is moment bounded and attain optimal rates when uniformly bounded.  Finally, we note that the target sample size has a subtle influence on all the estimators, as demonstrated in Panels (c) and (f) of Figures \ref{KQR_dimension1_bounded_main} and  \ref{KQR_dimension1_unbounded_main}.

\subsection{Real case study}
We consider the binary classification problem on the Raisin dataset, which is available in \url{https://archive.ics.uci.edu/ml/datasets.php}. The dataset contains 900 instances and 7 attributes. After standardization, the data are first randomly split into source and target datasets.  We introduce a binary random variable $s$ that serves as a labeling indicator and assign it to the source dataset if $s=0$. To ensure that the conditional distribution of different datasets remains invariant, we require $s$ to be conditionally independent with the response $y$ given the covariate $\bx$ \citep{zadrozny2004learning,quinonero2008dataset}. We implement KSVM in which the covariate shift only exists in the first covariate. Specifically, for $\ell>0$, we conduct the splitting rule as $P(s_i=1\mid {\bx}_i)=\min(1,(x_{i,0}-c)^2/\ell)$, where  $x_{i,0}$ is the first element of ${\bx}_i$ and $c=\min_{i}\{x_{i,0}\}$.  We refer to $\ell$ as the shift level that a smaller value of $\ell$ favors a greater distinction between source and target datasets.  We set the turning parameter $C_\lambda=(n\lambda)^{-1}$.  For the TIRW estimator, we use importance weighted cross validation (IWCV) \citep{sugiyama2007covariate} to tune the truncation parameter $\gamma_n$. We refer to Section A in the supplementary material for the details of the adopted method. 

\begin{figure}
\graphicspath{{KSVM/}}
    \centering
    \subfigure[$\ell=6$]{
    \includegraphics[width=1.6in,  height=1.09in]{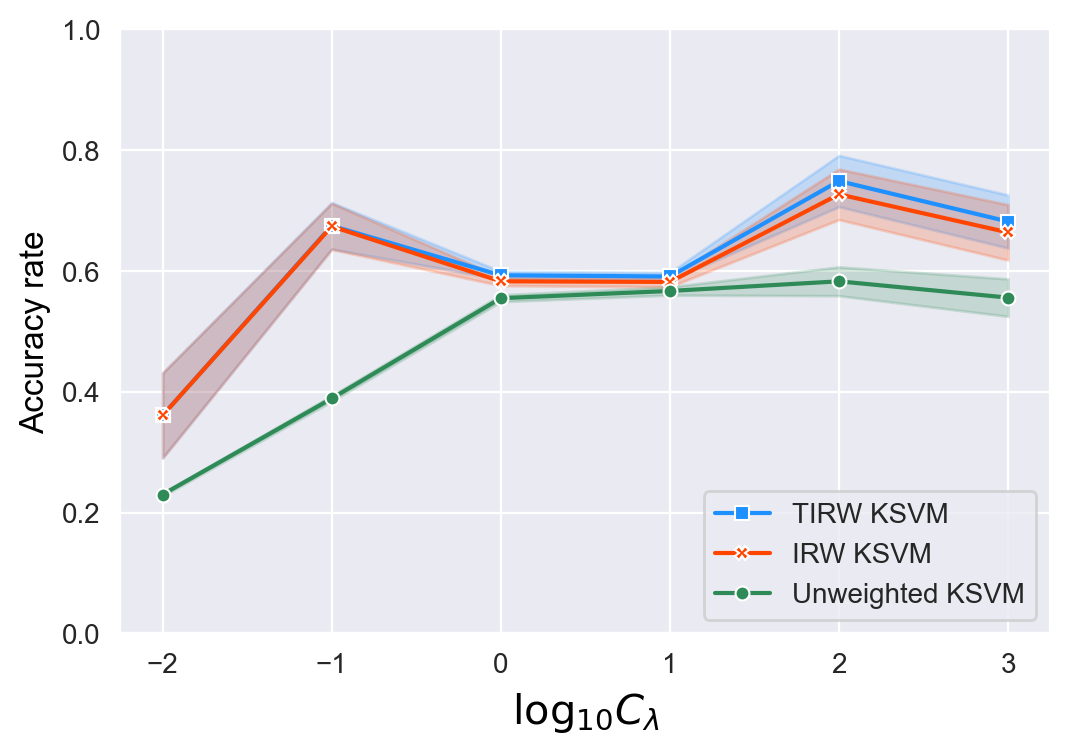}
    \label{KSVM_1}}
    \subfigure[$\ell=8$]{
	\includegraphics[width=1.6in,  height=1.09in]{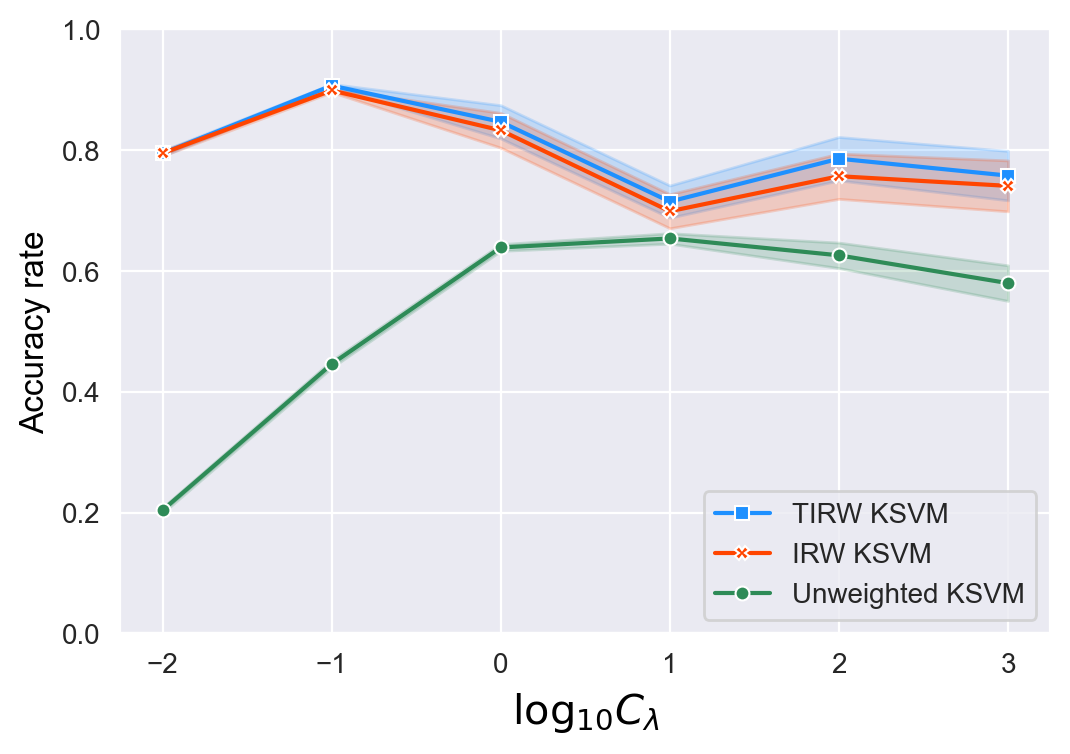}
\label{KSVM_2}}
   \subfigure[$\ell=10$]{
    \includegraphics[width=1.6in,  height=1.09in]{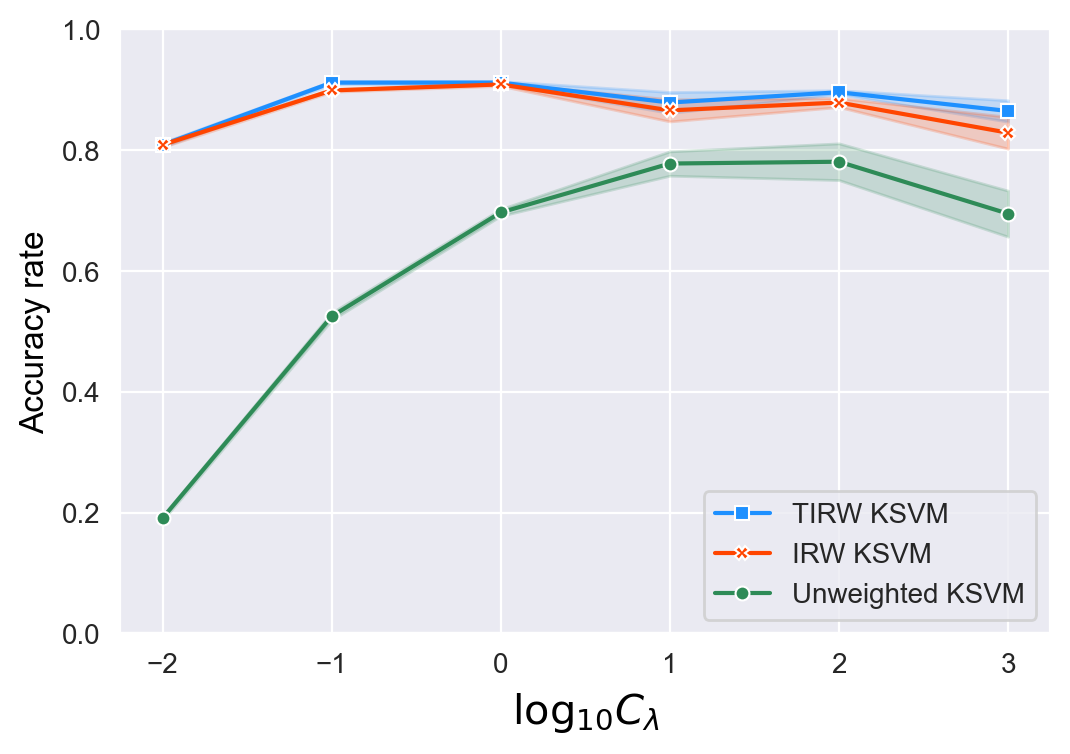}
        \label{KSVM_3}}
 \caption{\footnotesize{Averaged accuracy rate for unweighted KSVM, IRW KSVM and TIRW KSVM  with different shift levels $\ell$, respectively.}}
 \label{KSVM}
\end{figure}

Figure \ref{KSVM} summarizes the accuracy rate and standard error in different settings. As shown in this figure, the unweighted estimator has poor performance on target data, even for the optimal choice of $C_\lambda$. Nevertheless, the weighted estimator has a significant improvement in performance. Surprisingly, the weighted estimator is relatively stable with respect to the choices of $C_\lambda$, which is consistent with our results on synthetic data. Moreover, we can find that a proper truncation slightly improves the accuracy rate of the weighted estimator while reducing its variation.

\section{Conclusion}\label{sec:con}
In this work, we propose a unified analyzing framework for kernel-based methods under covariate shift and show that the predictive function can be better estimated by using the distribution information of the covariates from both aspects of theory and numerical performance.  The established theoretical results fulfill the gap in understanding the influence of the covariate shift under various kernel-based learning problems, both researchers and practitioners may benefit from our theoretical findings. Extensive numerical studies also provide empirical evidence and confirm the established theoretical results. Note that it is very interesting to test if there exists a distribution shift in practical applications and if the distribution shift satisfies the uniformly bounded or moment bounded assumptions. Unfortunately, the relevant approaches have remained lacking to our best knowledge. And it is interesting to note that as shown in our real applications, the TIRW estimator always outperforms the unweighted estimator, and thus we suggest using the TIRW estimator to analyze the real-life dataset. Moreover, in the theory, we directly use the true importance ratio, and it is difficult to derive the consistency of the estimated ratio and plug it into our theoretical analysis. In the machine learning literature, there exist several measures to quantify the divergence between two distributions, including but not limited to $f$-divergence, Wasserstein distance and kernel mean matching. It is still an open problem if it still works when we use other measures rather than the importance ratio. We leave such promising and interesting topics as potential future work.

\section*{Acknowledgment}
Xingdong Feng's research is supported in part by  NSFC-12371270. Xin He's research is supported in part by NSFC-11901375  and Program for Innovative Research Team of Shanghai University of Finance and Economics.  This research is also supported by Shanghai Research Center for Data Science and Decision Technology and the Fundamental Research Funds for the Central Universities.

\bibliographystyle{model5-names}
\bibliography{ref.bib}

\newpage

\appendix

\section{Additional numerical results} \label{A}
In this section, we provide more experiment results, including KRR (Section \ref{sec A.1}), KQR for various $\tau$ and $r$ (Section \ref{sec A.2}), kernel logistic regression (KLR) (Section \ref{sec A.3}) and multi-source real data studies (Section \ref{sec A.4}), that further confirm our theoretical findings. Section \ref{sec A.7} and Section \ref{sec A.8} are devoted to introducing KLIEP and IWCV. We also add some discussion on Assumption 2 of the main text in Section \ref{sec A.9}. The python code implementing the proposed method is available at \url{https://github.com/WangCaixing-96/Kernel_CS}.

\subsection{Kernel ridge regression} \label{sec A.1}
For the squared loss, we consider the following two examples. 

\begin{figure}
\graphicspath{{bounded_KRR_1_d/}}
    \centering
    \subfigure[]{
    \includegraphics[width=1.6in,  height=1.09in]{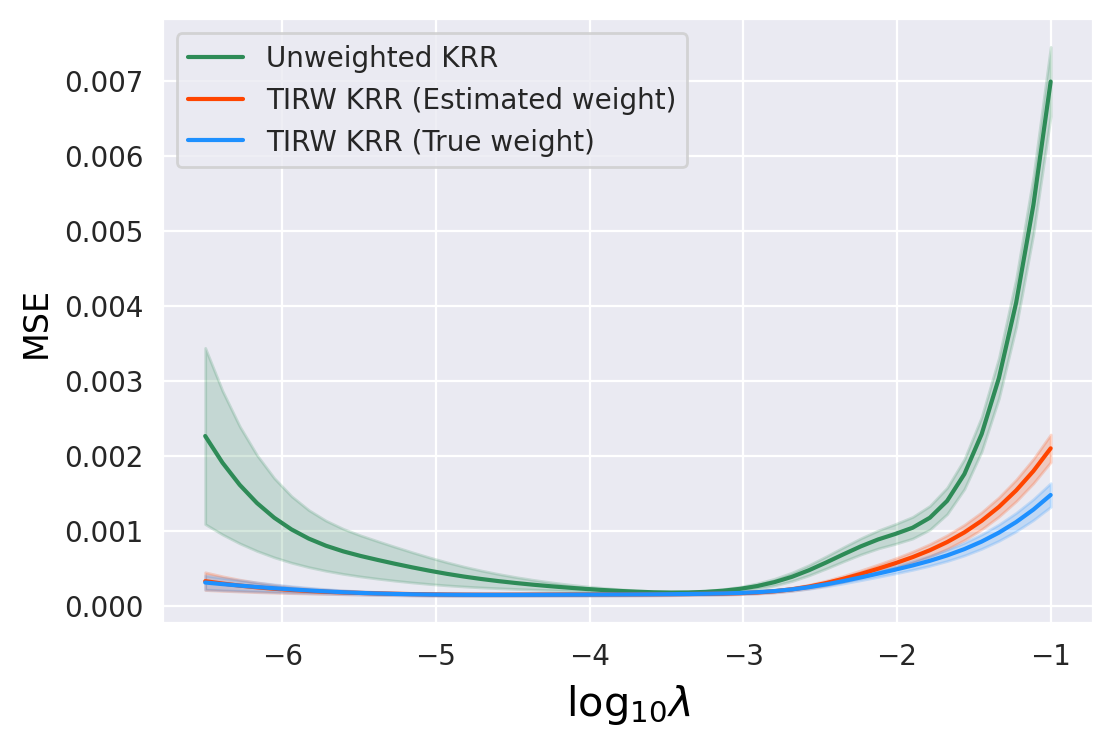}\label{KRR_dimension1_uniformbounded_MSE_smooth_1}}
    \subfigure[]{
	\includegraphics[width=1.6in,  height=1.09in]{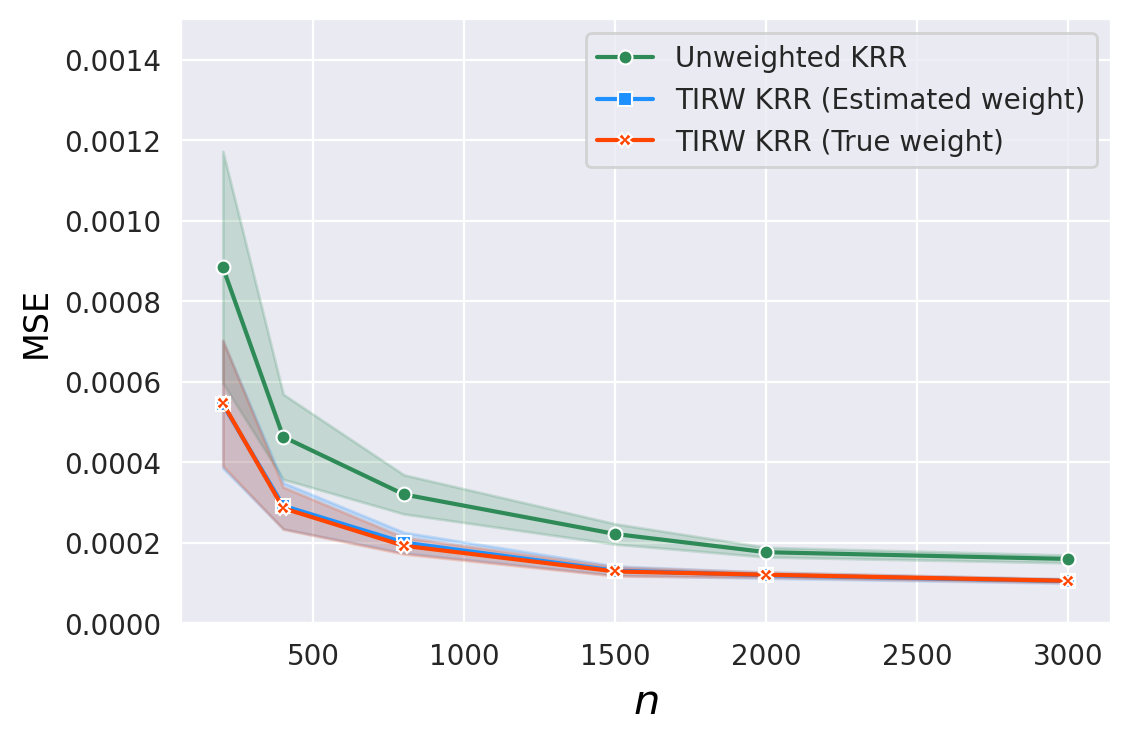}
\label{KRR_dimension1_uniformbounded_MSE_2}}
   \subfigure[]{
\includegraphics[width=1.6in,height=1.09in]{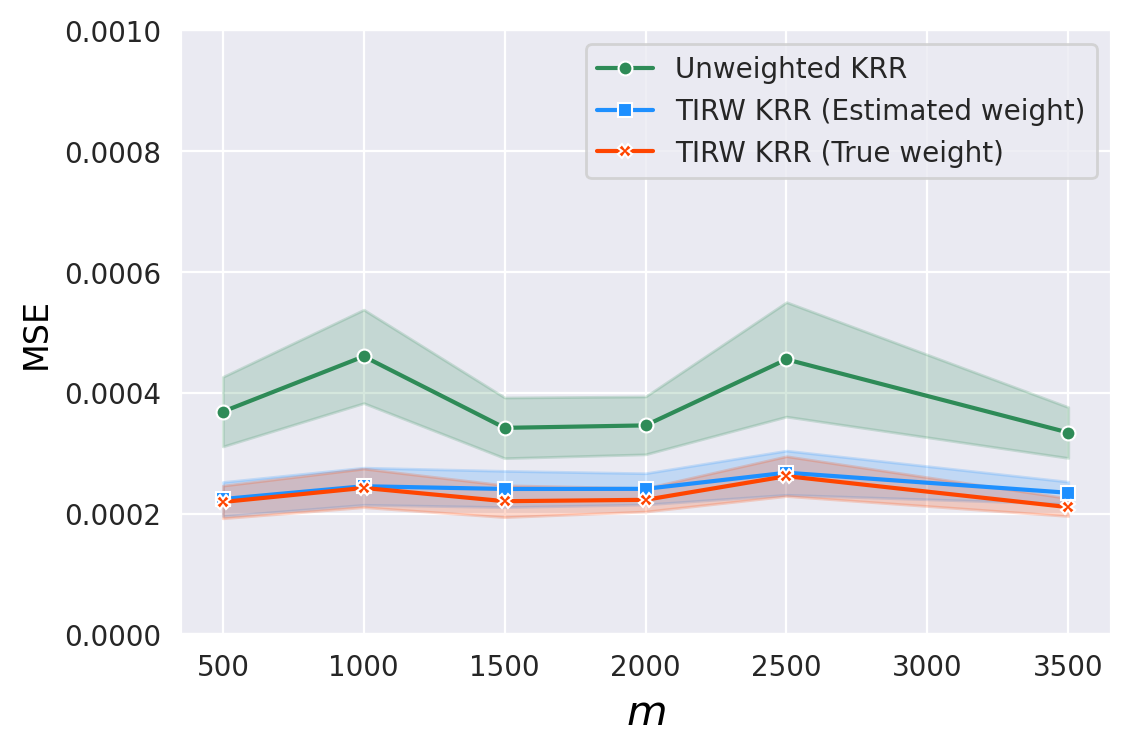}\label{KRR_dimension1_uniformbounded_MSE_3}}
 \graphicspath{{unbounded_KRR_1_d/}}
    \centering
    \subfigure[]{
    \includegraphics[width=1.6in,  height=1.09in]{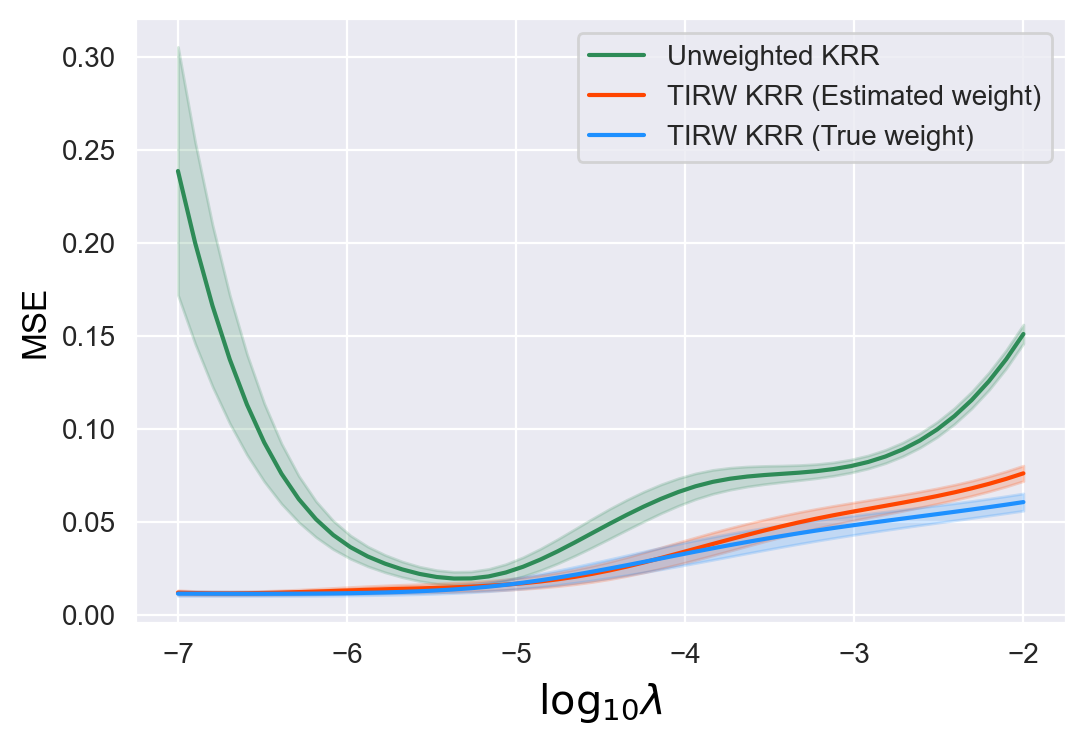}\label{KRR_dimension1_MSE_smooth_1}}
    \subfigure[]{
	\includegraphics[width=1.6in,  height=1.09in]{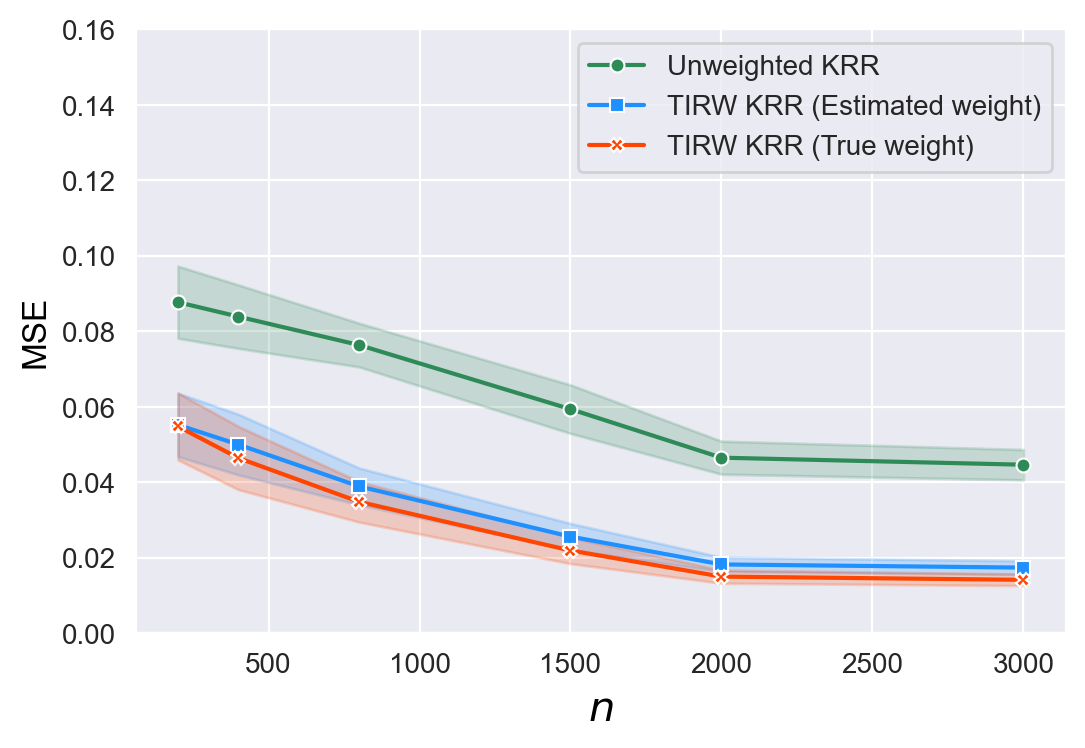}
\label{KRR_dimension1_MSE_2}}
   \subfigure[]{
\includegraphics[width=1.6in,height=1.09in]{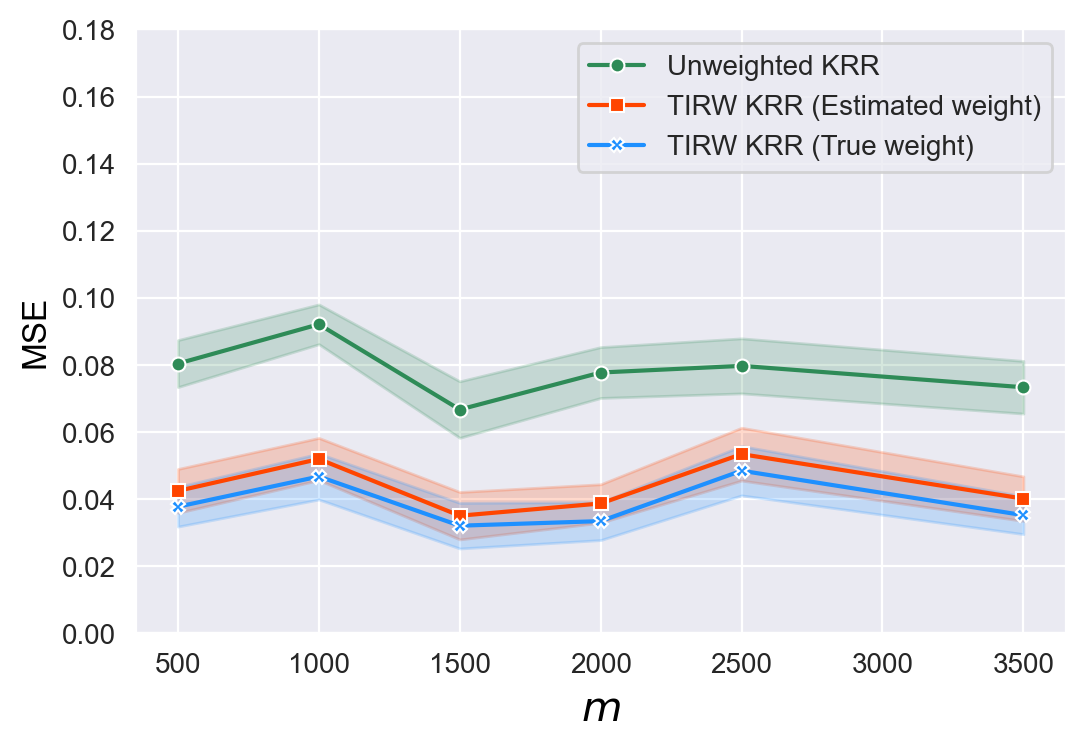}\label{KRR_dimension1_MSE_3}}
\caption{\footnotesize{Average MSE for unweighted KRR, TIRW KRR with true weight and estimated weight, respectively (The top panel is for the bounded case and the bottom is for the moment bounded case; in (a) and (d), the curves are plotted with respect to $\log_{10} \lambda$ with $n=500, m=1000$; in (b) and (e) the curves are plotted with respect to $n$ with fixed $m = 1000,\lambda=10^{-4}$; in (c) and (f), the curves are plotted with respect to $m$ with fixed $n=500,\lambda=10^{-4}$)}}
\label{KRR_dimension1}
\end{figure}

\noindent{\bf Example S1}: The response $y$ is generated by  $ y=f_0(x)+\sigma\varepsilon,$  
where $f_0(x)=e^{-\frac{1}{x^2}}$ and $\varepsilon\sim N(0,1)$.  The source and target distributions are  $\rho^S_{\bx}({\bx})\sim N(\mu_1,\sigma^2_1)$ and   $\rho^T_{\bx}({\bx})\sim N(\mu_2,\sigma^2_2)$ with   $\mu_1=0,\mu_2=0.8,\sigma^2_1=0.5,\sigma^2_2=0.3$ for the uniformly bounded case and $\mu_1=0,\mu_2=1.5,\sigma^2_1=0.3,\sigma^2_2=0.5$ for the moment bounded case, respectively. The noise level $\sigma$ is set to  $0.05$ for both uniformly bounded and moment bounded cases. The results are shown in Figure \ref{KRR_dimension1}

\begin{figure}
\graphicspath{{bounded_KRR_3_d/}}
    \centering
    \subfigure[]{
    \includegraphics[width=1.6in,  height=1.09in]{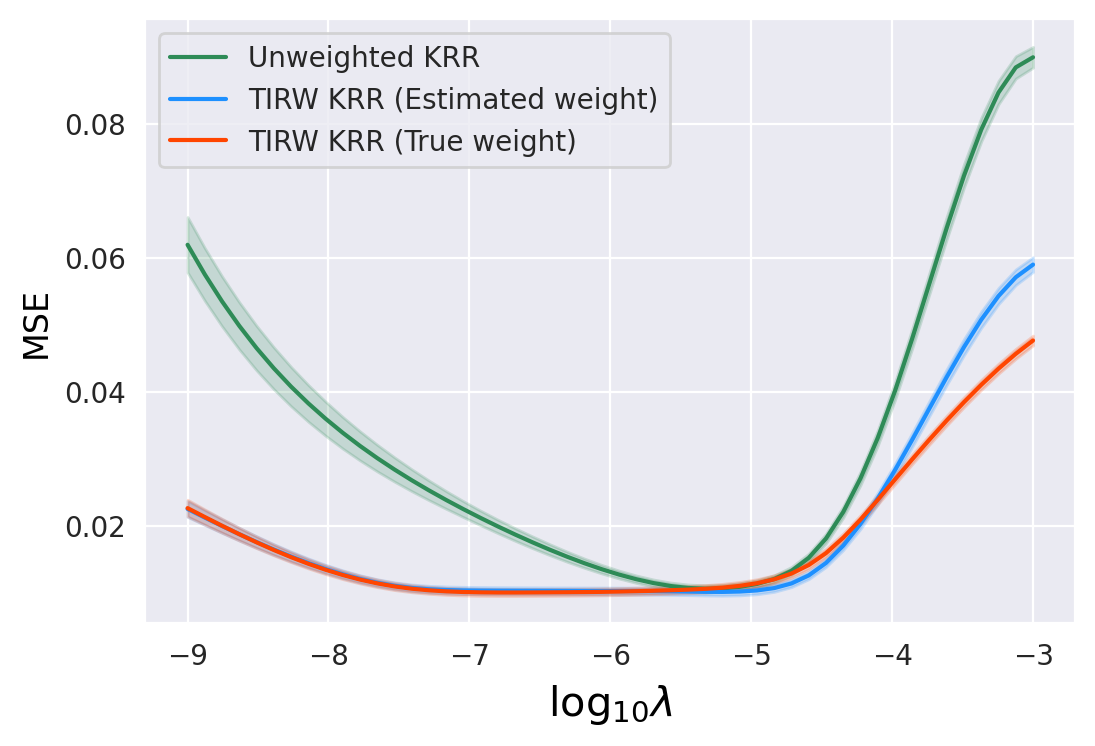}\label{KRR_dimension3_uniformbounded_MSE_smooth_1}}
    \subfigure[]{
\includegraphics[width=1.6in,  height=1.09in]{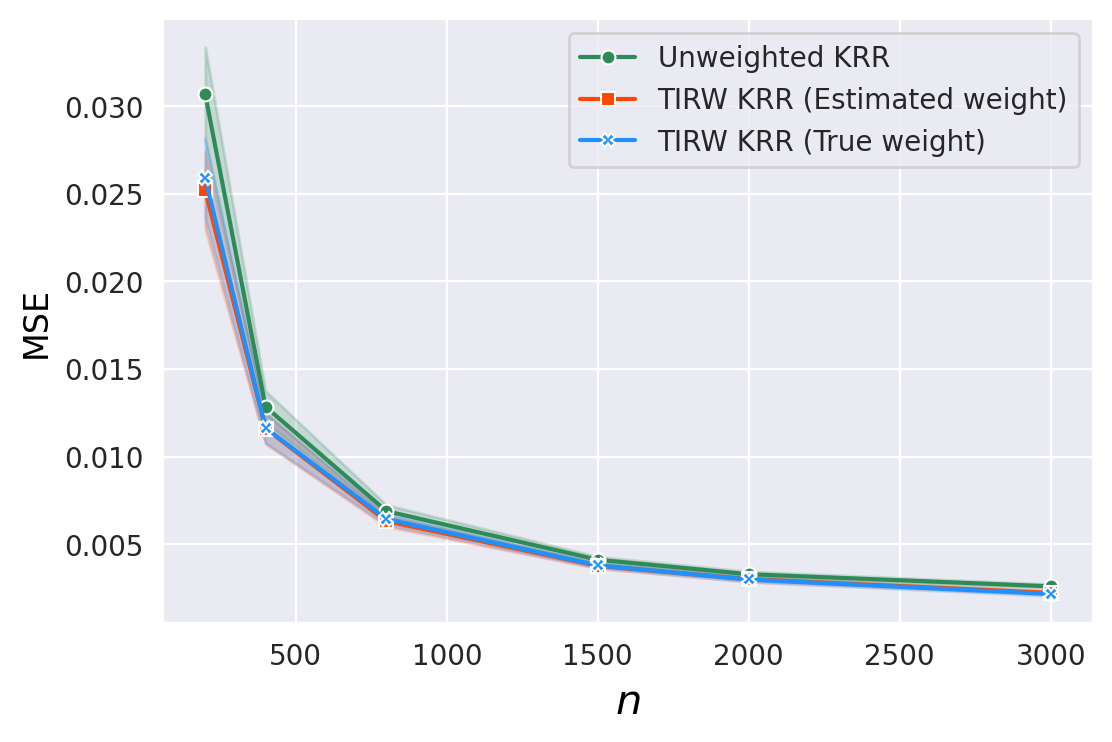}
\label{KRR_dimension3_uniformbounded_MSE_2}}
   \subfigure[]{
\includegraphics[width=1.6in,height=1.09in]{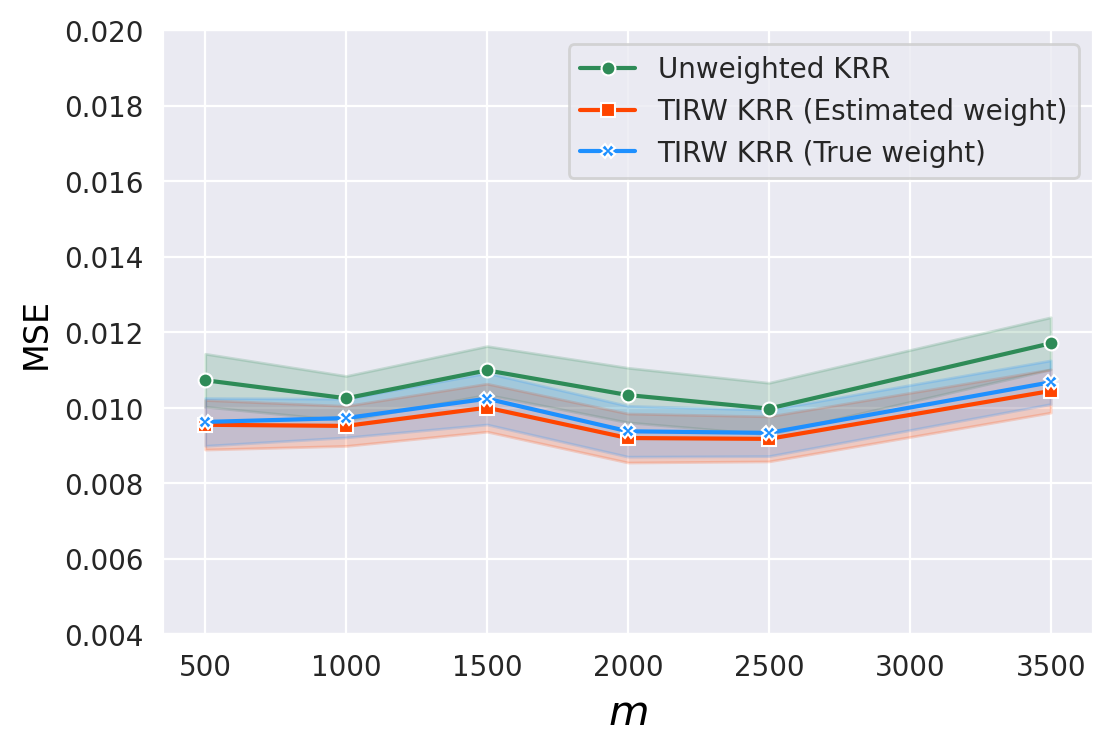}\label{KRR_dimension3_uniformbounded_MSE_3}}
\graphicspath{{unbounded_KRR_3_d/}}
    \centering
    \subfigure[]{
    \includegraphics[width=1.6in,  height=1.09in]{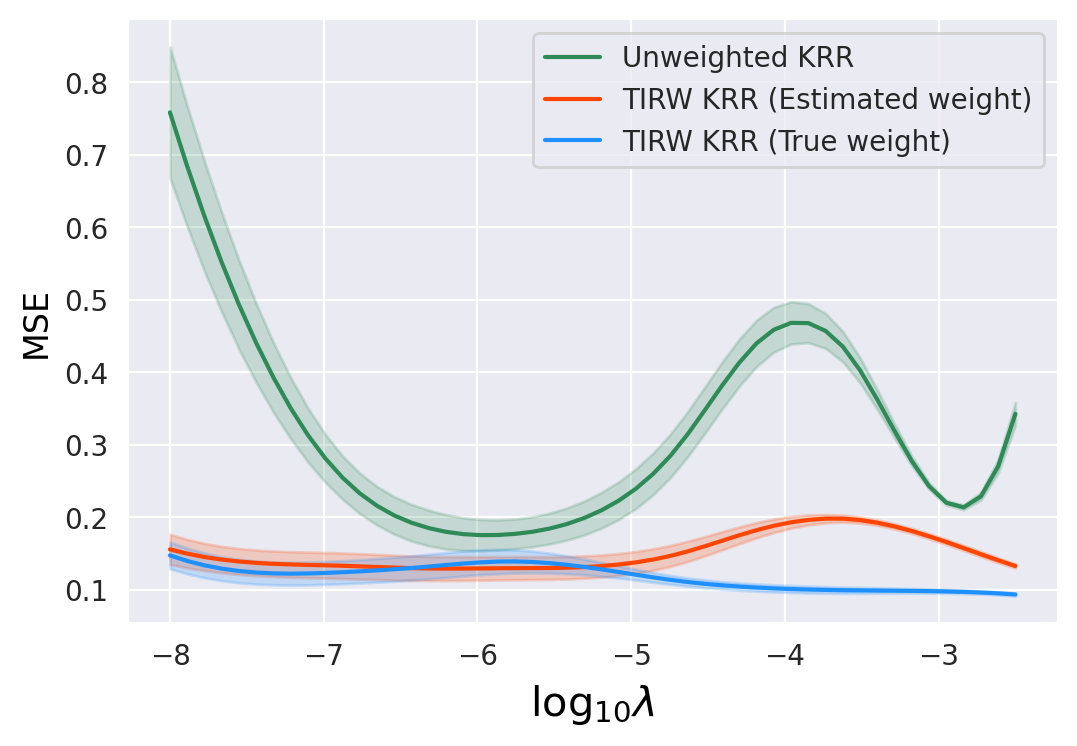}\label{KRR_dimension3_MSE_smooth1}}
    \subfigure[]{
	\includegraphics[width=1.6in,  height=1.09in]{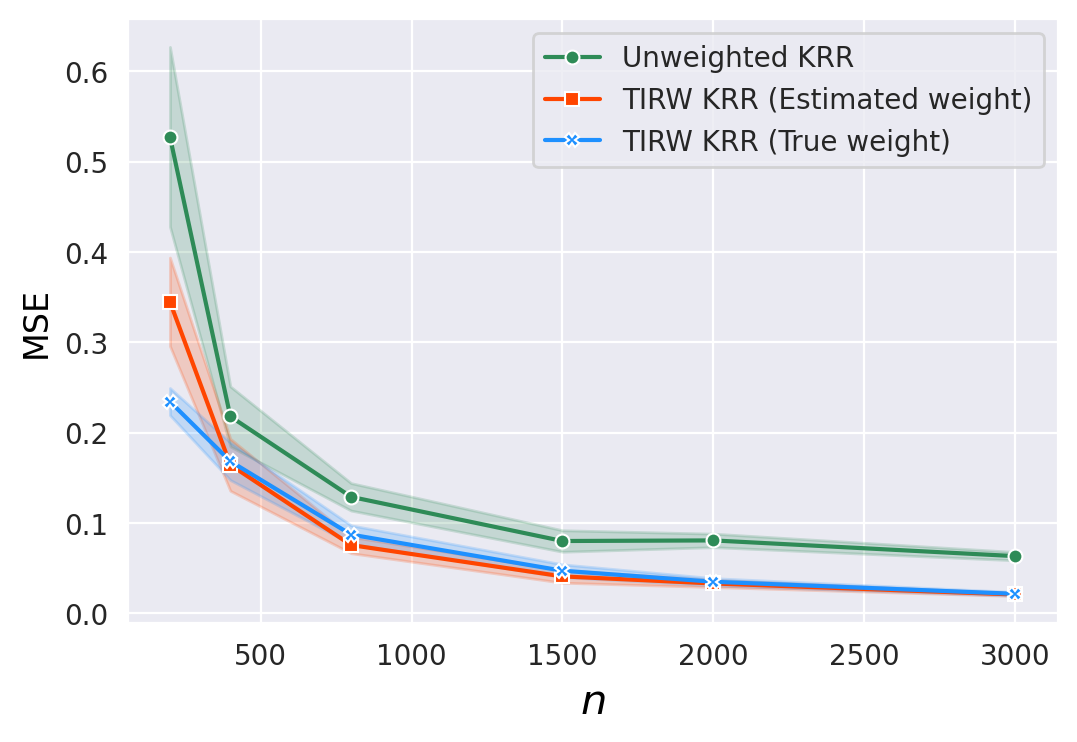}
\label{KRR_dimension3_MSE_2}}
   \subfigure[]{
\includegraphics[width=1.6in,height=1.09in]{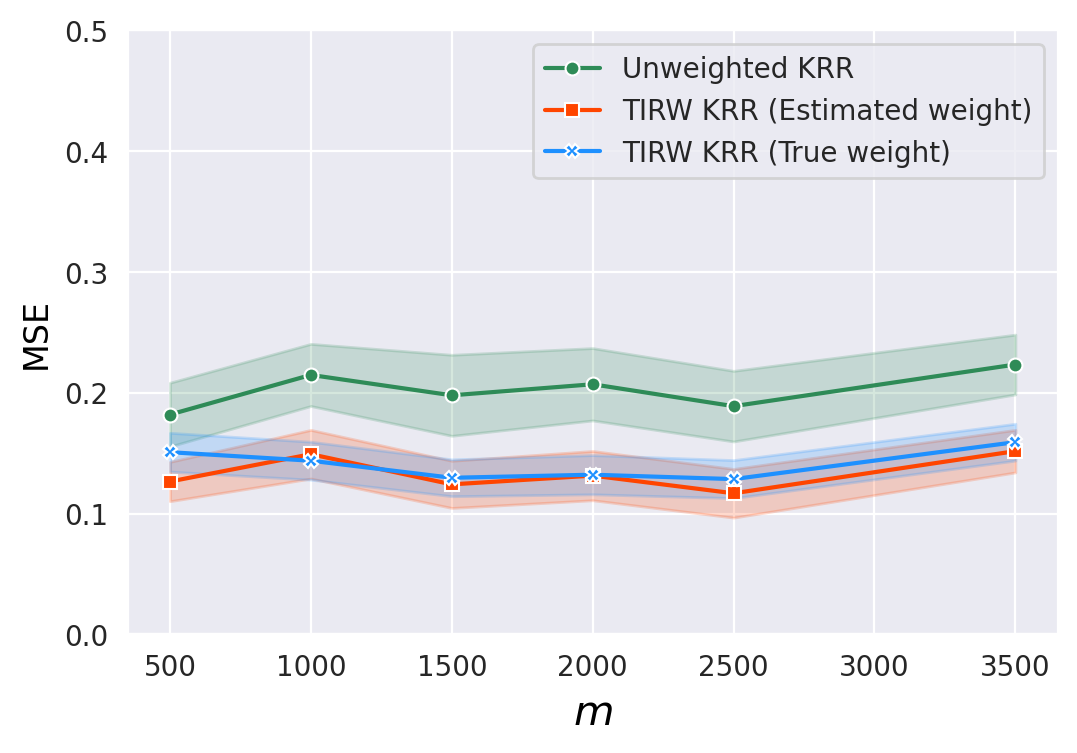}\label{KRR_dimension3_MSE_3}}
\caption{\footnotesize{Average MSE for unweighted KRR, TIRW KRR with true weight and estimated weight, respectively (The top panel is for the bounded case and the bottom is for the moment bounded case; in (a) and (d), the curves are plotted with respect to $\log_{10} \lambda$ with $n=500, m=1000$; in (b) and (e) the curves are plotted with respect to $n$ with fixed $m = 1000,\lambda=5\times 10^{-5}$; in (c) and (f), the curves are plotted with respect to $m$ with fixed $n=500,\lambda=5\times 10^{-5}$)}}
\label{KRR_dimension3}
\end{figure}

\noindent{\bf Example S2}:  The response $y$ is generated by 
$
y=f_0({\bx})+\sigma\varepsilon,
$
where ${\bx}=(x_0,x_1,x_2)^\top\in \mathcal{R}^3,  f_0({\bx})=\sin(2\pi x_0)-e^{-x_1^2-x_2^2}$ and $\varepsilon\sim N(0,1)$.  
Let $g(x_0; \alpha,\beta)$ denote the probability density function of Beta distribution with parameters $\alpha,\beta$. It is easy to see that the importance ratio $\phi({\bx})$ for this case is not uniformly bounded but second moment bounded if and only if $\alpha_s<\alpha_t, 2\alpha_s\ge\alpha_t, 2\beta_s\ge\beta_t$ or $\beta_s<\beta_t, 2\alpha_s\ge\alpha_t, 2\beta_s\ge\beta_t$. In the 3-dimensional KRR experiment, we consider  $\rho^S_{\bx}({\bx})=g(x_0; \alpha_s,\beta_s)$ and   $\rho^T_{\bx}({\bx})=g(x_0; \alpha_t,\beta_t)$ with   $\alpha_s=2.5,\beta_s=1.5,\alpha_t=3,\beta_t=4$ for the uniformly bounded case and $\alpha_s=4,\beta_s=1,\alpha_t=3,\beta_t=6$ for the moment bounded case, respectively. The noise level $\sigma$ is set to  $0.3$ for both uniformly bounded and moment bounded cases. The results are shown in Figure \ref{KRR_dimension3}

From (d) in Figures \ref{KRR_dimension1} and  \ref{KRR_dimension3}, we observe that for the moment bounded case, the TIRW estimator has a great improvement compared to the unweighted estimator, even for the choice of $\lambda$ that is far away from the optimum. Nevertheless, for the bounded case, we can see from (a) in Figure \ref{KRR_dimension1} and Figure \ref{KRR_dimension3} that there has a negligible gap between the performance of the unweighted estimator and that of the TIRW estimator as long as we choose $\lambda$ that is close to optimum. For the poor choice of $\lambda$, the TIRW estimator still performs significantly better. From (b) and (e) in Figures \ref{KRR_dimension1} and  \ref{KRR_dimension3}, it is shown that the error curve has an explicit gap with those of weighted estimators for the moment bounded case, whereas it is very close for the uniformly bounded case. From (c) and (f) in Figures \ref{KRR_dimension1} and  \ref{KRR_dimension3}, we observe that the target data size $m$ has a subtle influence on our estimators.

\subsection{Kernel quantile regression}\label{sec A.2}
For the check loss, we consider the following two examples. 

\noindent{\bf Example S3}: This example continues to study the KQR with a 1-dimensional covariate under the same setting as in the main text.  Here we further conduct experiments for various combinations of $\tau\in \{0.3,0.5,0.7\}$ and $r\in \{0,1\}$.

\noindent{\bf Example S4}: The response $y$ is generated by 
$
y=f_0(\bx)+(1+rx_0)\sigma(\varepsilon-t_4^{-1}(\tau)), \quad \bx=(x_0,x_1,x_2)^\top\in \mathcal{R}^3,    
$\
where $f_0(\bx)=\sin(1.5\pi x_0)-e^{-x_1^2-x_2^2}$, $t_4$ denotes the CDF function of t-distribution with $4$ degrees of freedom and $\varepsilon\sim t_4$.  
We consider  $\rho^S_{\bx}({\bx})=g(x_0; \alpha_s,\beta_s)$ and   $\rho^T_{\bx}({\bx})=g(x_0; \alpha_t,\beta_t)$ with   $\alpha_s=2.5,\beta_s=1.5,\alpha_t=3,\beta_t=6$ for the uniformly bounded case and $\alpha_s=5.5,\beta_s=1.5,\alpha_t=3,\beta_t=6$ for the moment bounded case, respectively. We set $r=0$ and $\sigma= 0.3$ for the homoscedastic case and $r=1$ and $\sigma= 0.3$ for the heteroscedastic case, respectively. 

Note that the numerical results are provided in Sections \ref{sec:a.2.1}--\ref{sec:a.2.4}. Specifically, Figures \ref{KQR_dimension1_bounded} in Section \ref{sec:a.2.1} present the results for the uniformly bounded case in Example S3, and the results for the moment bounded case in Example S3 are presented in Figure \ref{KQR_dimension1_unbounded} of Section \ref{sec:a.2.2}. Moreover, Figure \ref{KQR_dimension3_bounded} in Section \ref{sec:a.2.3} presents the results for the uniformly bounded case in Example S4 with various combinations of $\tau\in \{0.3,0.5,0.7\}$ and $r\in \{0,1\}$, and the results for the moment bounded case in Example S4 are presented in Figure \ref{KQR_dimension3_unbounded} of Section \ref{sec:a.2.4}.

It is thus clear that the TIRW estimator is robust for different combinations of $\tau$ and $r$. Specifically, the TIRW estimator takes a significant advantage over the unweighted estimator when the important ratio is indeed unbounded. Nevertheless, for the bounded case, the TIRW estimator seems to be not necessary since if the choice of turning parameter is nearly-optimal or the source data size is relatively enough, there is a  negligible gap between the TIRW estimator and the unweighted estimator.

\subsubsection{Uniformly bounded case in Example S3}\label{sec:a.2.1}
\renewcommand{\thesubfigure}{(1\alph{subfigure})}
\begin{figure}[H]
\graphicspath{{bounded_KQR_1_d/}}
    \centering
    \subfigure[$\tau=0.3$ and $r=0$]{
\includegraphics[width=1.6in,  height=1.09in]{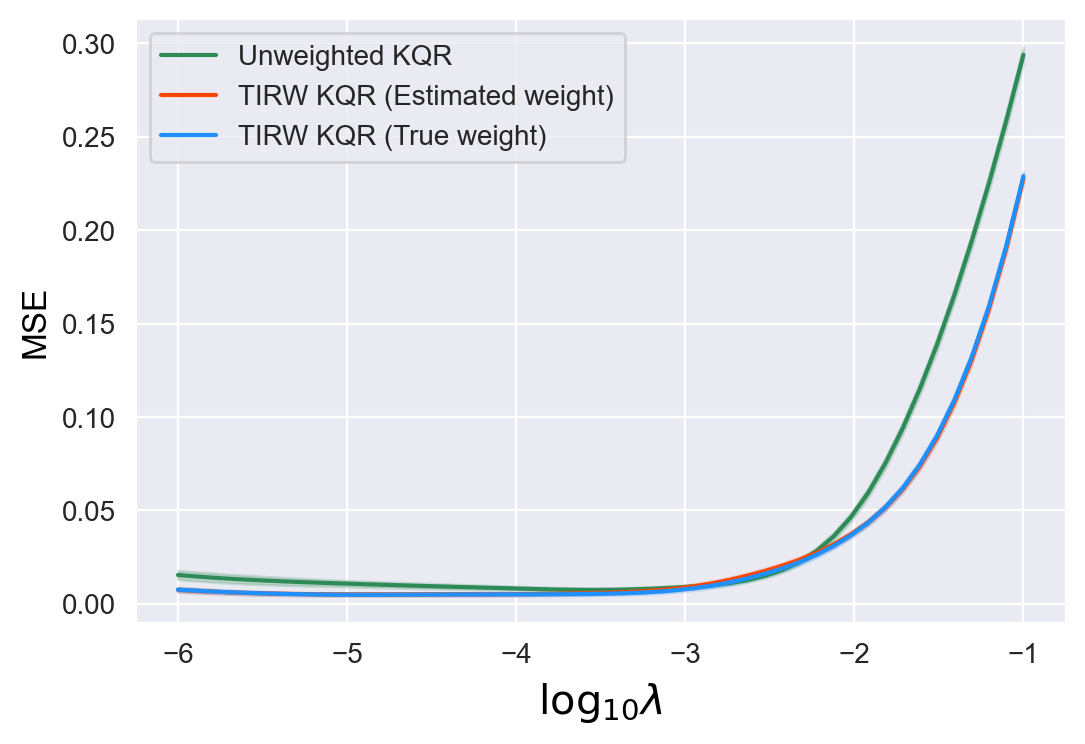}
\label{KQR_dimension1_uniformbounded_MSE_smooth_4}}
    \subfigure[$\tau=0.3$ and $r=0$]{
	\includegraphics[width=1.6in,  height=1.09in]{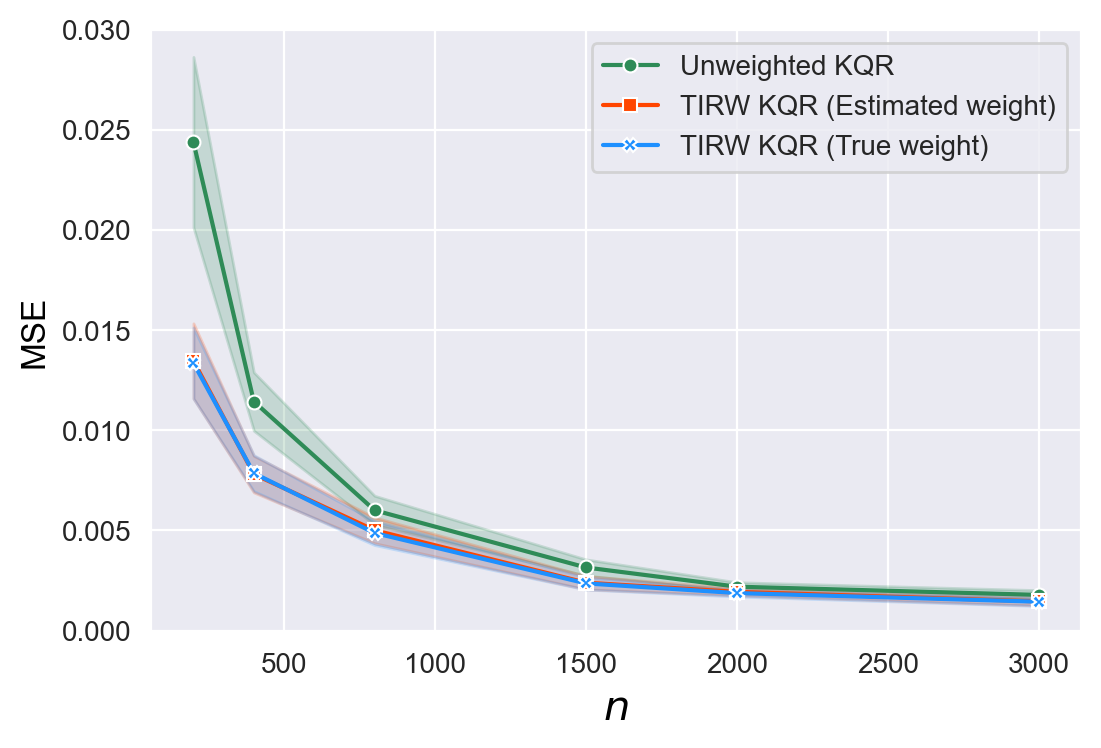}
\label{KQR_dimension1_uniformbounded_MSE_10}}
   \subfigure[$\tau=0.3$ and $r=0$]{
    \includegraphics[width=1.6in,  height=1.09in]{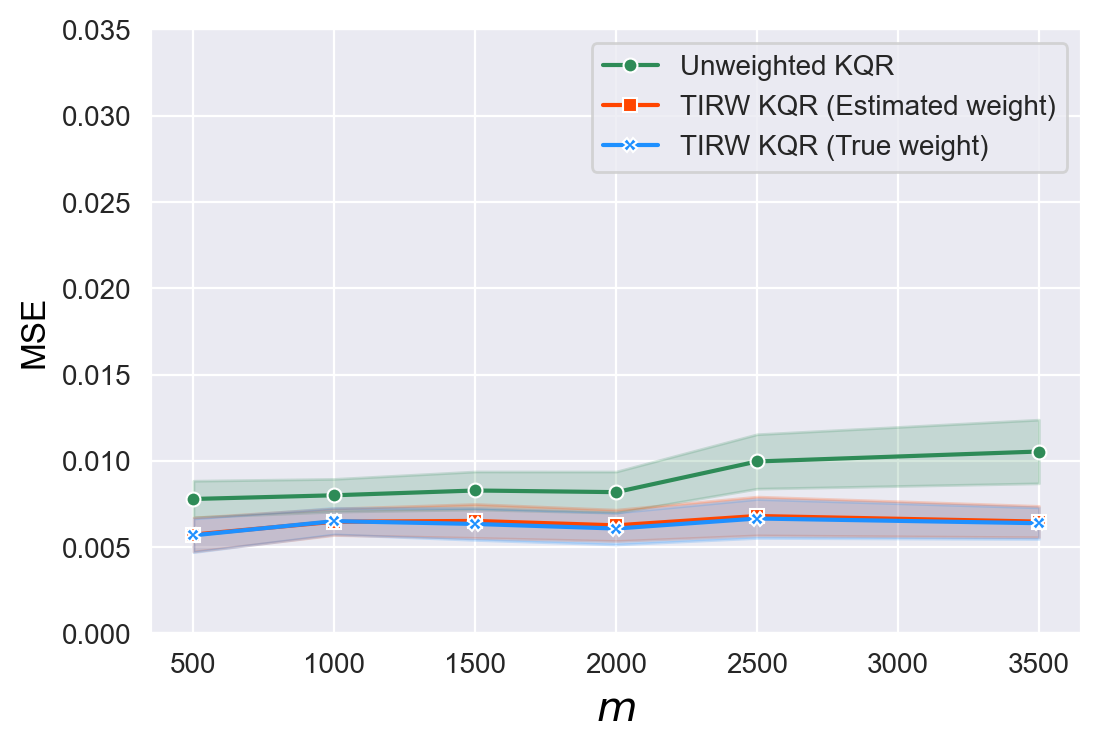}
\label{KQR_dimension1_uniformbounded_MSE_16}}
    \subfigure[$\tau=0.3$ and $r=0$]{
	\includegraphics[width=1.6in,  height=1.09in]{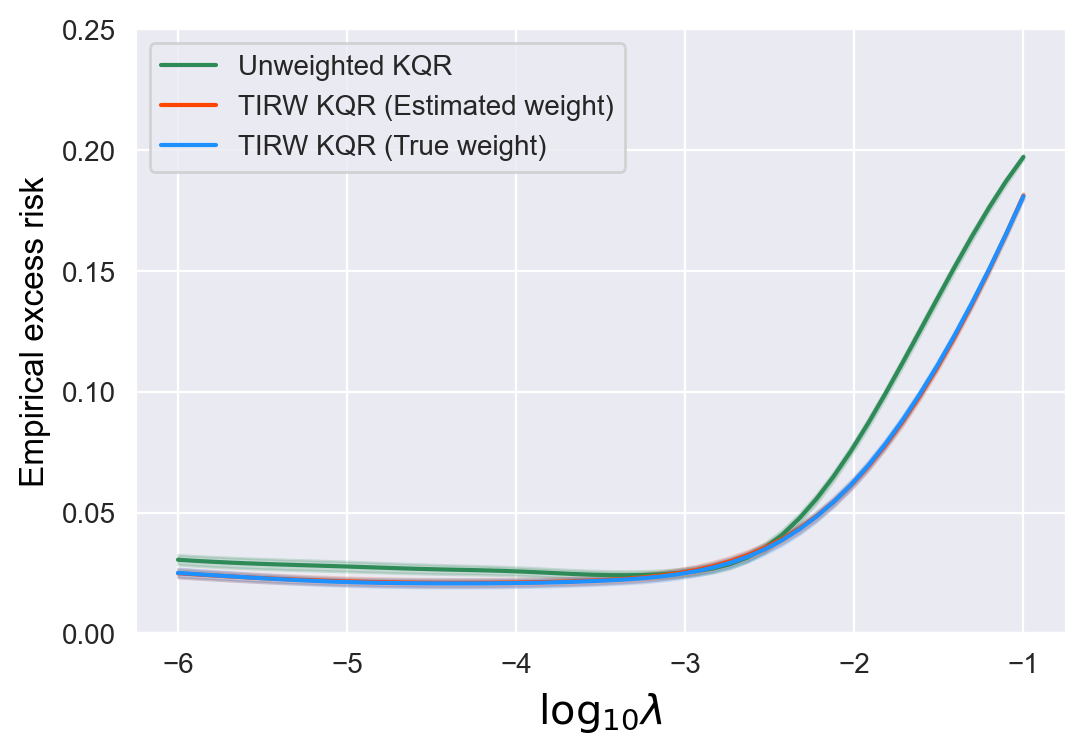}
\label{KQR_dimension1_uniformbounded_Empirical_smooth_4}}
 \subfigure[$\tau=0.3$ and $r=0$]{
    \includegraphics[width=1.6in,  height=1.09in]{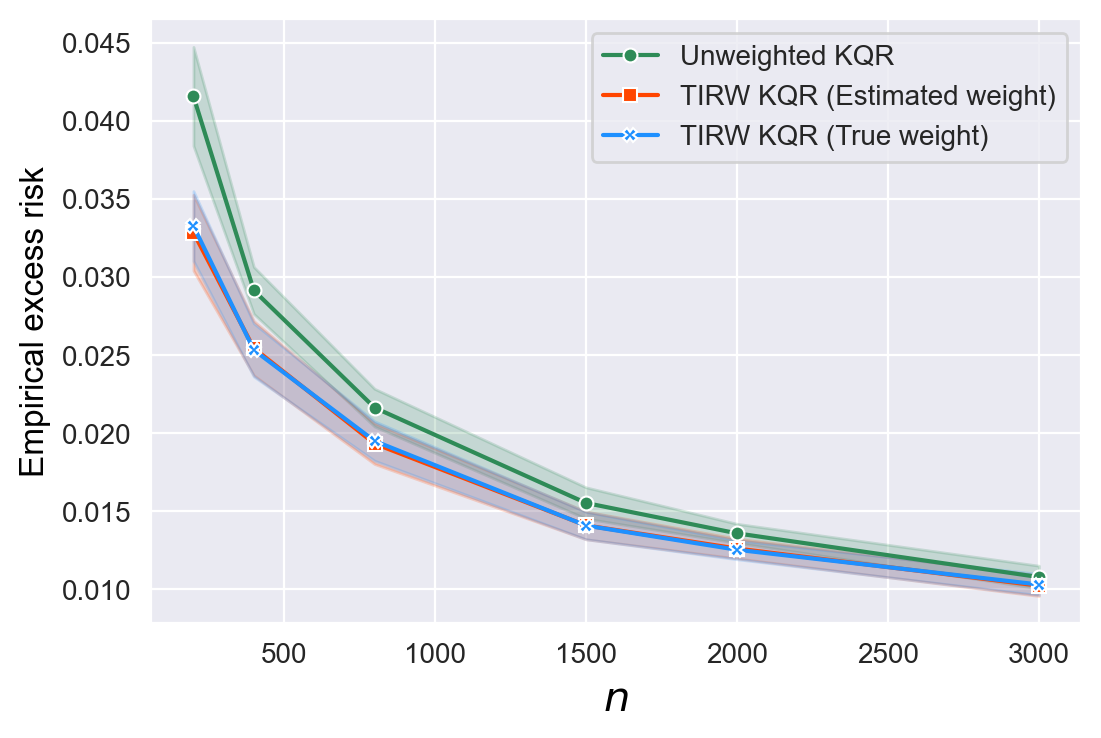}
\label{KQR_dimension1_uniformbounded_Empirical_10}}
    \subfigure[$\tau=0.3$ and $r=0$]{
	\includegraphics[width=1.6in,  height=1.09in]{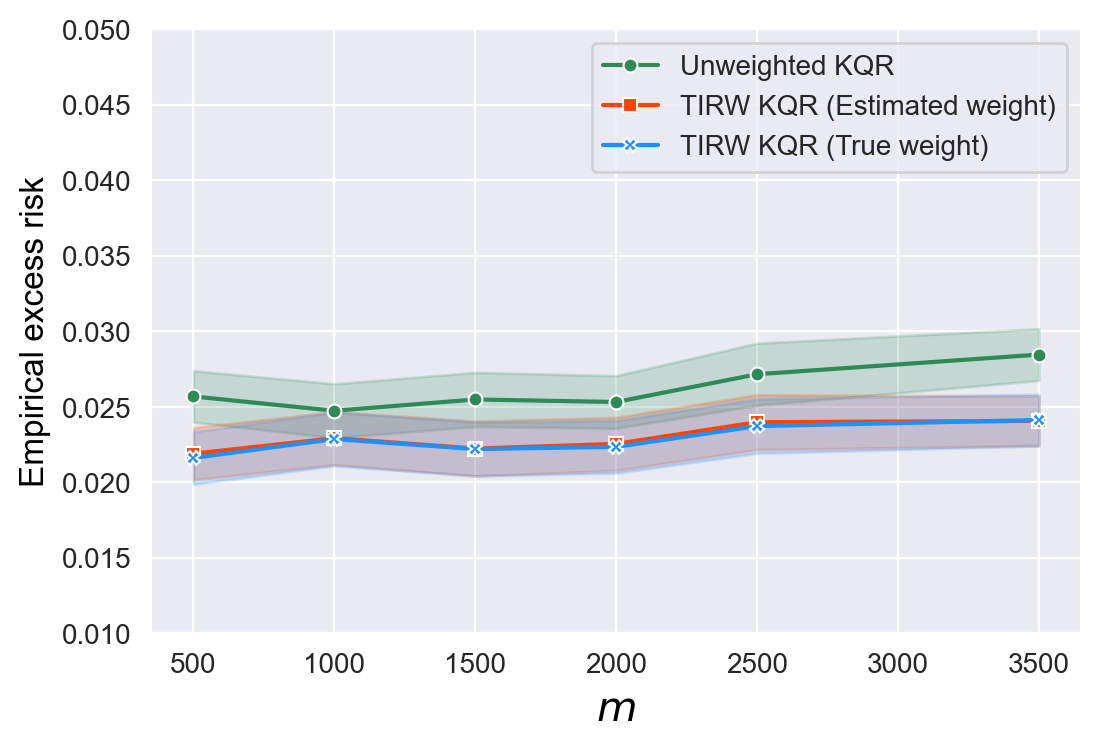}
\label{KQR_dimension1_uniformbounded_Empirical_16}}
\setcounter{subfigure}{0}
\renewcommand{\thesubfigure}{(2\alph{subfigure})}
  \subfigure[$\tau=0.5$ and $r=1$]{
    \includegraphics[width=1.6in,  height=1.09in]{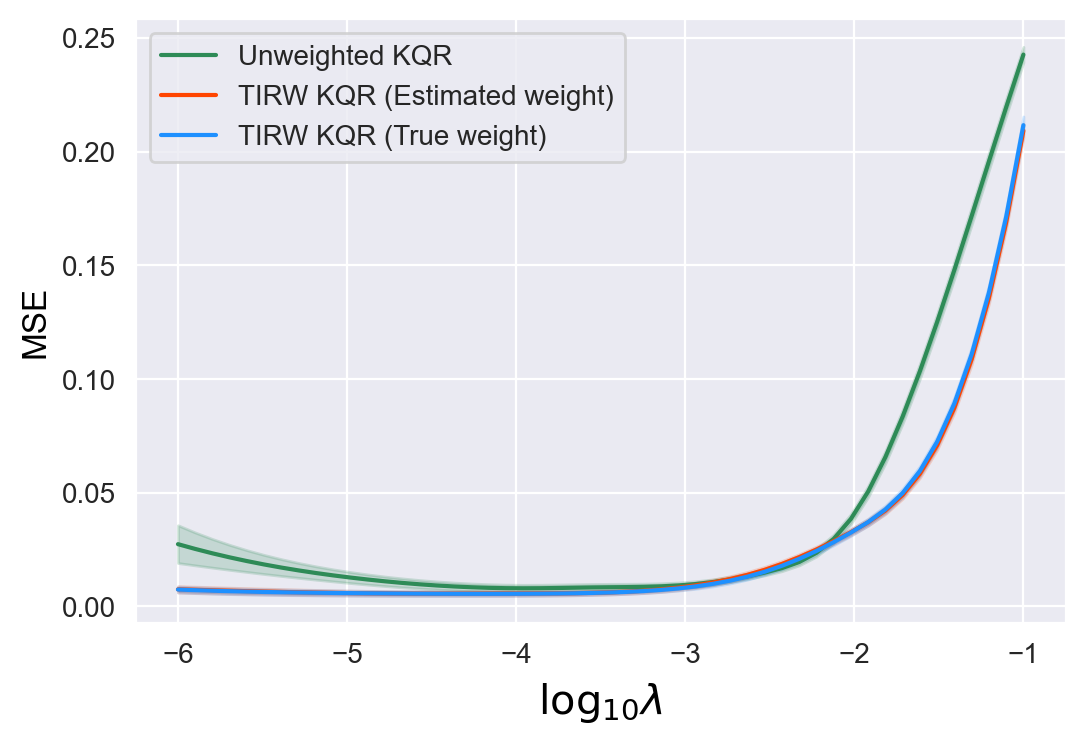}
\label{KQR_dimension1_uniformbounded_MSE_smooth_1}}
    \subfigure[$\tau=0.5$ and $r=1$]{
	\includegraphics[width=1.6in,  height=1.09in]{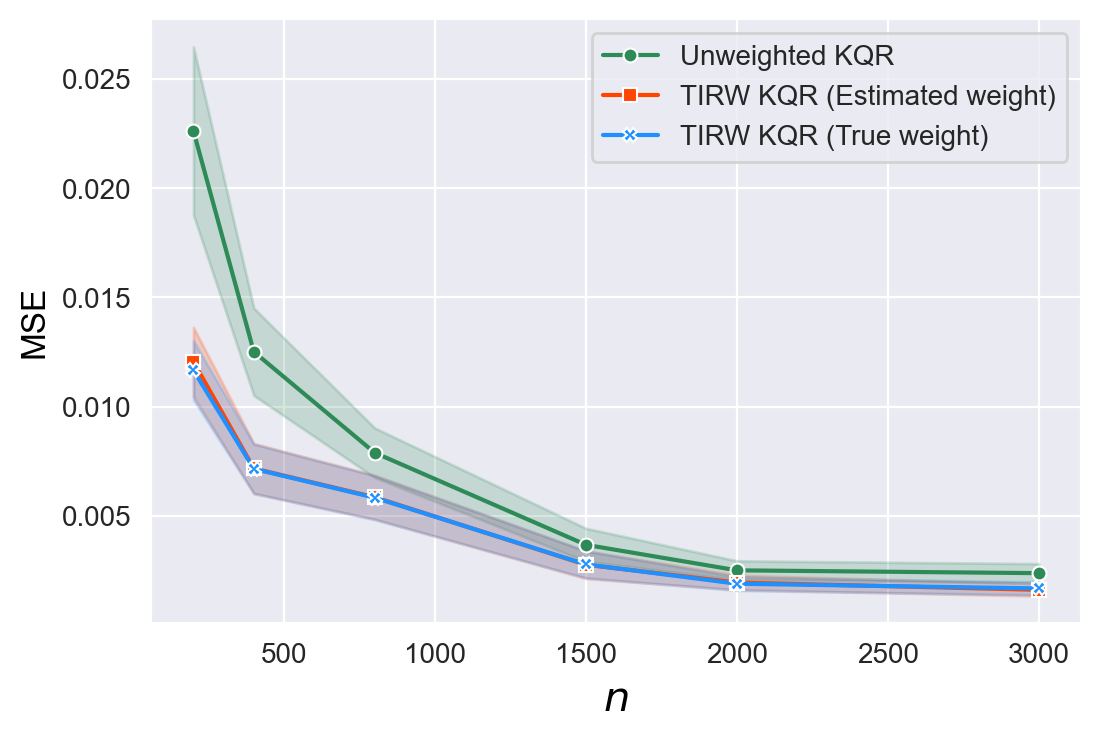}
\label{KQR_dimension1_uniformbounded_MSE_7}}
   \subfigure[$\tau=0.5$ and $r=1$]{
    \includegraphics[width=1.6in,  height=1.09in]{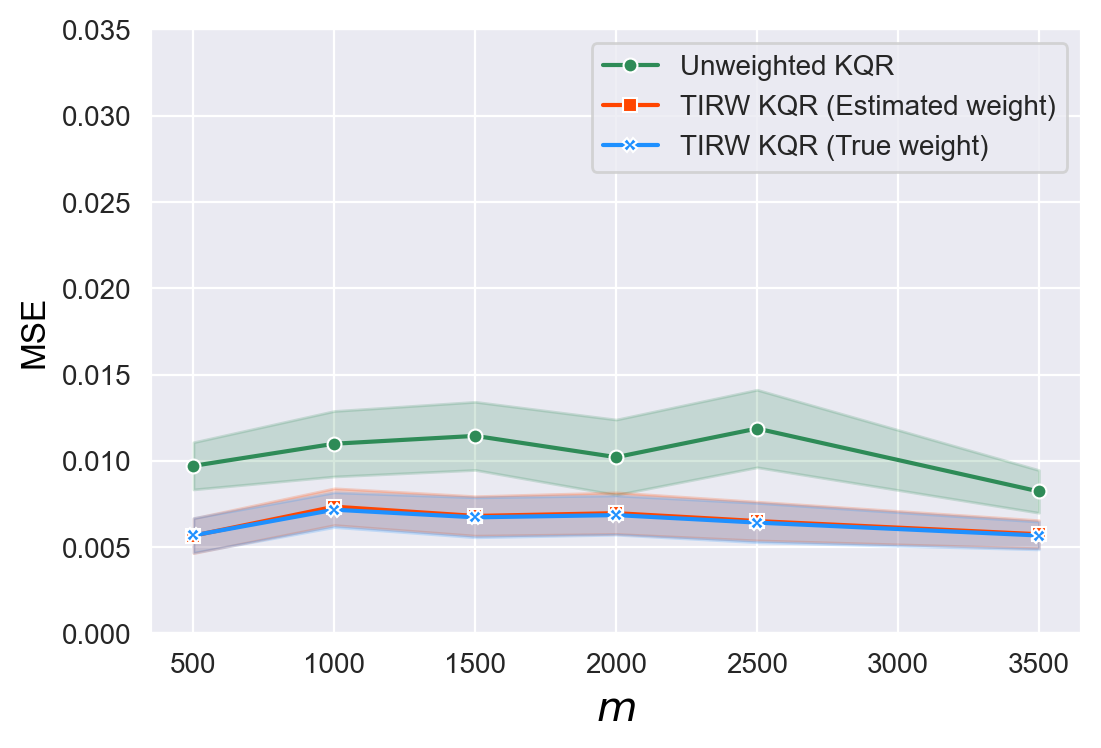}
\label{KQR_dimension1_uniformbounded_MSE_13}}
\setcounter{subfigure}{3}
\renewcommand{\thesubfigure}{(2\alph{subfigure})}
    \subfigure[$\tau=0.5$ and $r=1$]{
	\includegraphics[width=1.6in,  height=1.09in]{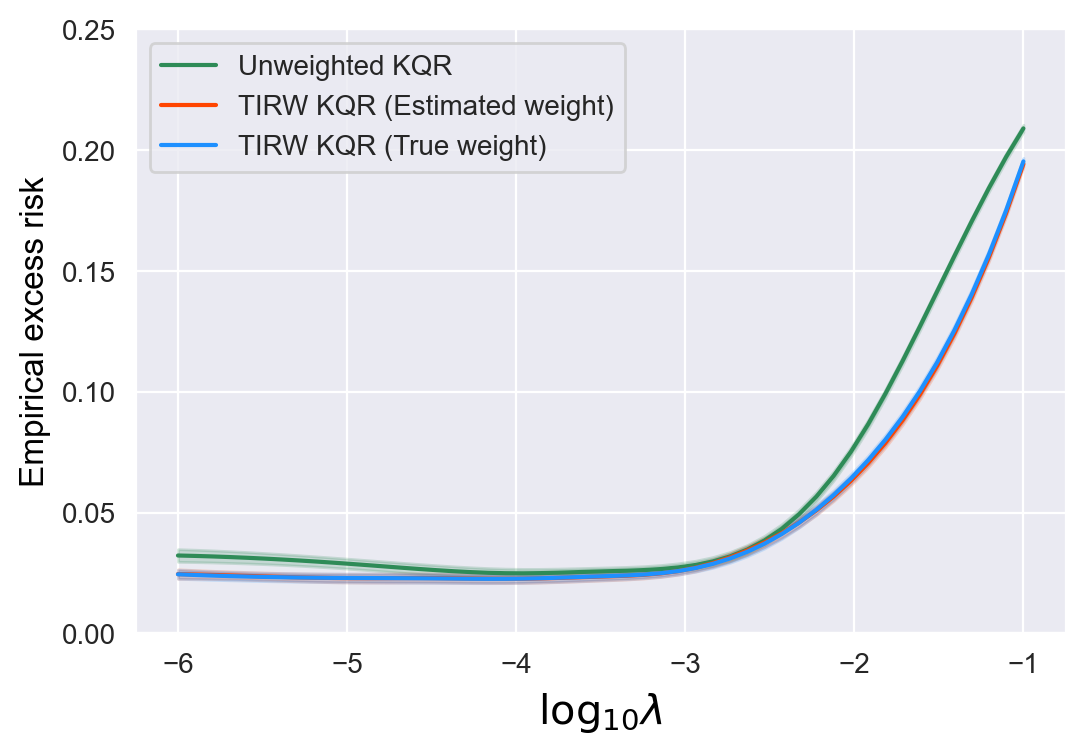}
\label{KQR_dimension1_uniformbounded_Empirical_smooth_1}}
 \subfigure[$\tau=0.5$ and $r=1$]{
    \includegraphics[width=1.6in,  height=1.09in]{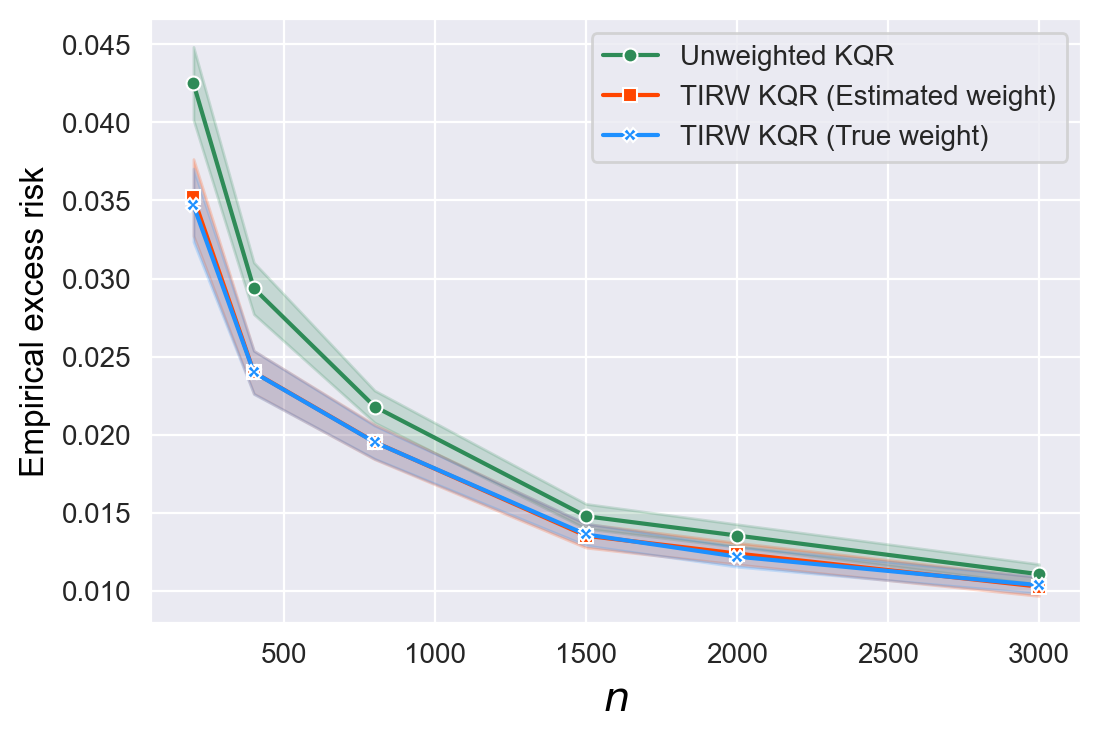}
\label{KQR_dimension1_uniformbounded_Empirical_7}}
    \subfigure[$\tau=0.5$ and $r=1$]{
	\includegraphics[width=1.6in,  height=1.09in]{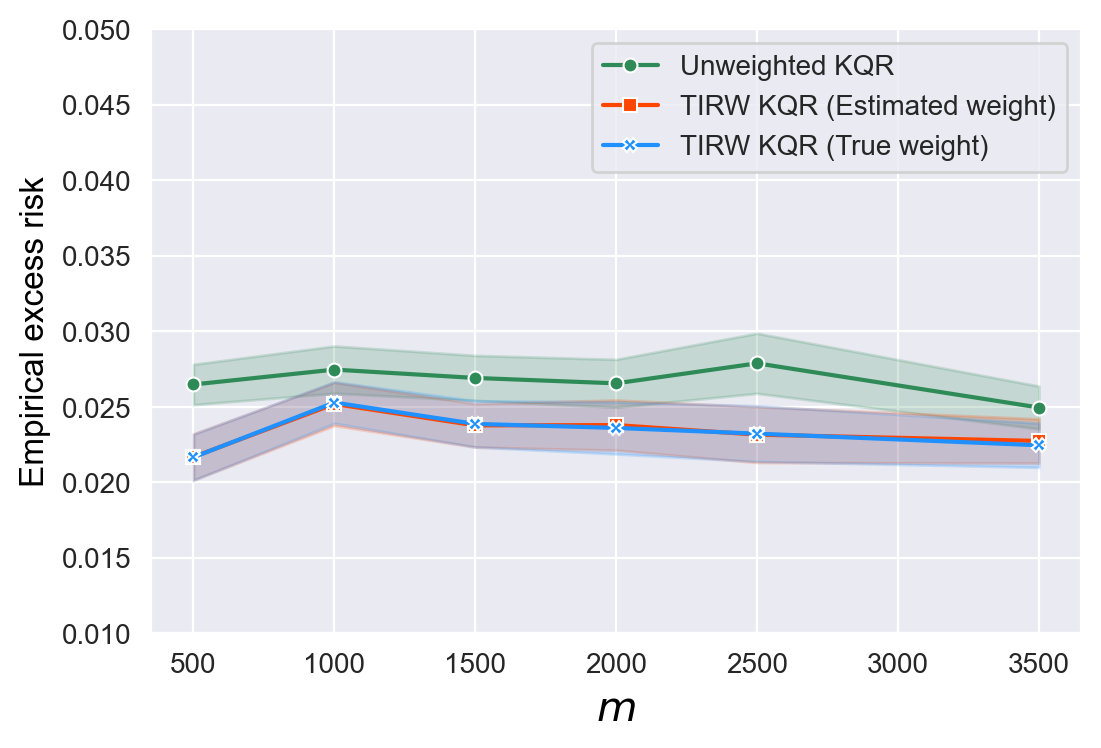}
\label{KQR_dimension1_uniformbounded_Empirical_13}}
\end{figure}

\begin{figure}[H]
\graphicspath{{bounded_KQR_1_d/}}
    \centering  
    
    \setcounter{subfigure}{0}
\renewcommand{\thesubfigure}{(3\alph{subfigure})}
\subfigure[$\tau=0.5$ and $r=0$]{
    \includegraphics[width=1.6in,  height=1.09in]{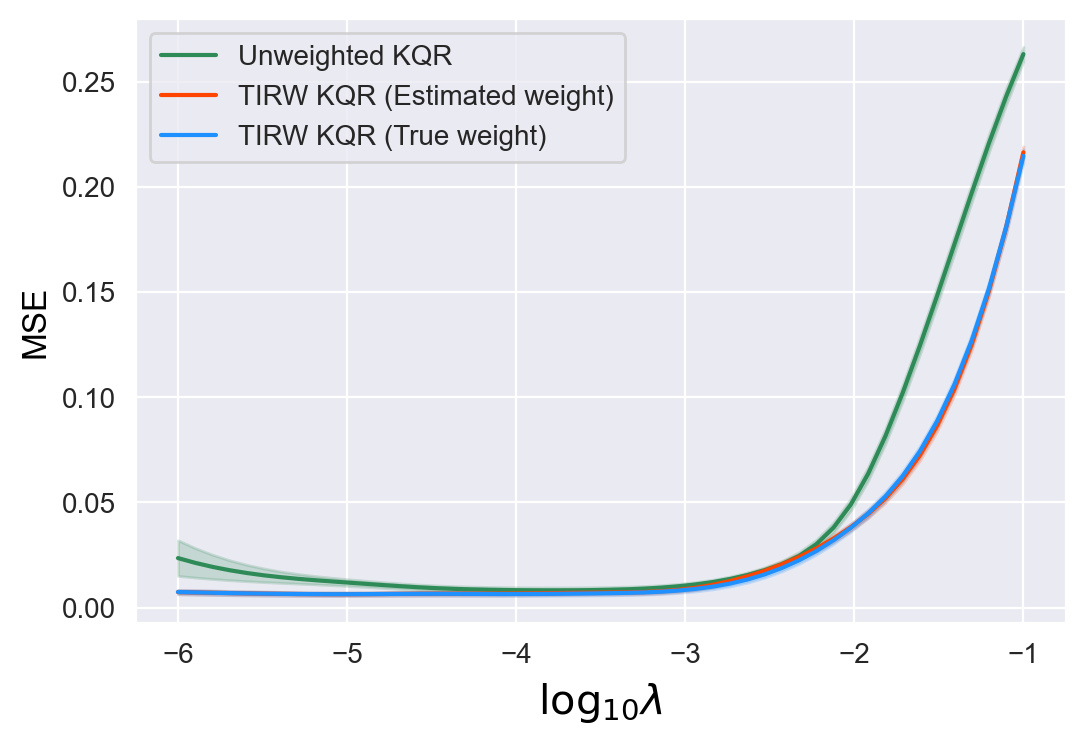}
    \label{KQR_dimension1_uniformbounded_MSE_smooth_2}}
    \subfigure[$\tau=0.5$ and $r=0$]{
	\includegraphics[width=1.6in,  height=1.09in]{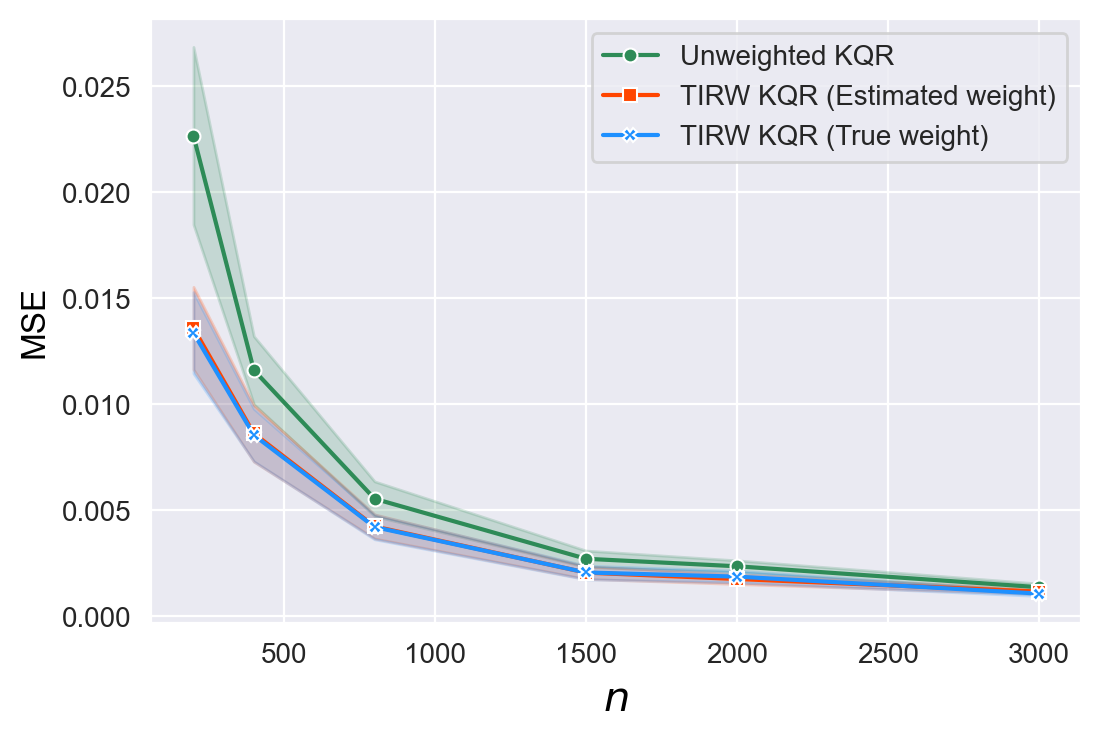}
\label{KQR_dimension1_uniformbounded_MSE_8}}
   \subfigure[$\tau=0.5$ and $r=0$]{
    \includegraphics[width=1.6in,  height=1.09in]{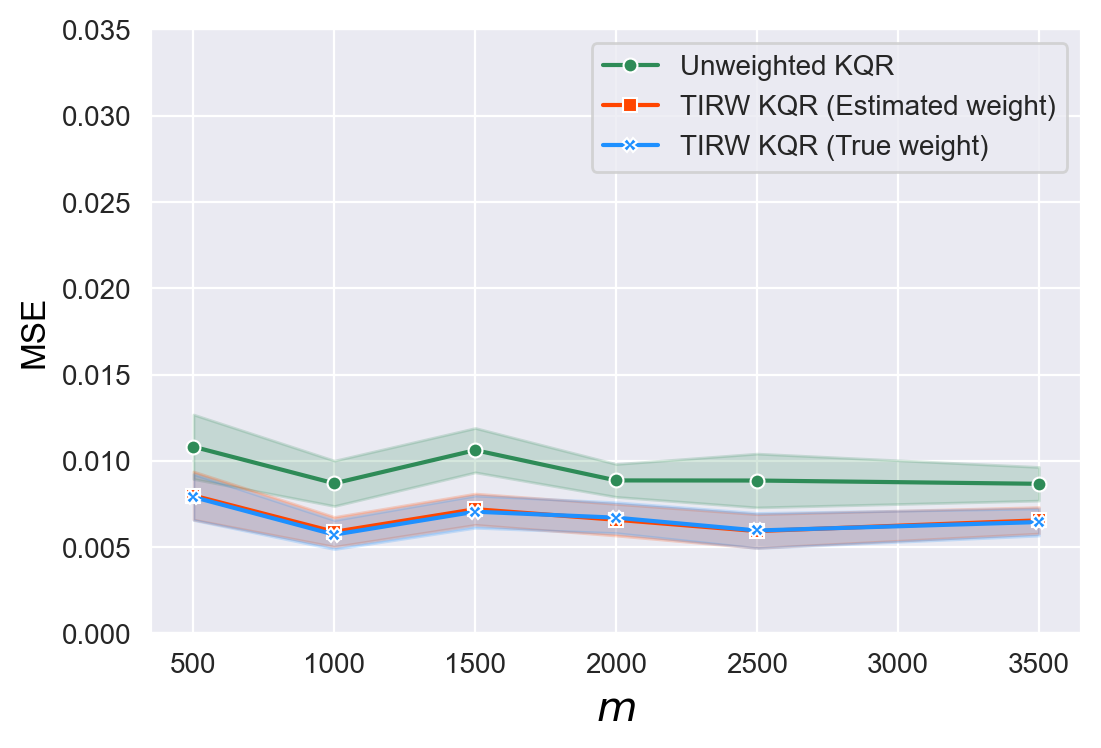}
    \label{KQR_dimension1_uniformbounded_MSE_14}}
    \subfigure[$\tau=0.5$ and $r=0$]{
	\includegraphics[width=1.6in,  height=1.09in]{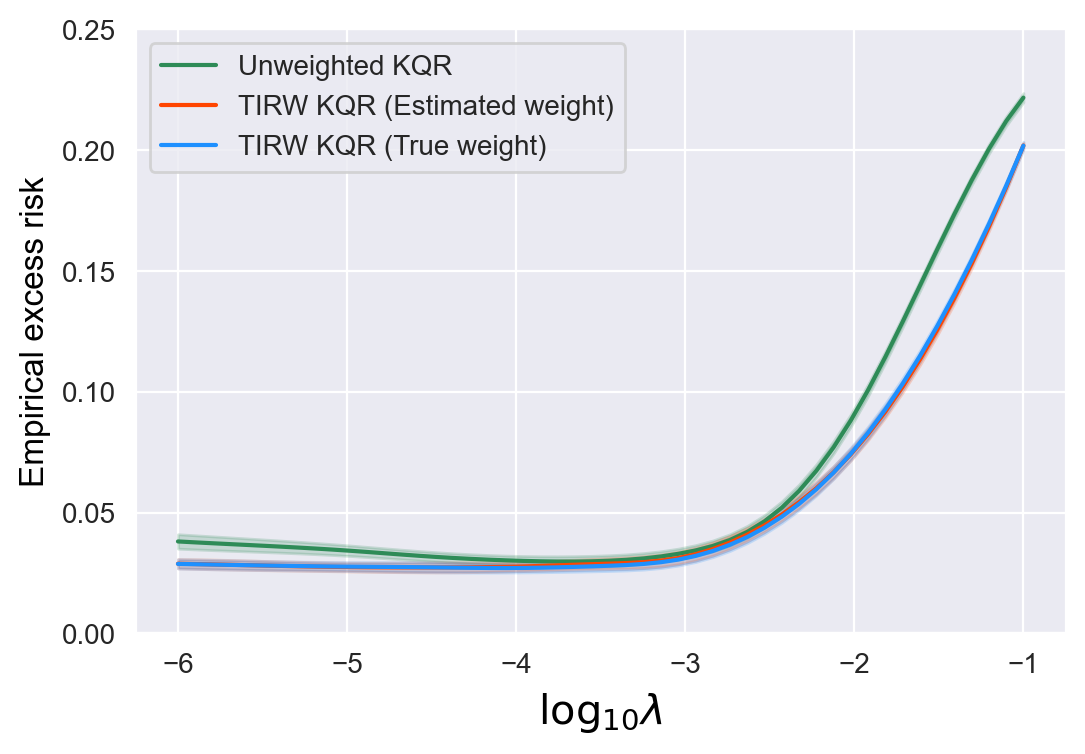}
\label{KQR_dimension1_uniformbounded_Empirical_smooth_2}}
 \subfigure[$\tau=0.5$ and $r=0$]{
    \includegraphics[width=1.6in,  height=1.09in]{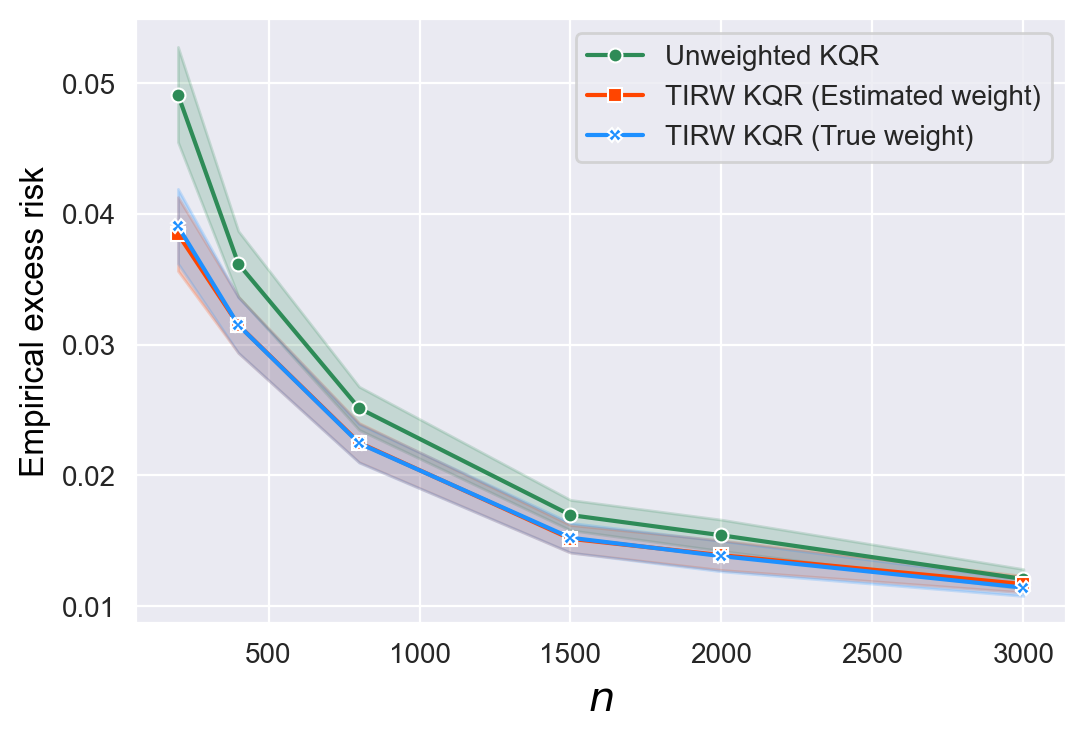}
\label{KQR_dimension1_uniformbounded_Empirical_8}}
    \subfigure[$\tau=0.5$ and $r=0$]{
	\includegraphics[width=1.6in,  height=1.09in]{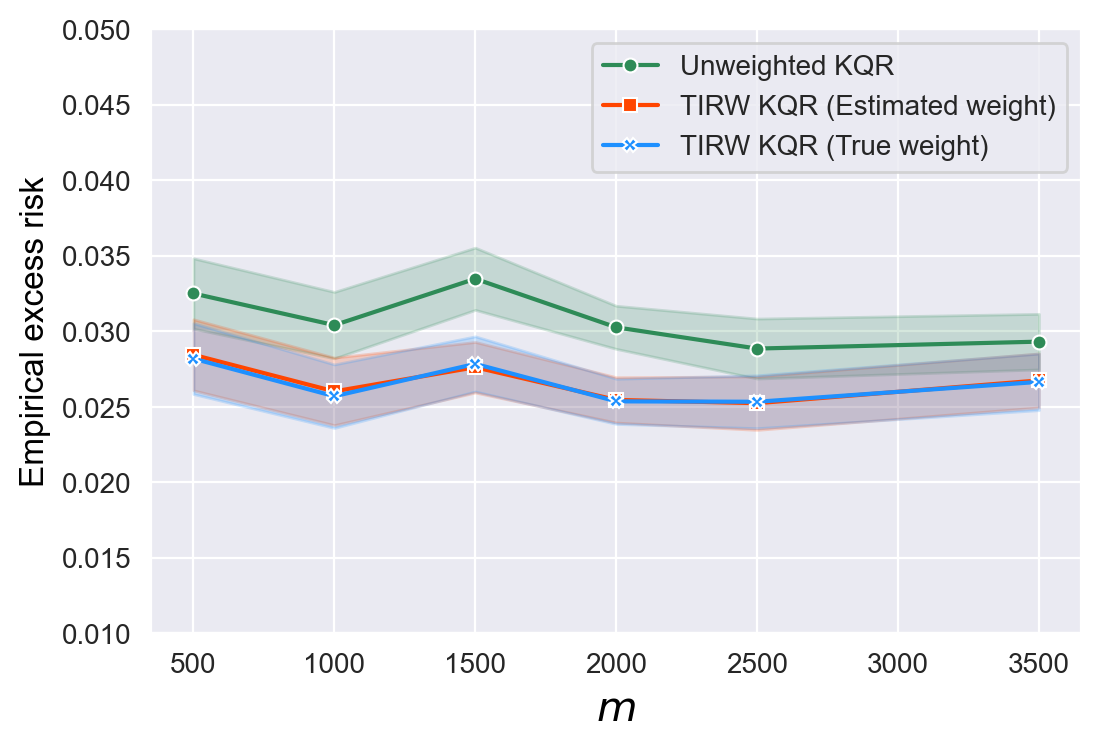}
 \label{KQR_dimension1_uniformbounded_Empirical_14}}
 \setcounter{subfigure}{0}
\renewcommand{\thesubfigure}{(4\alph{subfigure})}
  \subfigure[$\tau=0.7$ and $r=1$]{
    \includegraphics[width=1.6in,  height=1.09in]{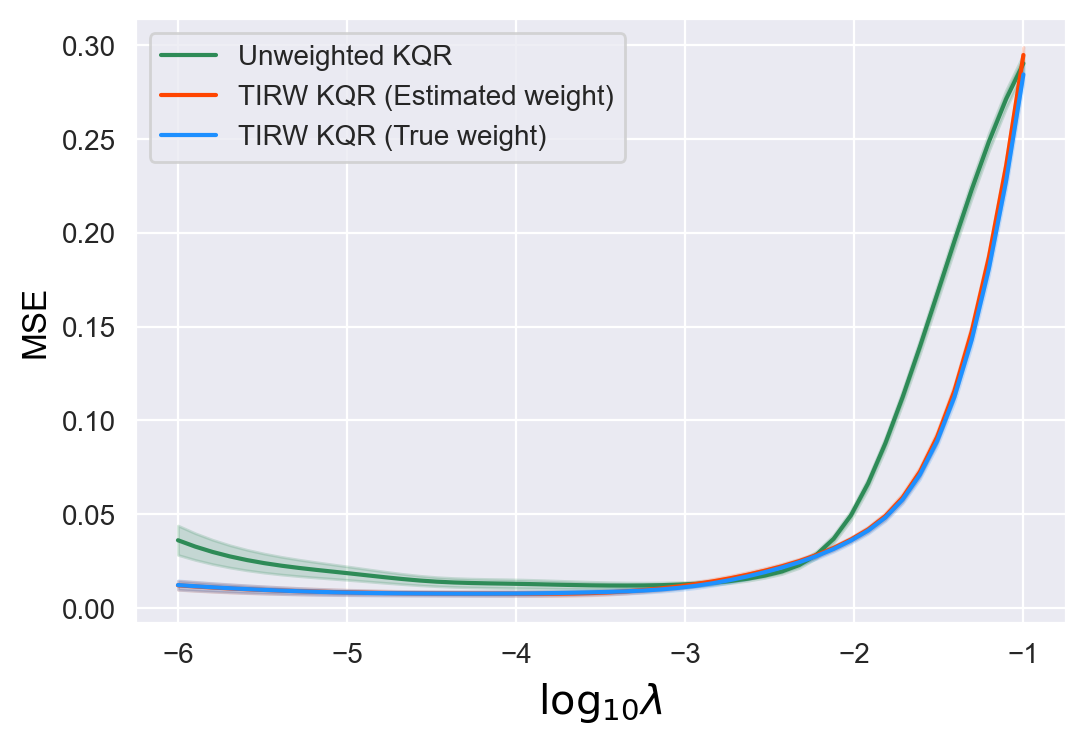}
    \label{KQR_dimension1_uniformbounded_MSE_smooth_5}}
    \subfigure[$\tau=0.7$ and $r=1$]{
	\includegraphics[width=1.6in,  height=1.09in]{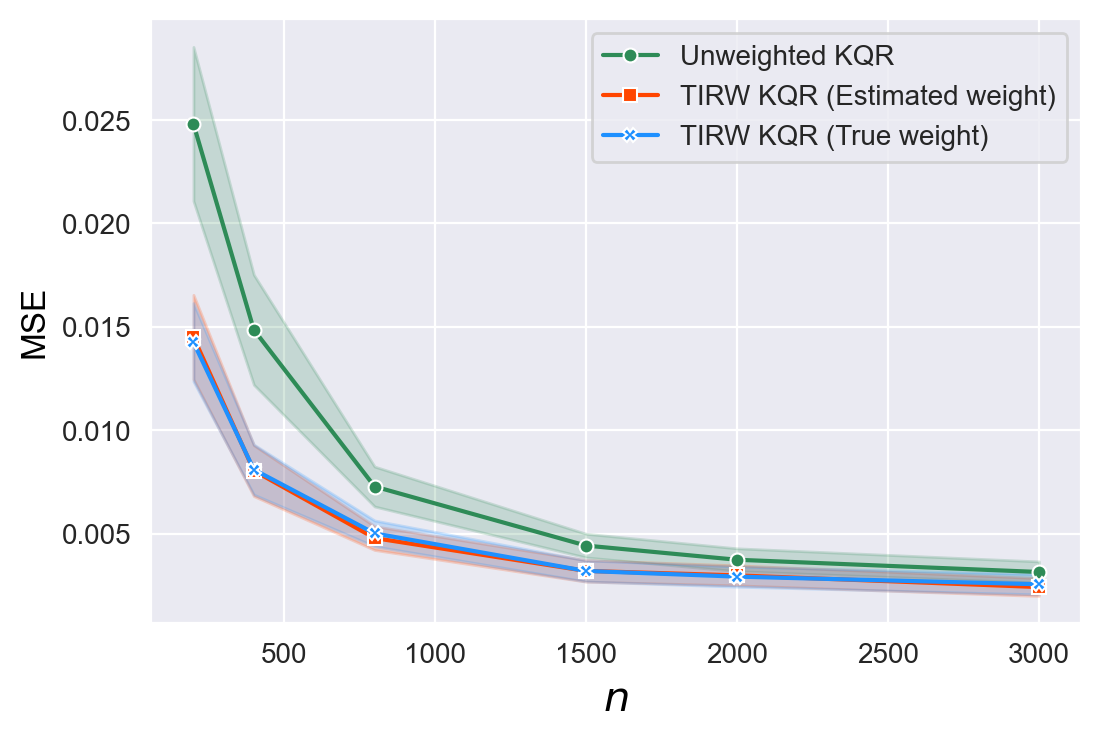}
\label{KQR_dimension1_uniformbounded_MSE_11}}
   \subfigure[$\tau=0.7$ and $r=1$]{
    \includegraphics[width=1.6in,  height=1.09in]{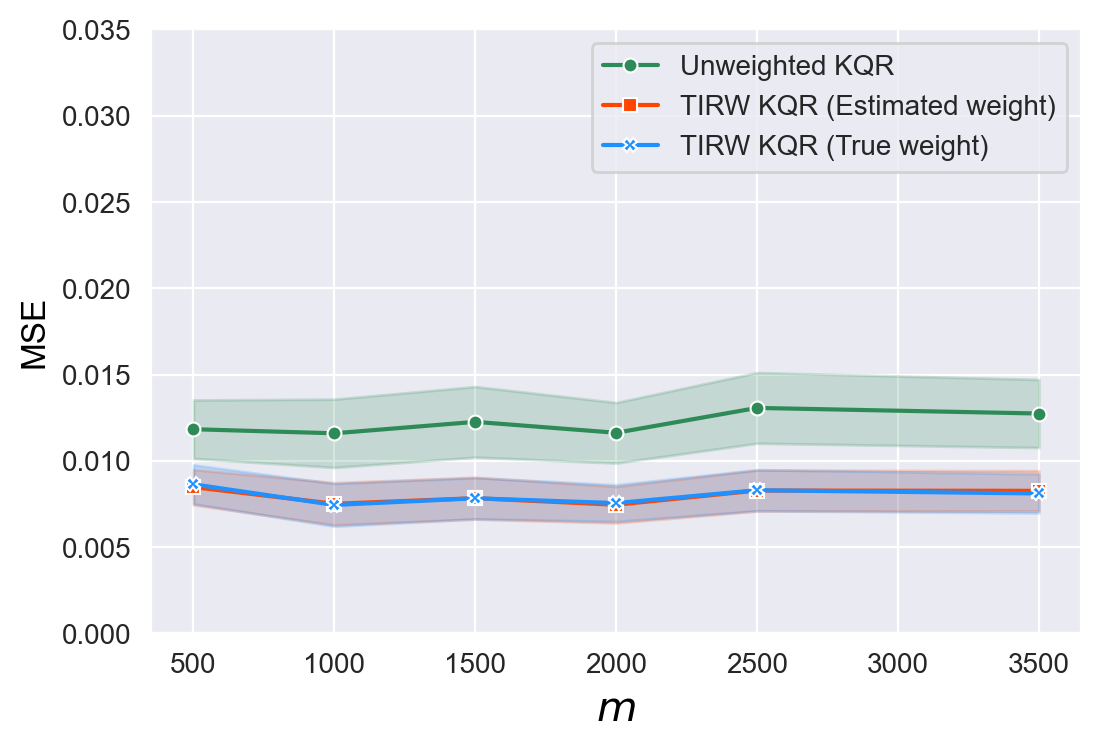}
    \label{KQR_dimension1_uniformbounded_MSE_17}}
    \subfigure[$\tau=0.7$ and $r=1$]{
	\includegraphics[width=1.6in,  height=1.09in]{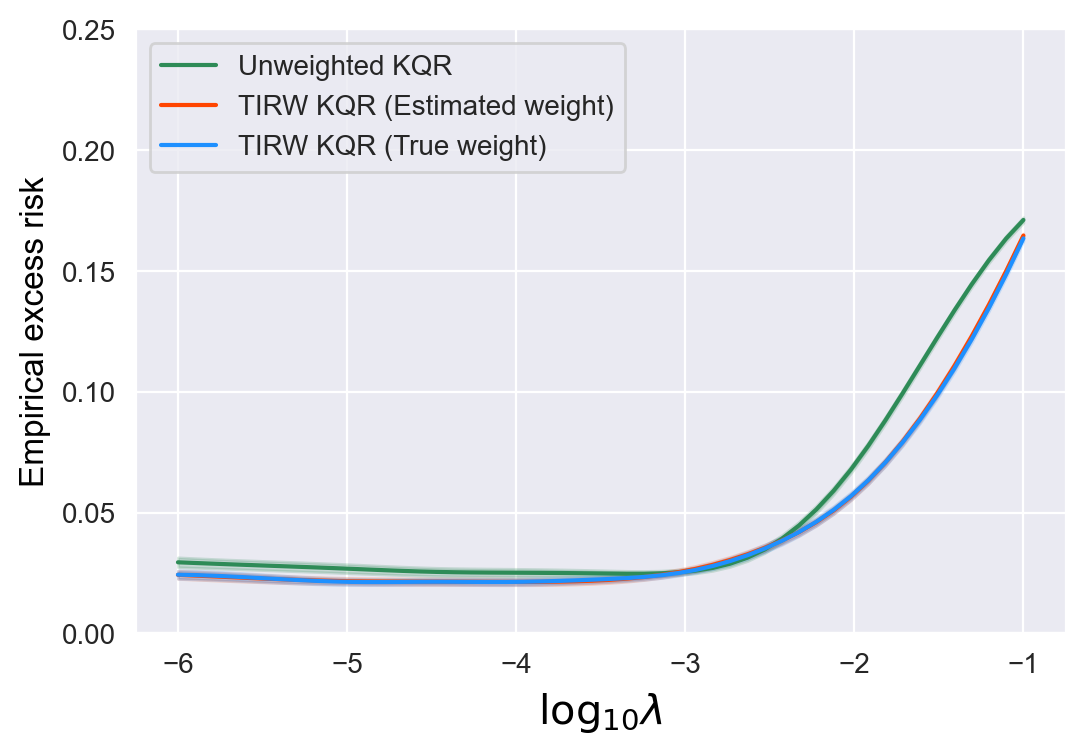}
\label{KQR_dimension1_uniformbounded_Empirical_smooth_5}}
 \subfigure[$\tau=0.7$ and $r=1$]{
    \includegraphics[width=1.6in,  height=1.09in]{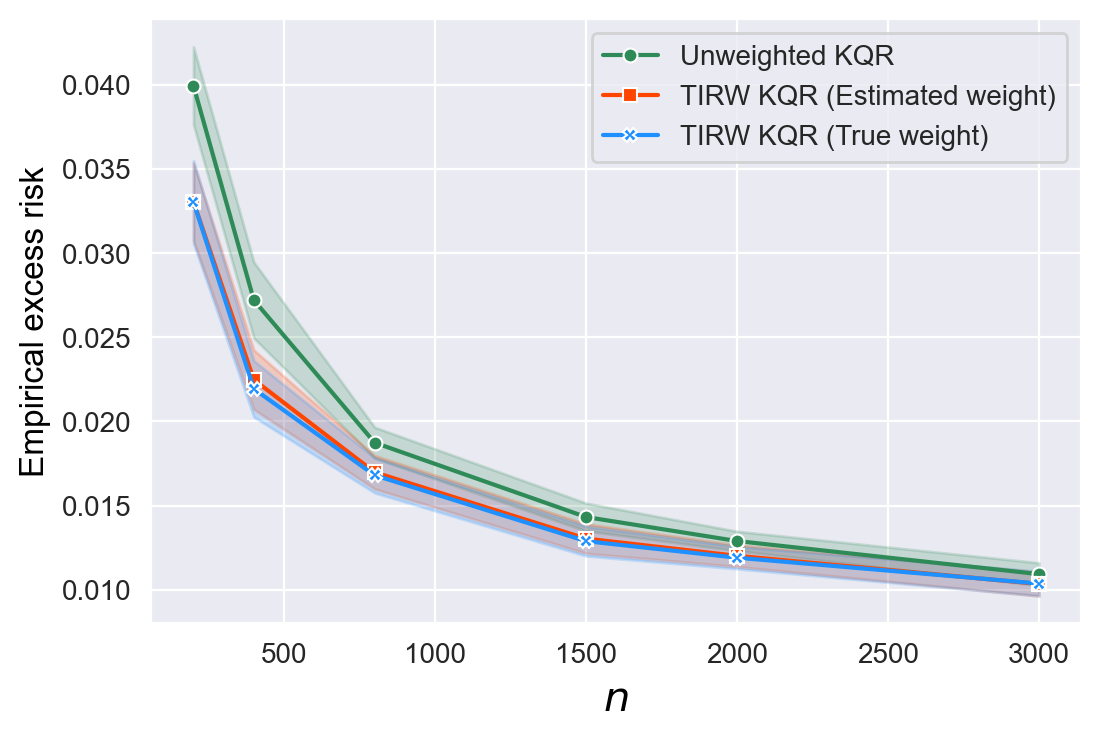}
\label{KQR_dimension1_uniformbounded_Empirical_11}}
    \subfigure[$\tau=0.7$ and $r=1$]{
	\includegraphics[width=1.6in,  height=1.09in]{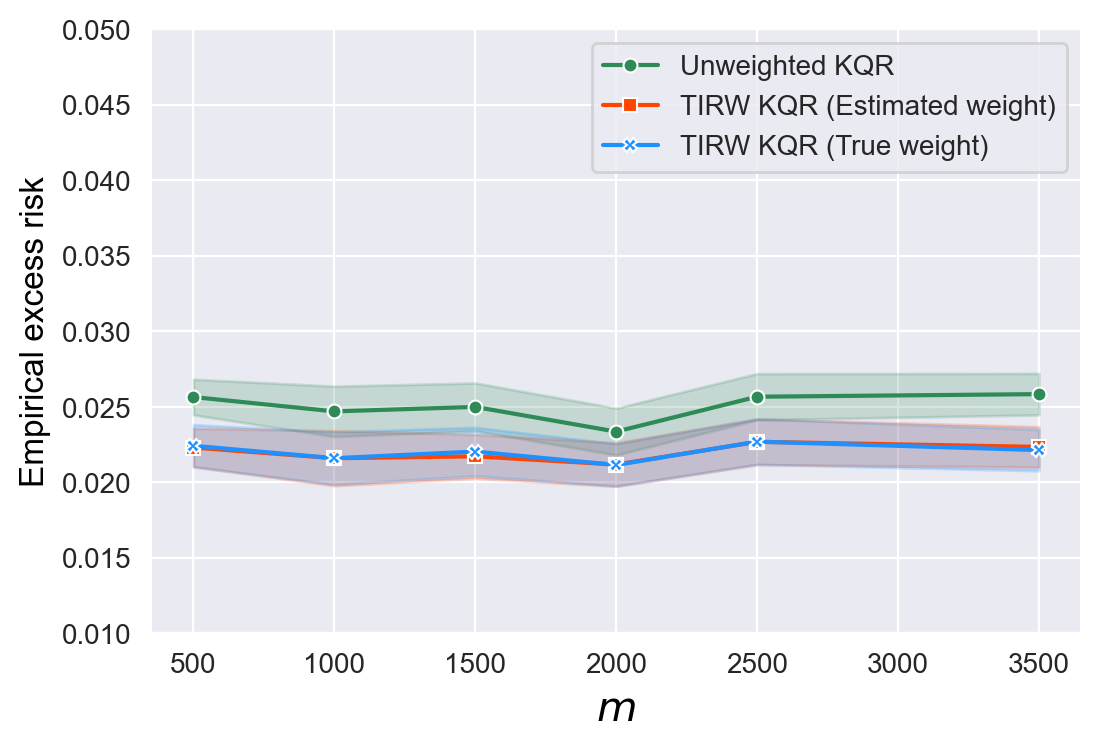}
 \label{KQR_dimension1_uniformbounded_Empirical_17}}
      \setcounter{subfigure}{0}
\renewcommand{\thesubfigure}{(5\alph{subfigure})}
 \subfigure[$\tau=0.7$ and $r=0$]{
    \includegraphics[width=1.6in,  height=1.09in]{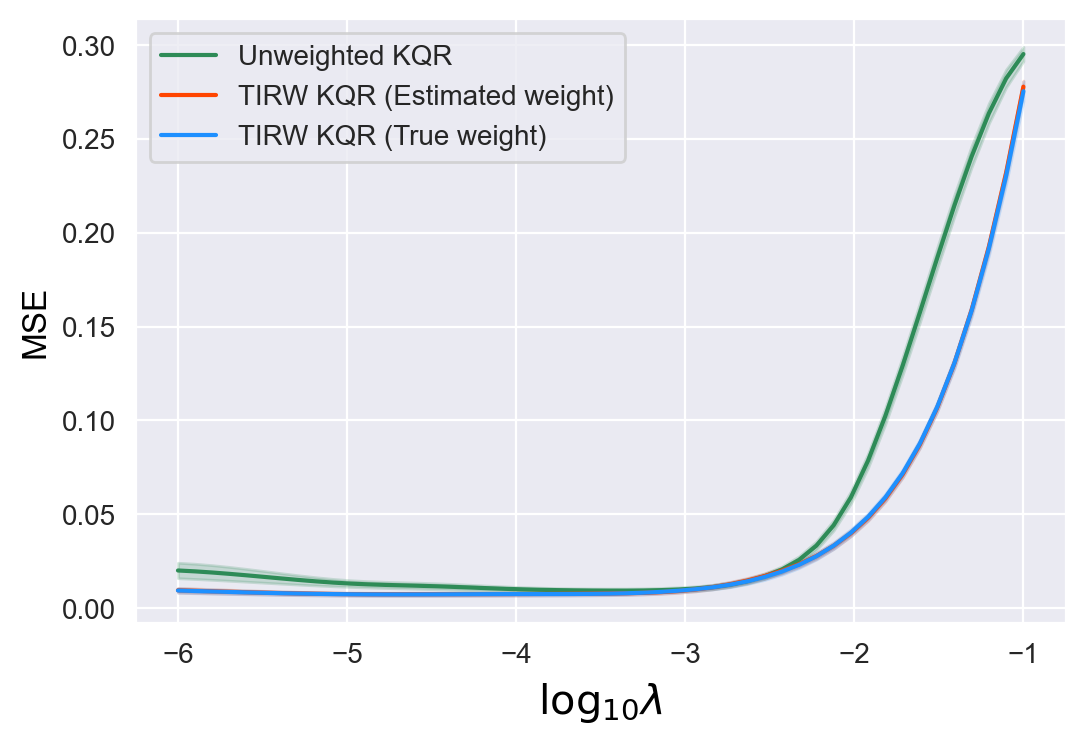}
    \label{KQR_dimension1_uniformbounded_MSE_smooth_6}}
    \subfigure[$\tau=0.7$ and $r=0$]{
	\includegraphics[width=1.6in,  height=1.09in]{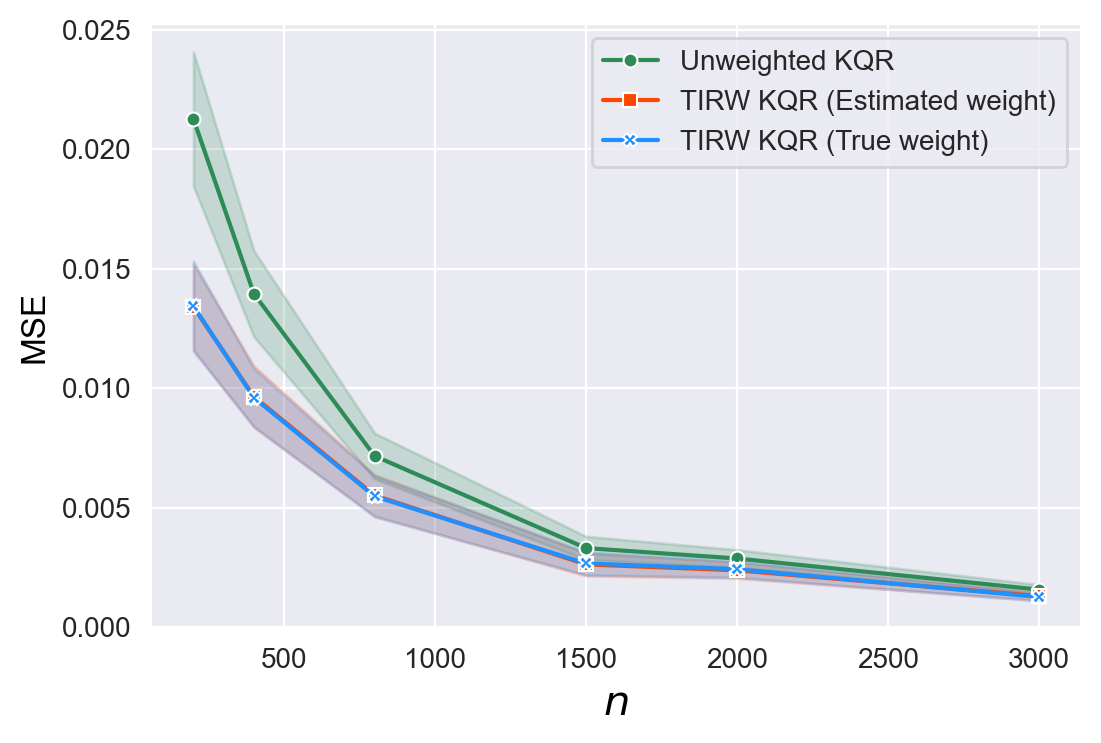}
\label{KQR_dimension1_uniformbounded_MSE_12}}
   \subfigure[$\tau=0.7$ and $r=0$]{
    \includegraphics[width=1.6in,  height=1.09in]{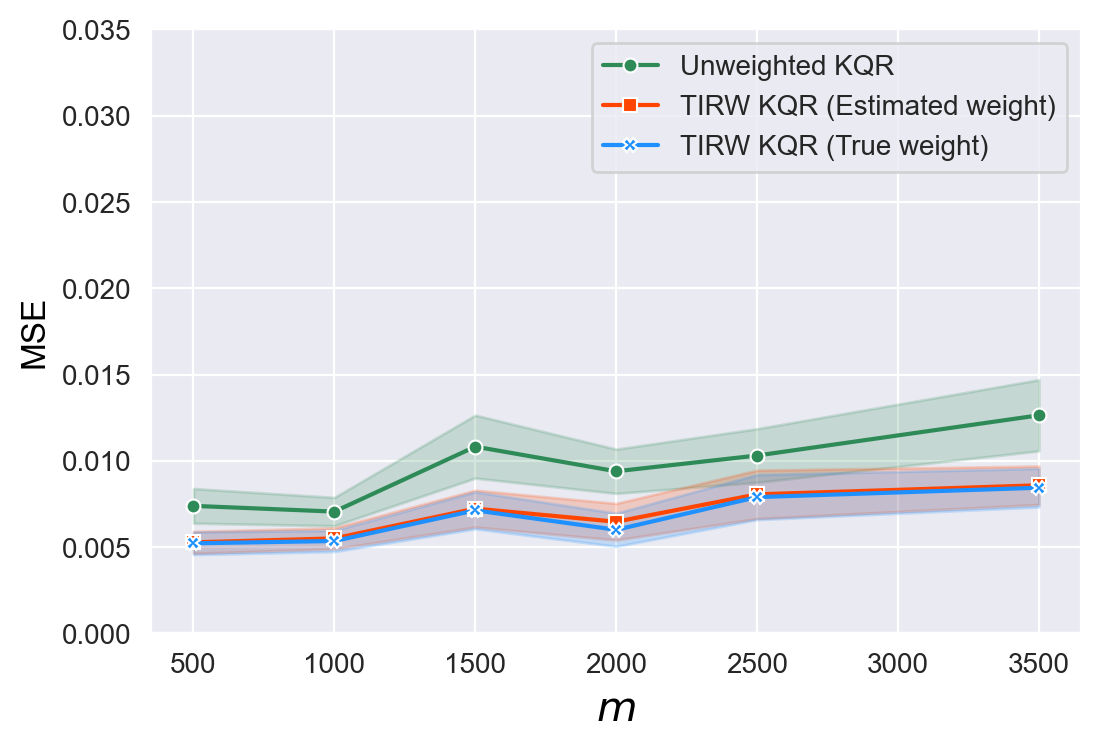}
    \label{KQR_dimension1_uniformbounded_MSE_18}}
     \setcounter{subfigure}{3}
\renewcommand{\thesubfigure}{(5\alph{subfigure})}
    \centering
    \subfigure[$\tau=0.7$ and $r=0$]{
	\includegraphics[width=1.6in,  height=1.09in]{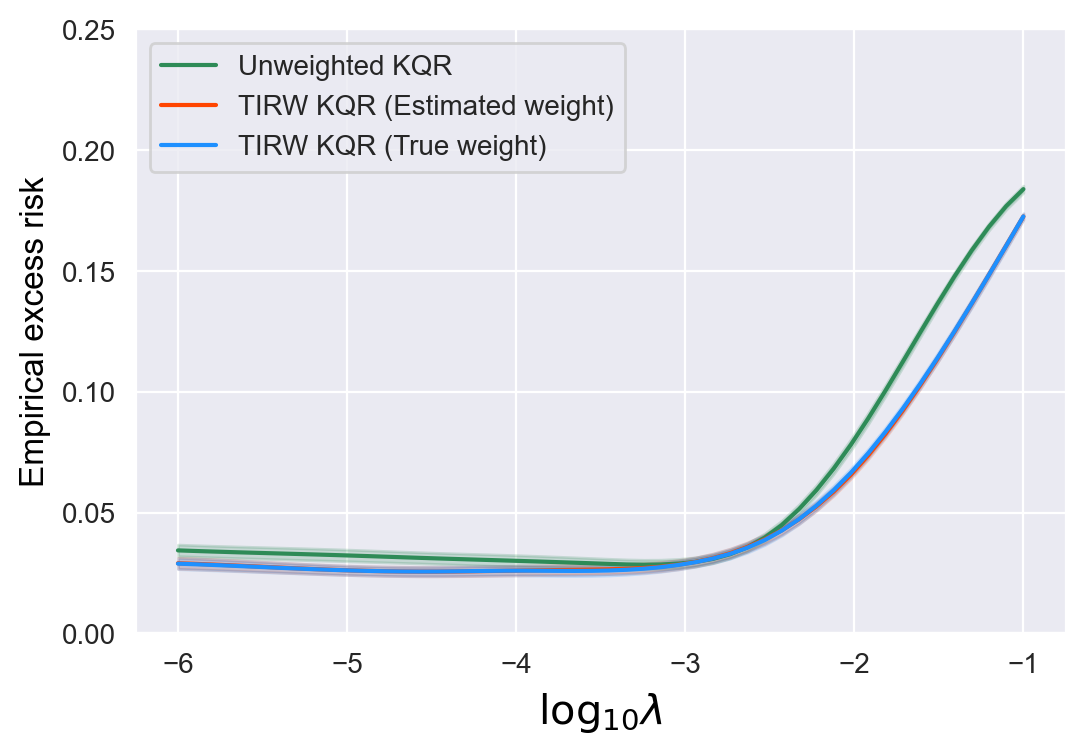}
\label{KQR_dimension1_uniformbounded_Empirical_smooth_6}}
    \subfigure[$\tau=0.7$ and $r=0$]{
    \includegraphics[width=1.6in,  height=1.09in]{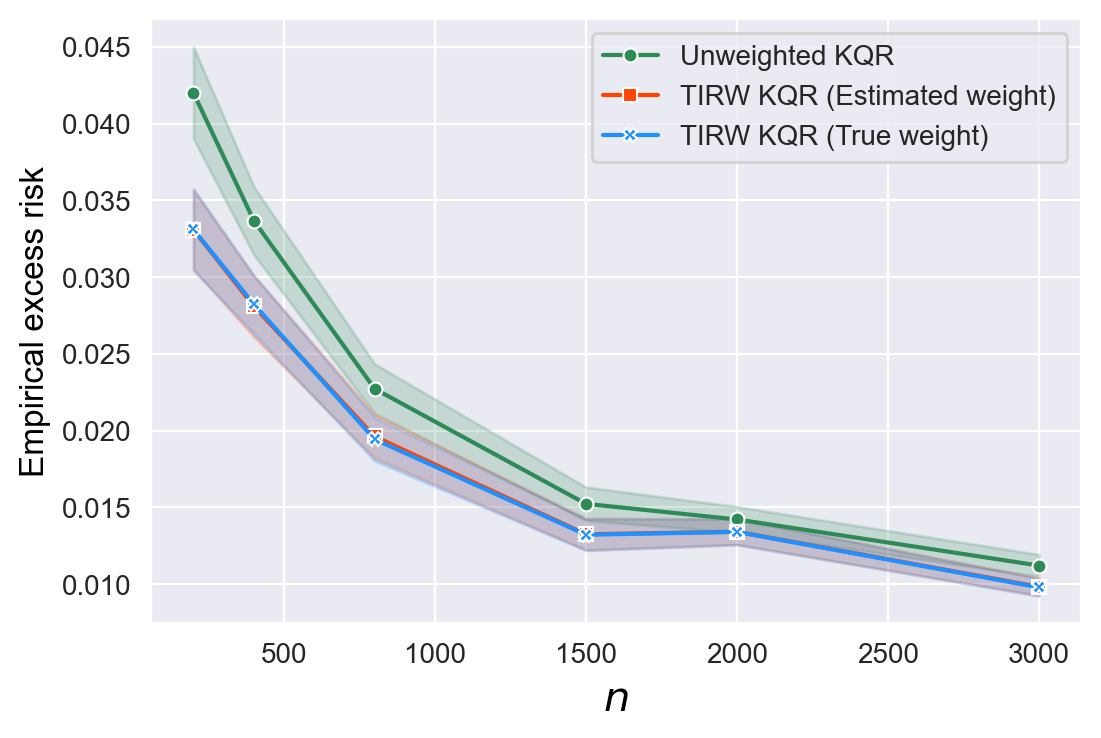}
   \label{KQR_dimension1_uniformbounded_Empirical_12}}
    \subfigure[$\tau=0.7$ and $r=0$]{
	\includegraphics[width=1.6in,  height=1.09in]{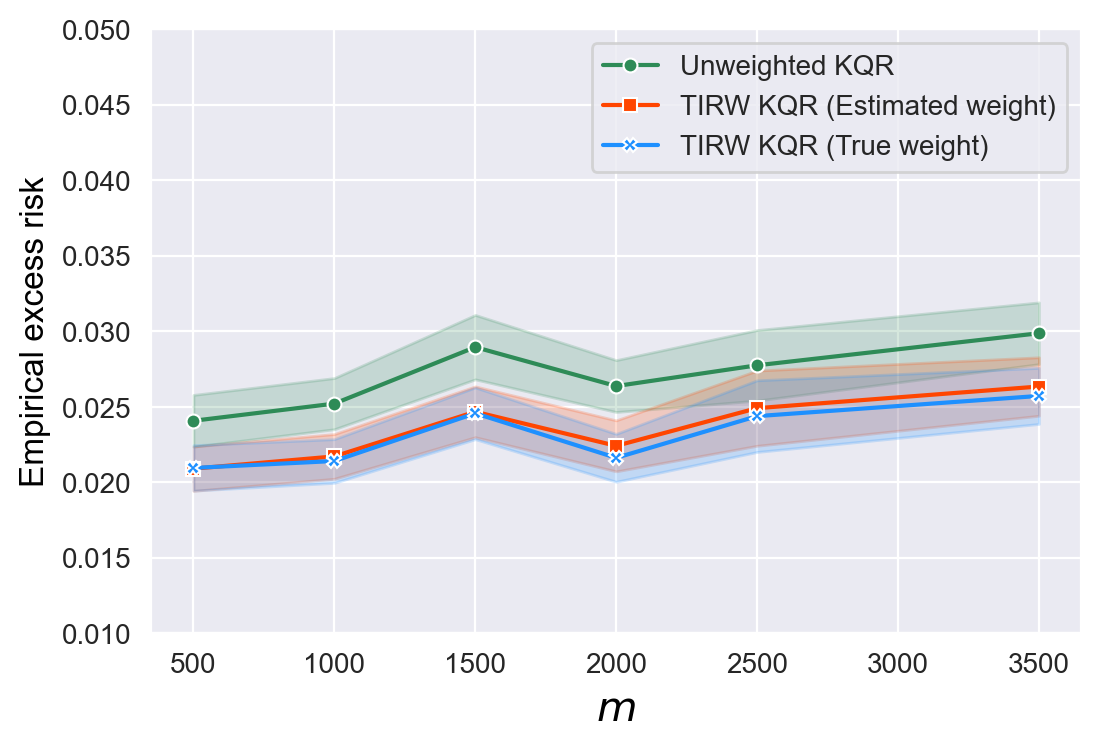}
 \label{KQR_dimension1_uniformbounded_Empirical_18}}
 \caption{\footnotesize{Average MSE  and empirical excess risk for unweighted KQR, TIRW KQR with true weight and estimated weight, respectively (in the left panel, the curves are plotted with respect to $\log_{10} \lambda$ with $n=500, m=1000$; in the middle panel, the curves are plotted with respect to $n$ with fixed $m = 1000,\lambda=10^{-4}$; in the right panel, the curves are plotted with respect to $m$ with fixed $n=500,\lambda=10^{-4}$)}}
\label{KQR_dimension1_bounded}
\end{figure}

\subsubsection{Moment bounded case in Example S3}\label{sec:a.2.2}
\begin{figure}[H]
\graphicspath{{unbounded_KQR_1_d/}}
    \centering
\setcounter{subfigure}{0}
\renewcommand{\thesubfigure}{(1\alph{subfigure})}
    \subfigure[$\tau=0.3$ and $r=0$]{
    \includegraphics[width=1.6in,  height=1.09in]{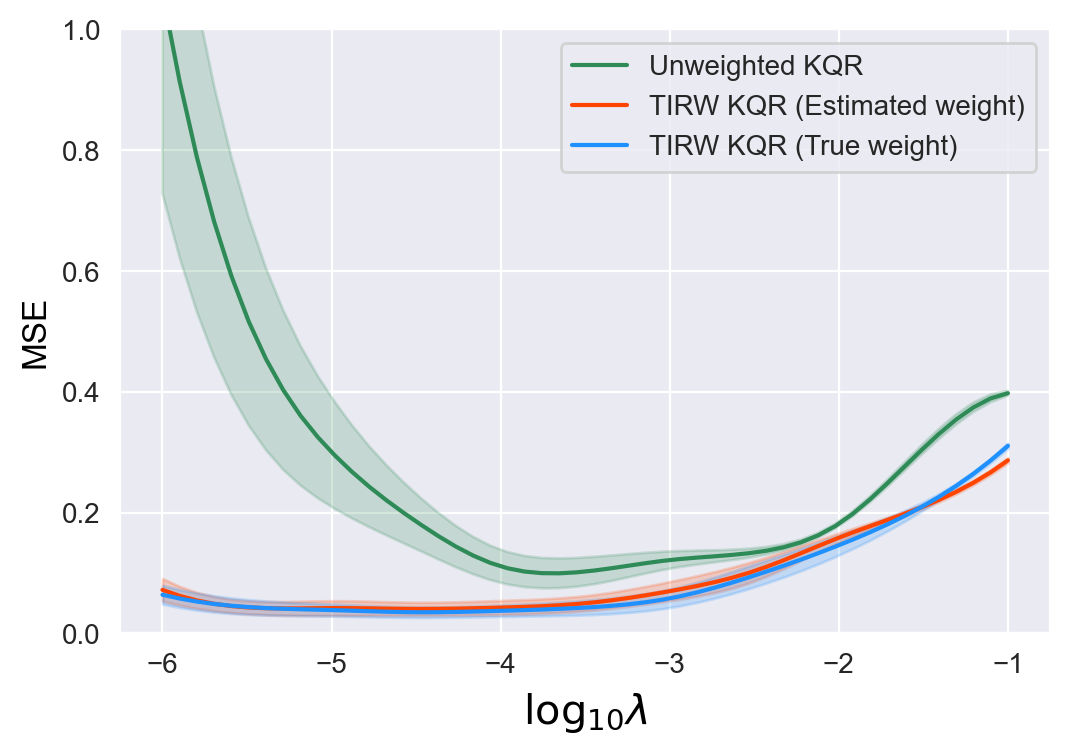}
    \label{KQR_dimension1_MSE_smooth_4}}
    \subfigure[$\tau=0.3$ and $r=0$]{
	\includegraphics[width=1.6in,  height=1.09in]{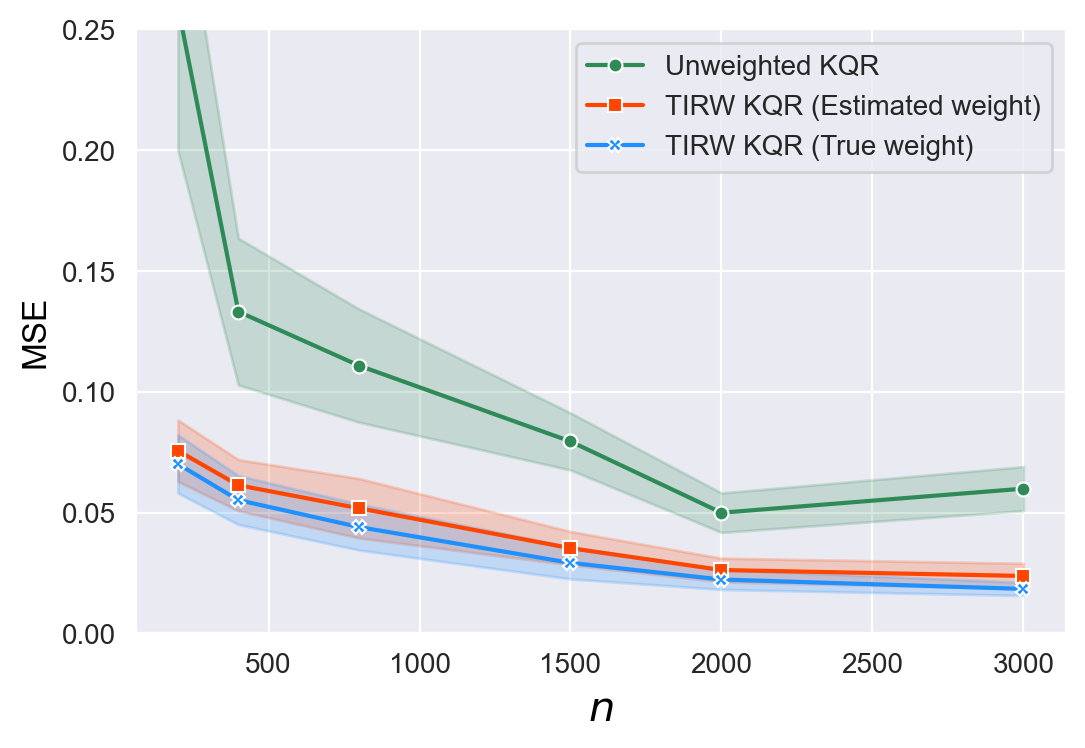}
\label{KQR_dimension1_MSE_10}}
   \subfigure[$\tau=0.3$ and $r=0$]{
    \includegraphics[width=1.6in,  height=1.09in]{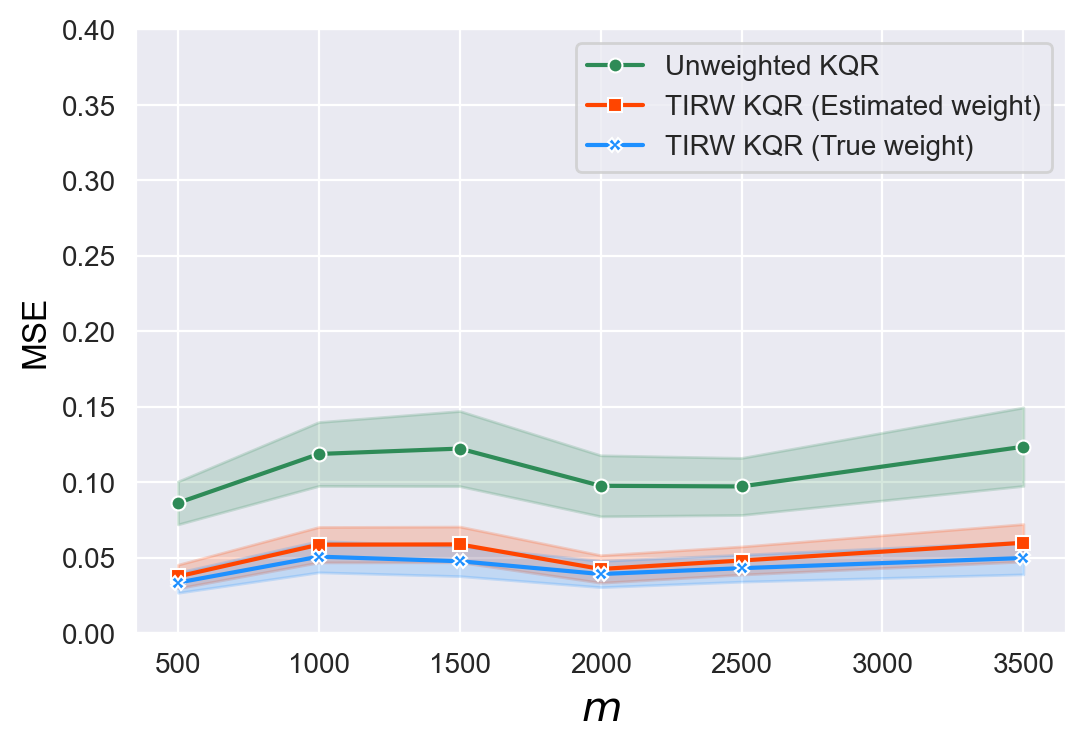}
    \label{KQR_dimension1_MSE_16}}
    \subfigure[$\tau=0.3$ and $r=0$]{
	\includegraphics[width=1.6in,  height=1.09in]{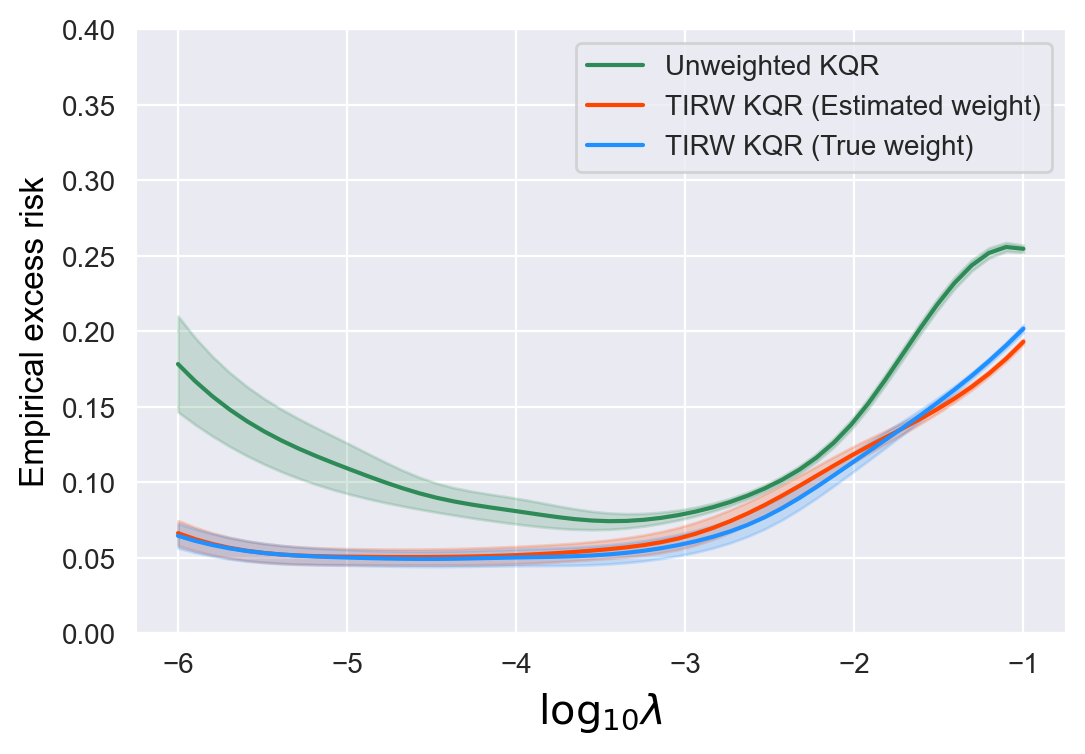}
\label{KQR_dimension1_Empirical_smooth_4}}
 \subfigure[$\tau=0.3$ and $r=0$]{
    \includegraphics[width=1.6in,  height=1.09in]{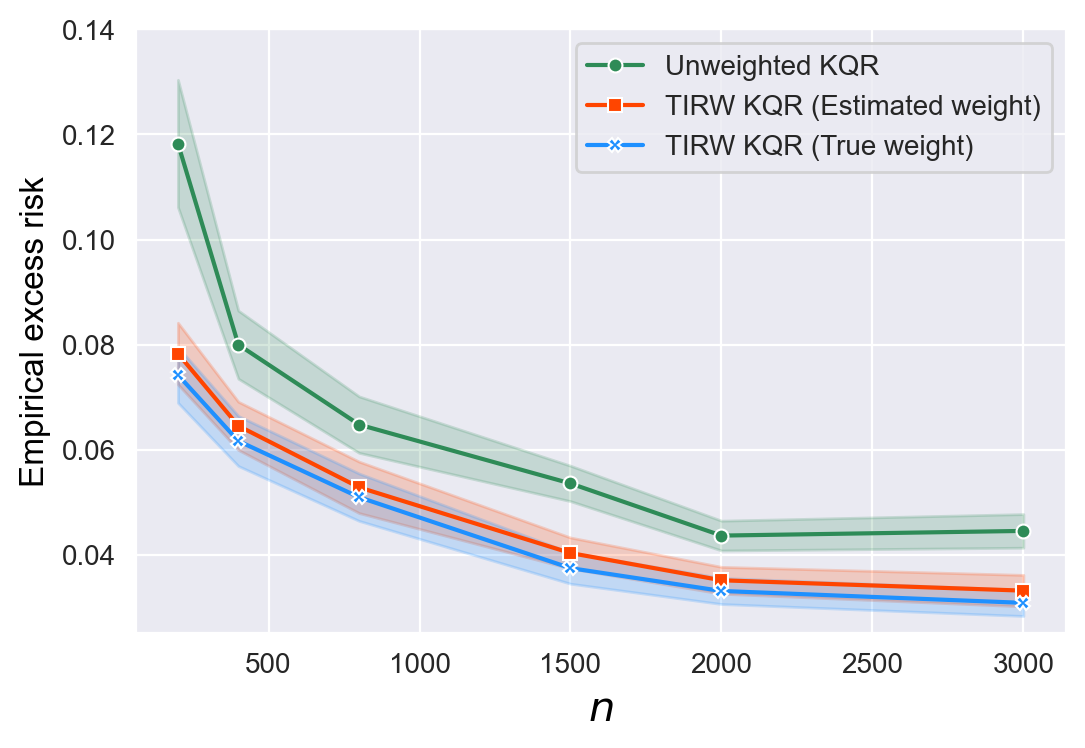}
\label{KQR_dimension1_Empirical_10}}
    \subfigure[$\tau=0.3$ and $r=0$]{
	\includegraphics[width=1.6in,  height=1.09in]{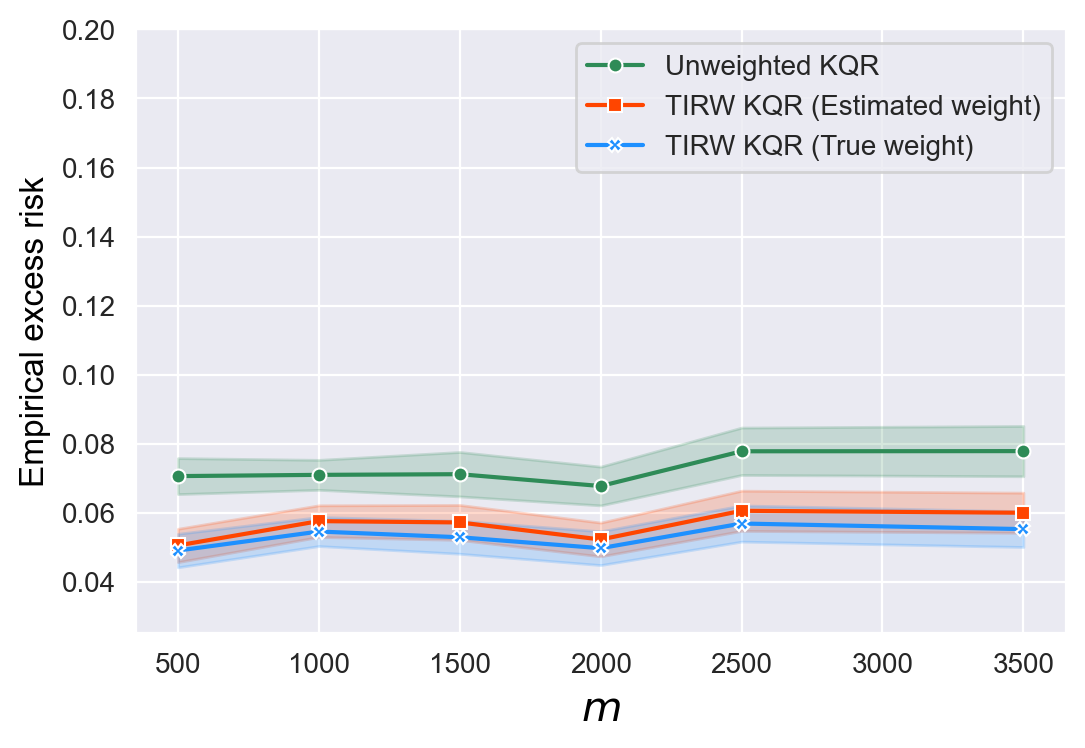}
 \label{KQR_dimension1_Empirical_16}}
 \setcounter{subfigure}{0}
\renewcommand{\thesubfigure}{(2\alph{subfigure})}
 \subfigure[$\tau=0.5$ and $r=1$]{
    \includegraphics[width=1.6in,  height=1.09in]{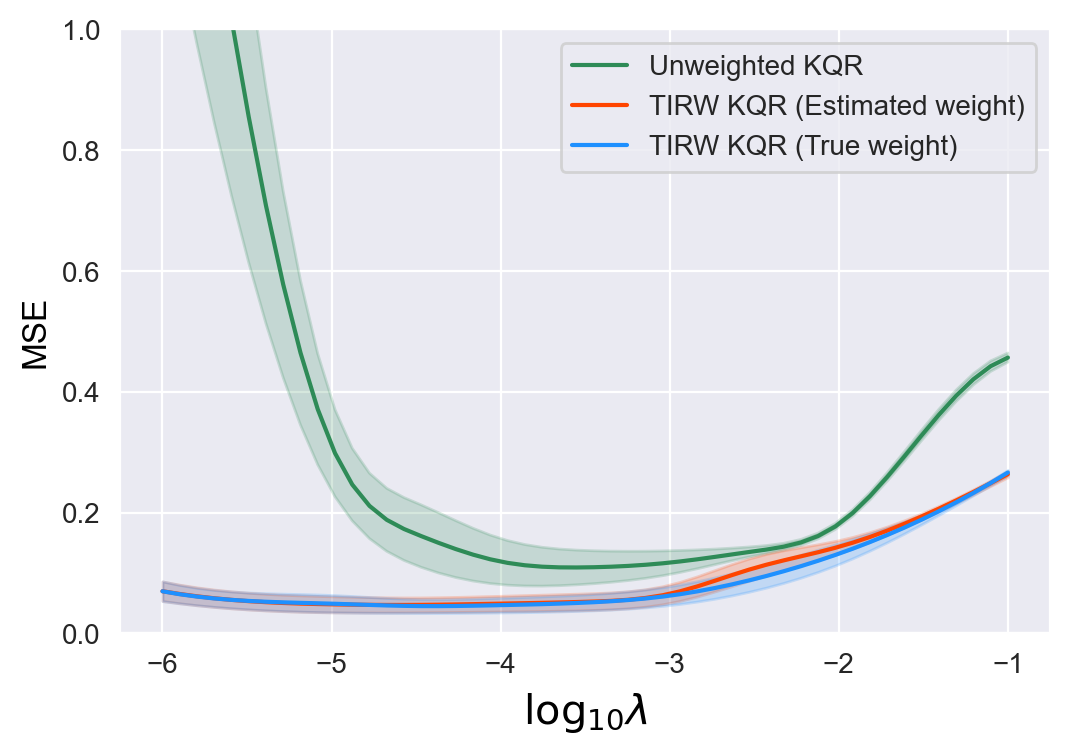}
    \label{KQR_dimension1_MSE_smooth_1}}
    \subfigure[$\tau=0.5$ and $r=1$]{
	\includegraphics[width=1.6in,  height=1.09in]{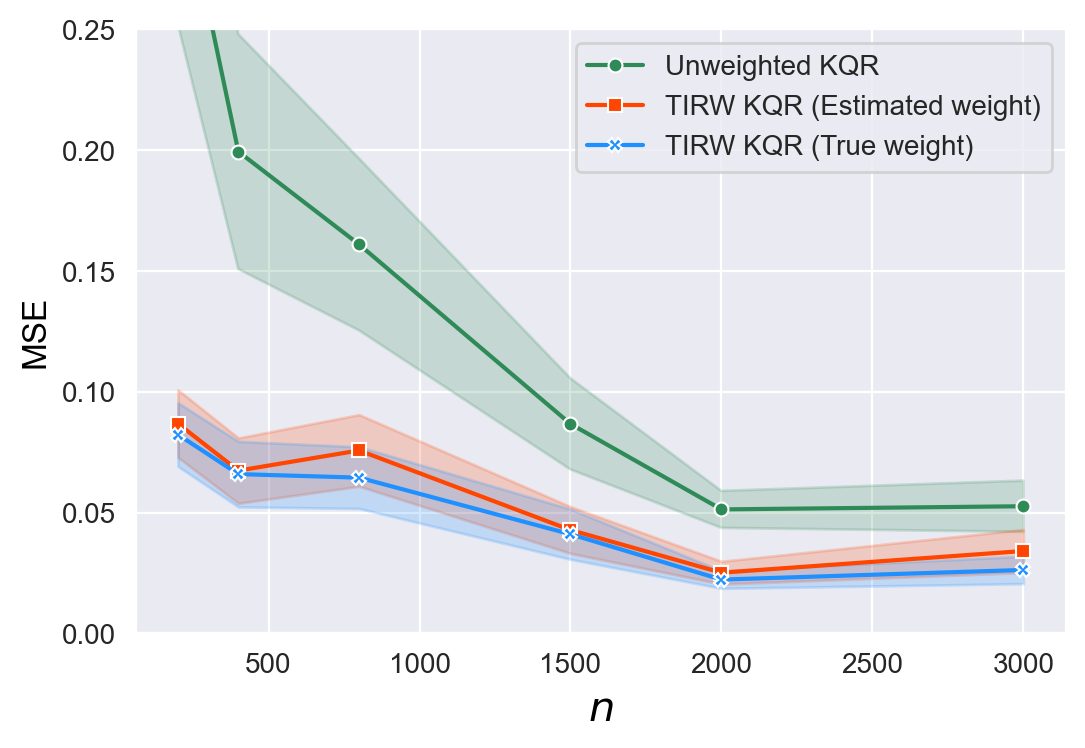}
\label{KQR_dimension1_MSE_7}}
   \subfigure[$\tau=0.5$ and $r=1$]{
    \includegraphics[width=1.6in,  height=1.09in]{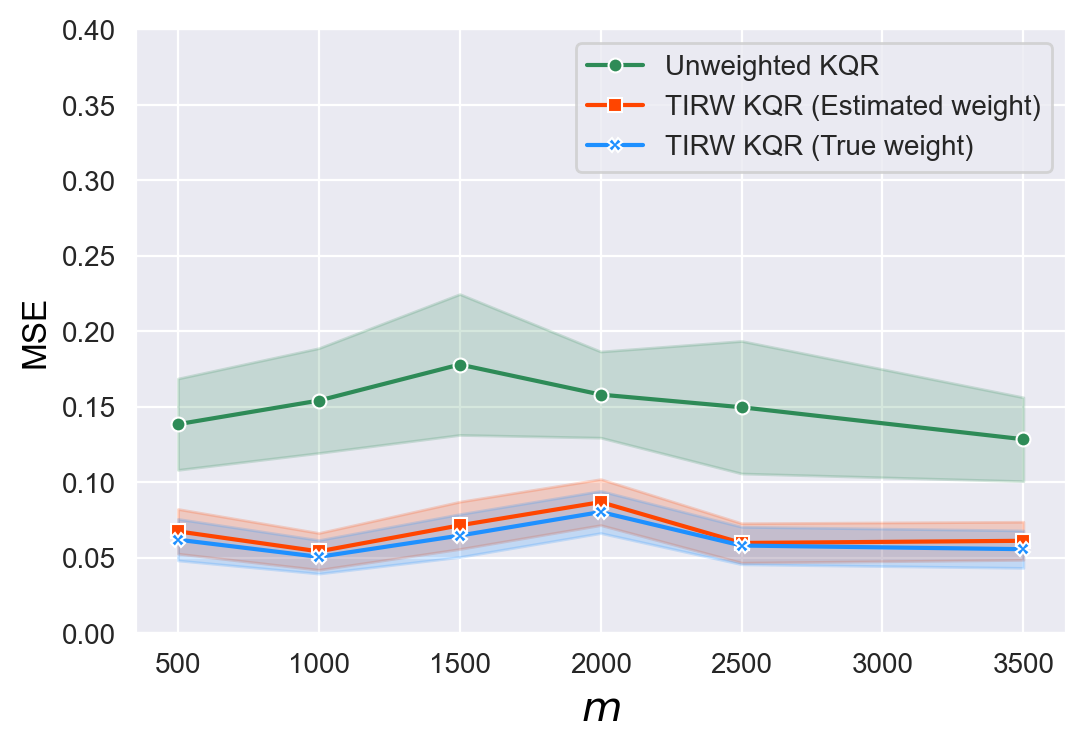}
    \label{KQR_dimension1_MSE_13}}
    \subfigure[$\tau=0.5$ and $r=1$]{
	\includegraphics[width=1.6in,  height=1.09in]{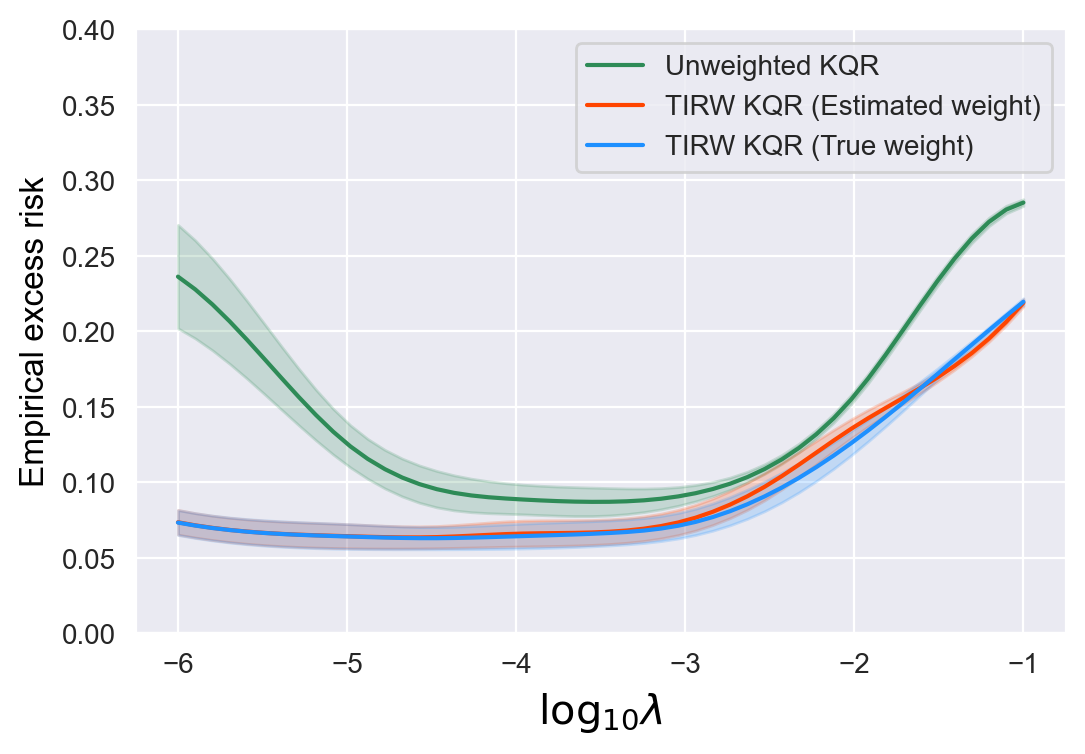}
        \label{KQR_dimension1_Empirical_smooth_1}}
 \subfigure[$\tau=0.5$ and $r=1$]{
    \includegraphics[width=1.6in,  height=1.09in]{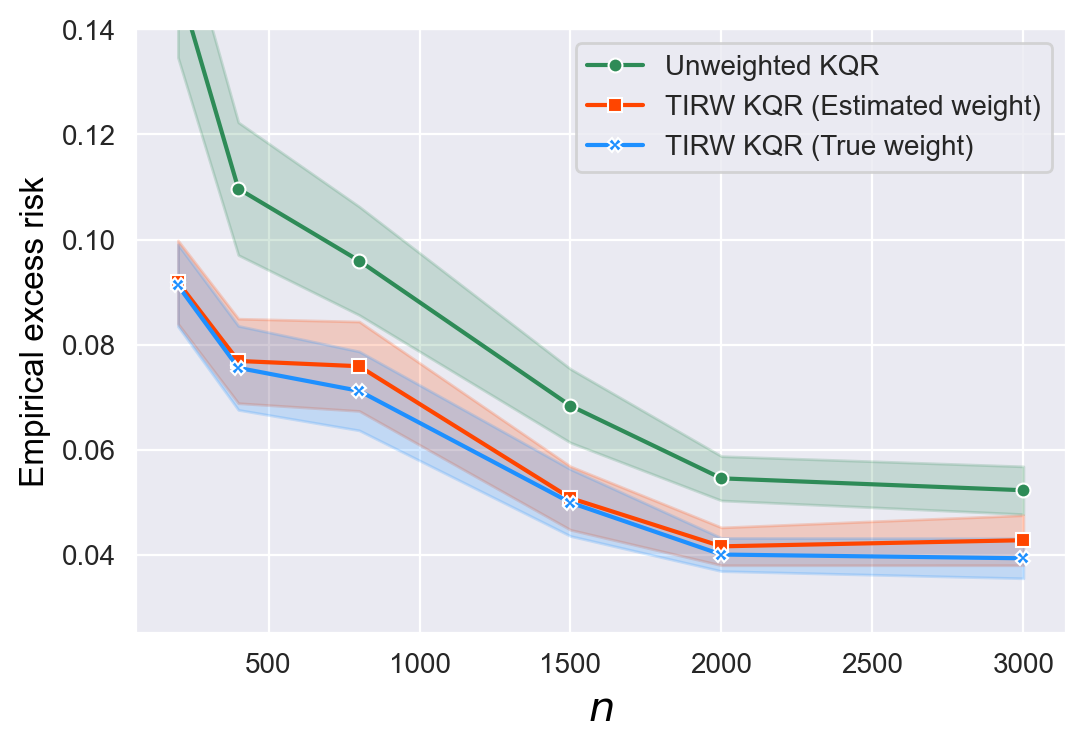}
\label{KQR_dimension1_Empirical_7}}
    \subfigure[$\tau=0.5$ and $r=1$]{
	\includegraphics[width=1.6in,  height=1.09in]{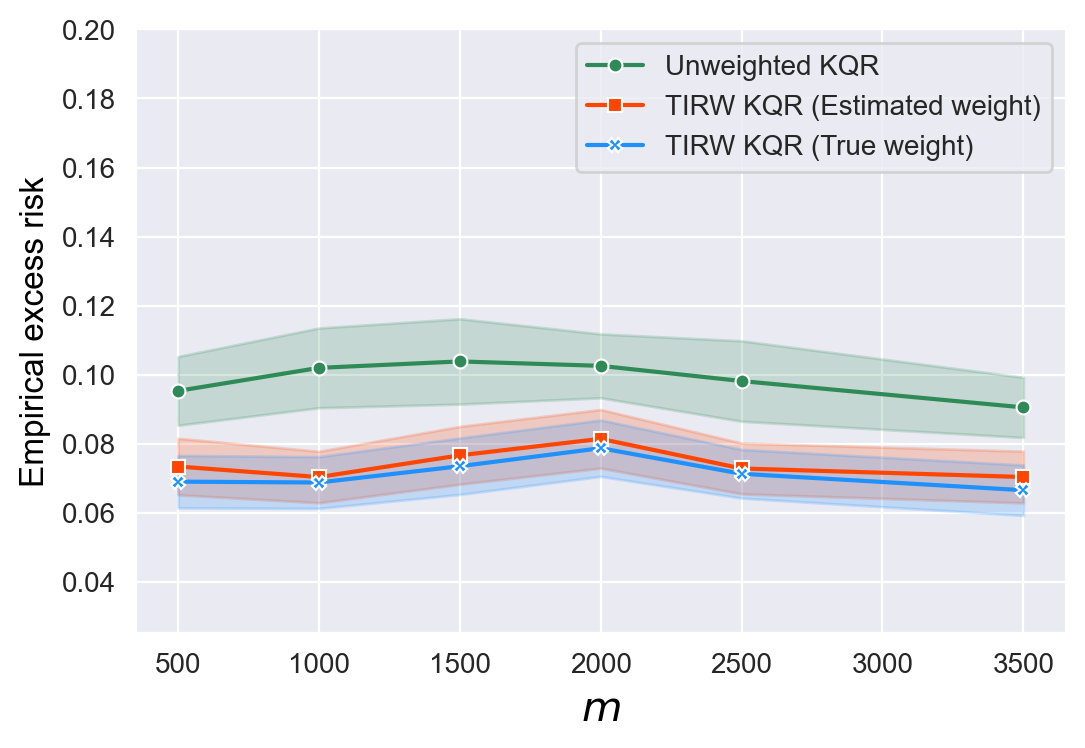}
 \label{KQR_dimension1_Empirical_13}}
 \setcounter{subfigure}{0}
\renewcommand{\thesubfigure}{(3\alph{subfigure})}
 \subfigure[$\tau=0.5$ and $r=0$]{
    \includegraphics[width=1.6in,  height=1.09in]{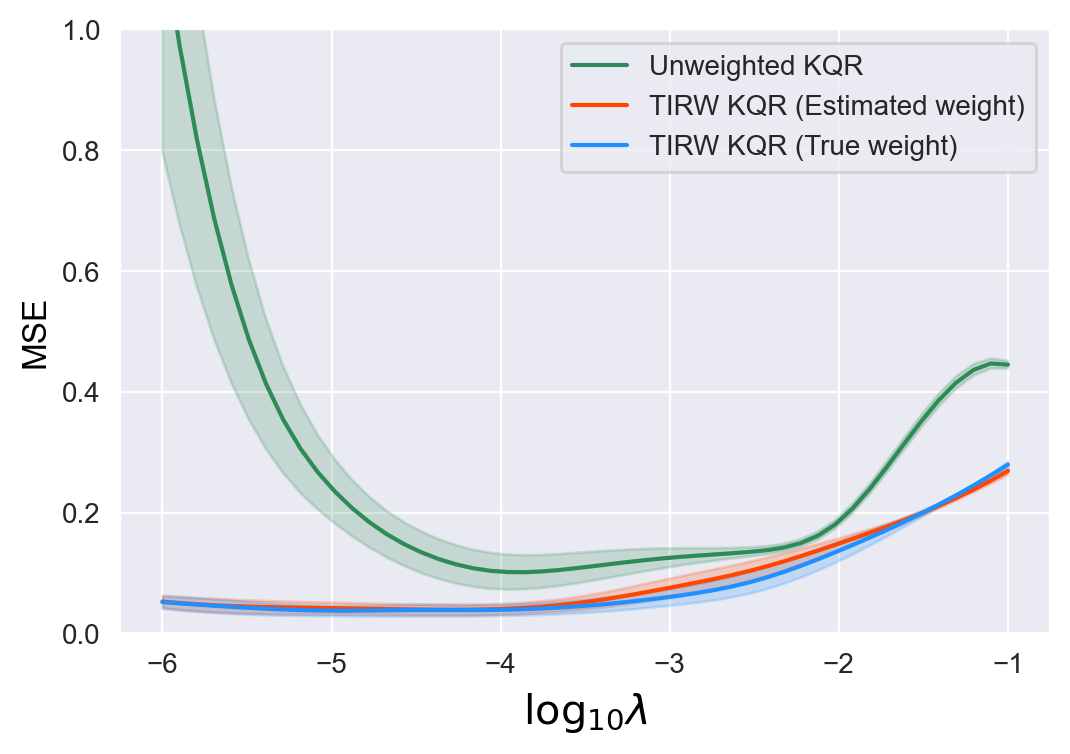}
    \label{KQR_dimension1_MSE_smooth_2}}
    \subfigure[$\tau=0.5$ and $r=0$]{
	\includegraphics[width=1.6in,  height=1.09in]{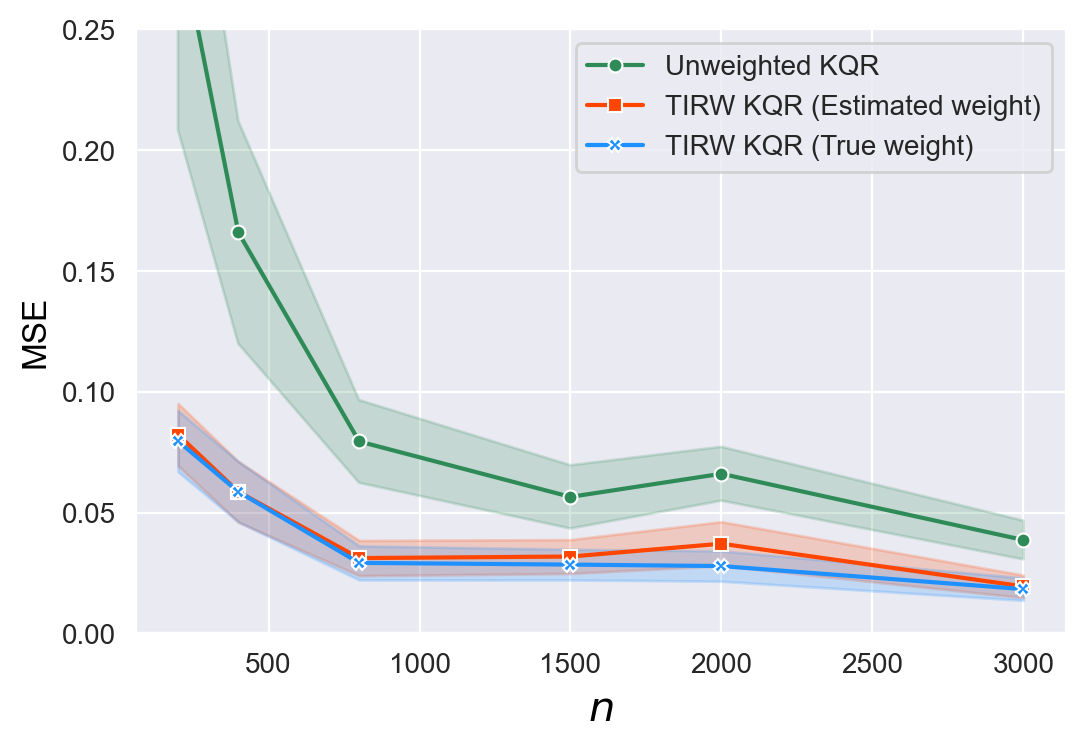}
\label{KQR_dimension1_MSE_8}}
   \subfigure[$\tau=0.5$ and $r=0$]{
    \includegraphics[width=1.6in,  height=1.09in]{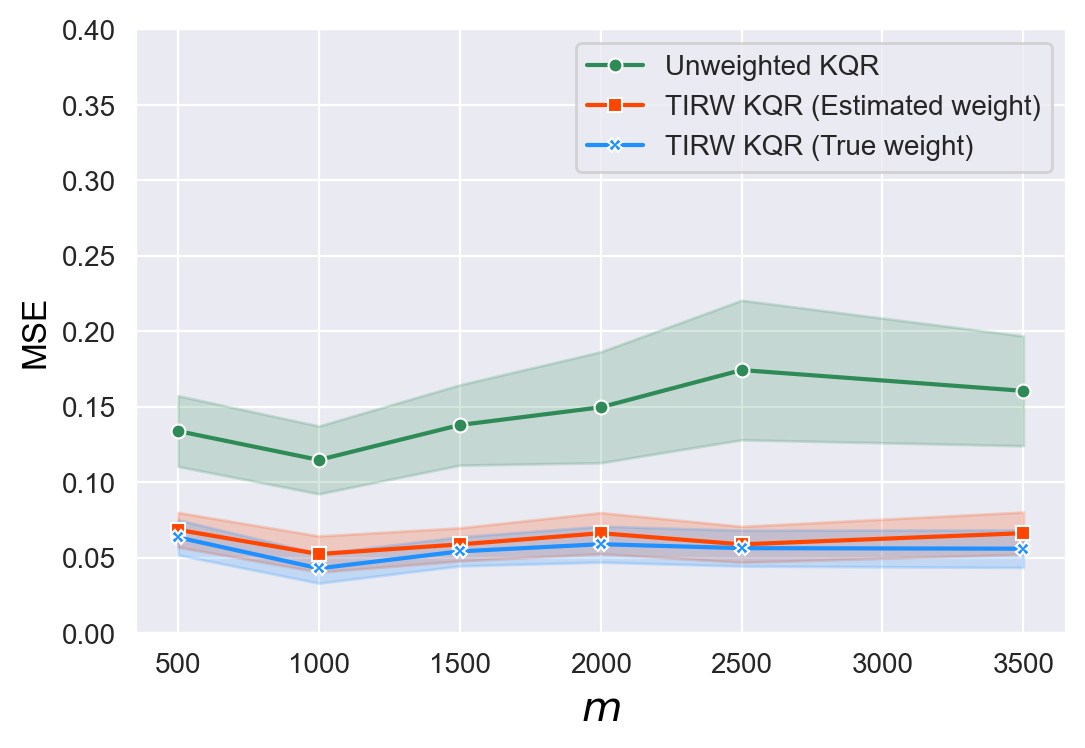}
    \label{KQR_dimension1_MSE_14}}
    \subfigure[$\tau=0.5$ and $r=0$]{
	\includegraphics[width=1.6in,  height=1.09in]{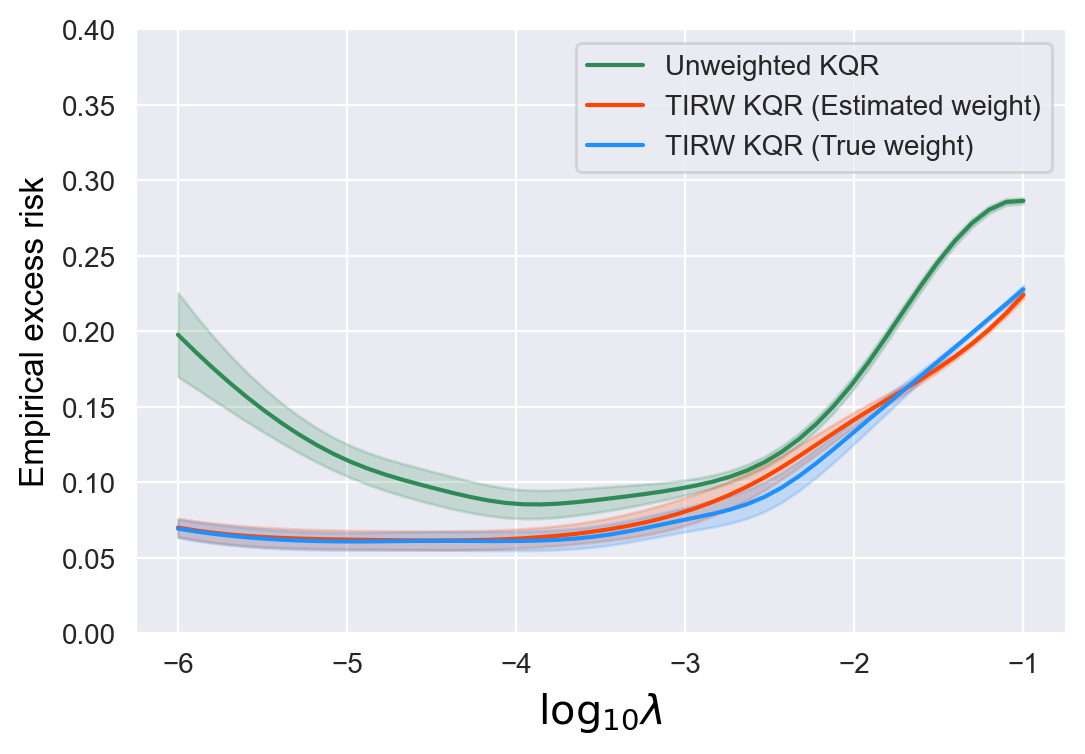}
        \label{KQR_dimension1_Empirical_smooth_2}}
 \subfigure[$\tau=0.5$ and $r=0$]{
    \includegraphics[width=1.6in,  height=1.09in]{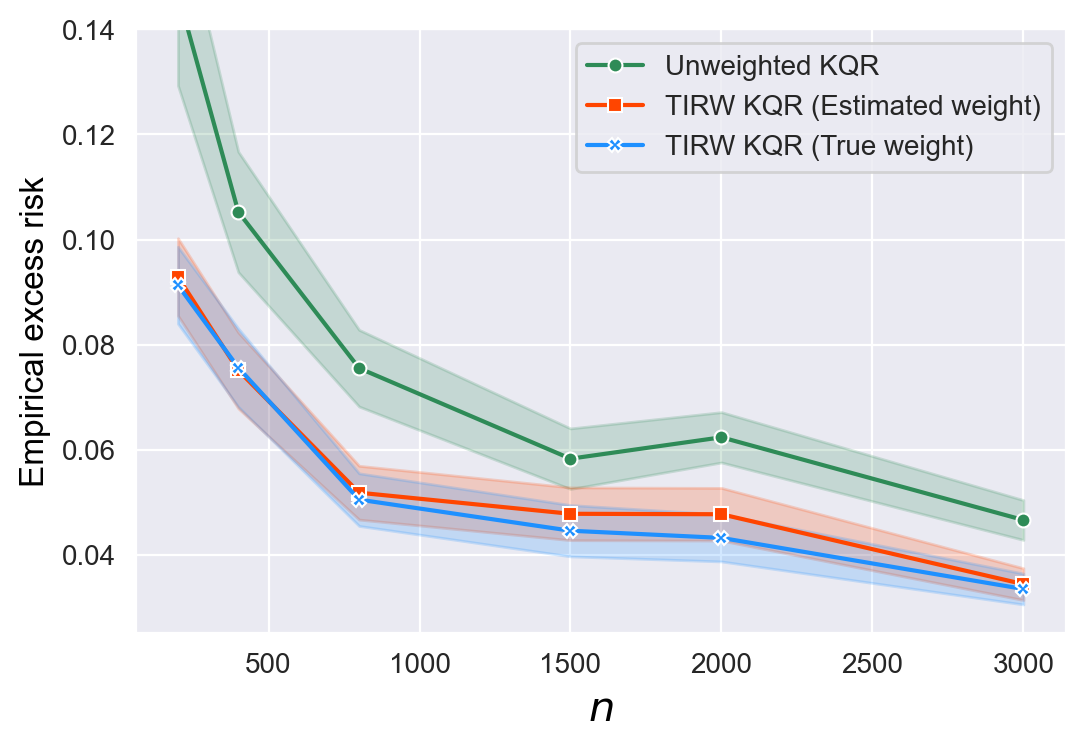}
\label{KQR_dimension1_Empirical_8}}
    \subfigure[$\tau=0.5$ and $r=0$]{
	\includegraphics[width=1.6in,  height=1.09in]{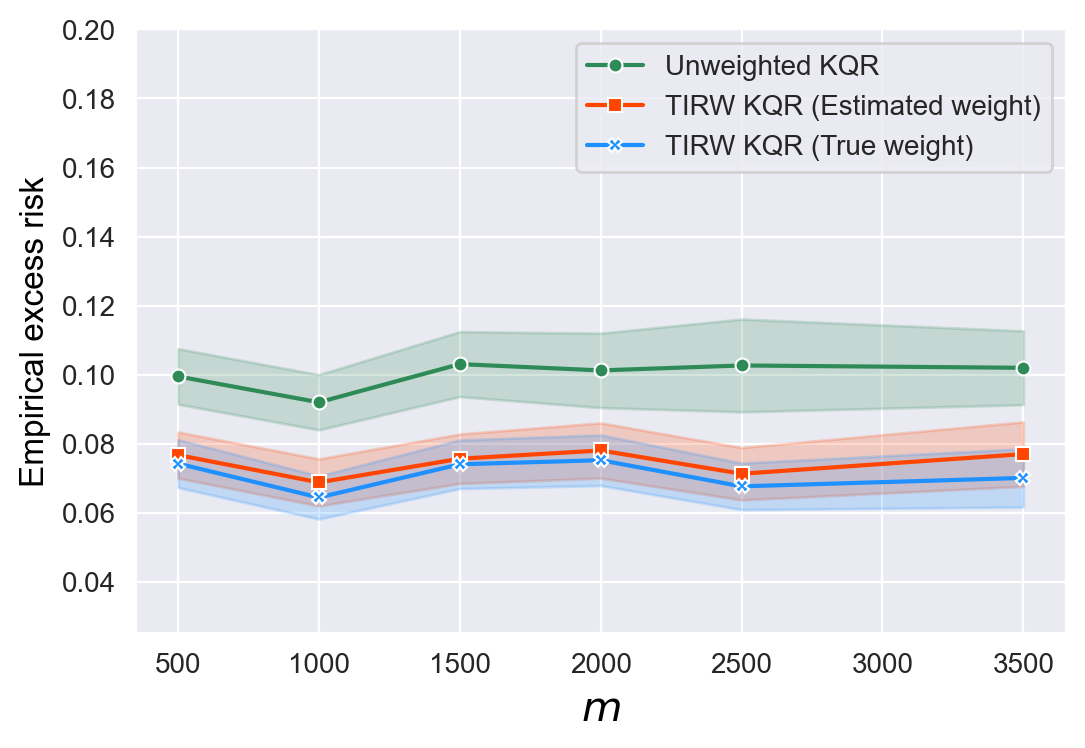}
 \label{KQR_dimension1_Empirical_14}}
\end{figure}

\begin{figure}[H]
\graphicspath{{unbounded_KQR_1_d/}}
 \centering
    
 \setcounter{subfigure}{0}
\renewcommand{\thesubfigure}{(4\alph{subfigure})}
  \subfigure[$\tau=0.7$ and $r=1$]{
  \includegraphics[width=1.6in,  height=1.09in]{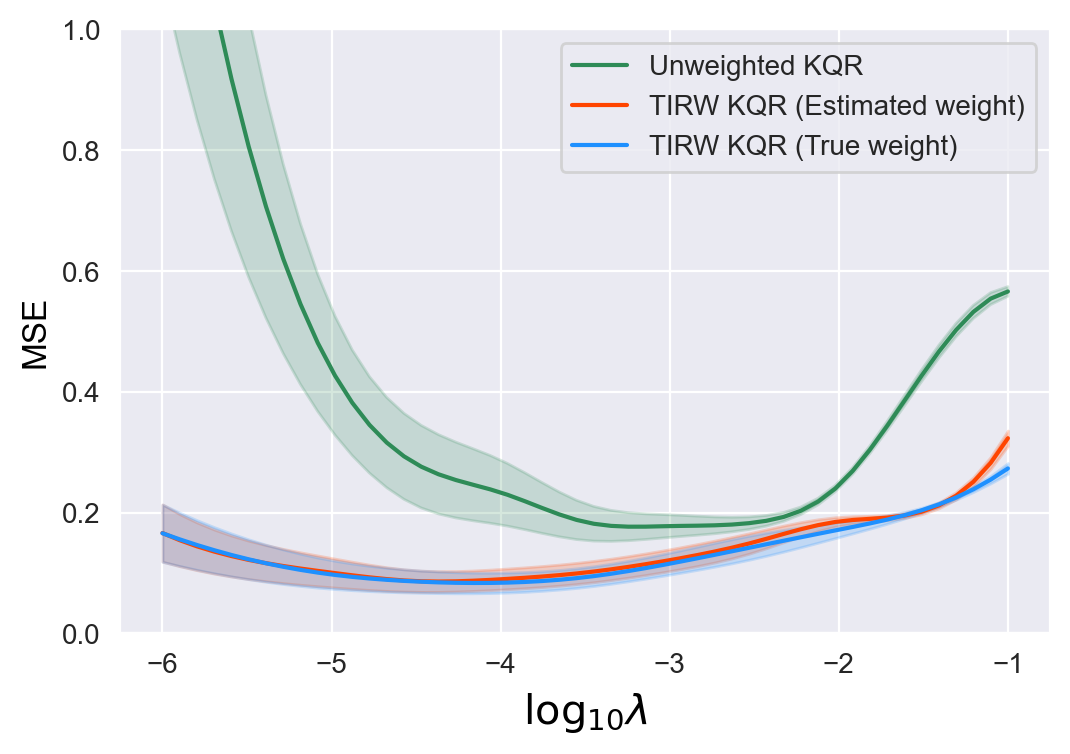}
\label{KQR_dimension1_MSE_smooth_5}}
    \subfigure[$\tau=0.7$ and $r=1$]{
	\includegraphics[width=1.6in,  height=1.09in]{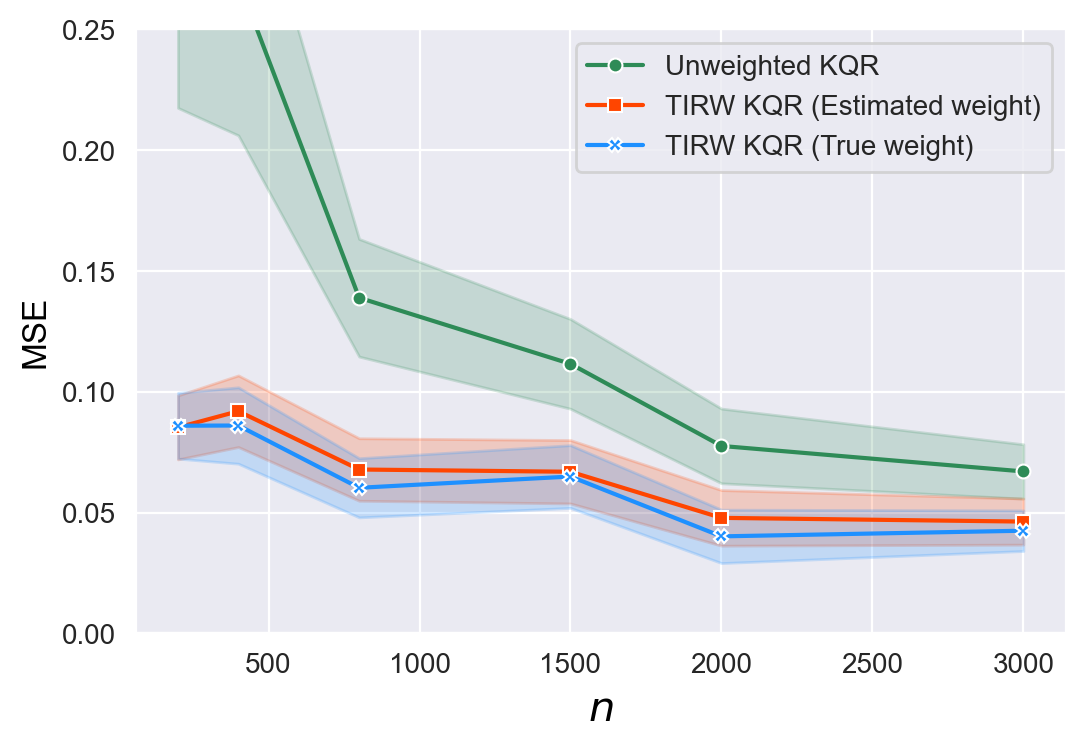}
\label{KQR_dimension1_MSE_11}}
   \subfigure[$\tau=0.7$ and $r=1$]{
    \includegraphics[width=1.6in,  height=1.09in]{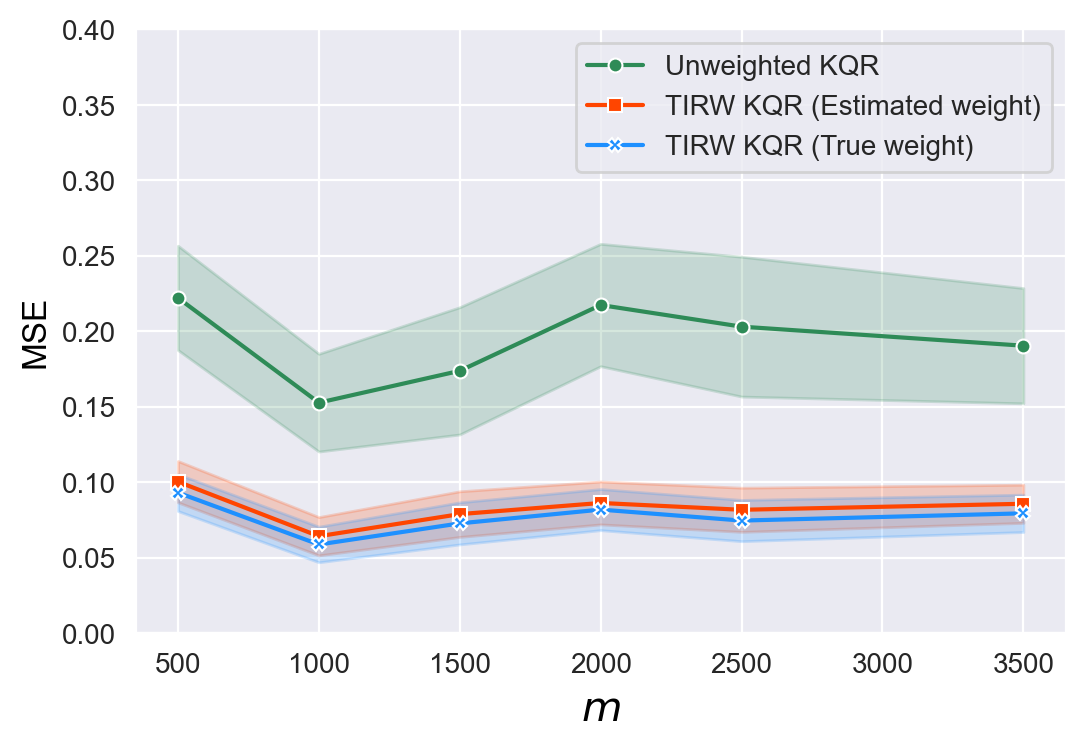}
    \label{KQR_dimension1_MSE_17}}
    \subfigure[$\tau=0.7$ and $r=1$]{
	\includegraphics[width=1.6in,  height=1.09in]{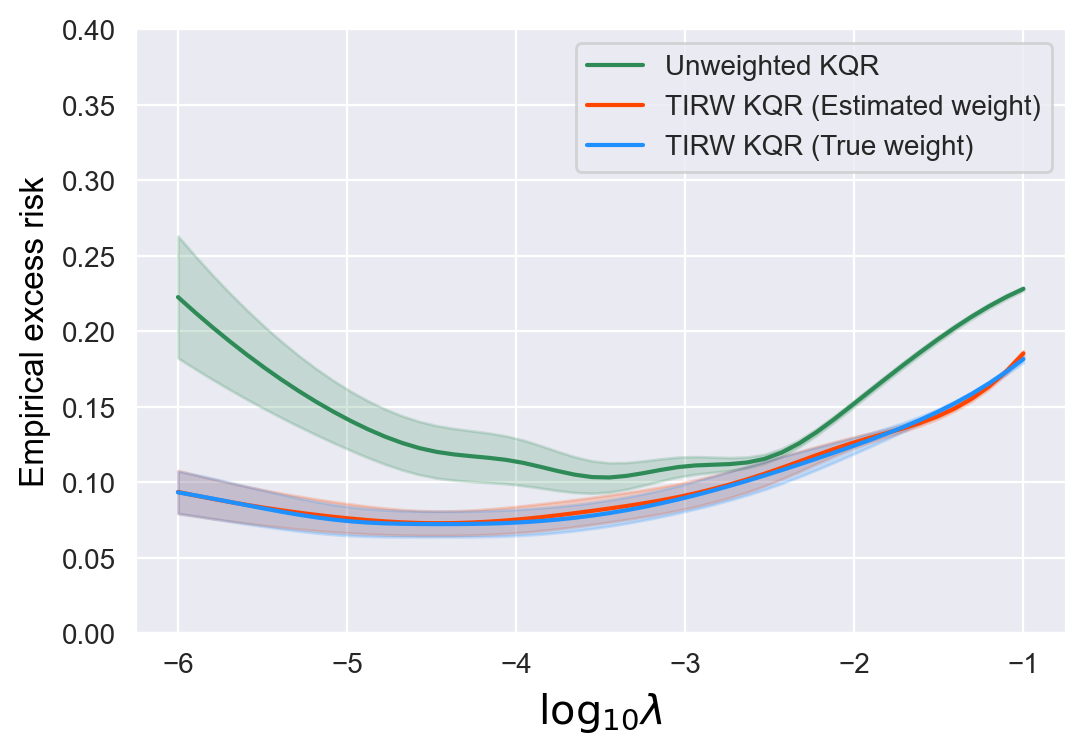} \label{KQR_dimension1_Empirical_smooth_5}}
 \subfigure[$\tau=0.7$ and $r=1$]{
    \includegraphics[width=1.6in,  height=1.09in]{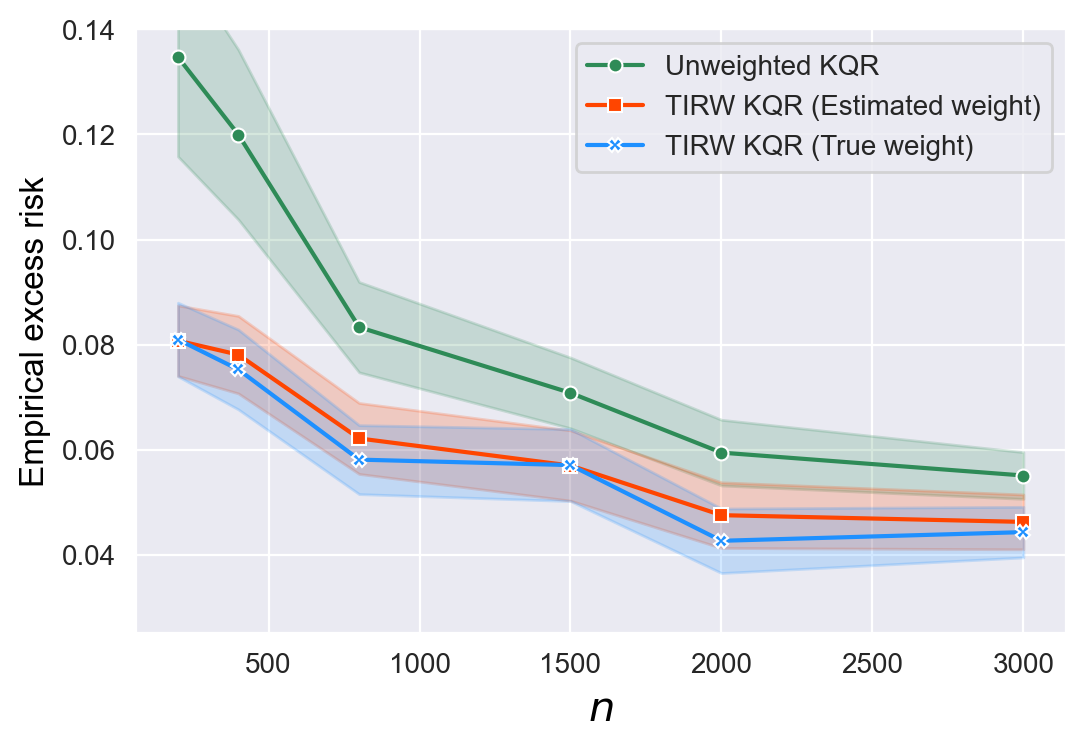}
\label{KQR_dimension1_Empirical_11}}
    \subfigure[$\tau=0.7$ and $r=1$]{
	\includegraphics[width=1.6in,  height=1.09in]{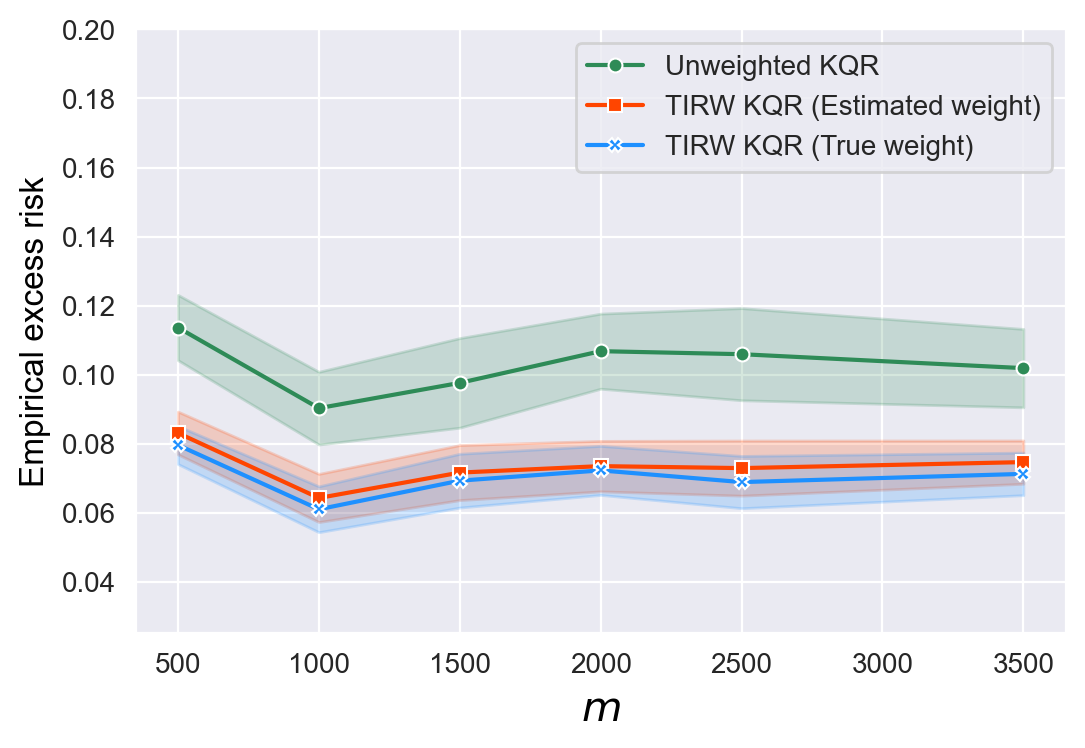}
 \label{KQR_dimension1_Empirical_17}}
 \setcounter{subfigure}{0}
\renewcommand{\thesubfigure}{(5\alph{subfigure})}
     \subfigure[$\tau=0.7$ and $r=0$]{
    \includegraphics[width=1.6in,  height=1.09in]{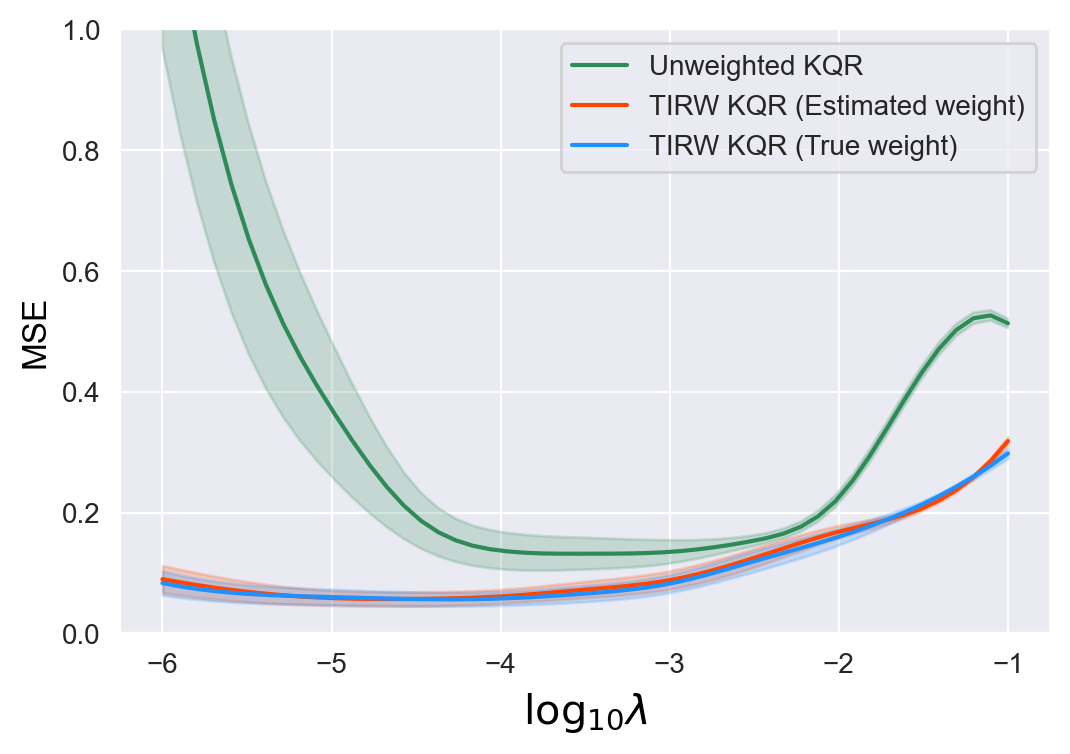}
\label{KQR_dimension1_MSE_smooth_6}}
    \subfigure[$\tau=0.7$ and $r=0$]{
	\includegraphics[width=1.6in,  height=1.09in]{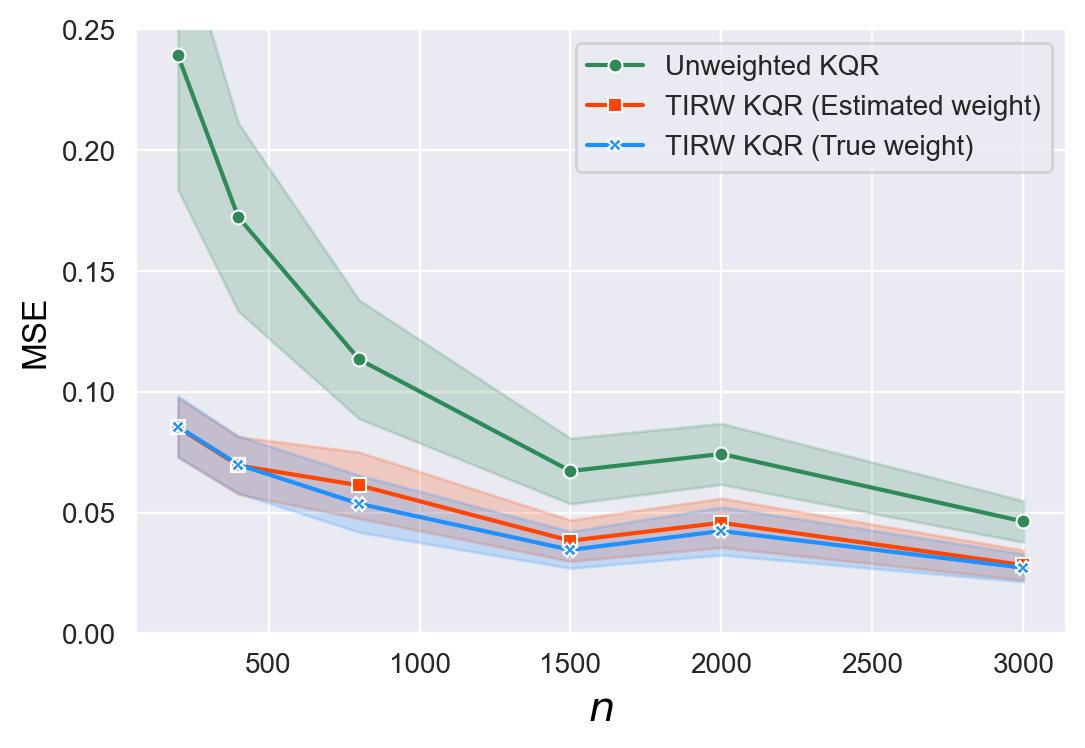}
\label{KQR_dimension1_MSE_12}}
   \subfigure[$\tau=0.7$ and $r=0$]{
    \includegraphics[width=1.6in,  height=1.09in]{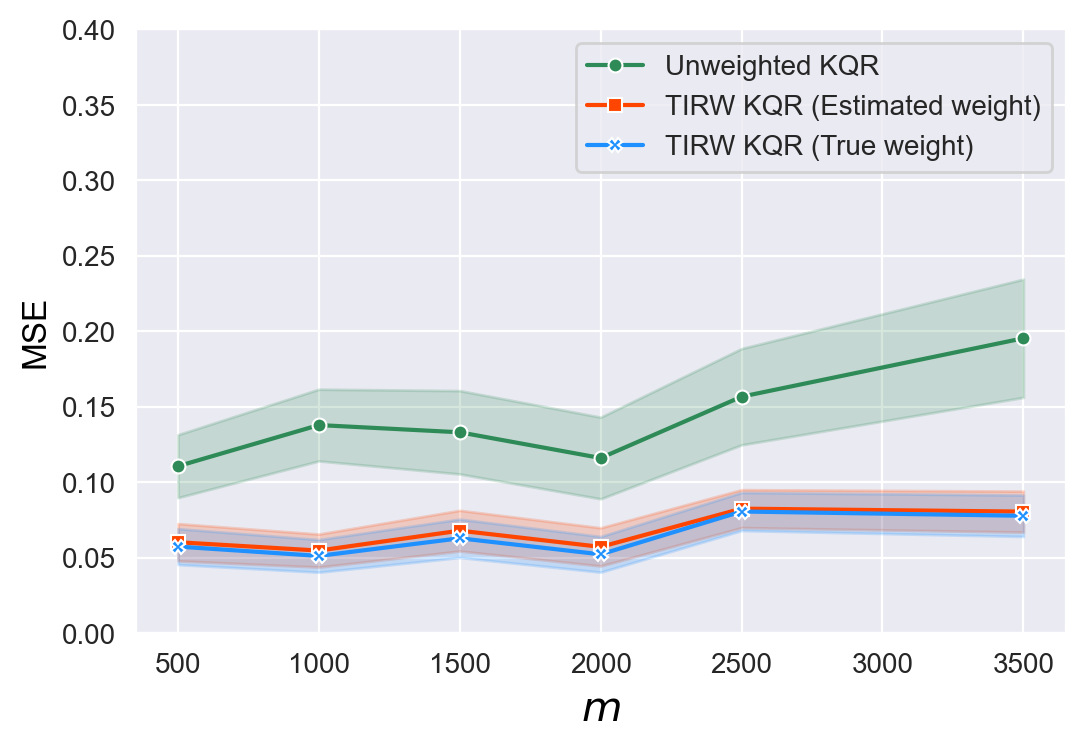}
    \label{KQR_dimension1_MSE_18}}
    \subfigure[$\tau=0.7$ and $r=0$]{
	\includegraphics[width=1.6in,  height=1.09in]{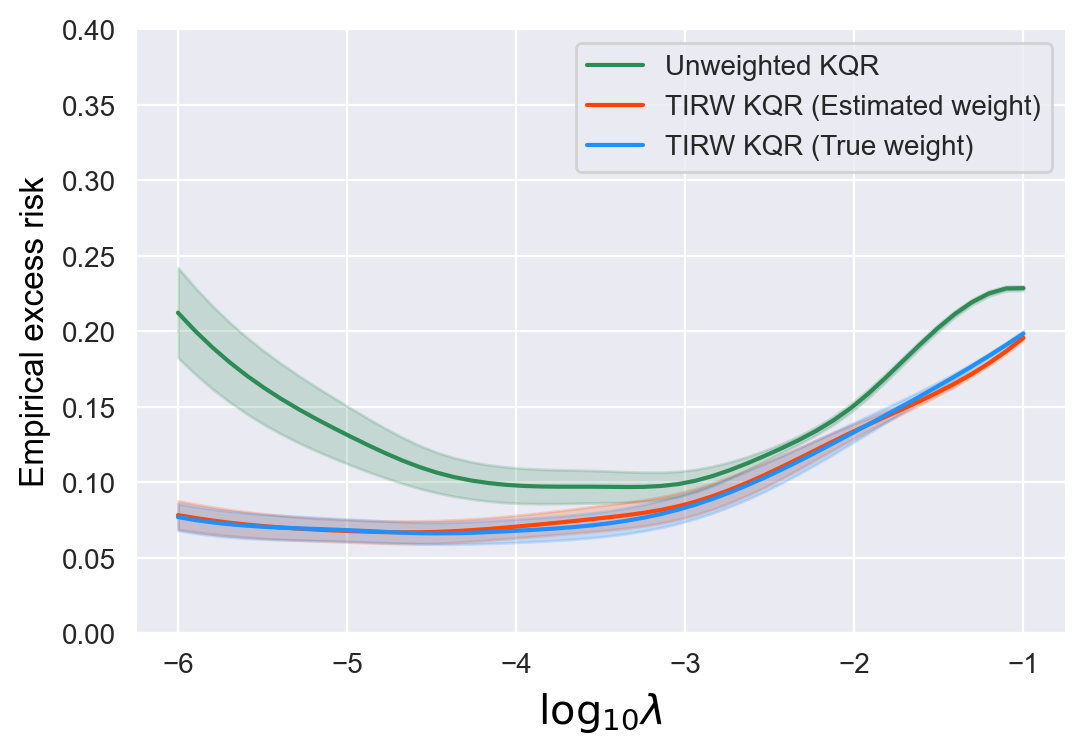} \label{KQR_dimension1_Empirical_smooth_6}}
 \subfigure[$\tau=0.7$ and $r=0$]{
    \includegraphics[width=1.6in,  height=1.09in]{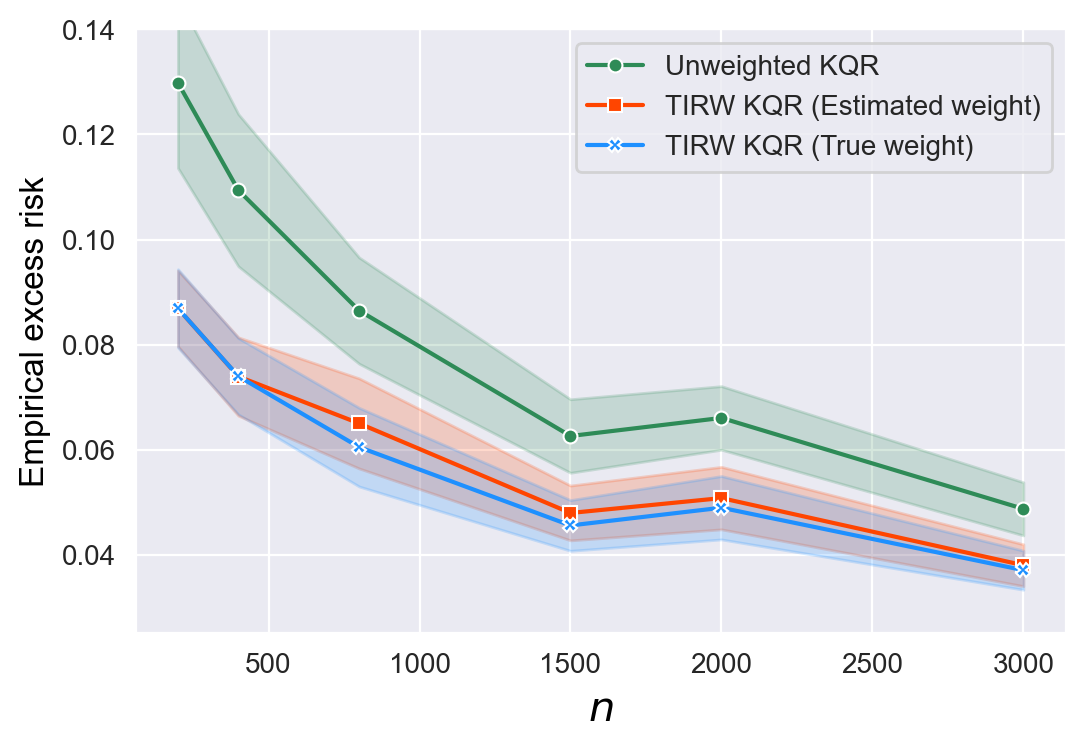}
\label{KQR_dimension1_Empirical_12}}
    \subfigure[$\tau=0.7$ and $r=0$]{
	\includegraphics[width=1.6in,  height=1.09in]{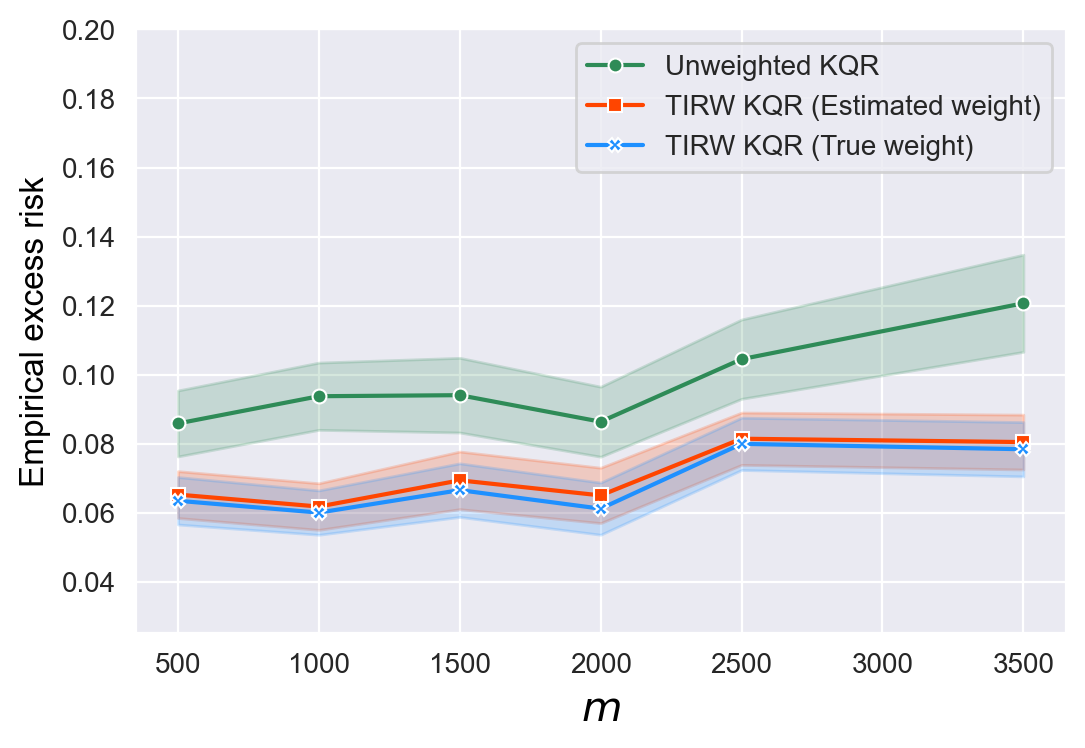}
 \label{KQR_dimension1_Empirical_18}}
 \caption{\footnotesize{Average MSE  and empirical excess risk for unweighted KQR, TIRW KQR with true weight and estimated weight, respectively (in the left panel, the curves are plotted with respect to $\log_{10} \lambda$ with $n=500, m=1000$; in the middle panel, the curves are plotted with respect to $n$ with fixed $m = 1000,\lambda=10^{-4}$; in the right panel, the curves are plotted with respect to $m$ with fixed $n=500,\lambda=10^{-4}$)}}
\label{KQR_dimension1_unbounded}
\end{figure}

\subsubsection{Uniformly bounded case in Example S4}\label{sec:a.2.3}
\begin{figure}[H]
\graphicspath{{bounded_KQR_3_d/}}
    \centering
    \setcounter{subfigure}{0}
\renewcommand{\thesubfigure}{(1\alph{subfigure})}
    \subfigure[$\tau=0.3$ and $r=1$]{
    \includegraphics[width=1.6in,  height=1.09in]{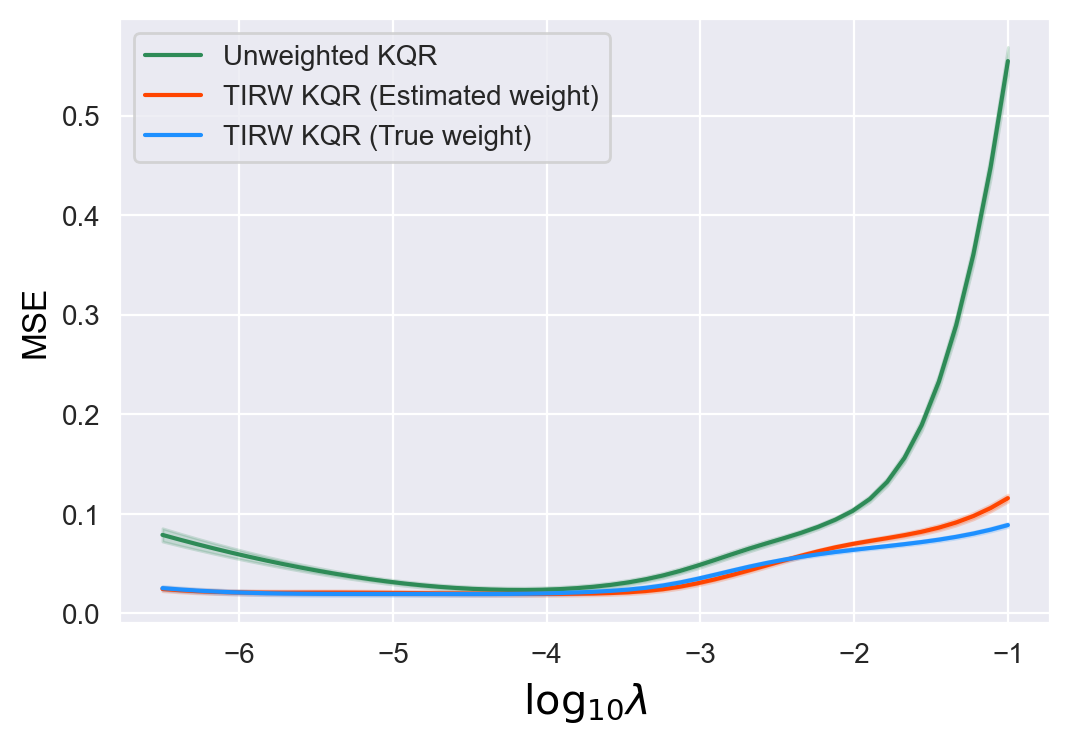}
\label{KQR_dimension3_uniformbounded_MSE_smooth_3}}
    \subfigure[$\tau=0.3$ and $r=1$]{
	\includegraphics[width=1.6in,  height=1.09in]{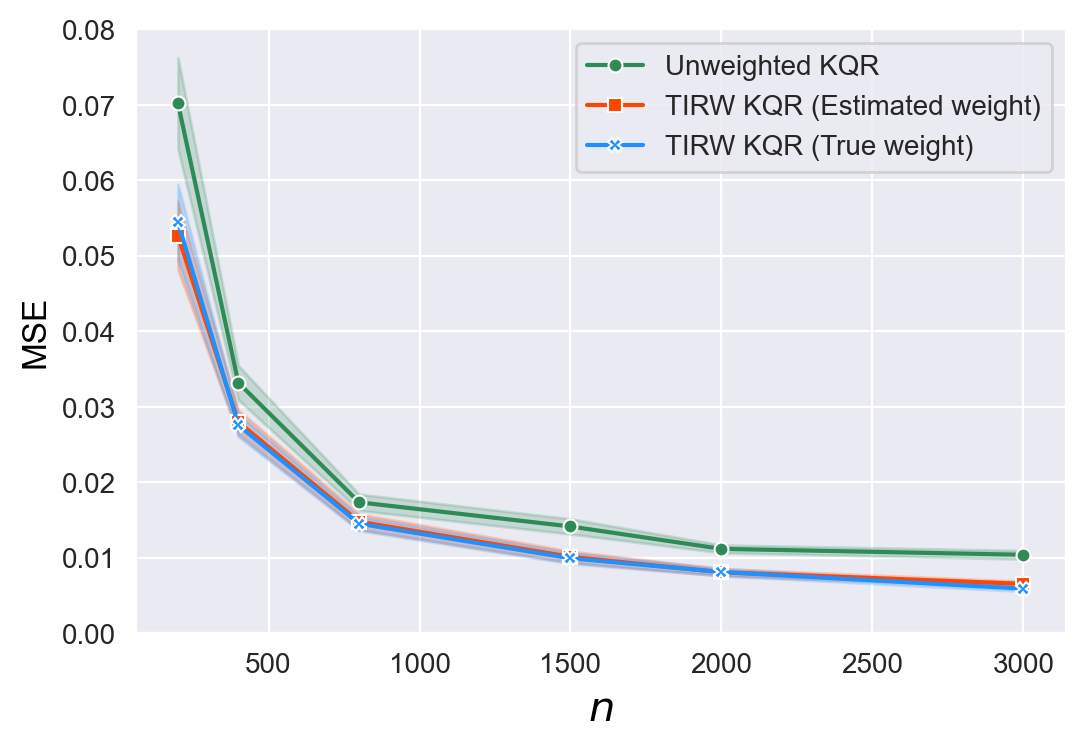}
\label{KQR_dimension3_uniformbounded_MSE_9}}
   \subfigure[$\tau=0.3$ and $r=1$]{
    \includegraphics[width=1.6in,  height=1.09in]{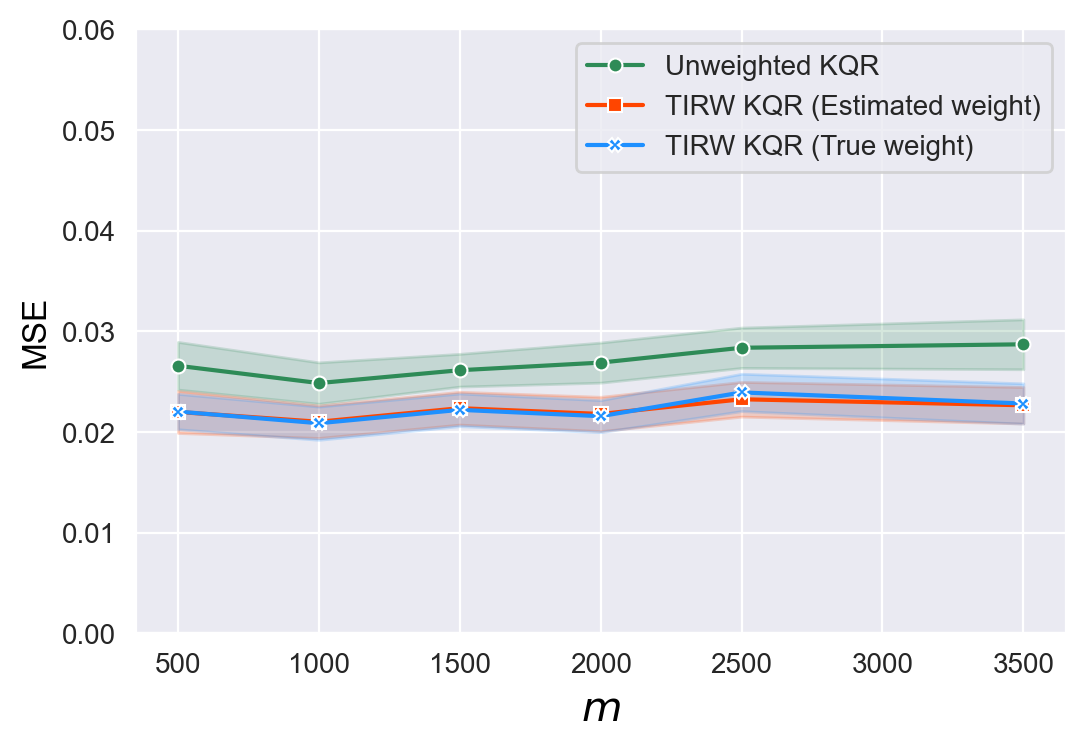}
\label{KQR_dimension3_uniformbounded_MSE_15}}
    \subfigure[$\tau=0.3$ and $r=1$]{
	\includegraphics[width=1.6in,  height=1.09in]{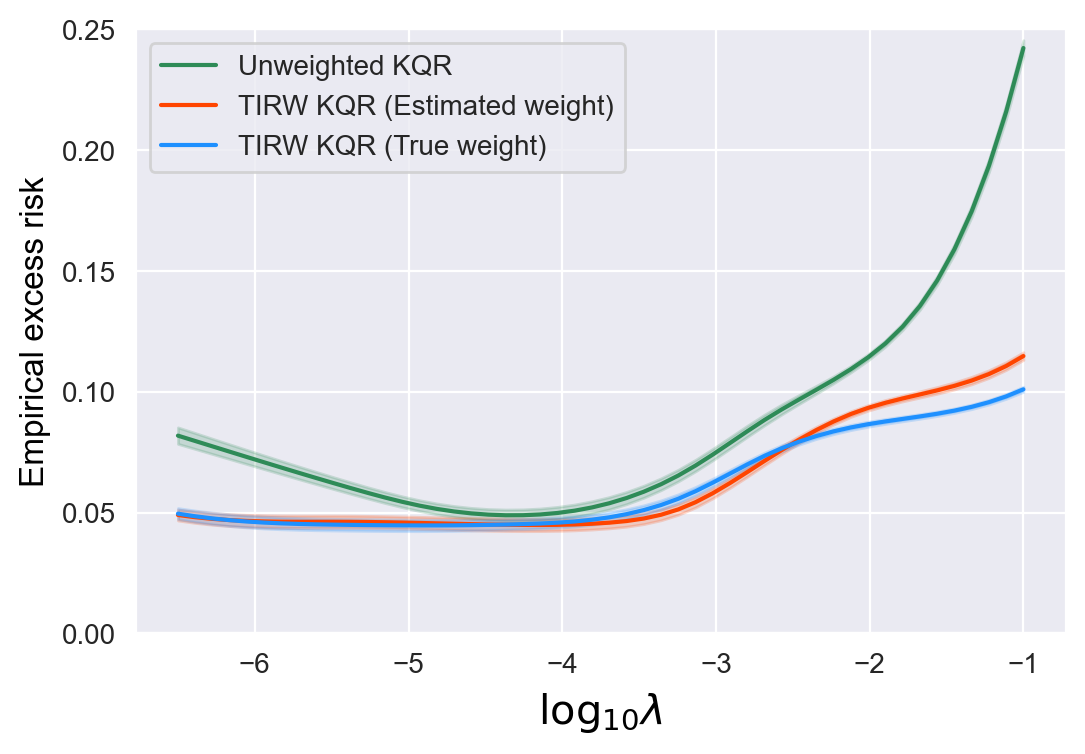}
\label{KQR_dimension3_uniformbounded_Empirical_smooth_3}}
 \subfigure[$\tau=0.3$ and $r=1$]{
    \includegraphics[width=1.6in,  height=1.09in]{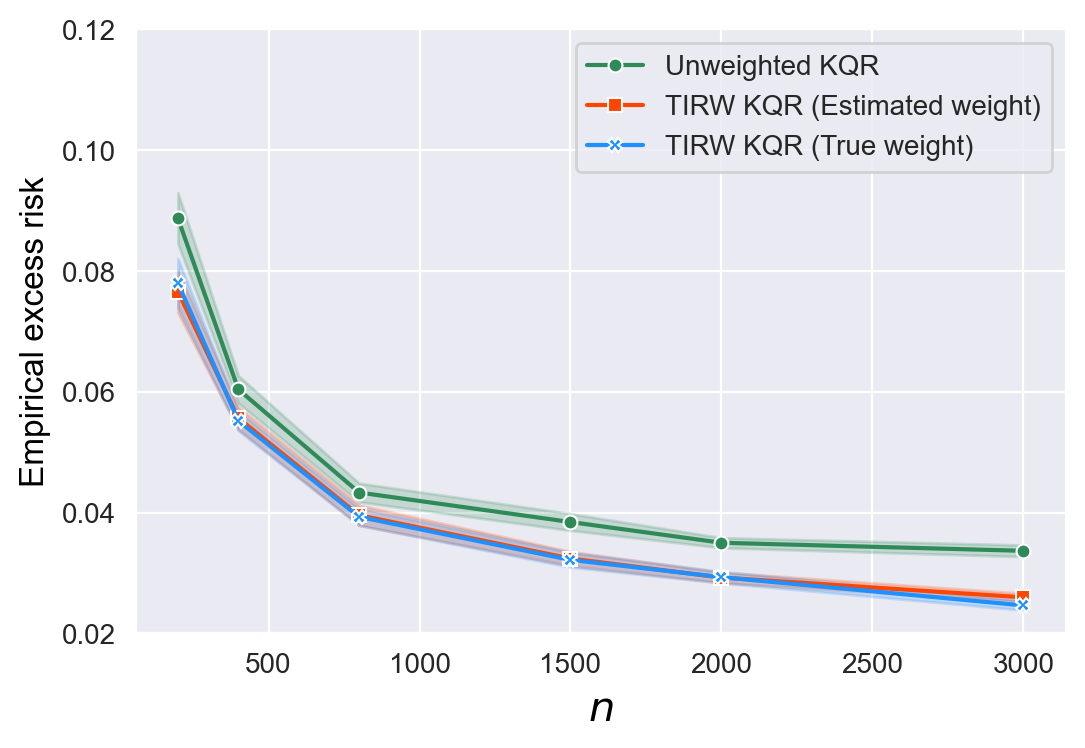}
\label{KQR_dimension3_uniformbounded_Empirical_9}}
    \subfigure[$\tau=0.3$ and $r=1$]{
	\includegraphics[width=1.6in,  height=1.09in]{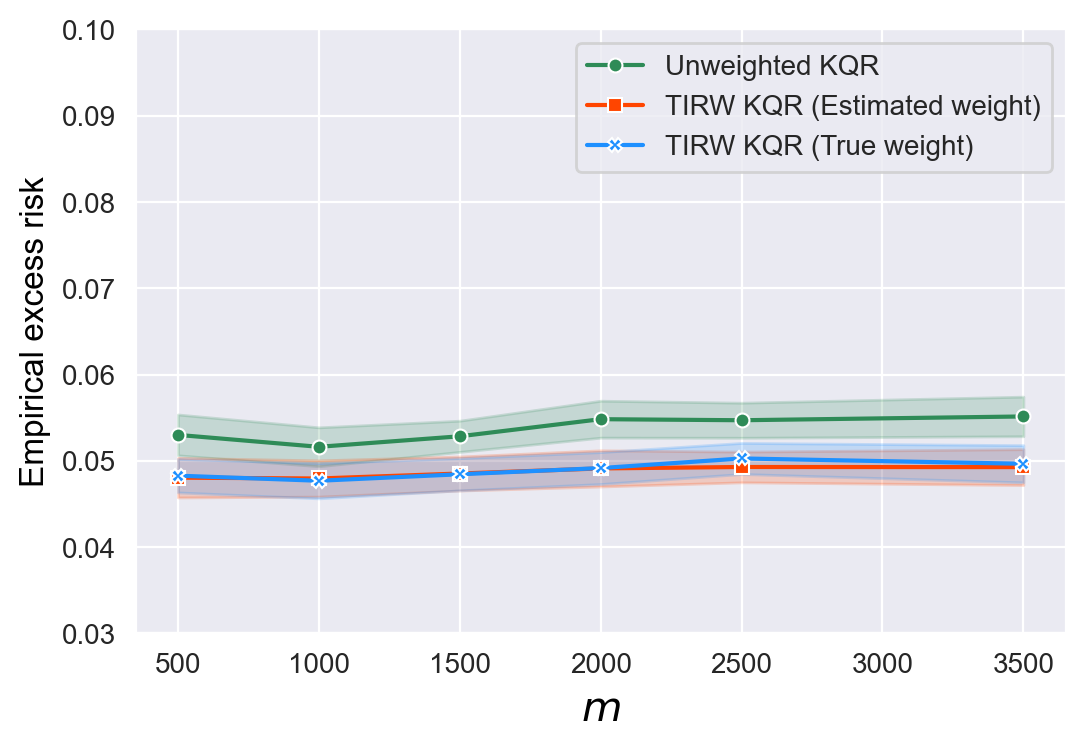}
\label{KQR_dimension3_uniformbounded_Empirical_15}}
\setcounter{subfigure}{0}
\renewcommand{\thesubfigure}{(2\alph{subfigure})}
\subfigure[$\tau=0.3$ and $r=0$]{
    \includegraphics[width=1.6in,  height=1.09in]{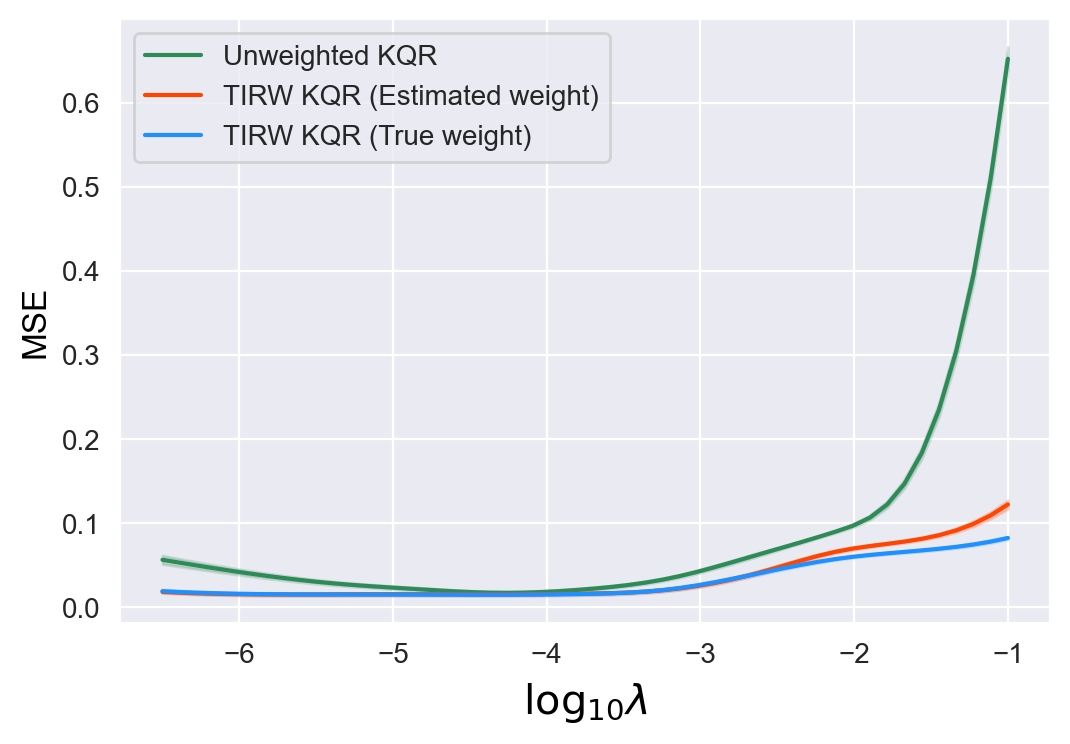}
\label{KQR_dimension3_uniformbounded_MSE_smooth_4}}
    \subfigure[$\tau=0.3$ and $r=0$]{
	\includegraphics[width=1.6in,  height=1.09in]{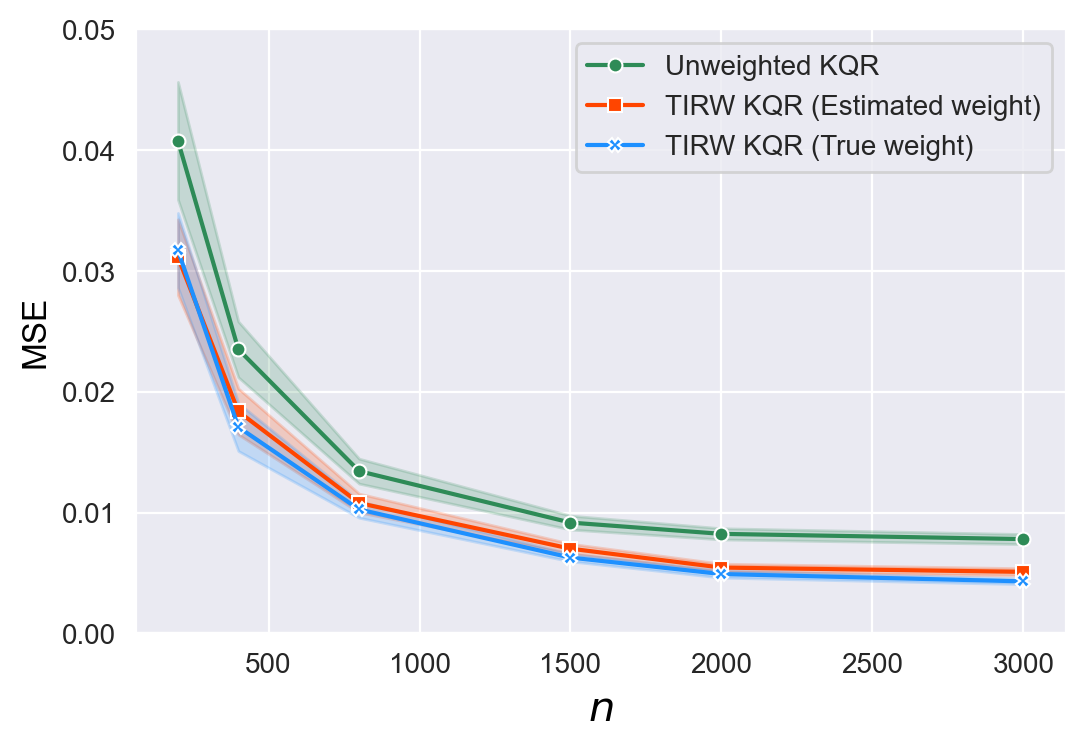}
\label{KQR_dimension3_uniformbounded_MSE_10}}
   \subfigure[$\tau=0.3$ and $r=0$]{
    \includegraphics[width=1.6in,  height=1.09in]{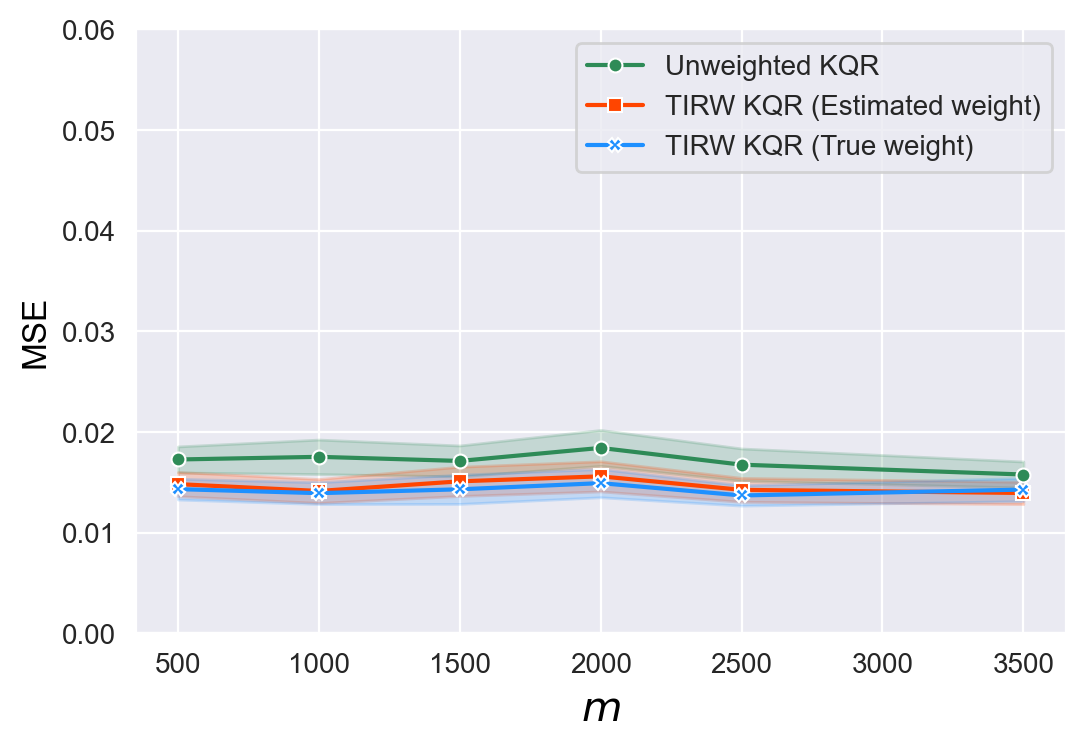}
\label{KQR_dimension3_uniformbounded_MSE_16}}
    \subfigure[$\tau=0.3$ and $r=0$]{
	\includegraphics[width=1.6in,  height=1.09in]{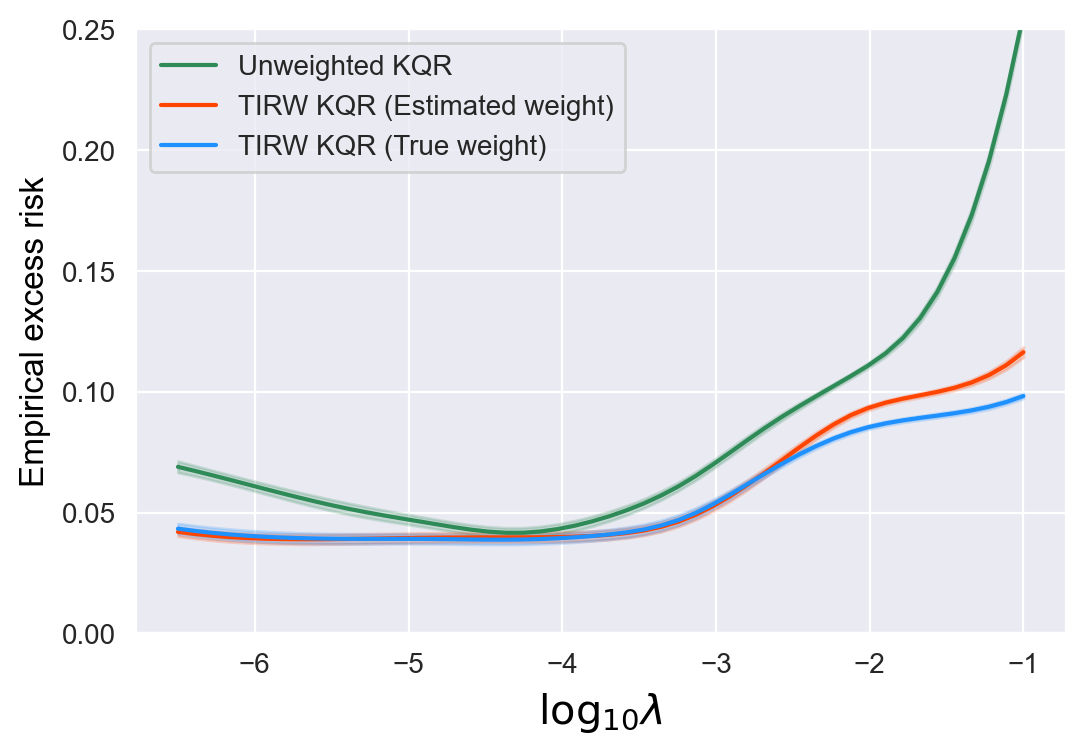}
\label{KQR_dimension3_uniformbounded_Empirical_smooth_4}}
 \subfigure[$\tau=0.3$ and $r=0$]{
    \includegraphics[width=1.6in,  height=1.09in]{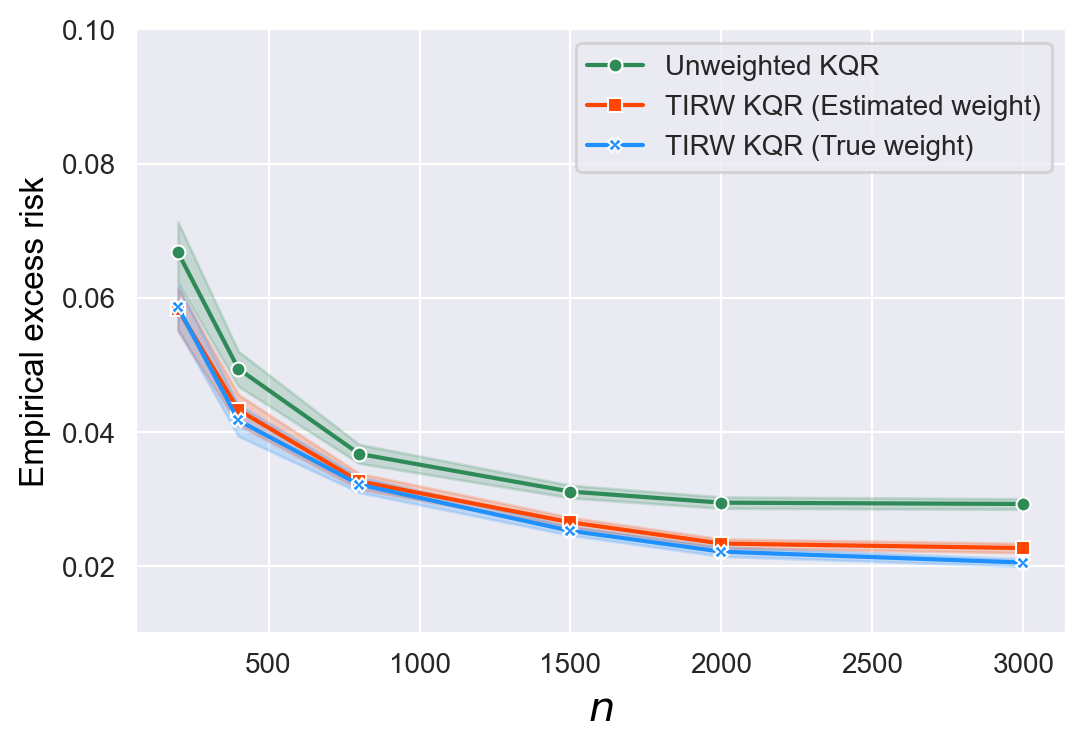}
\label{KQR_dimension3_uniformbounded_Empirical_10}}
    \subfigure[$\tau=0.3$ and $r=0$]{
	\includegraphics[width=1.6in,  height=1.09in]{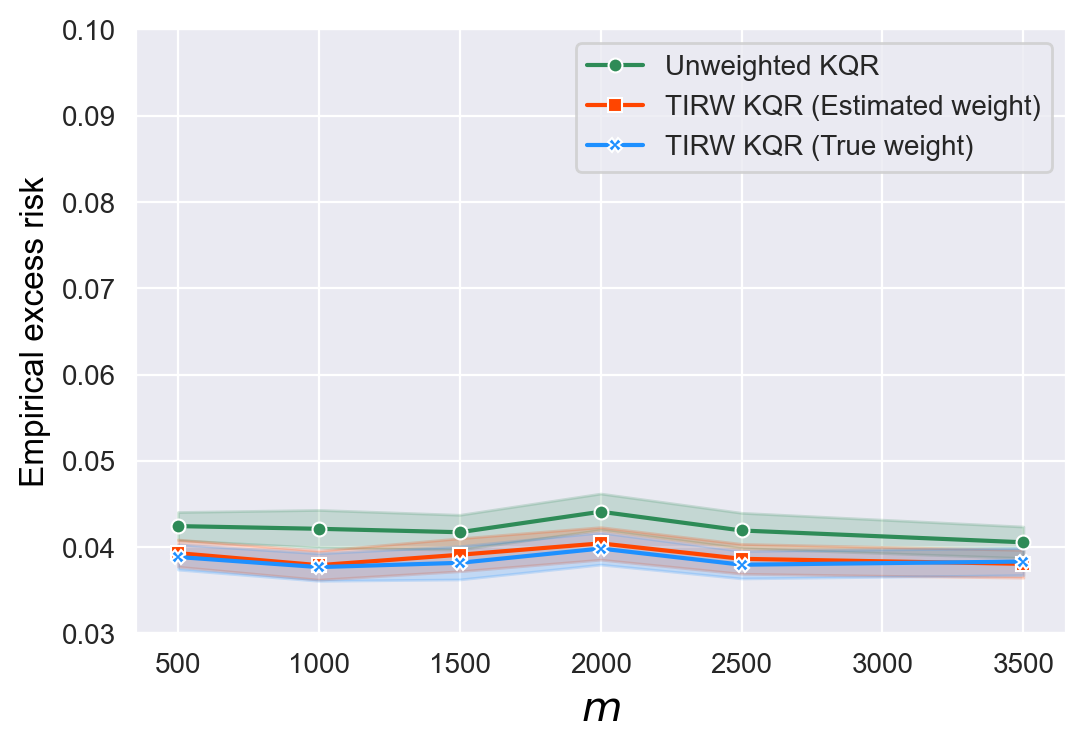}
\label{KQR_dimension3_uniformbounded_Empirical_16}}
\setcounter{subfigure}{0}
\renewcommand{\thesubfigure}{(3\alph{subfigure})}
\subfigure[$\tau=0.5$ and $r=1$]{
    \includegraphics[width=1.6in,  height=1.09in]{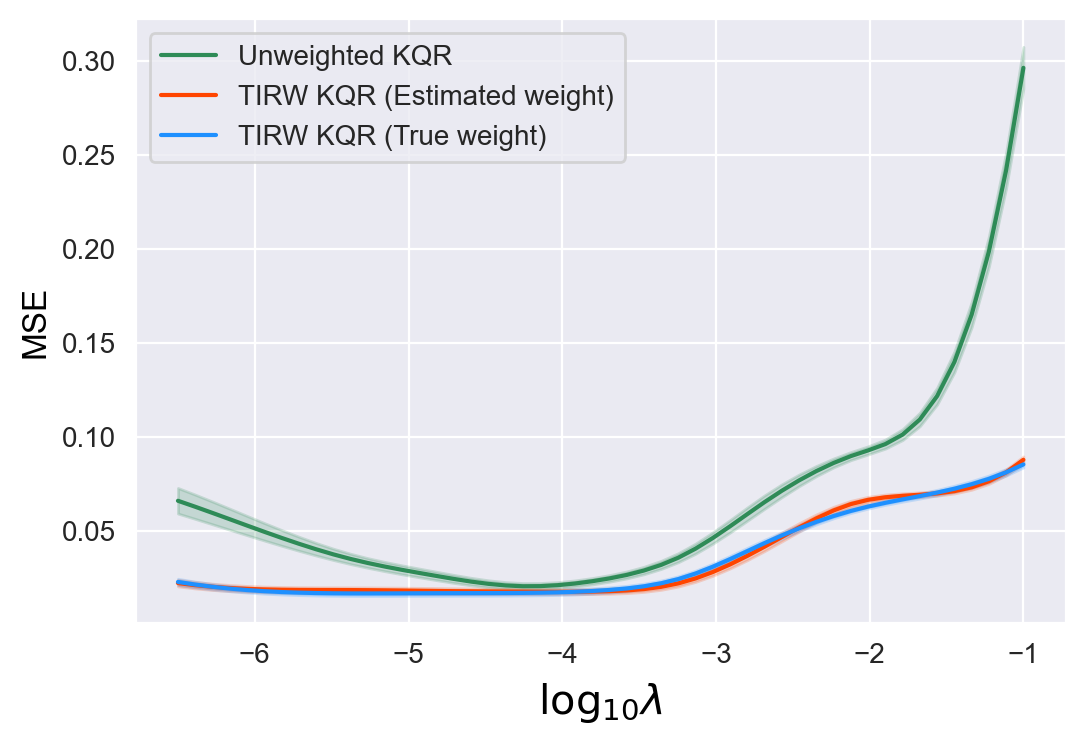}
\label{KQR_dimension3_uniformbounded_MSE_smooth_1}}
    \subfigure[$\tau=0.5$ and $r=1$]{
	\includegraphics[width=1.6in,  height=1.09in]{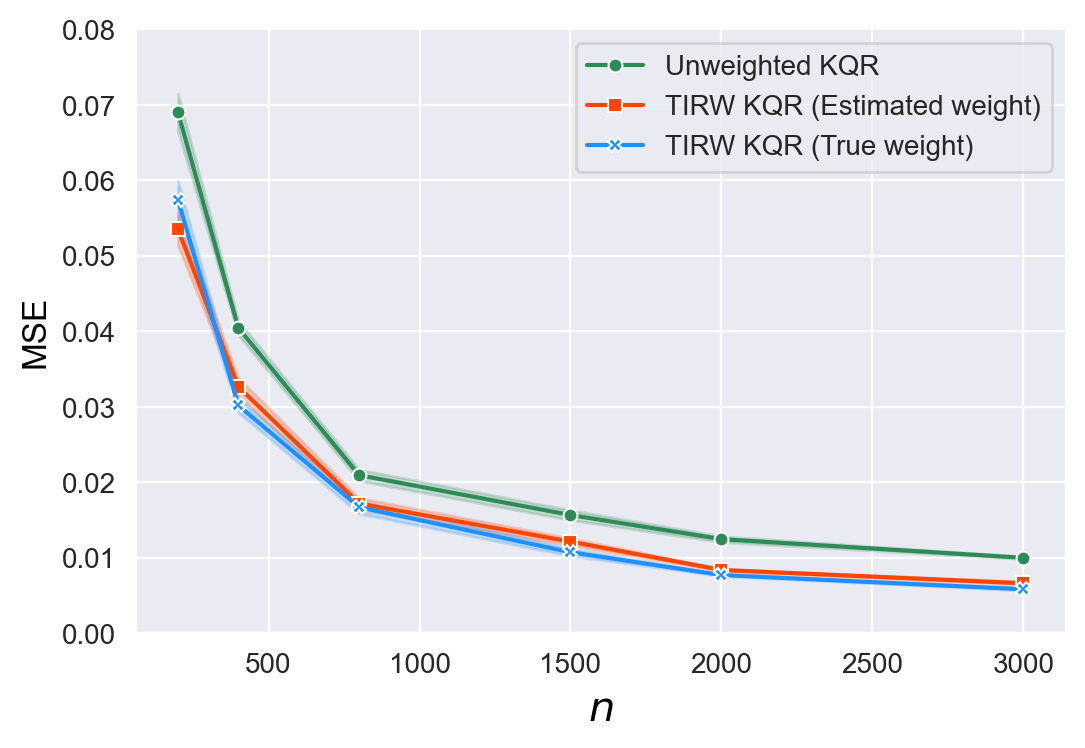}
\label{KQR_dimension3_uniformbounded_MSE_7}}
   \subfigure[$\tau=0.5$ and $r=1$]{
    \includegraphics[width=1.6in,  height=1.09in]{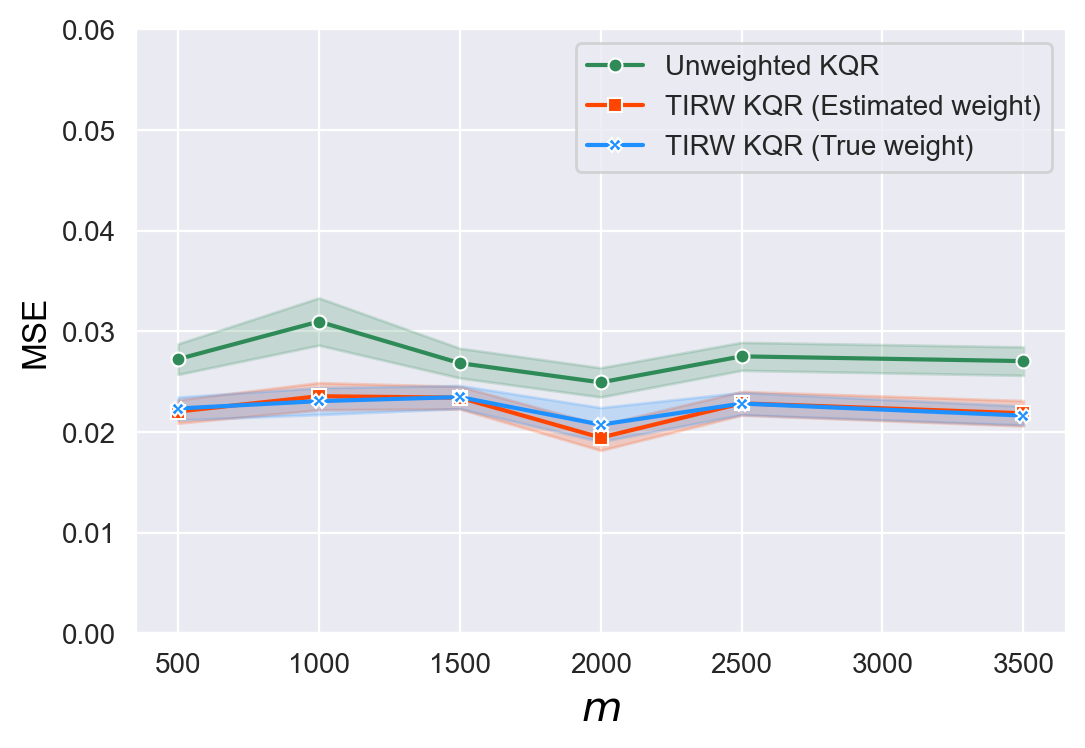}
\label{KQR_dimension3_uniformbounded_MSE_13}}
    \subfigure[$\tau=0.5$ and $r=1$]{
	\includegraphics[width=1.6in,  height=1.09in]{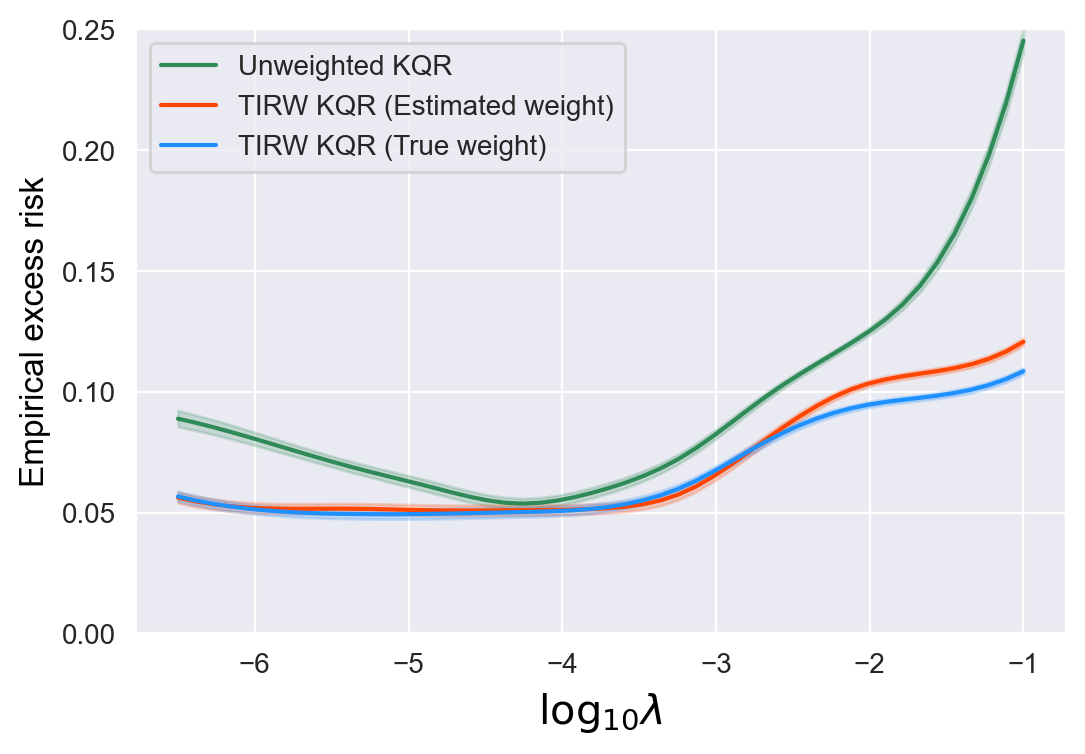}
\label{KQR_dimension3_uniformbounded_Empirical_smooth_1}}
 \subfigure[$\tau=0.5$ and $r=1$]{
    \includegraphics[width=1.6in,  height=1.09in]{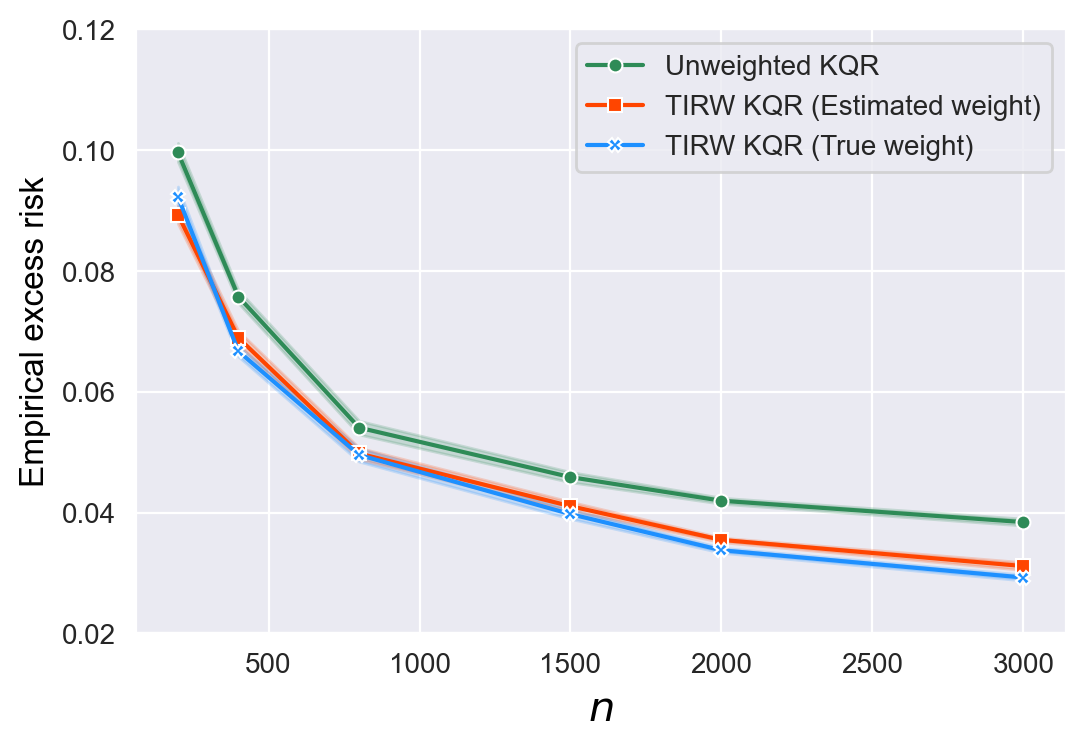}
\label{KQR_dimension3_uniformbounded_Empirical_7}}
    \subfigure[$\tau=0.5$ and $r=1$]{
	\includegraphics[width=1.6in,  height=1.09in]{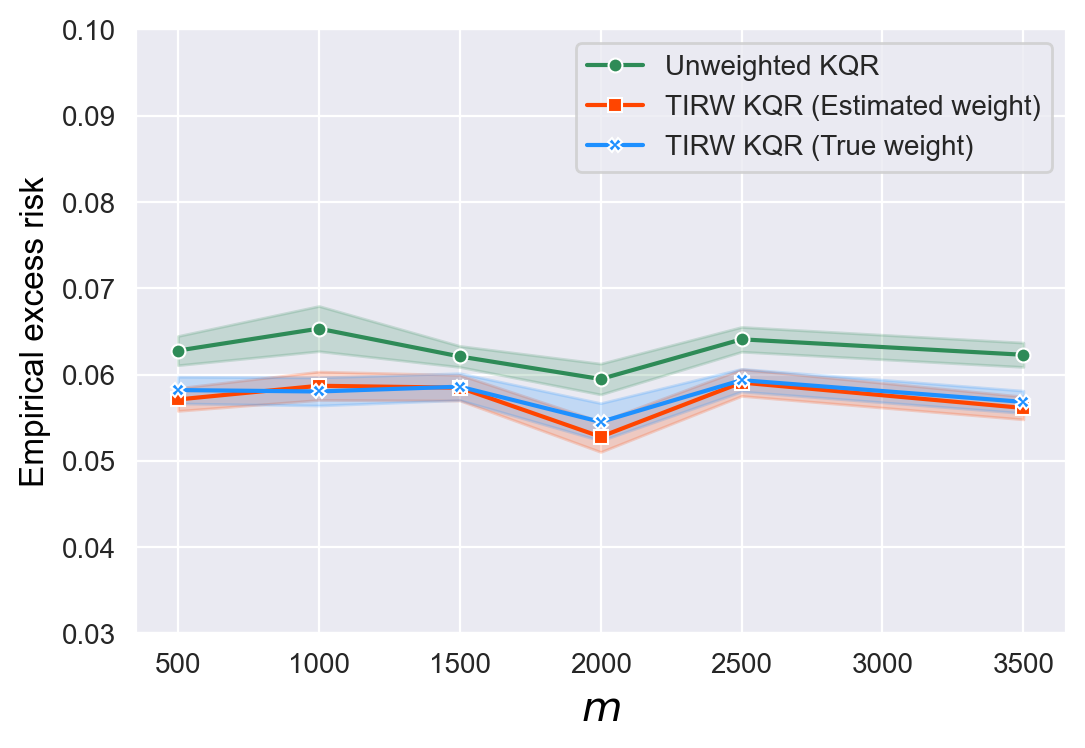}
\label{KQR_dimension3_uniformbounded_Empirical_13}}

\end{figure}

\begin{figure}[H]
\graphicspath{{bounded_KQR_3_d/}}
    \centering
    \setcounter{subfigure}{0}
\renewcommand{\thesubfigure}{(4\alph{subfigure})}
\subfigure[$\tau=0.5$ and $r=0$]{
    \includegraphics[width=1.6in,  height=1.09in]{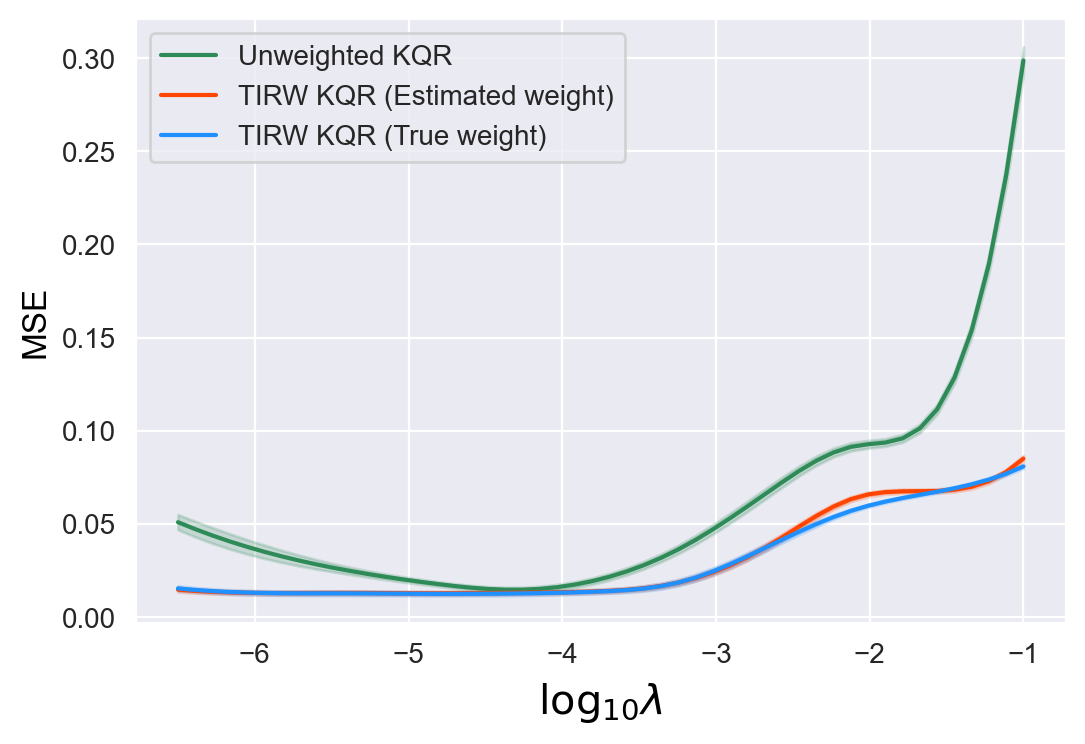}
    \label{KQR_dimension3_uniformbounded_MSE_smooth_2}}
    \subfigure[$\tau=0.5$ and $r=0$]{
	\includegraphics[width=1.6in,  height=1.09in]{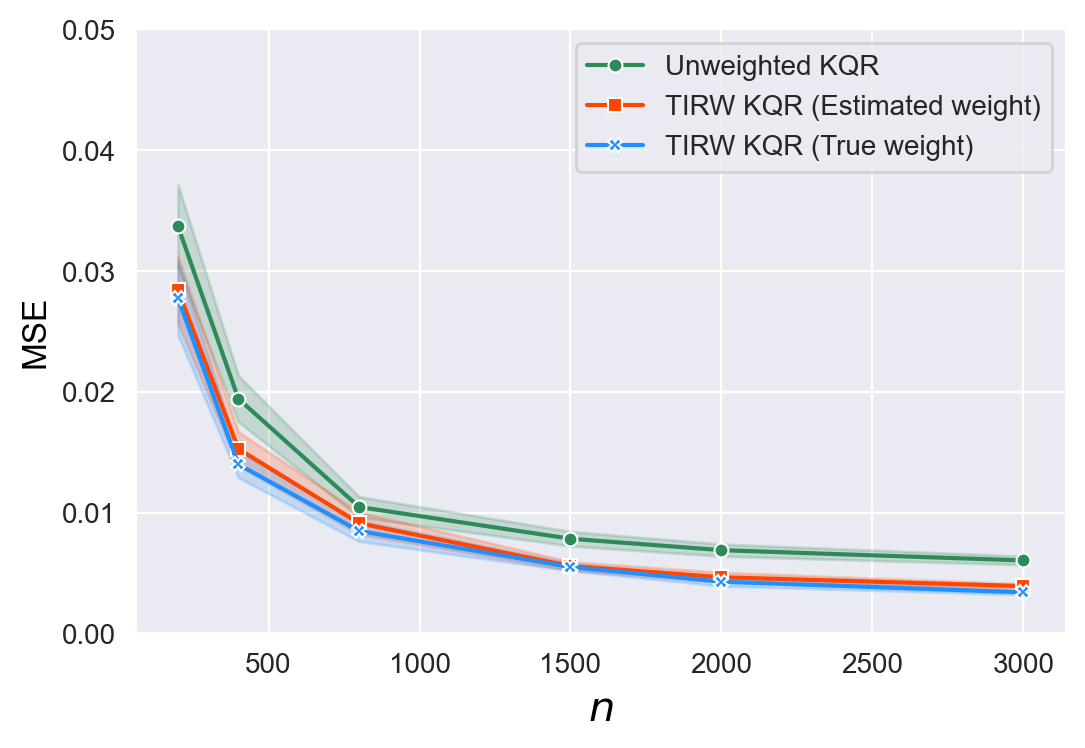}
\label{KQR_dimension3_uniformbounded_MSE_8}}
   \subfigure[$\tau=0.5$ and $r=0$]{
    \includegraphics[width=1.6in,  height=1.09in]{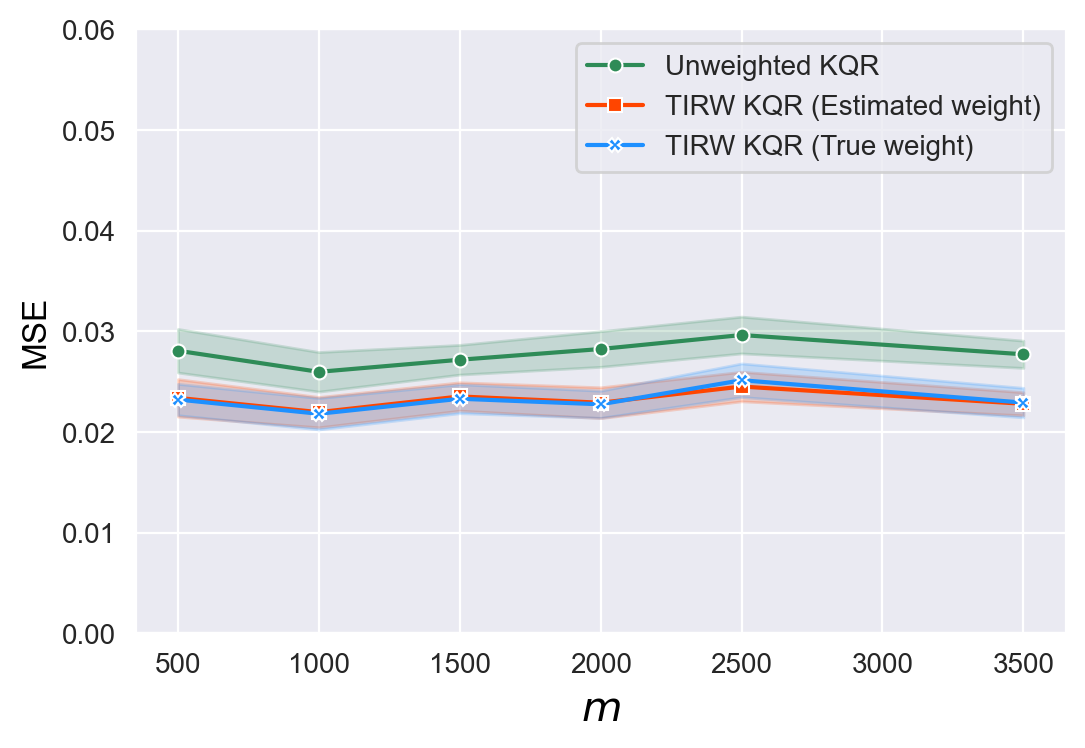}
    \label{KQR_dimension3_uniformbounded_MSE_14}}
    \subfigure[$\tau=0.5$ and $r=0$]{
	\includegraphics[width=1.6in,  height=1.09in]{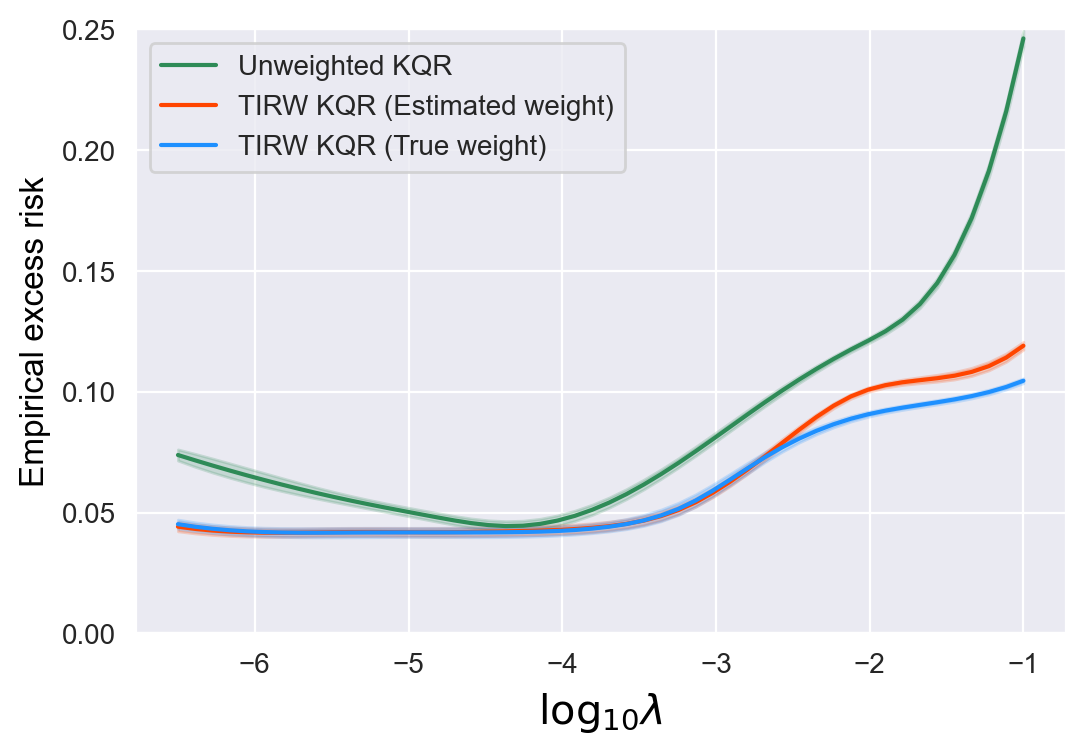}
\label{KQR_dimension3_uniformbounded_Empirical_smooth_2}}
 \subfigure[$\tau=0.5$ and $r=0$]{
    \includegraphics[width=1.6in,  height=1.09in]{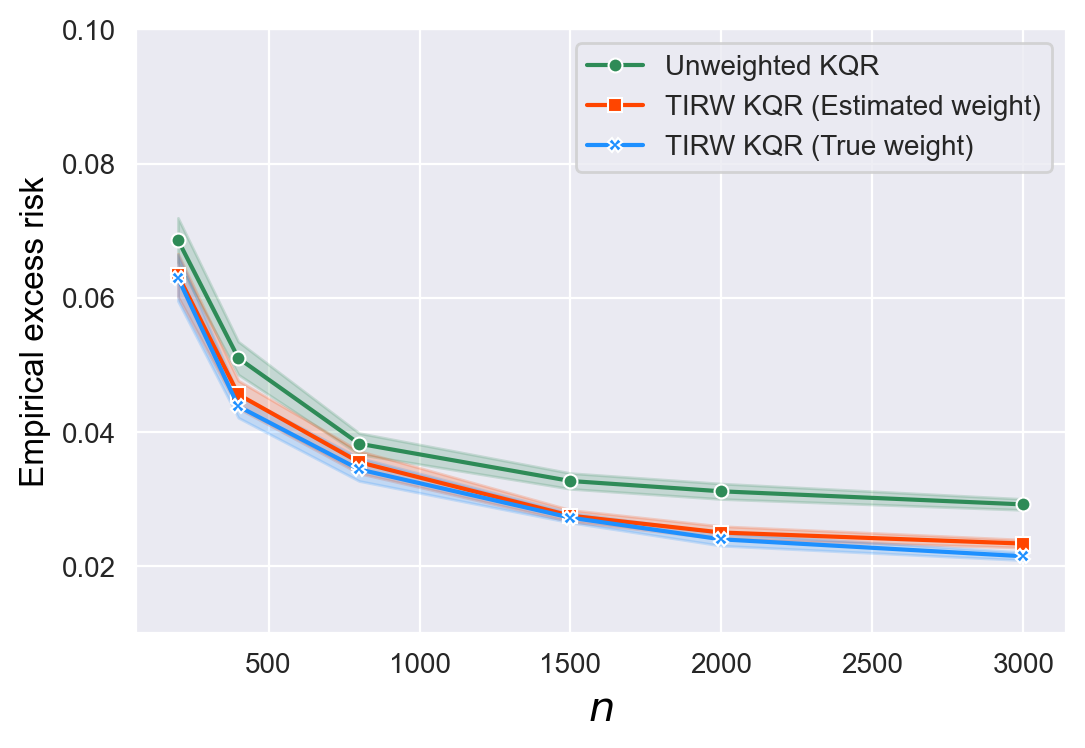}
\label{KQR_dimension3_uniformbounded_Empirical_8}}
    \subfigure[$\tau=0.5$ and $r=0$]{
	\includegraphics[width=1.6in,  height=1.09in]{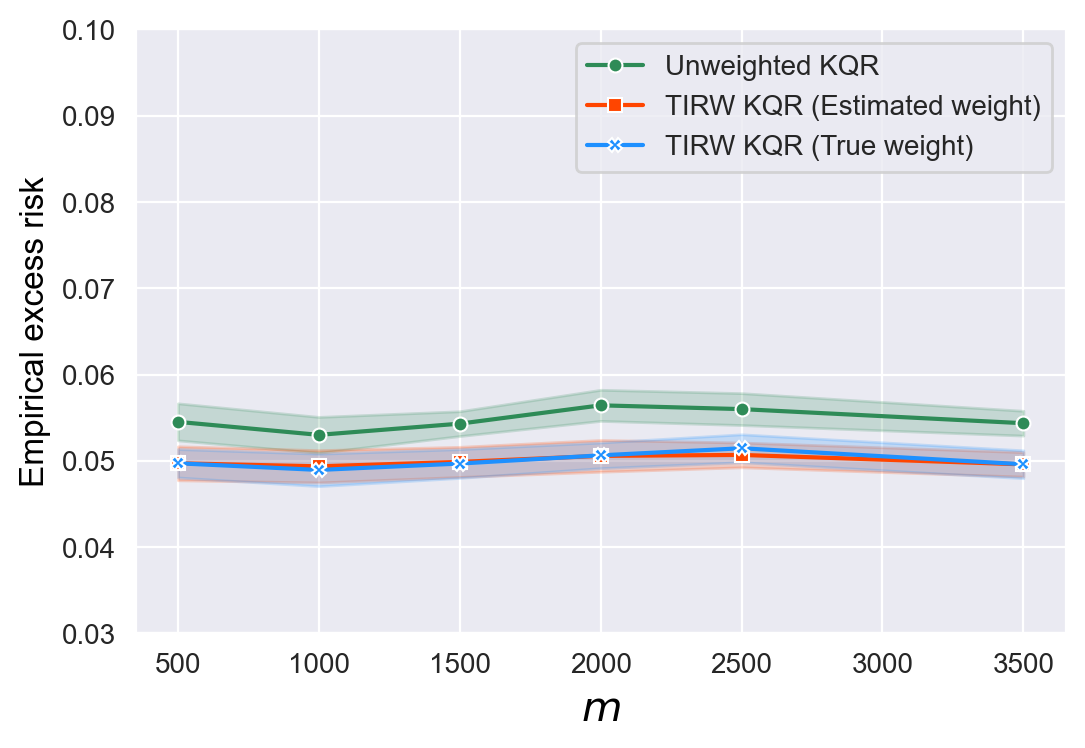}
 \label{KQR_dimension3_uniformbounded_Empirical_14}}
 \setcounter{subfigure}{0}
\renewcommand{\thesubfigure}{(5\alph{subfigure})}
  \subfigure[$\tau=0.7$ and $r=1$]{
    \includegraphics[width=1.6in,  height=1.09in]{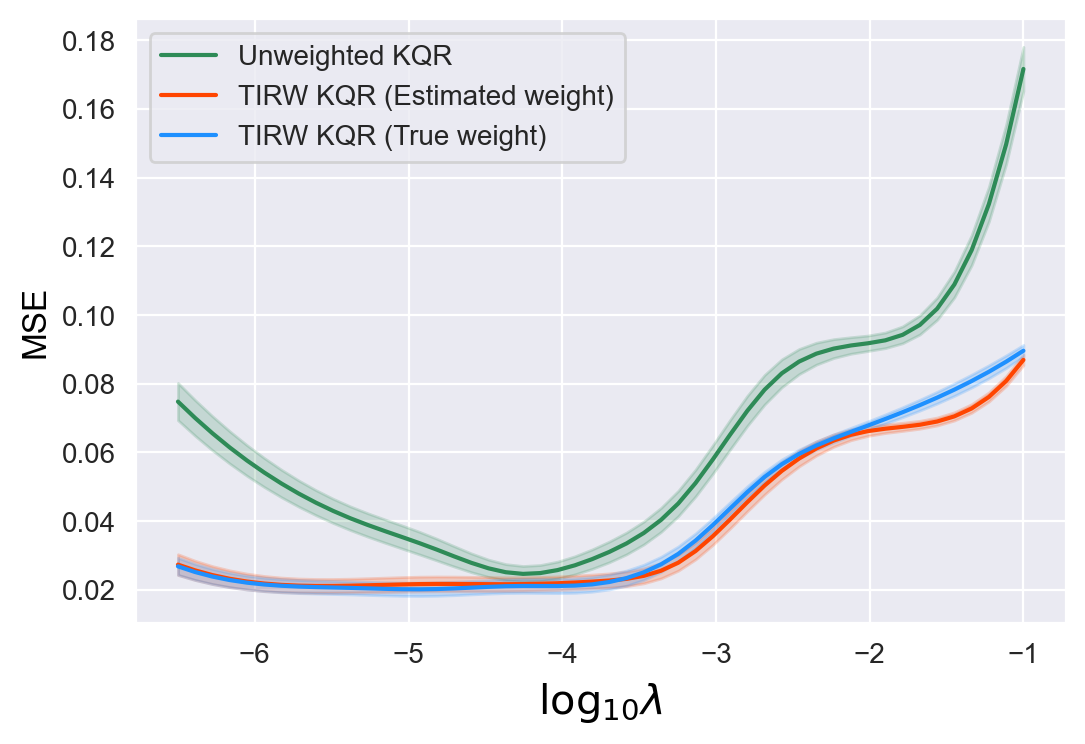}
    \label{KQR_dimension3_uniformbounded_MSE_smooth_5}}
    \subfigure[$\tau=0.7$ and $r=1$]{
	\includegraphics[width=1.6in,  height=1.09in]{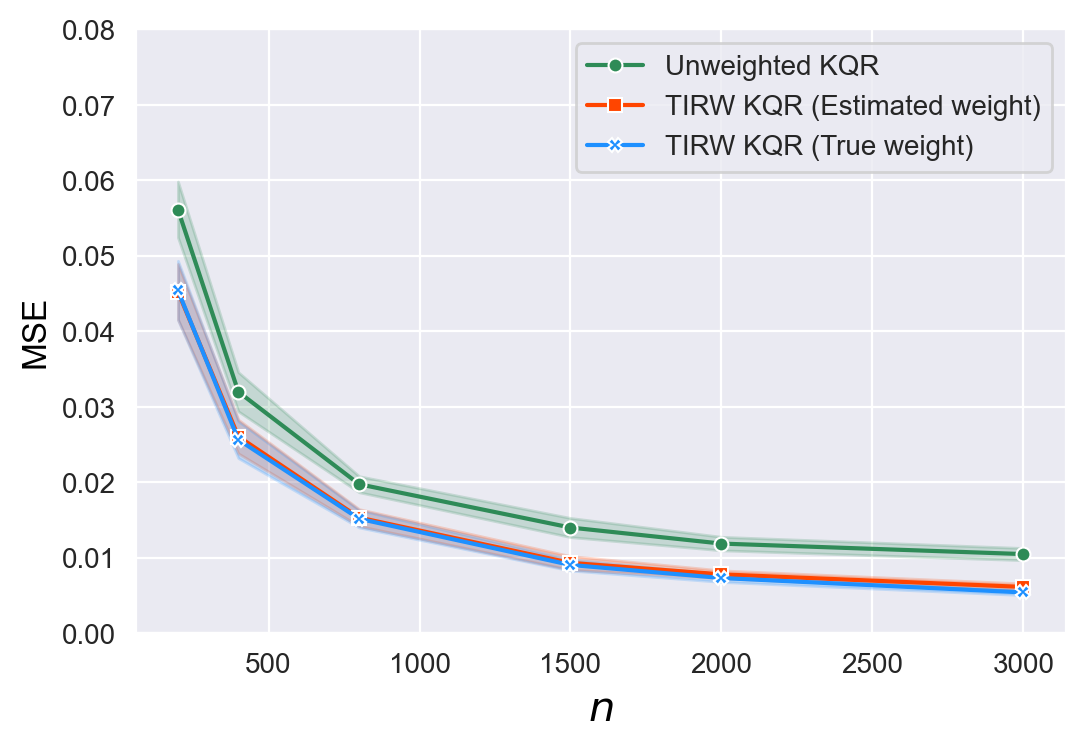}
\label{KQR_dimension3_uniformbounded_MSE_11}}
   \subfigure[$\tau=0.7$ and $r=1$]{
    \includegraphics[width=1.6in,  height=1.09in]{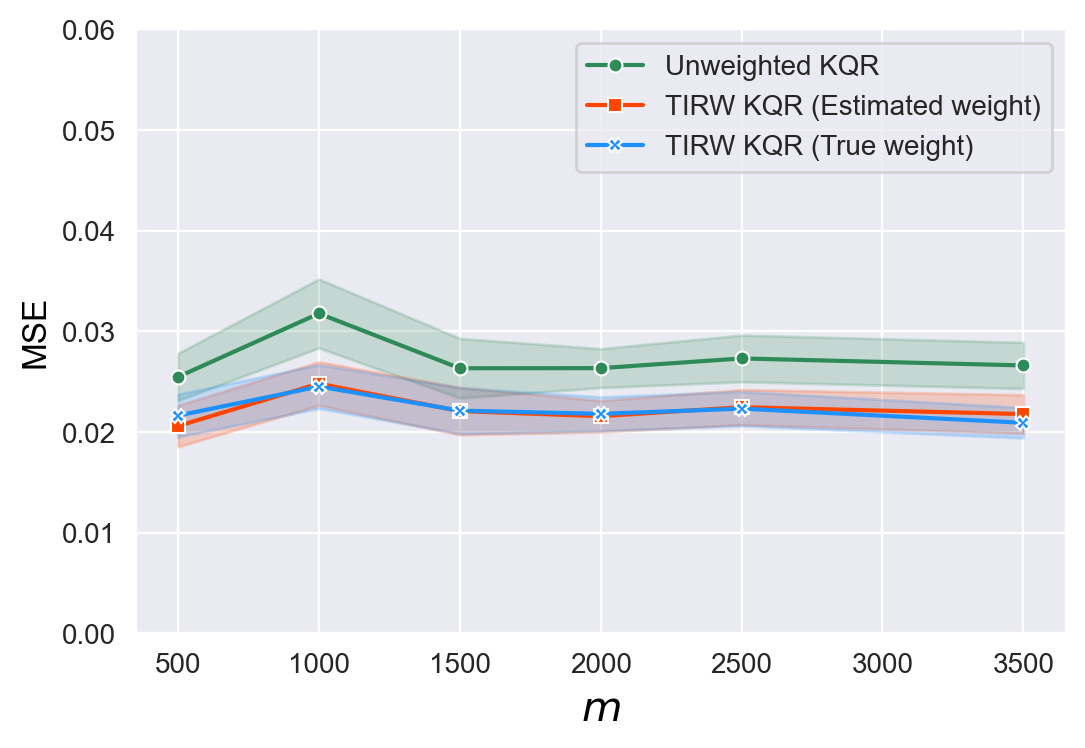}
    \label{KQR_dimension3_uniformbounded_MSE_17}}
    \subfigure[$\tau=0.7$ and $r=1$]{
	\includegraphics[width=1.6in,  height=1.09in]{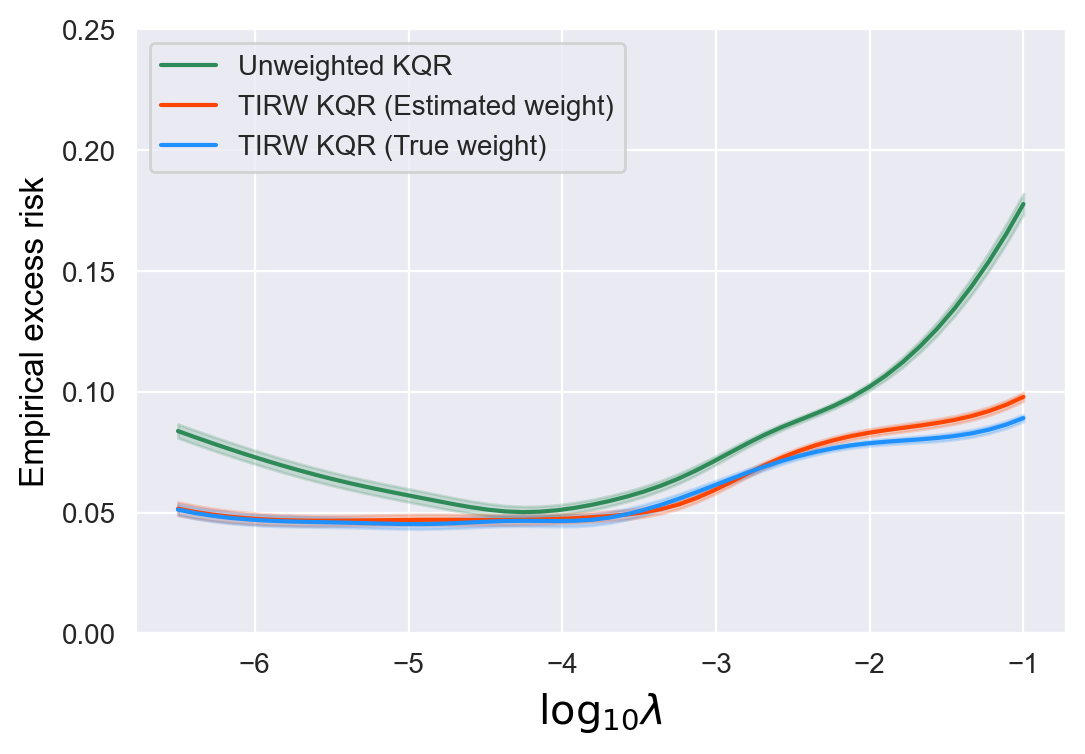}
\label{KQR_dimension3_uniformbounded_Empirical_smooth_5}}
 \subfigure[$\tau=0.7$ and $r=1$]{
    \includegraphics[width=1.6in,  height=1.09in]{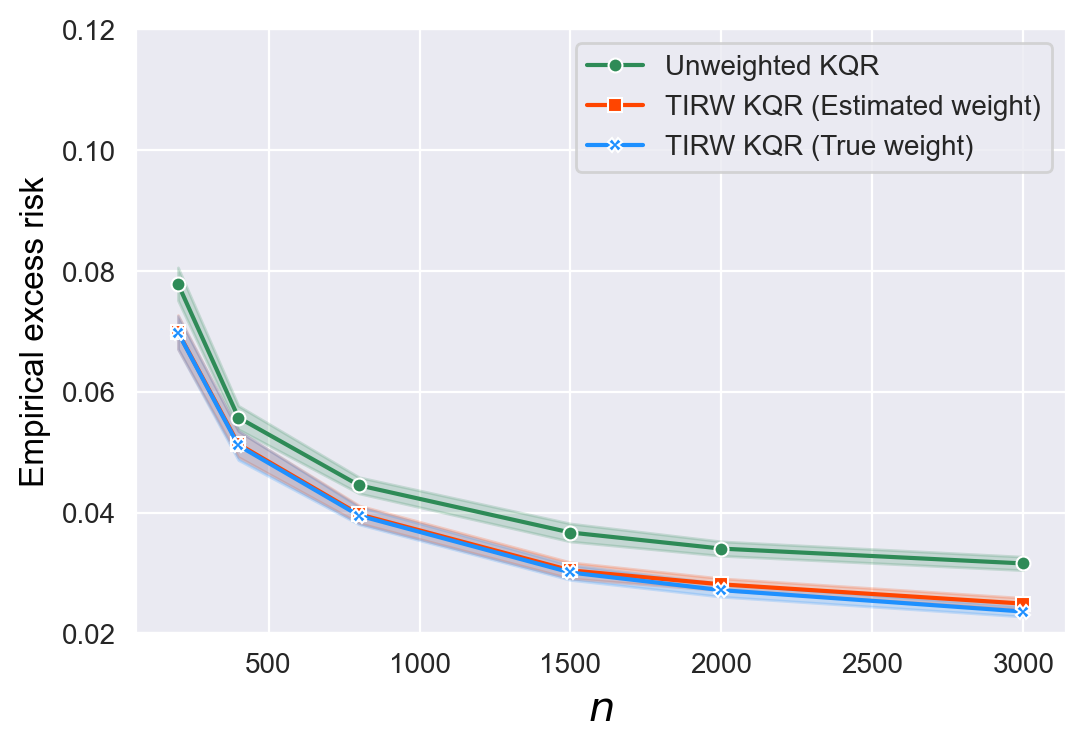}
\label{KQR_dimension3_uniformbounded_Empirical_11}}
    \subfigure[$\tau=0.7$ and $r=1$]{
	\includegraphics[width=1.6in,  height=1.09in]{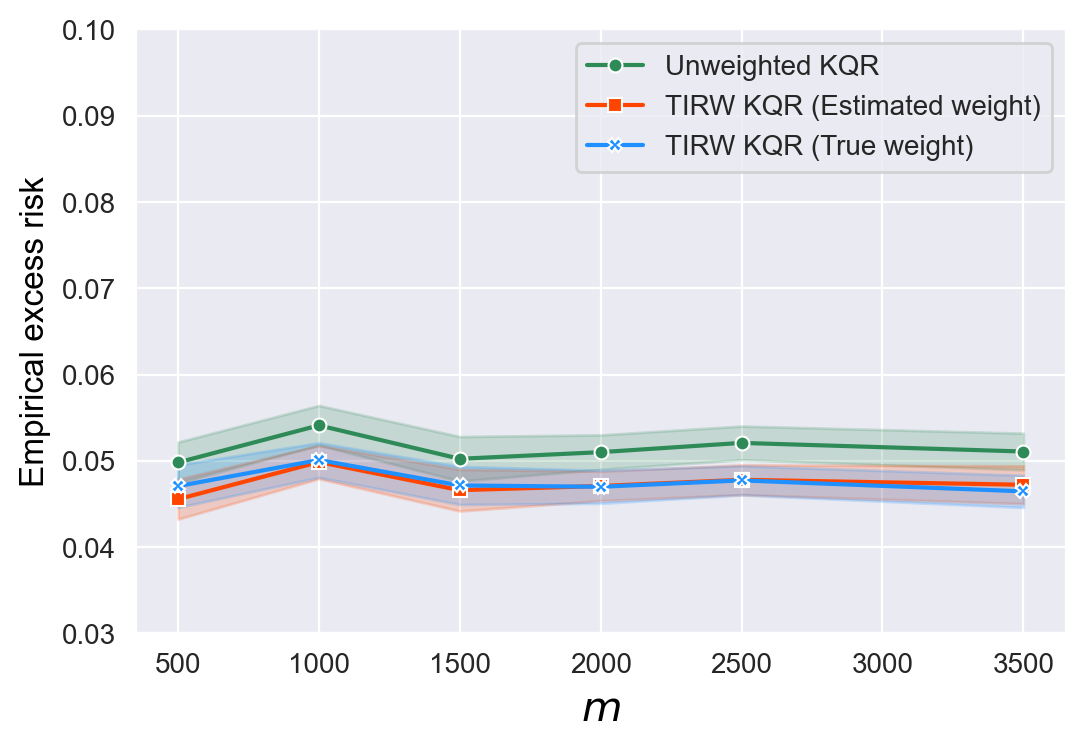}
 \label{KQR_dimension3_uniformbounded_Empirical_17}}
 \setcounter{subfigure}{0}
\renewcommand{\thesubfigure}{(6\alph{subfigure})}
  \subfigure[$\tau=0.7$ and $r=0$]{
    \includegraphics[width=1.6in,  height=1.09in]{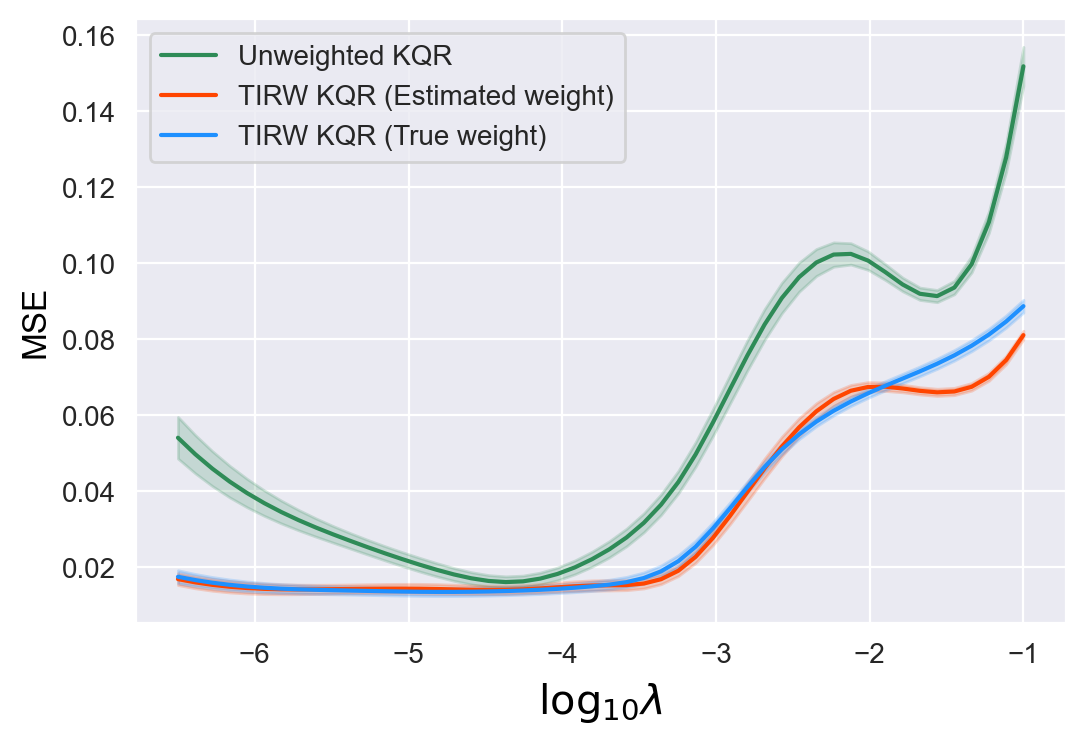}
    \label{KQR_dimension3_uniformbounded_MSE_smooth_6}}
    \subfigure[$\tau=0.7$ and $r=0$]{
	\includegraphics[width=1.6in,  height=1.09in]{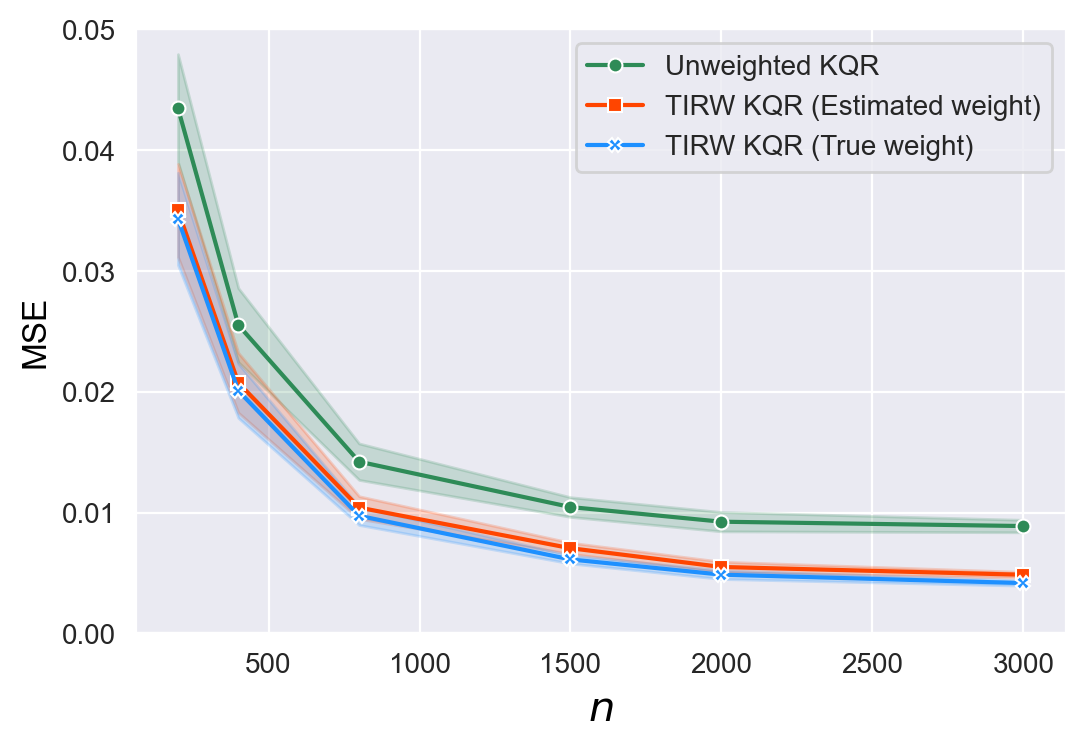}
\label{KQR_dimension3_uniformbounded_MSE_12}}
   \subfigure[$\tau=0.7$ and $r=0$]{
    \includegraphics[width=1.6in,  height=1.09in]{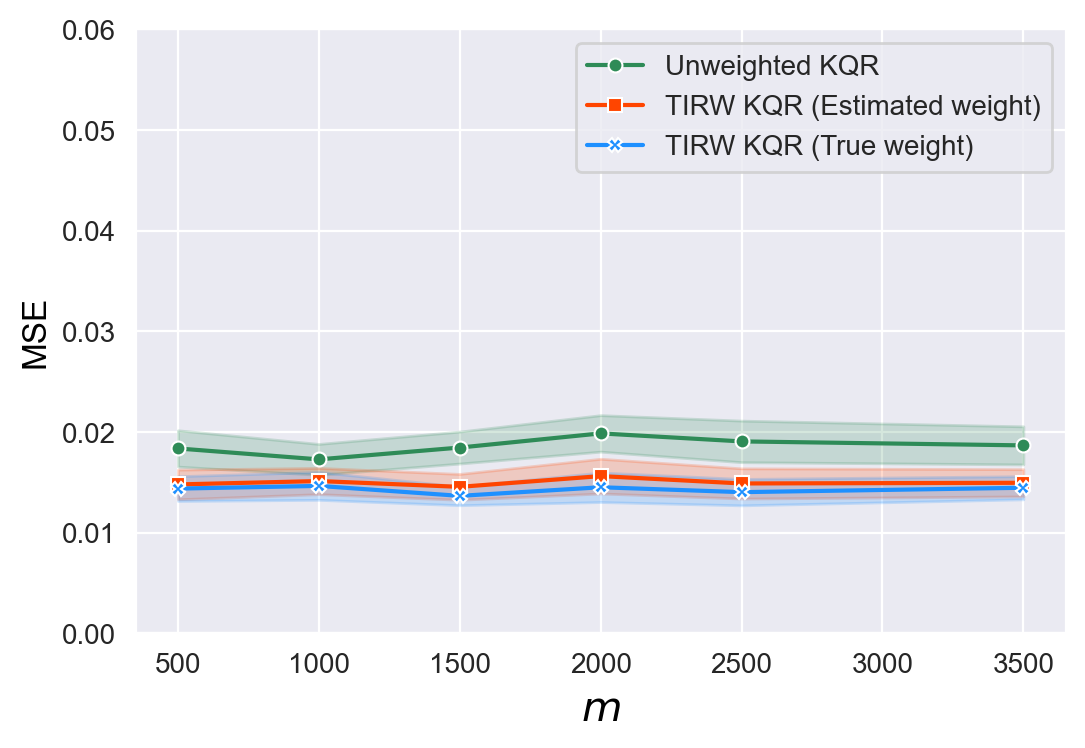}
    \label{KQR_dimension3_uniformbounded_MSE_18}}
    \subfigure[$\tau=0.7$ and $r=0$]{
	\includegraphics[width=1.6in,  height=1.09in]{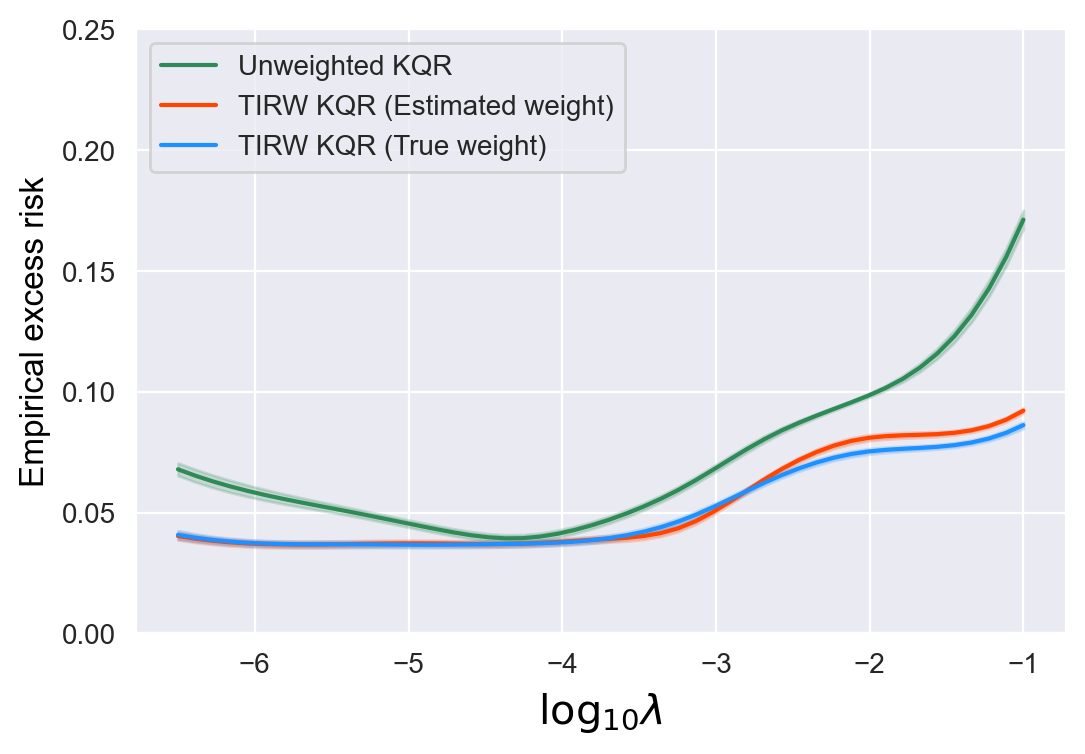}
\label{KQR_dimension3_uniformbounded_Empirical_smooth_6}}
 \subfigure[$\tau=0.7$ and $r=0$]{
    \includegraphics[width=1.6in,  height=1.09in]{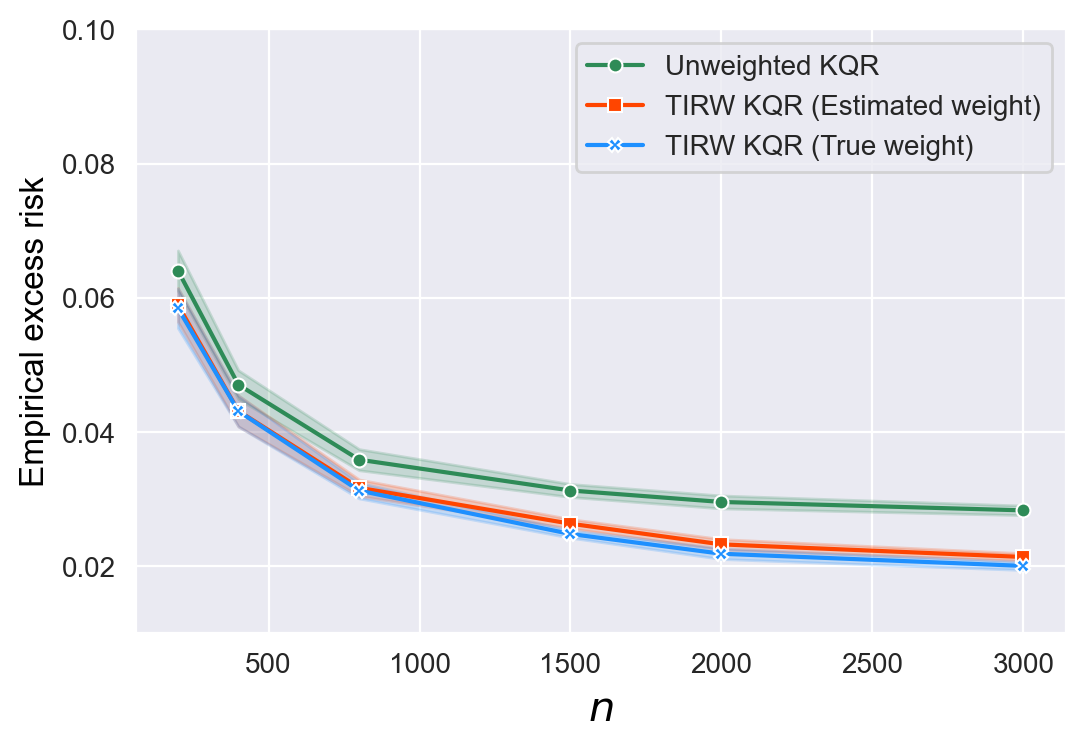}
\label{KQR_dimension3_uniformbounded_Empirical_12}}
    \subfigure[$\tau=0.7$ and $r=0$]{
	\includegraphics[width=1.6in,  height=1.09in]{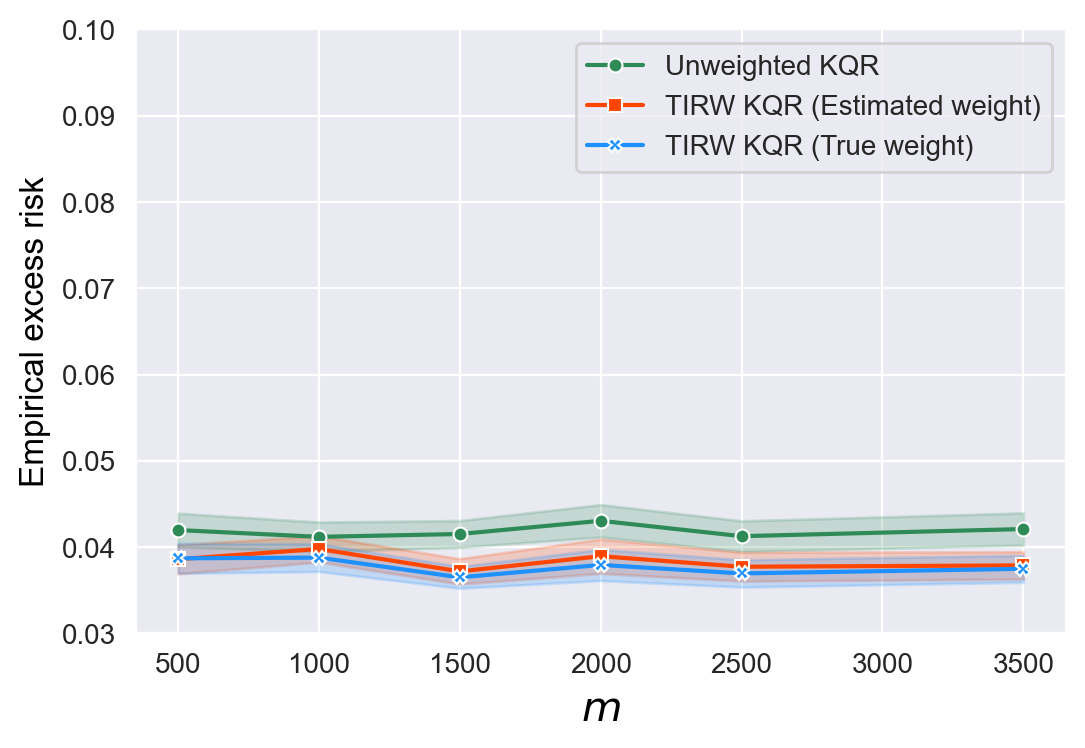}
 \label{KQR_dimension3_uniformbounded_Empirical_18}}
 \caption{\footnotesize{Average MSE  and empirical excess risk for unweighted KQR, TIRW KQR with true weight and estimated weight, respectively (in the left panel, the curves are plotted with respect to $\log_{10} \lambda$ with $n=500, m=1000$; in the middle panel, the curves are plotted with respect to $n$ with fixed $m = 1000,\lambda=10^{-4}$; in the right panel, the curves are plotted with respect to $m$ with fixed $n=500,\lambda=10^{-4}$)}}
\label{KQR_dimension3_bounded}
\end{figure}

\subsubsection{Moment bounded case in Example S4}\label{sec:a.2.4}

\graphicspath{{unbounded_KQR_3_d/}}
\setcounter{subfigure}{0}
\renewcommand{\thesubfigure}{(1\alph{subfigure})}
\begin{figure}[H]
    \centering
    \subfigure[$\tau=0.3$ and $r=1$]{
    \includegraphics[width=1.6in,  height=1.09in]{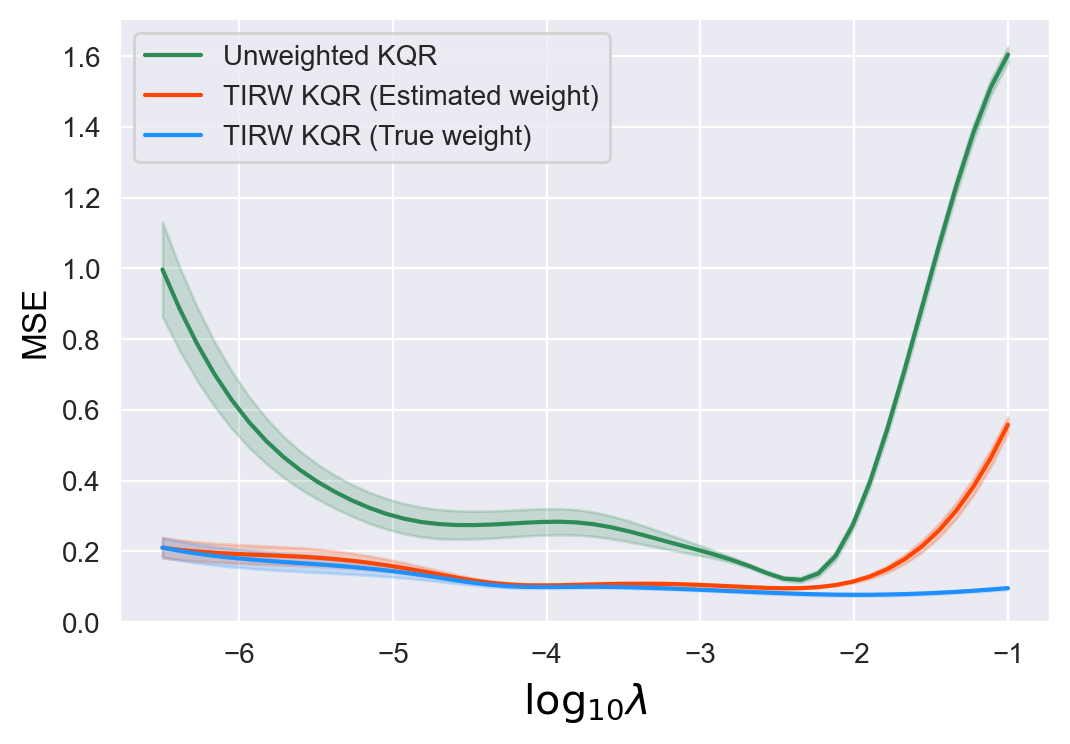}
    \label{KQR_dimension3_MSE_smooth_3}}
    \subfigure[$\tau=0.3$ and $r=1$]{
	\includegraphics[width=1.6in,  height=1.09in]{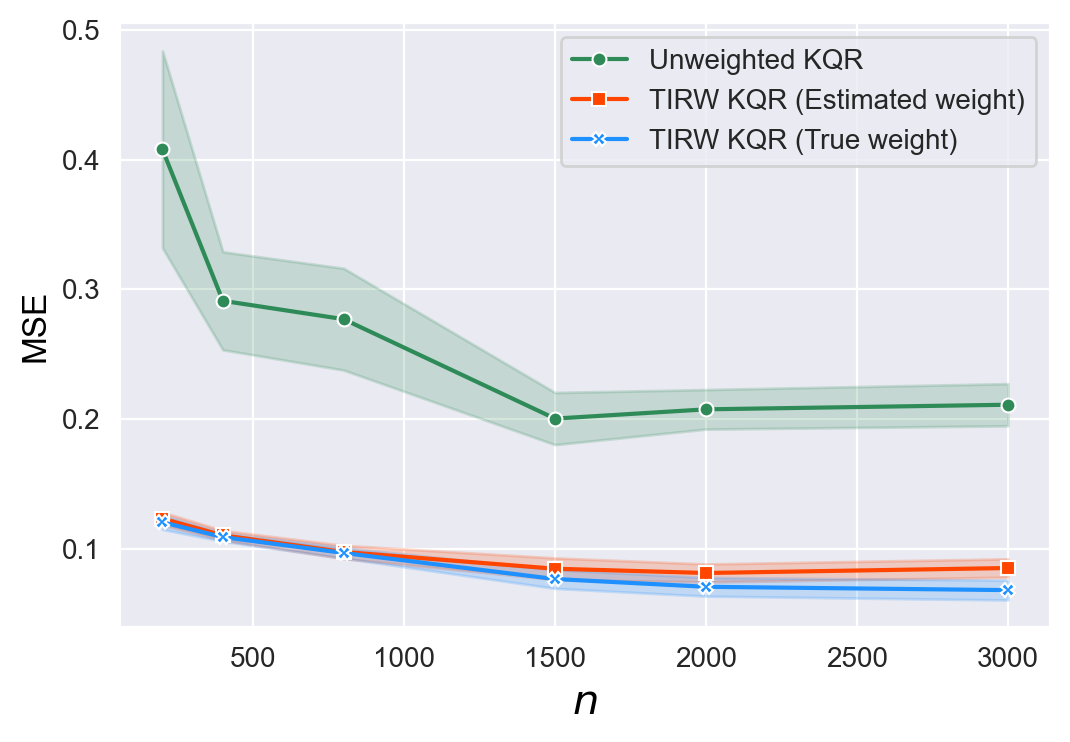}
\label{KQR_dimension3_MSE_9}}
   \subfigure[$\tau=0.3$ and $r=1$]{
    \includegraphics[width=1.6in,  height=1.09in]{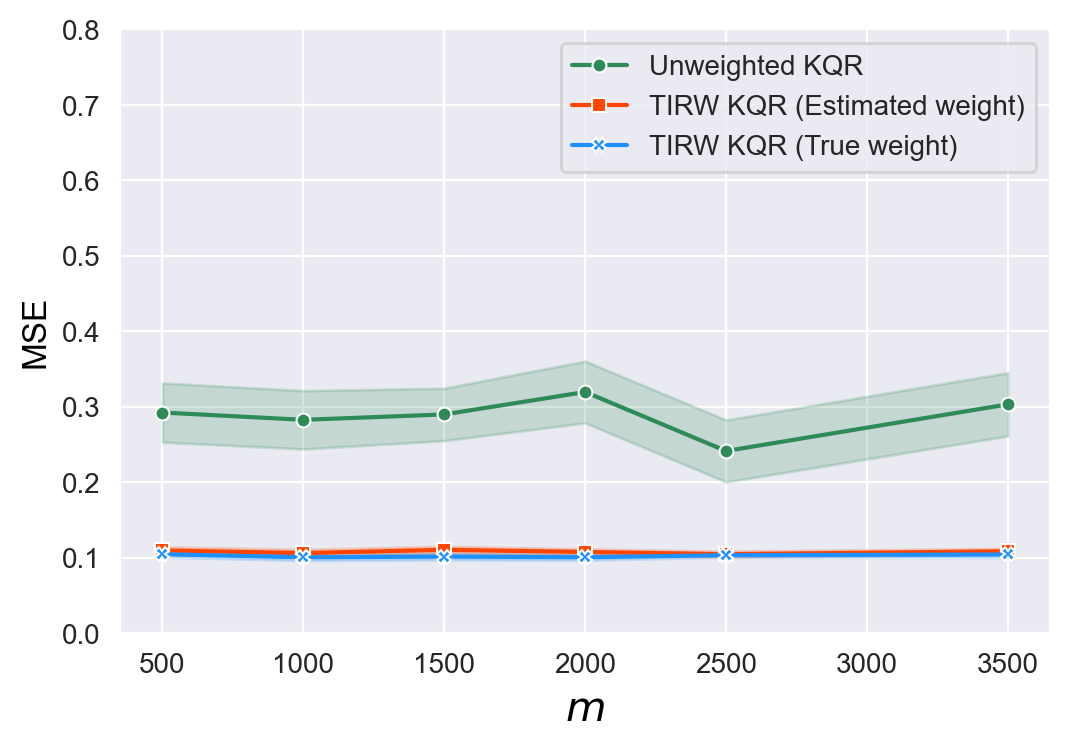}
    \label{KQR_dimension3_MSE_15}}
   \subfigure[$\tau=0.3$ and $r=1$]{
	\includegraphics[width=1.6in,  height=1.09in]{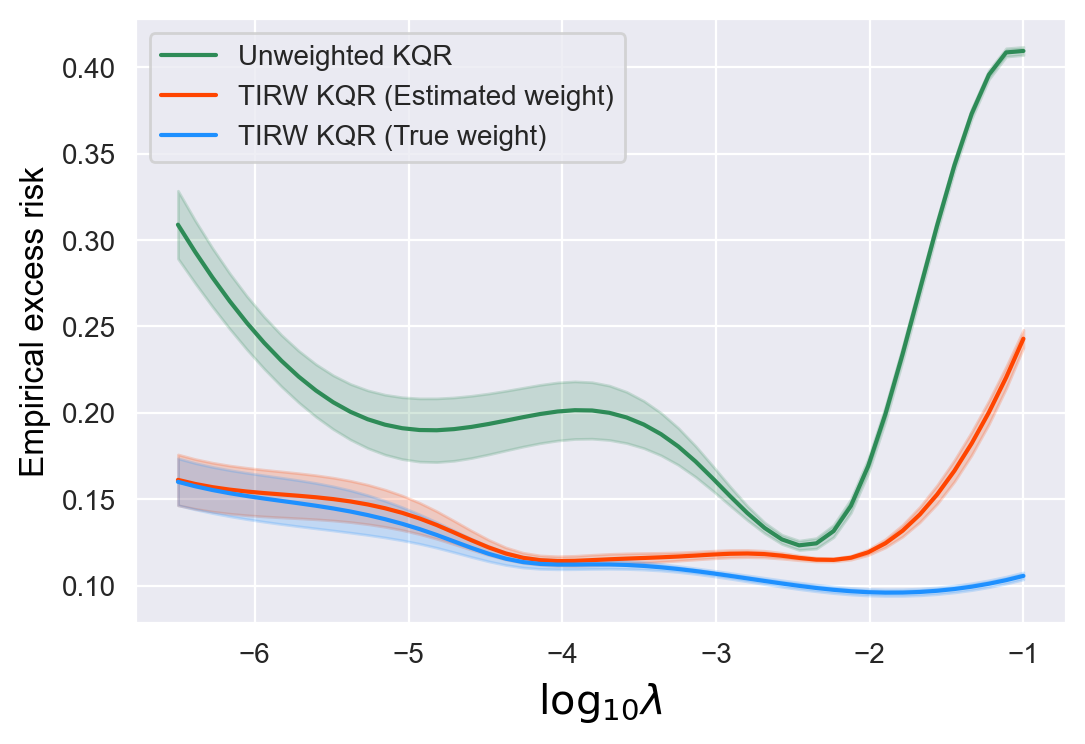}
        \label{KQR_dimension3_Empirical_smooth_3}}
 \subfigure[$\tau=0.3$ and $r=1$]{
    \includegraphics[width=1.6in,  height=1.09in]{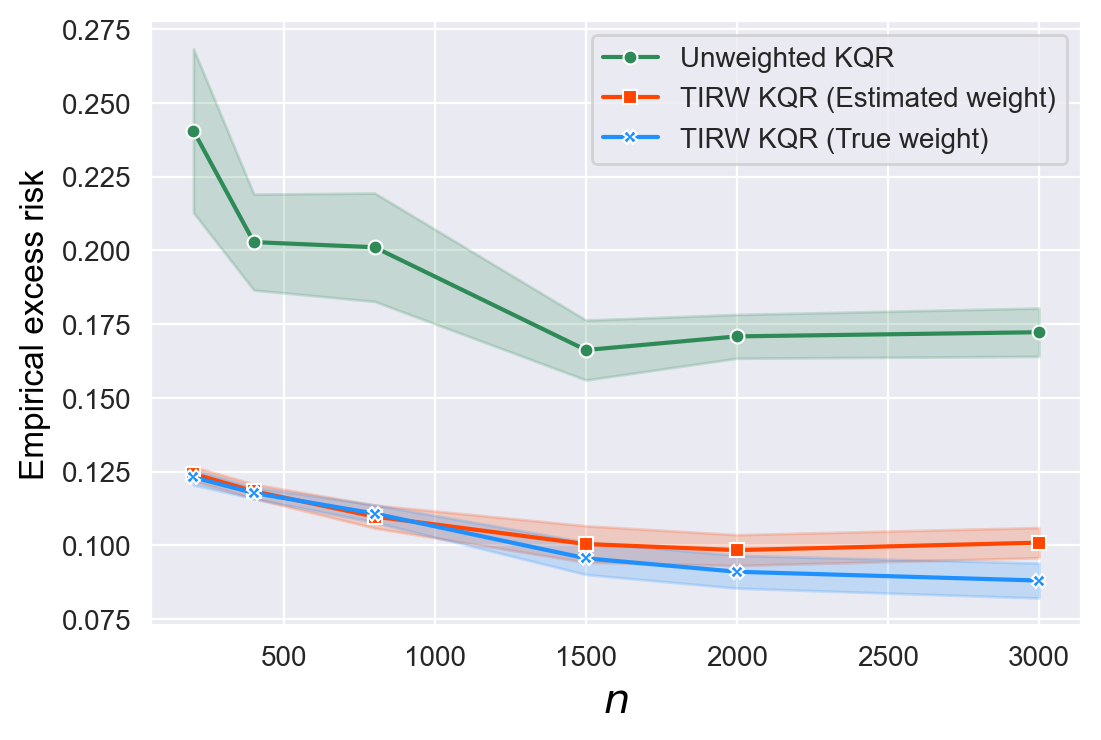}
\label{KQR_dimension3_Empirical_9}}
    \subfigure[$\tau=0.3$ and $r=1$]{
	\includegraphics[width=1.6in,  height=1.09in]{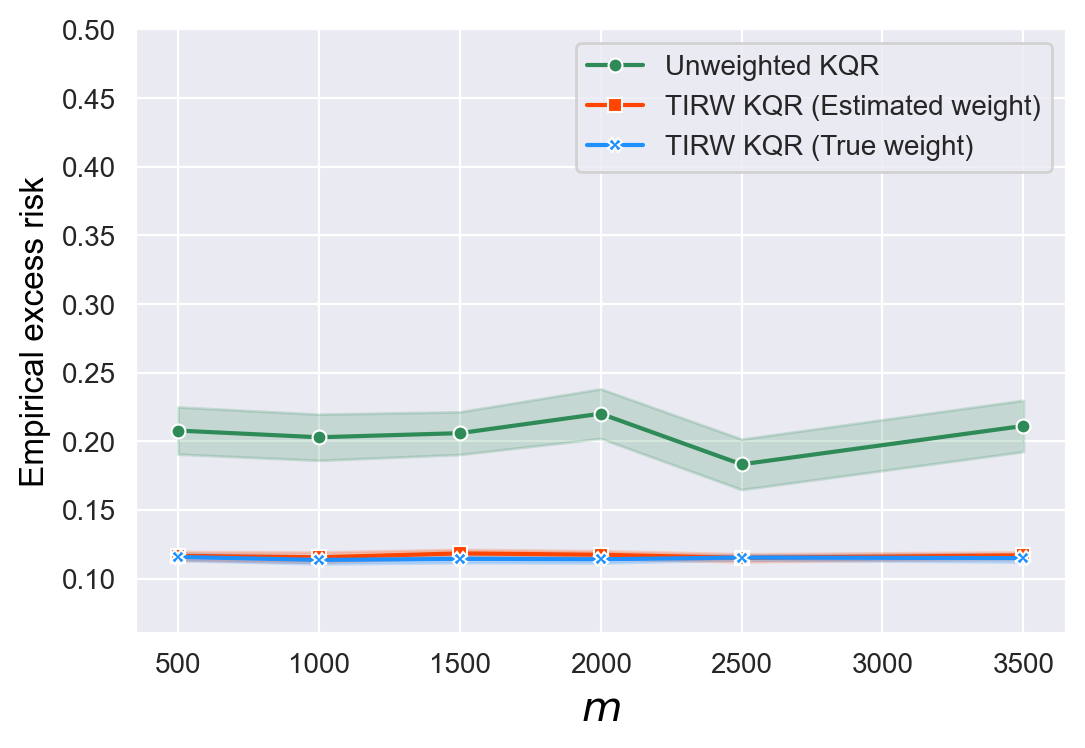}
 \label{KQR_dimension3_Empirical_15}}
 \setcounter{subfigure}{0}
\renewcommand{\thesubfigure}{(2\alph{subfigure})}
  \subfigure[$\tau=0.3$ and $r=0$]{
    \includegraphics[width=1.6in,  height=1.09in]{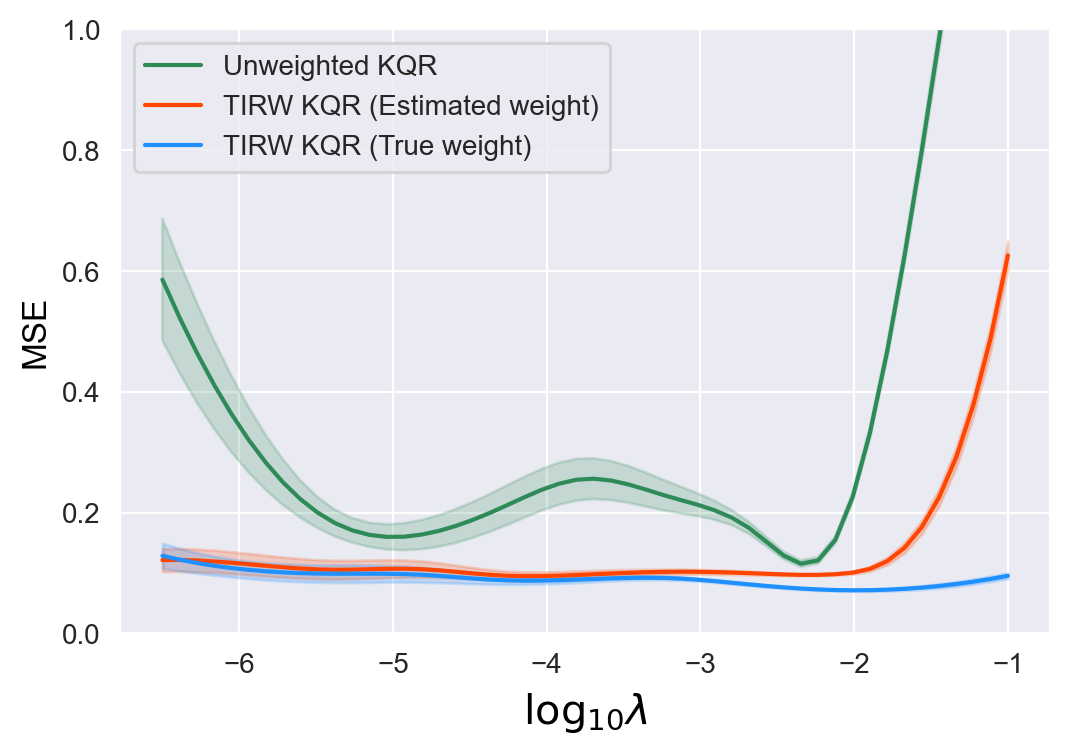}
    \label{KQR_dimension3_MSE_smooth_4}}
    \subfigure[$\tau=0.3$ and $r=0$]{
	\includegraphics[width=1.6in,  height=1.09in]{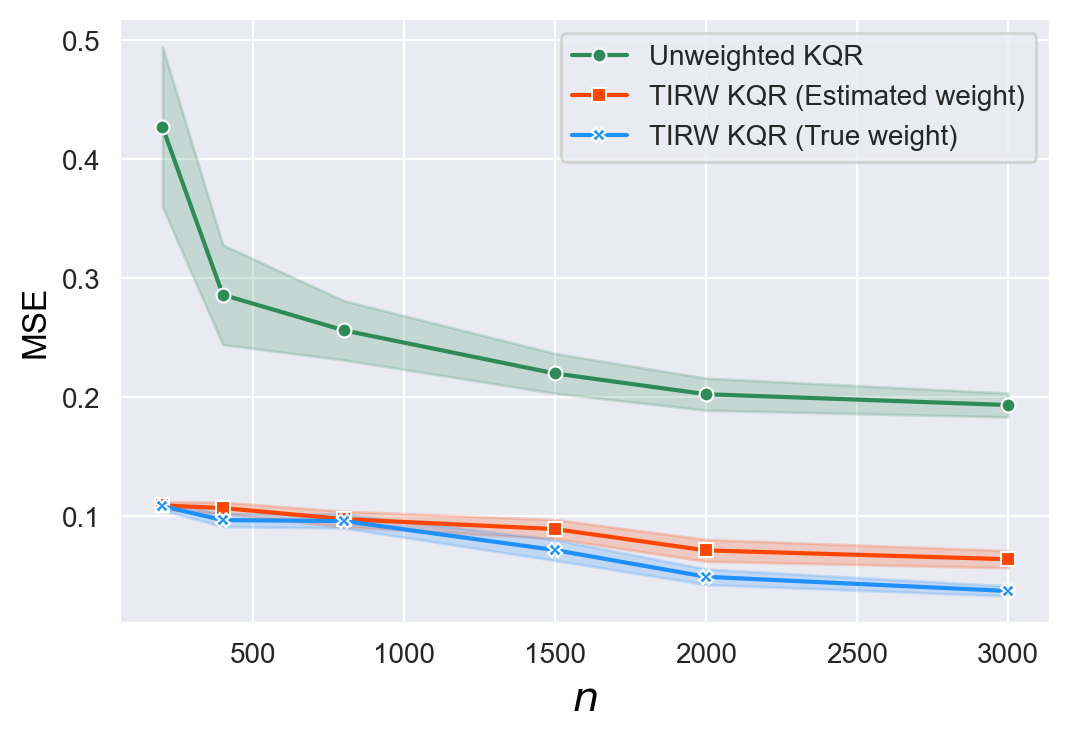}
\label{KQR_dimension3_MSE_10}}
   \subfigure[$\tau=0.3$ and $r=0$]{
    \includegraphics[width=1.6in,  height=1.09in]{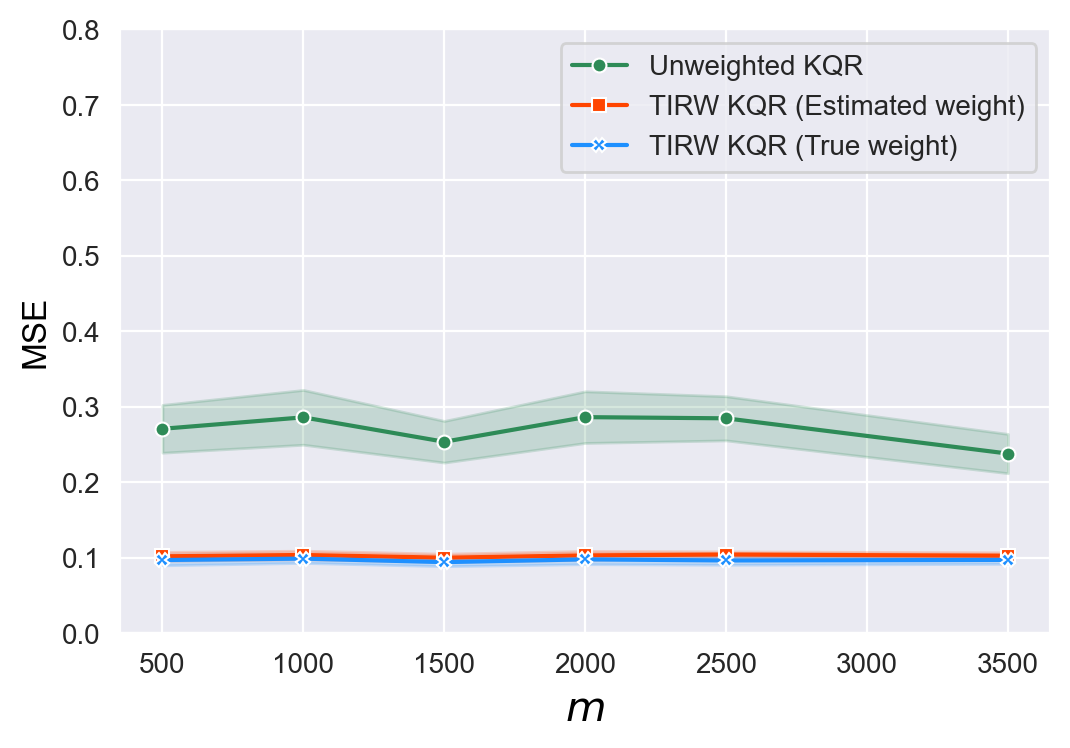}
    \label{KQR_dimension3_MSE_16}}
    \subfigure[$\tau=0.3$ and $r=0$]{
	\includegraphics[width=1.6in,  height=1.09in]{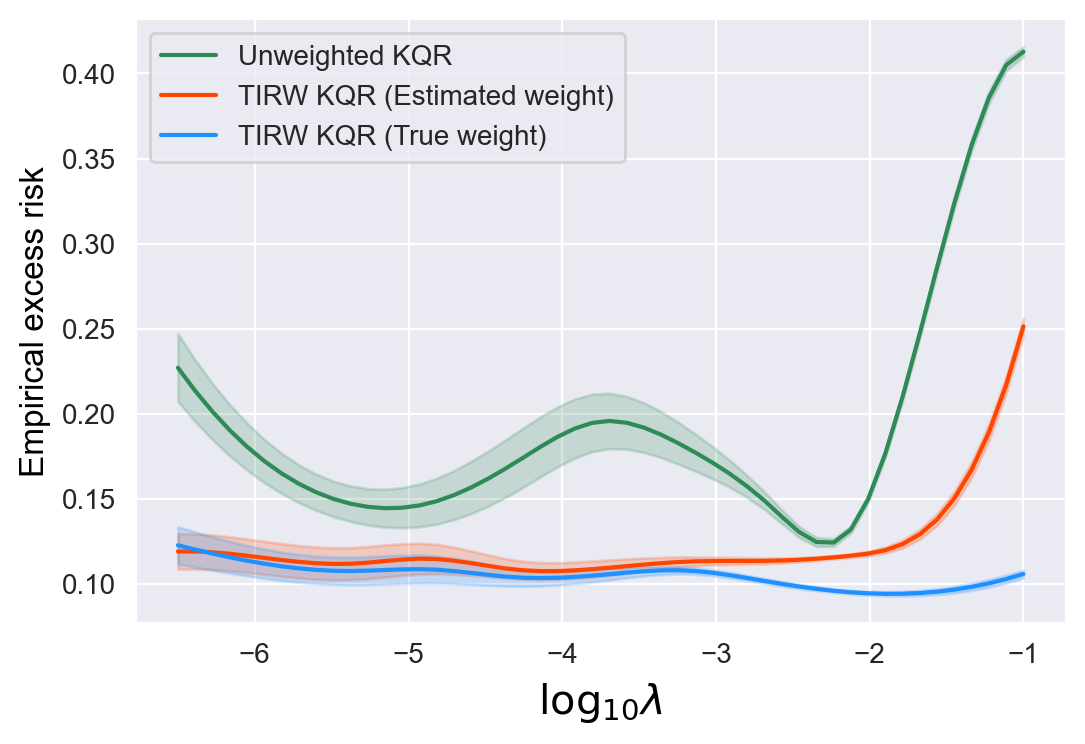}
\label{KQR_dimension3_Empirical_smooth_4}}
 \subfigure[$\tau=0.3$ and $r=0$]{
    \includegraphics[width=1.6in,  height=1.09in]{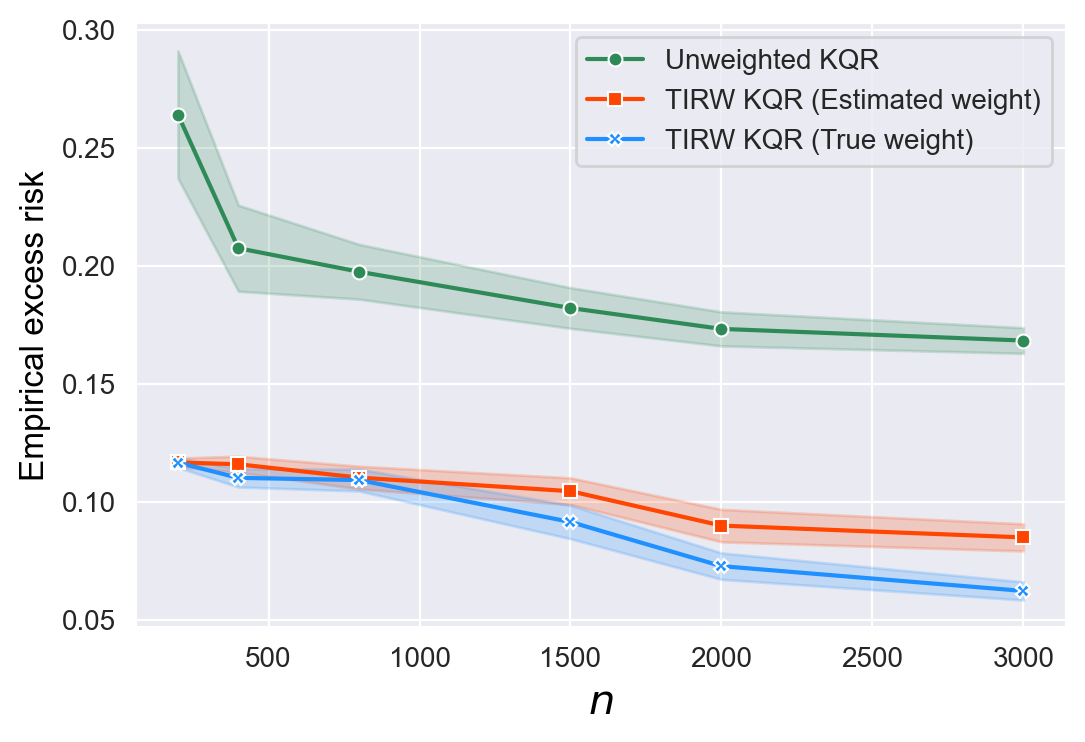}
\label{KQR_dimension3_Empirical_10}}
    \subfigure[$\tau=0.3$ and $r=0$]{
	\includegraphics[width=1.6in,  height=1.09in]{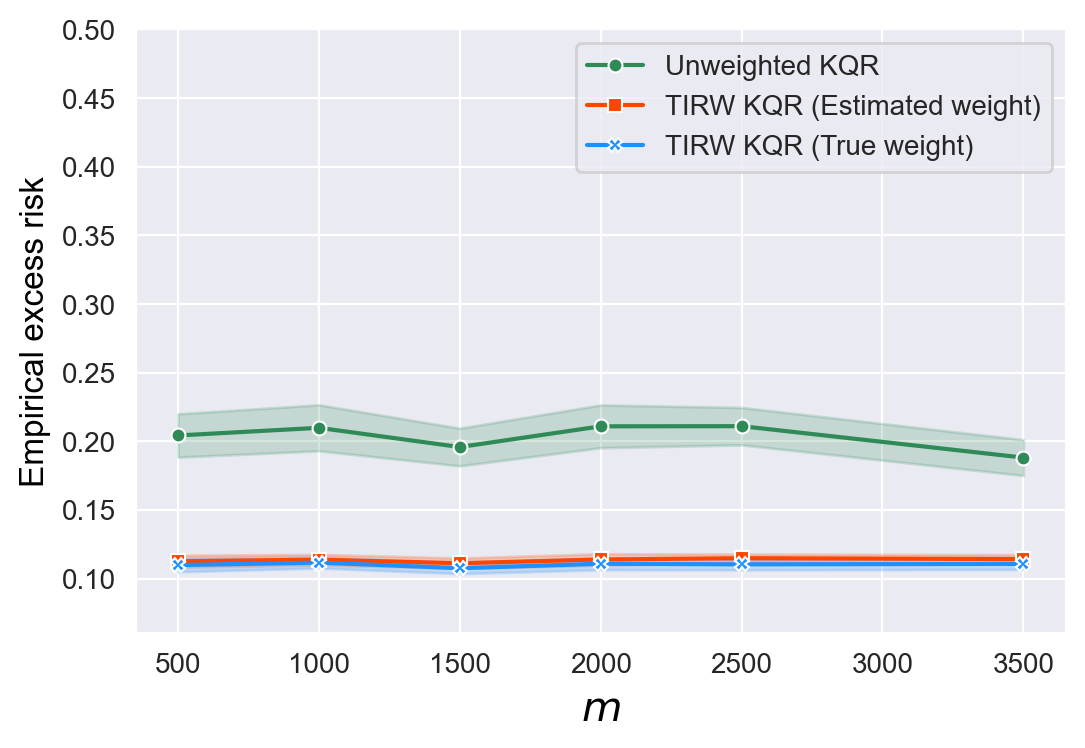}
 \label{KQR_dimension3_Empirical_16}}
 \setcounter{subfigure}{0}
\renewcommand{\thesubfigure}{(3\alph{subfigure})}
 \subfigure[$\tau=0.5$ and $r=1$]{
    \includegraphics[width=1.6in,  height=1.09in]{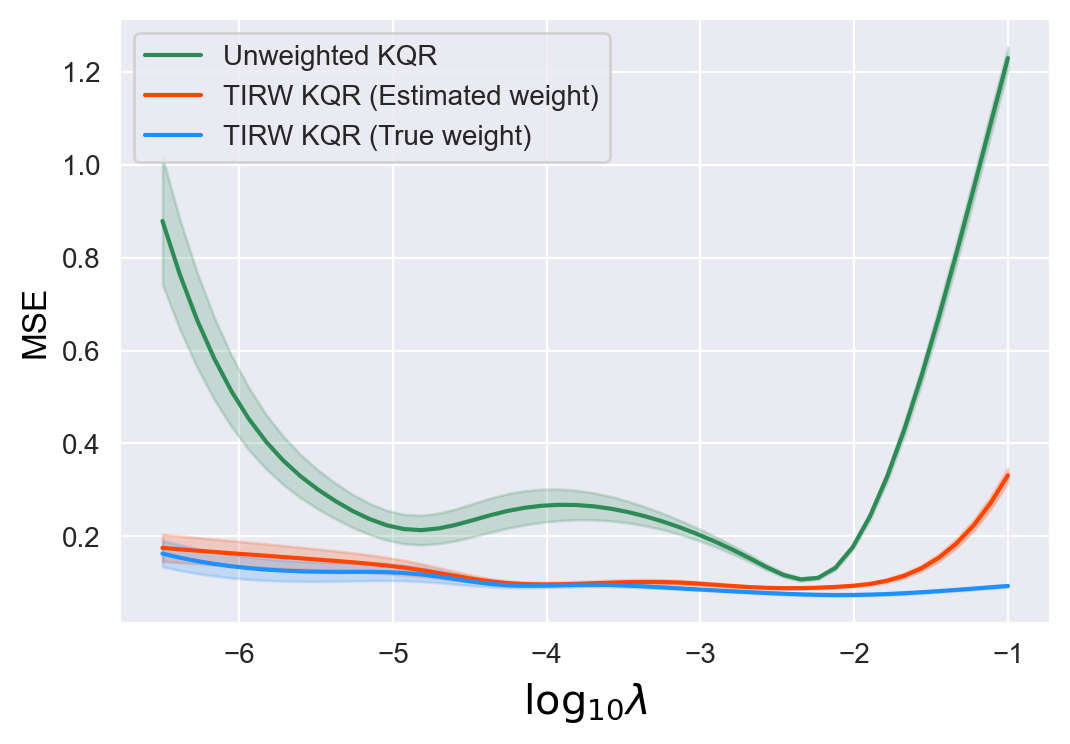}
    \label{KQR_dimension3_MSE_smooth_1}}
    \subfigure[$\tau=0.5$ and $r=1$]{
	\includegraphics[width=1.6in,  height=1.09in]{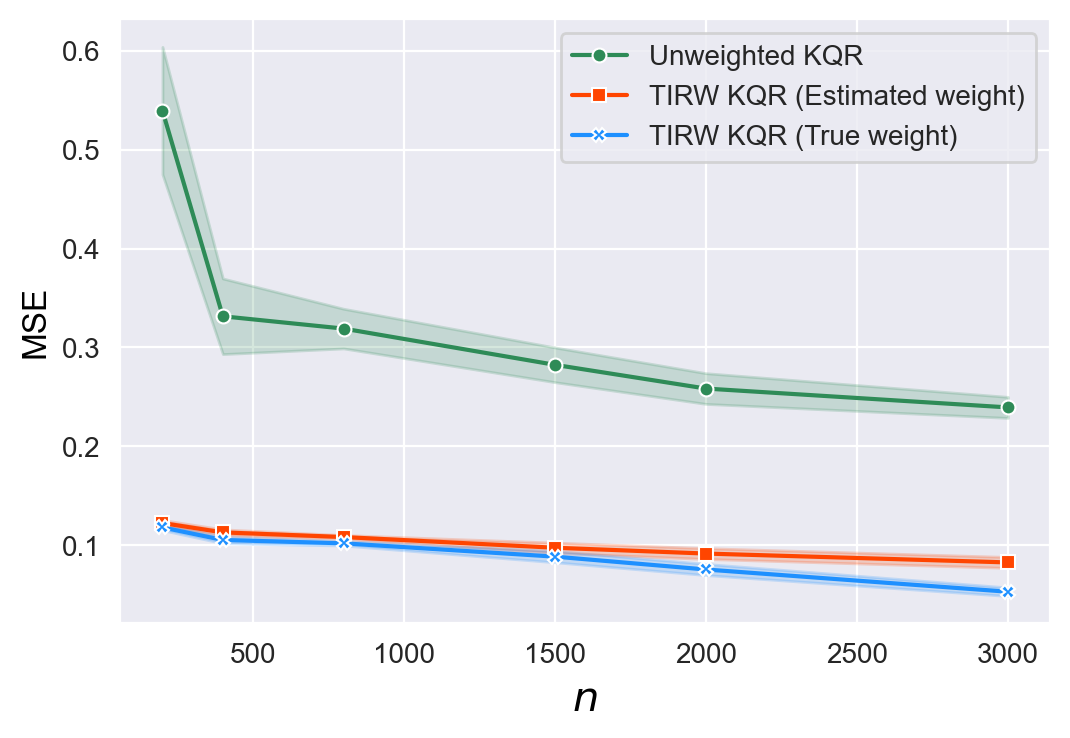}
\label{KQR_dimension3_MSE_7}}
   \subfigure[$\tau=0.5$ and $r=1$]{
    \includegraphics[width=1.6in,  height=1.09in]{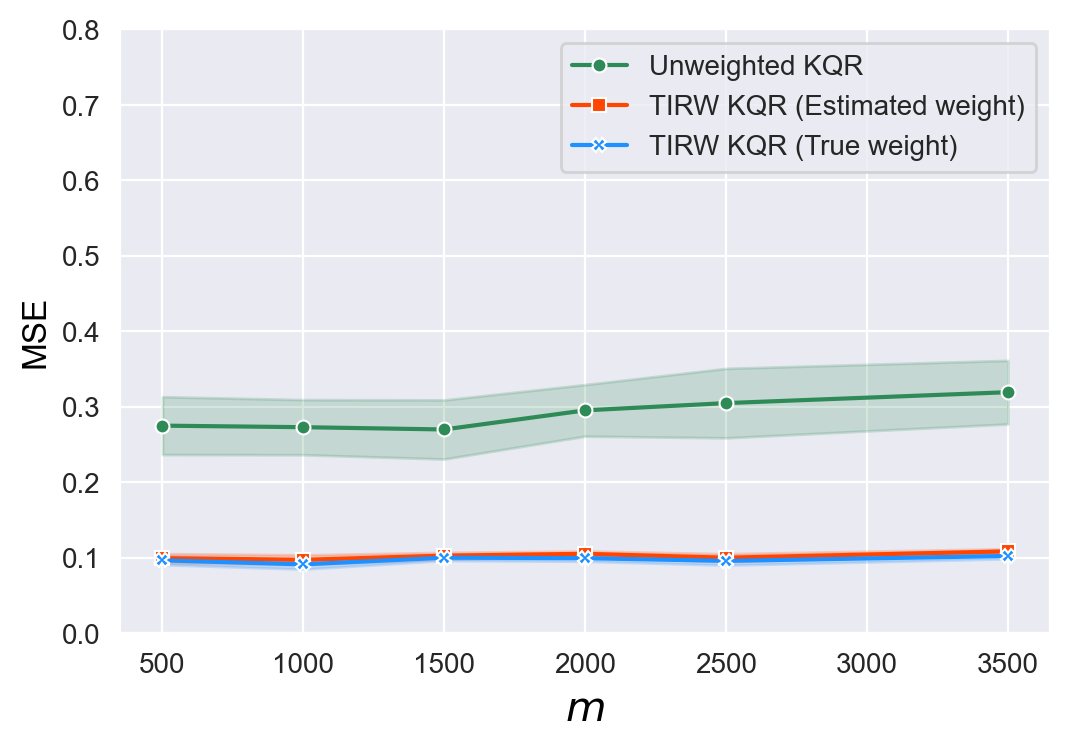}
    \label{KQR_dimension3_MSE_13}}
    \subfigure[$\tau=0.5$ and $r=1$]{
	\includegraphics[width=1.6in,  height=1.09in]{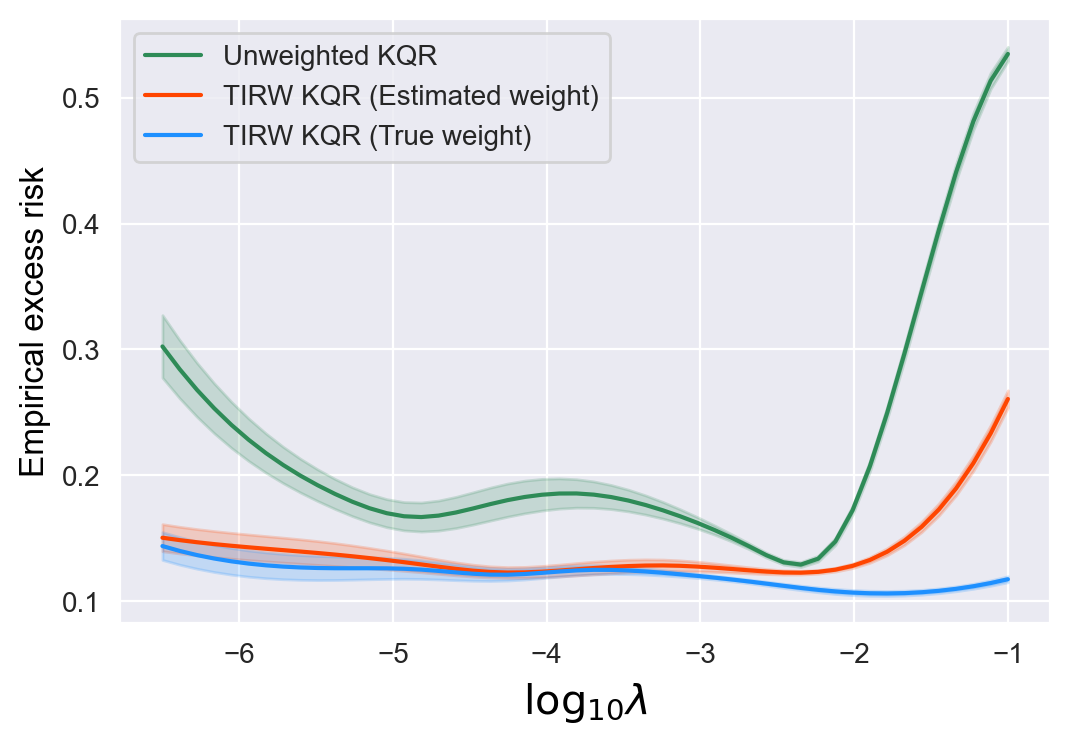}
        \label{KQR_dimension3_Empirical_smooth_1}}
 \subfigure[$\tau=0.5$ and $r=1$]{
    \includegraphics[width=1.6in,  height=1.09in]{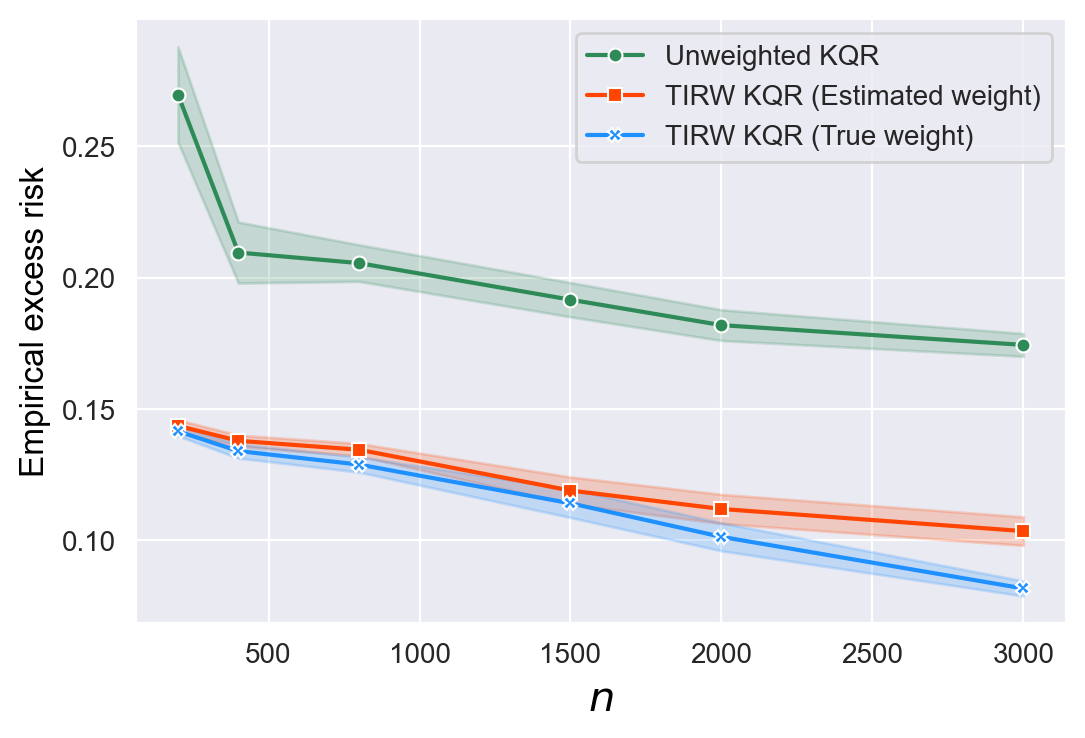}
\label{KQR_dimension3_Empirical_7}}
    \subfigure[$\tau=0.5$ and $r=1$]{
	\includegraphics[width=1.6in,  height=1.09in]{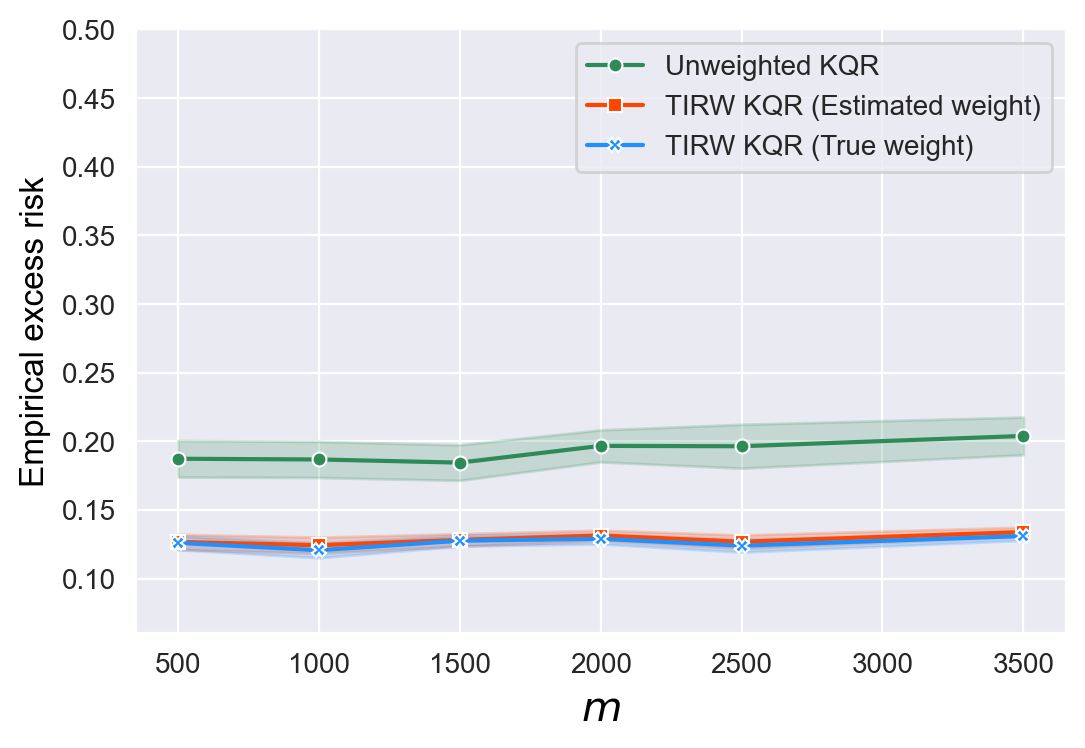}
 \label{KQR_dimension3_Empirical_13}} 

\end{figure}

\begin{figure}[H]
\graphicspath{{unbounded_KQR_3_d/}}
\setcounter{subfigure}{0}
\renewcommand{\thesubfigure}{(4\alph{subfigure})}
    \centering
    \subfigure[$\tau=0.5$ and $r=0$]{
    \includegraphics[width=1.6in,  height=1.09in]{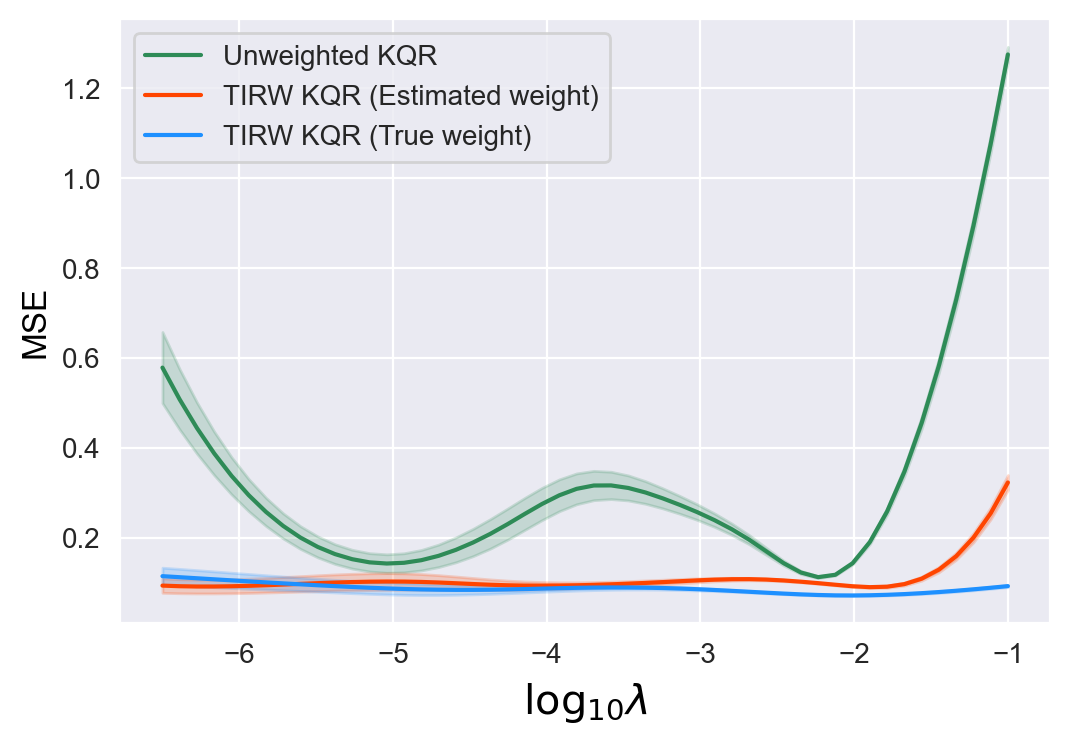}
    \label{KQR_dimension3_MSE_smooth_2}}
    \subfigure[$\tau=0.5$ and $r=0$]{
	\includegraphics[width=1.6in,  height=1.09in]{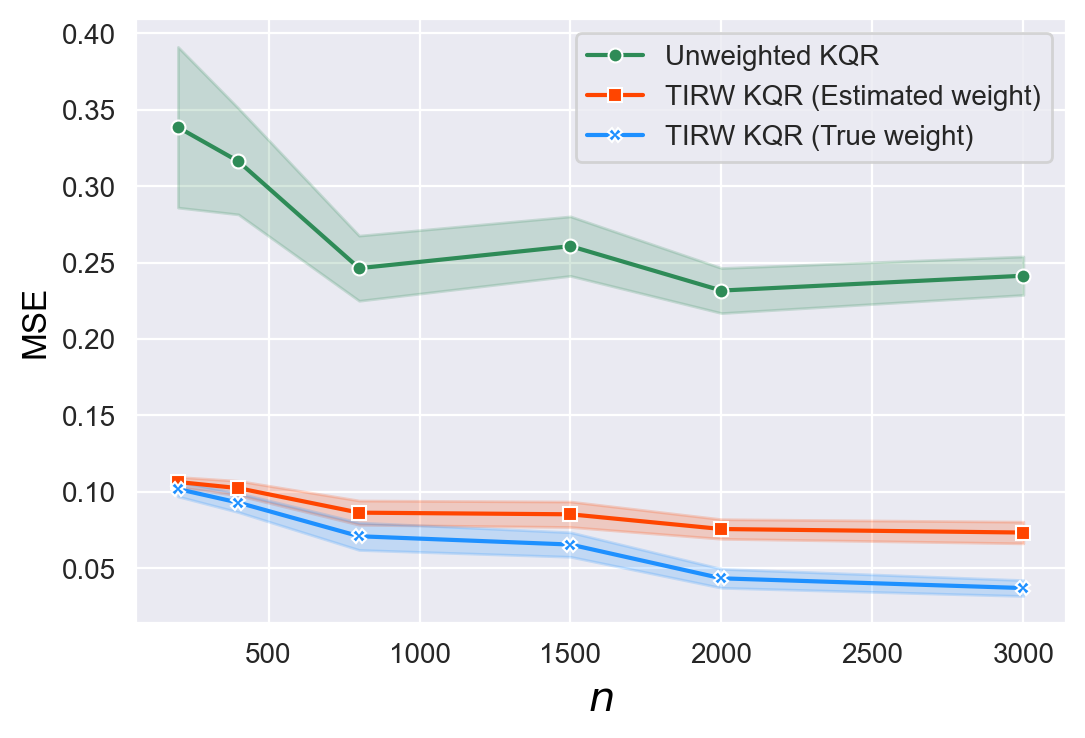}
\label{KQR_dimension3_MSE_8}}
   \subfigure[$\tau=0.5$ and $r=0$]{
    \includegraphics[width=1.6in,  height=1.09in]{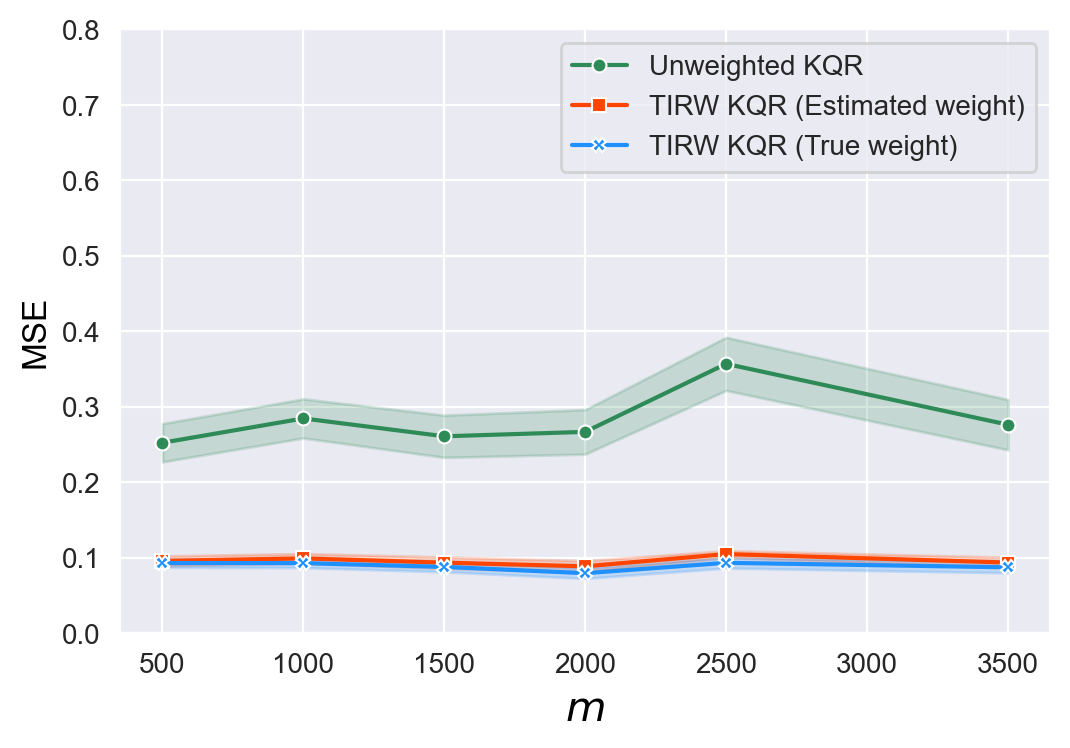}
    \label{KQR_dimension3_MSE_14}}
    \subfigure[$\tau=0.5$ and $r=0$]{
	\includegraphics[width=1.6in,  height=1.09in]{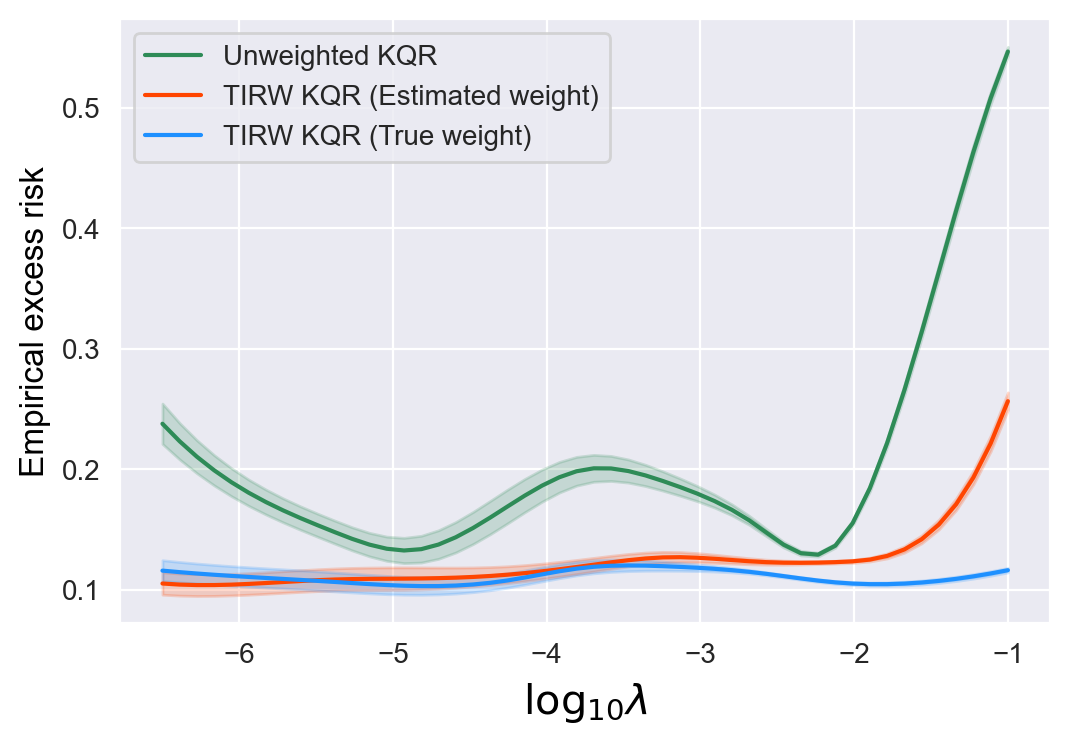}
        \label{KQR_dimension3_Empirical_smooth_2}}
 \subfigure[$\tau=0.5$ and $r=0$]{
    \includegraphics[width=1.6in,  height=1.09in]{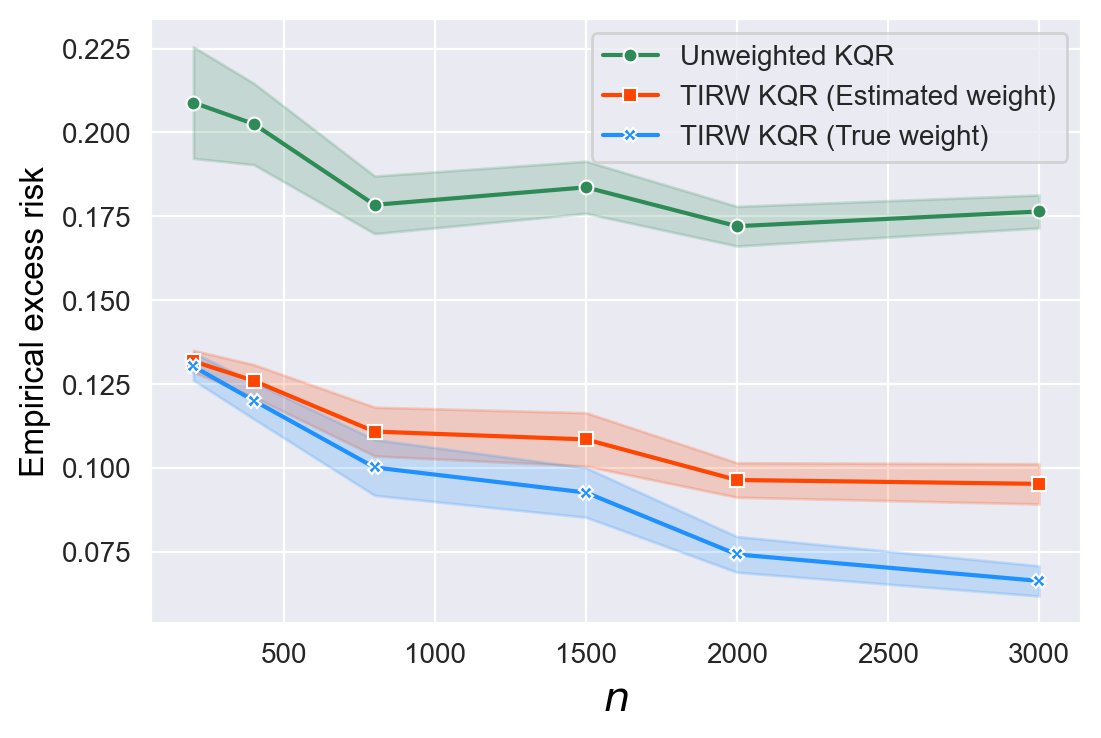}
\label{KQR_dimension3_Empirical_8}}
    \subfigure[$\tau=0.5$ and $r=0$]{
	\includegraphics[width=1.6in,  height=1.09in]{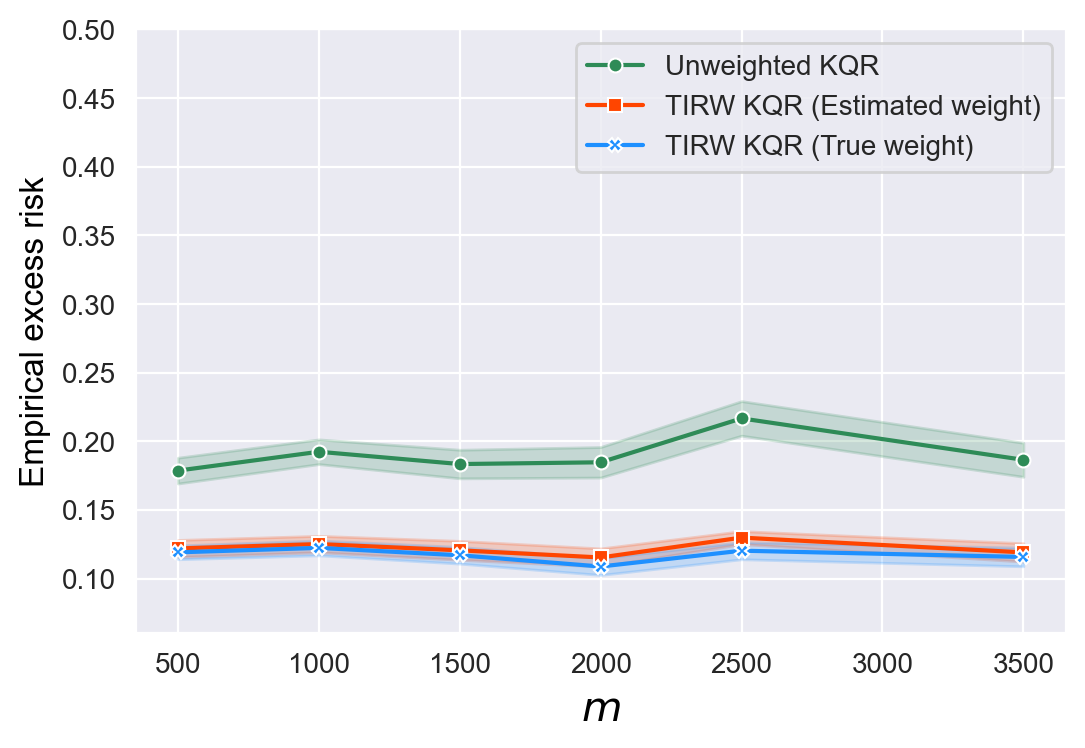}
 \label{KQR_dimension3_Empirical_14}}
 \setcounter{subfigure}{0}
\renewcommand{\thesubfigure}{5\alph{subfigure})}
    \subfigure[$\tau=0.7$ and $r=1$]{
    \includegraphics[width=1.6in,  height=1.09in]{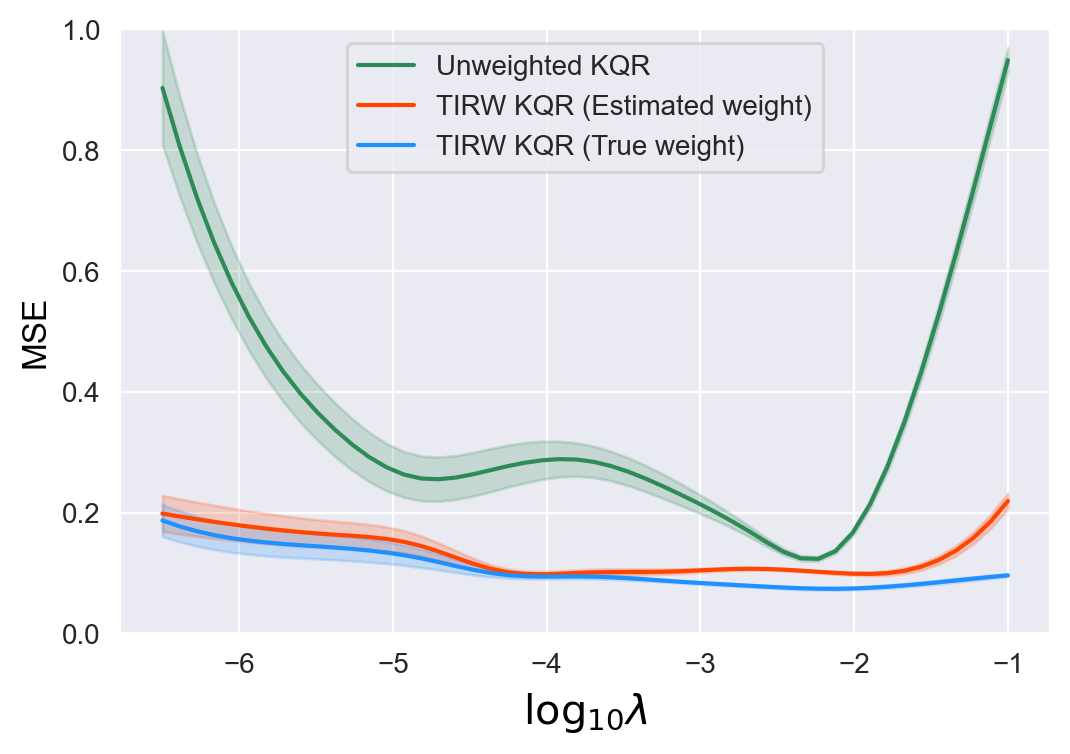}
\label{KQR_dimension3_MSE_smooth_5}}
    \subfigure[$\tau=0.7$ and $r=1$]{
	\includegraphics[width=1.6in,  height=1.09in]{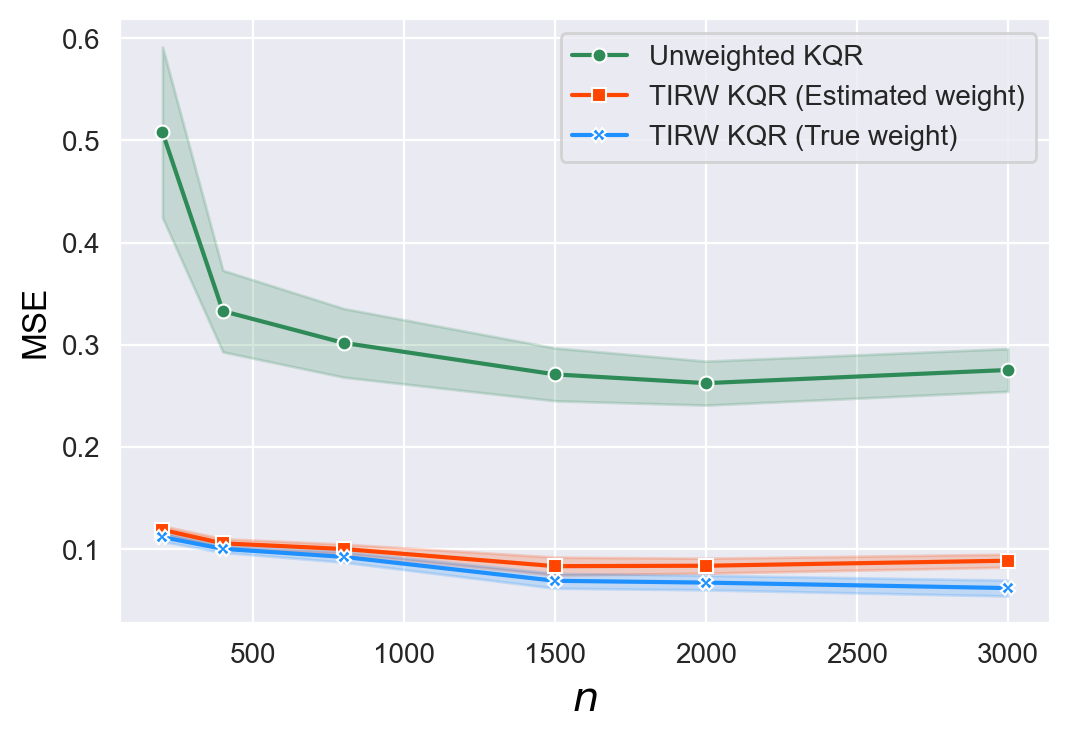}
\label{KQR_dimension3_MSE_11}}
   \subfigure[$\tau=0.7$ and $r=1$]{
    \includegraphics[width=1.6in,  height=1.09in]{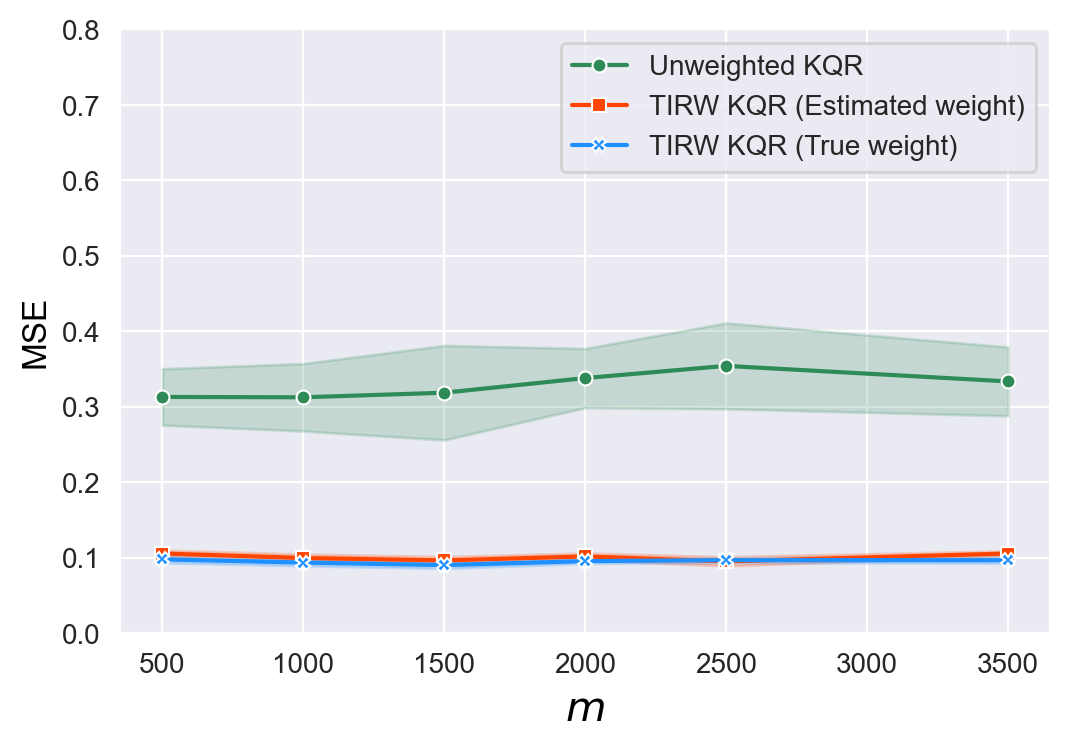}
    \label{KQR_dimension3_MSE_17}}
    \subfigure[$\tau=0.7$ and $r=1$]{
	\includegraphics[width=1.6in,  height=1.09in]{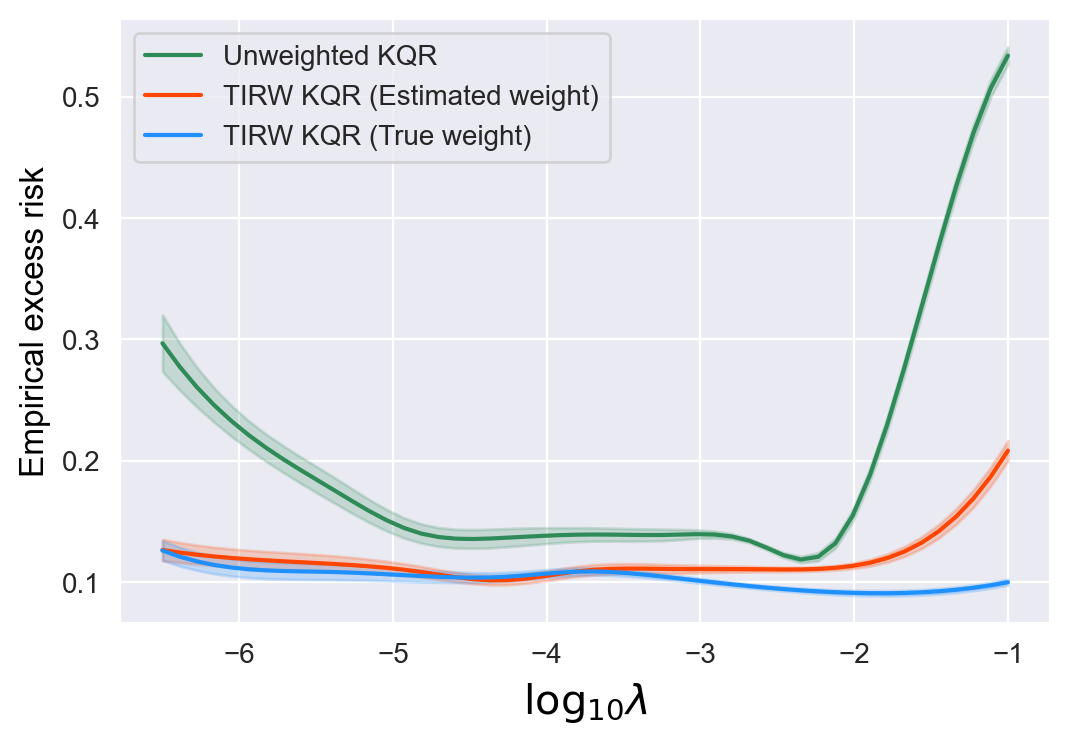} \label{KQR_dimension3_Empirical_smooth_5}}
 \subfigure[$\tau=0.7$ and $r=1$]{
    \includegraphics[width=1.6in,  height=1.09in]{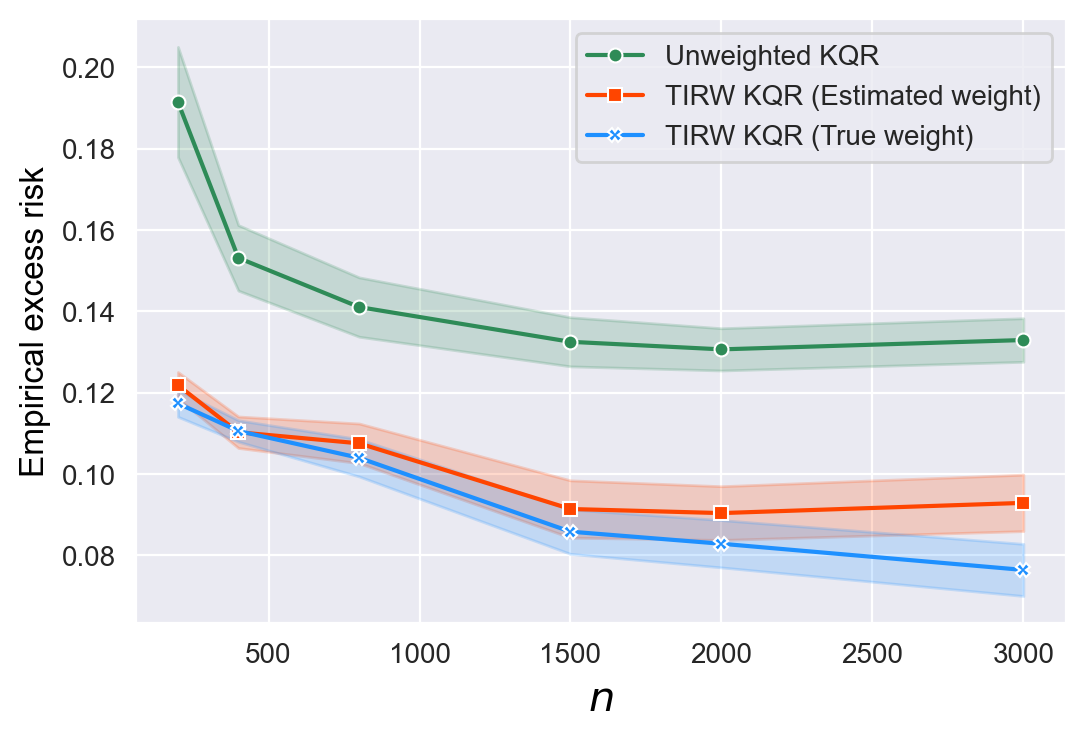}
\label{KQR_dimension3_Empirical_11}}
    \subfigure[$\tau=0.7$ and $r=1$]{
	\includegraphics[width=1.6in,  height=1.09in]{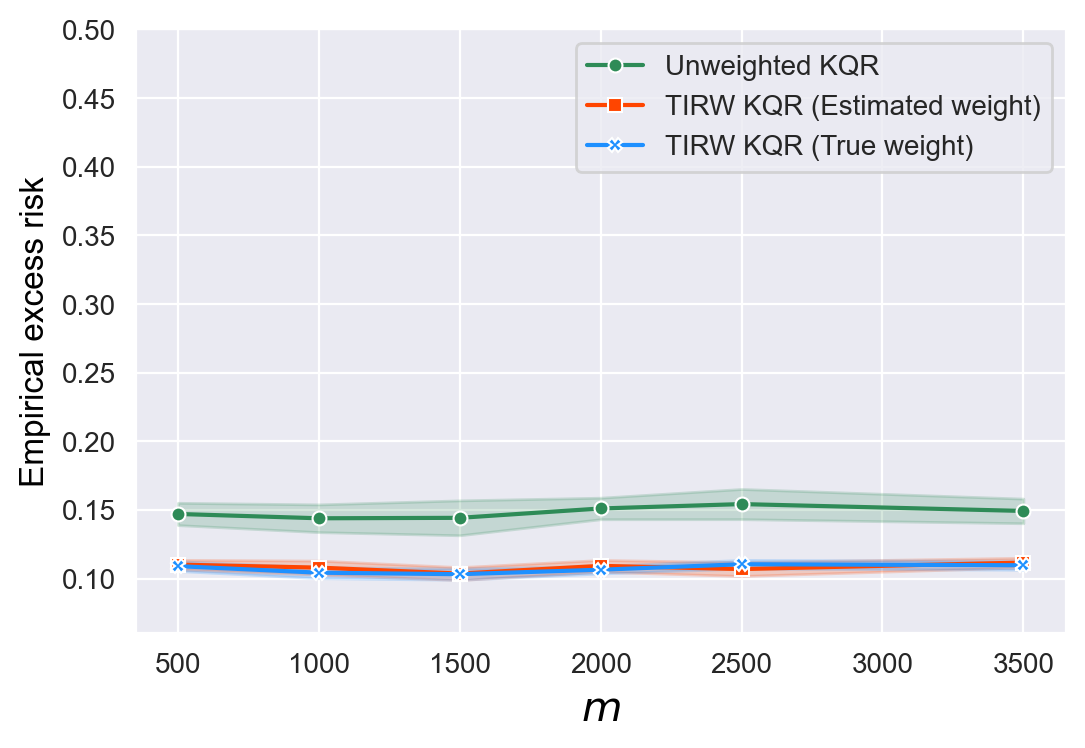}
 \label{KQR_dimension3_Empirical_17}}
 \setcounter{subfigure}{0}
\renewcommand{\thesubfigure}{(6\alph{subfigure})}
     \subfigure[$\tau=0.7$ and $r=0$]{
    \includegraphics[width=1.6in,  height=1.09in]{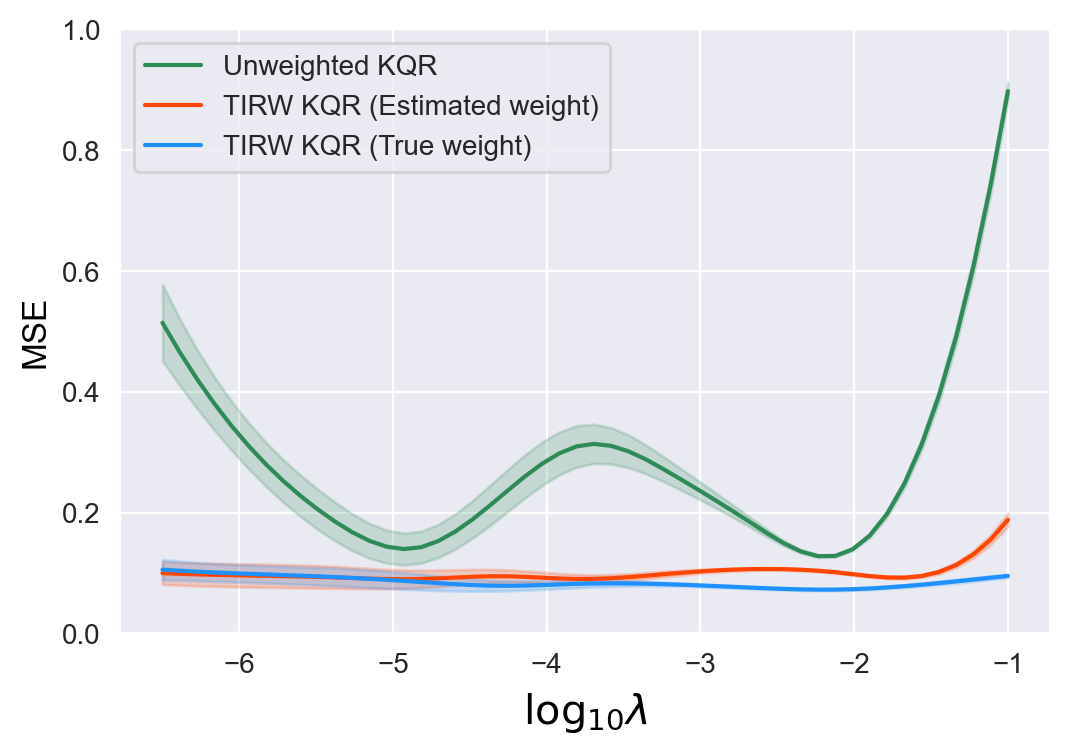}
\label{KQR_dimension3_MSE_smooth_6}}
    \subfigure[$\tau=0.7$ and $r=0$]{
	\includegraphics[width=1.6in,  height=1.09in]{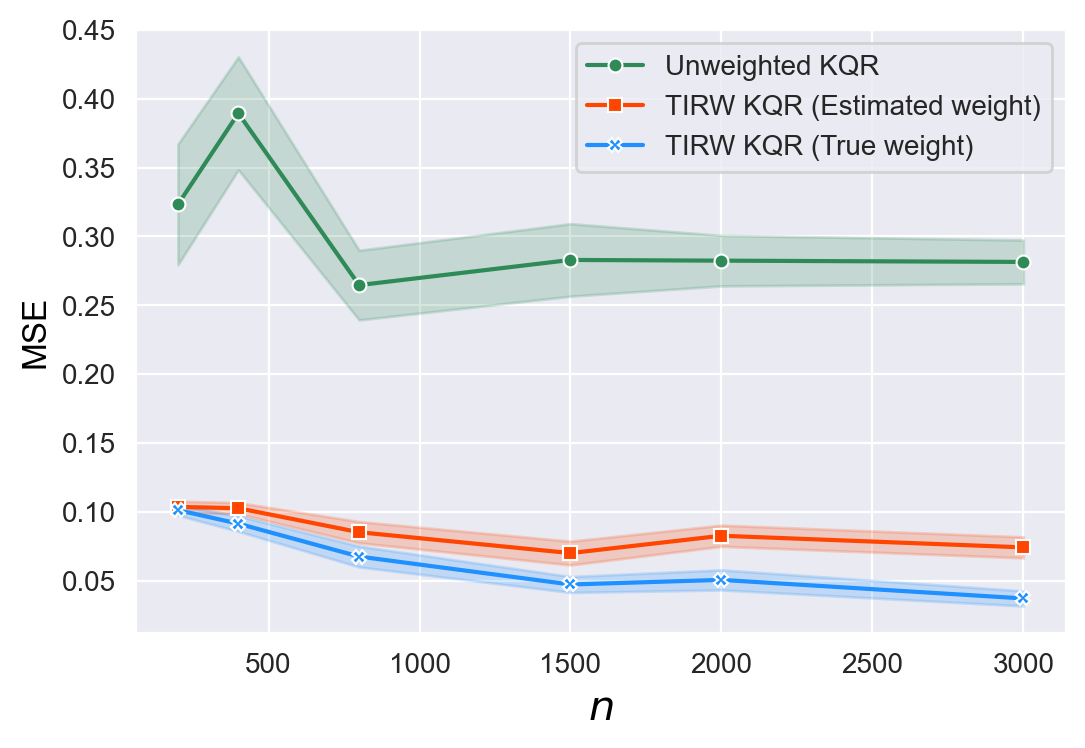}
\label{KQR_dimension3_MSE_12}}
   \subfigure[$\tau=0.7$ and $r=0$]{
    \includegraphics[width=1.6in,  height=1.09in]{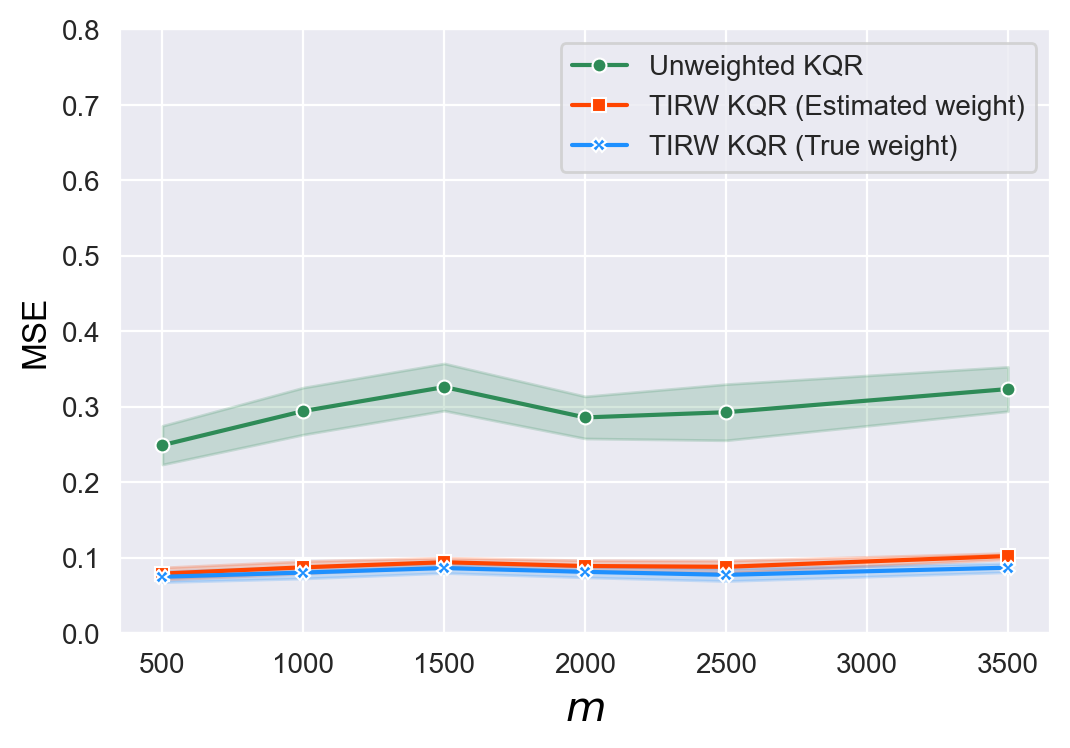}
    \label{KQR_dimension3_MSE_18}}
    \subfigure[$\tau=0.7$ and $r=0$]{
	\includegraphics[width=1.6in,  height=1.09in]{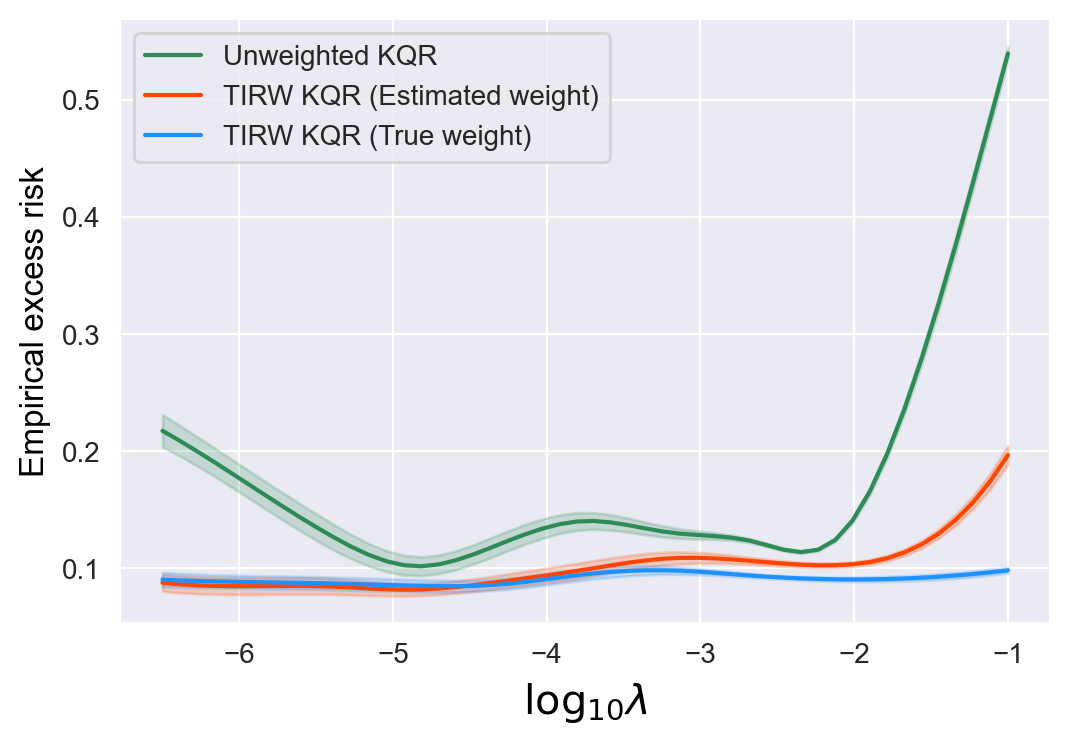} \label{KQR_dimension3_Empirical_smooth_6}}
 \subfigure[$\tau=0.7$ and $r=0$]{
    \includegraphics[width=1.6in,  height=1.09in]{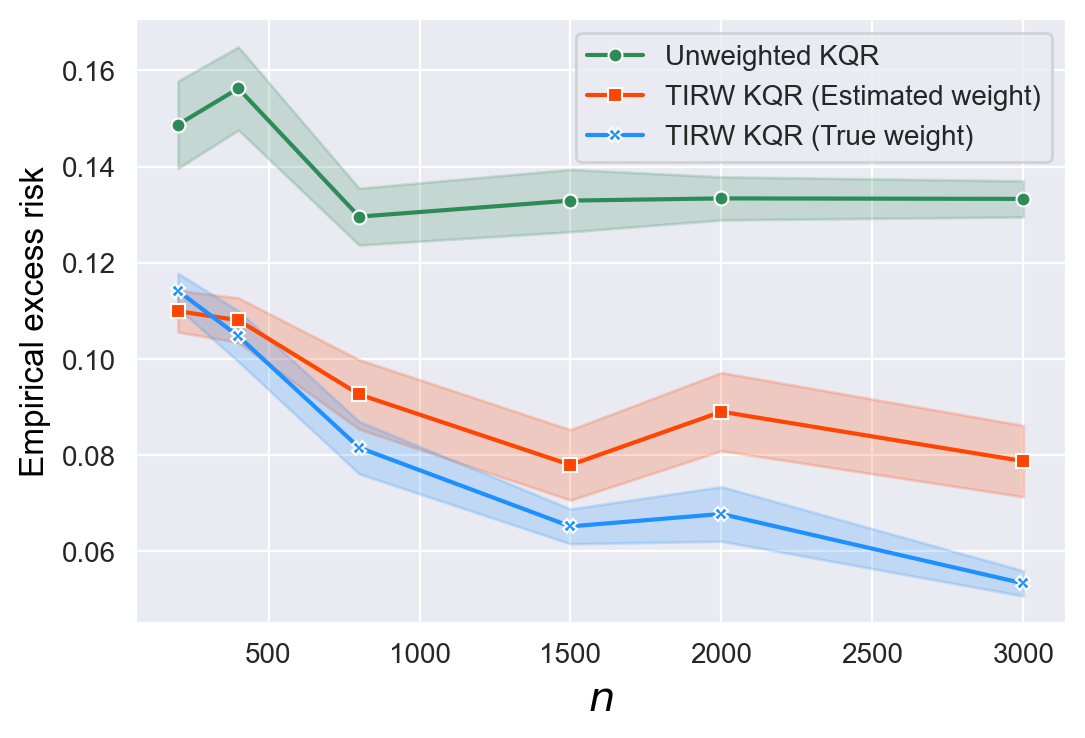}
\label{KQR_dimension3_Empirical_12}}
    \subfigure[$\tau=0.7$ and $r=0$]{
	\includegraphics[width=1.6in,  height=1.09in]{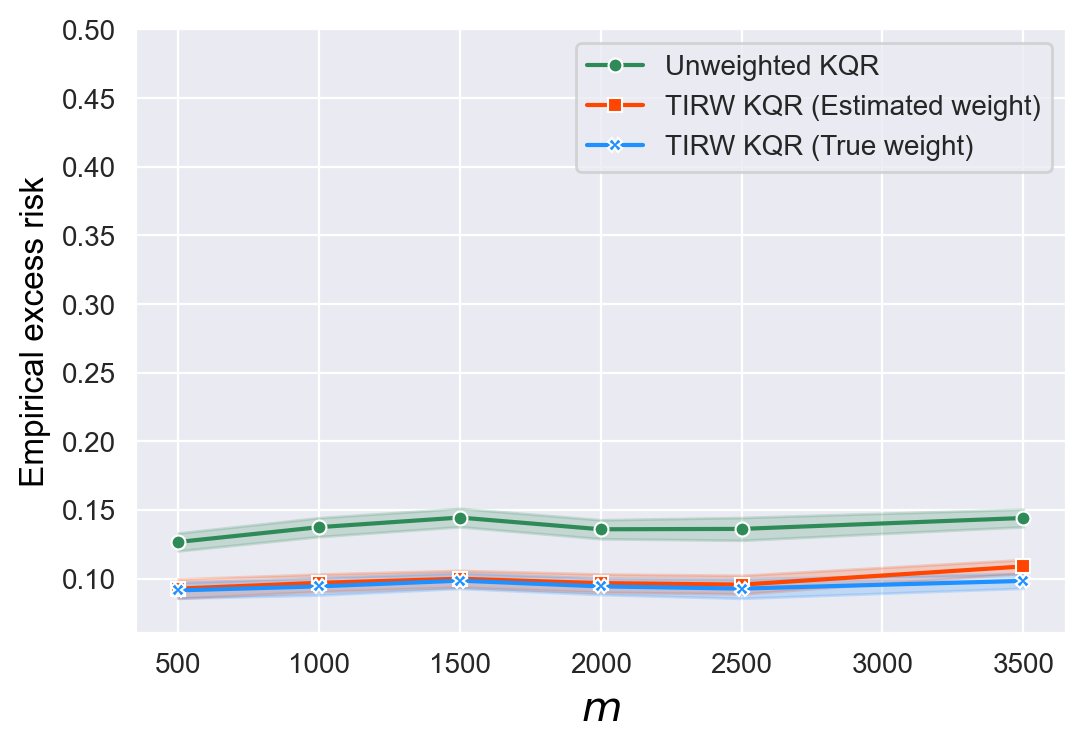}
 \label{KQR_dimension3_Empirical_18}}
 \caption{\footnotesize{Average MSE  and empirical excess risk for unweighted KQR, TIRW KQR with true weight and estimated weight, respectively (in the left panel, the curves are plotted with respect to $\log_{10} \lambda$ with $n=500, m=1000$; in the middle panel,  the curves are plotted with respect to $n$ with fixed $m = 1000,\lambda=10^{-4}$; in the right panel, the curves are plotted with respect to $m$ with fixed $n=500,\lambda=10^{-4}$)}}
\label{KQR_dimension3_unbounded}
\end{figure}

\subsection{Kernel logistic regression} \label{sec A.3}
\begin{figure}[H]
\graphicspath{{bounded_KLR_3_d/}}
    \centering
    \subfigure[]{
\includegraphics[width=1.6in,  height=1.09in]{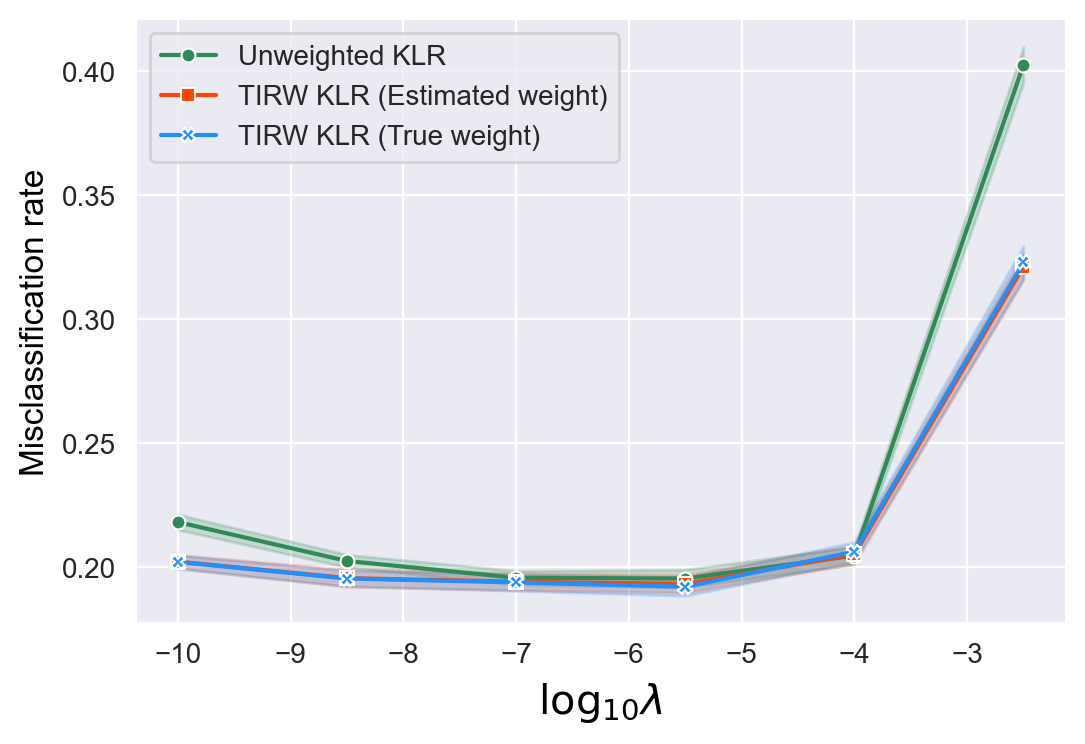}\label{KLR_dimension3_uniformbounded_Misclassification_1}}
    \subfigure[]{
	\includegraphics[width=1.6in,  height=1.09in]{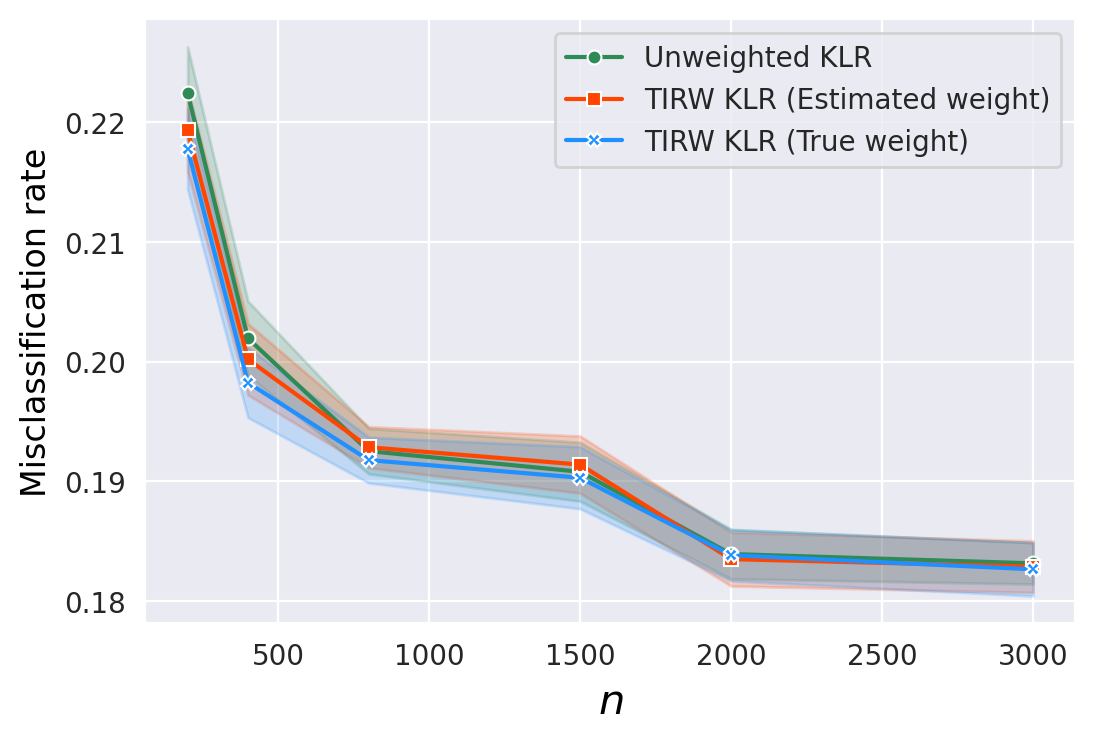}
\label{KLR_dimension3_uniformbounded_Misclassification_2}}
   \subfigure[]{
    \includegraphics[width=1.6in,height=1.09in]{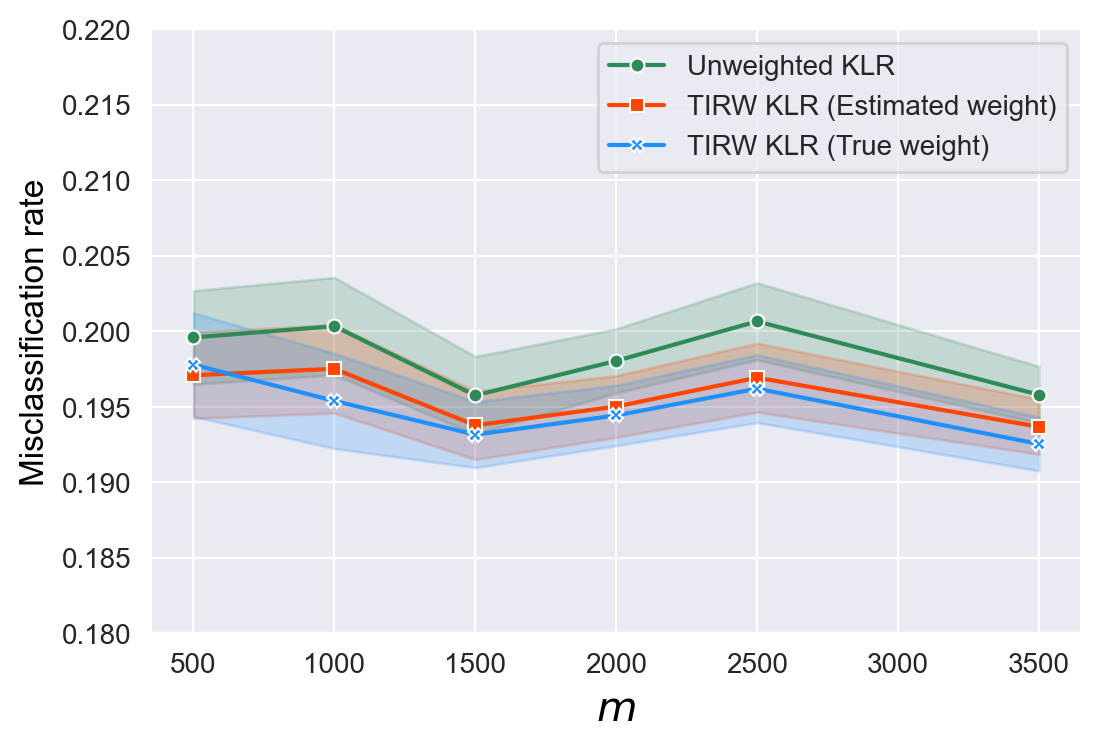}\label{KLR_dimension3_uniformbounded_Misclassification_3}}
      \subfigure[]{
\includegraphics[width=1.6in,  height=1.09in]{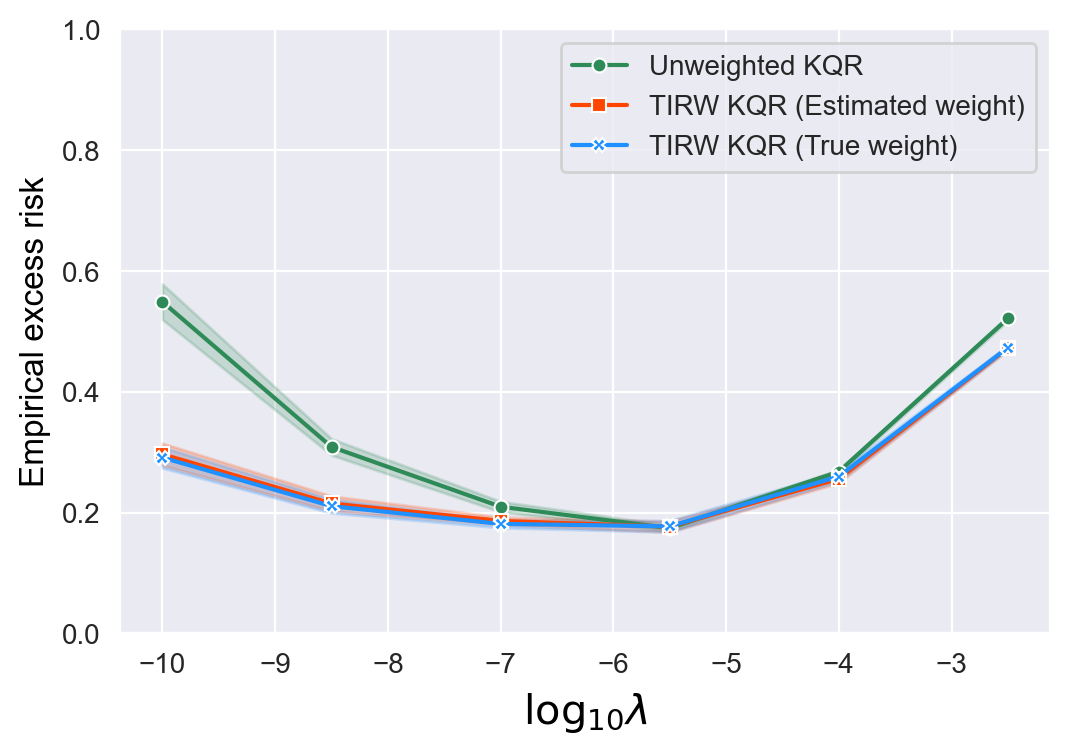}\label{KLR_dimension3_uniformbounded_Empirical_1}}
    \subfigure[]{
	\includegraphics[width=1.6in,  height=1.09in]{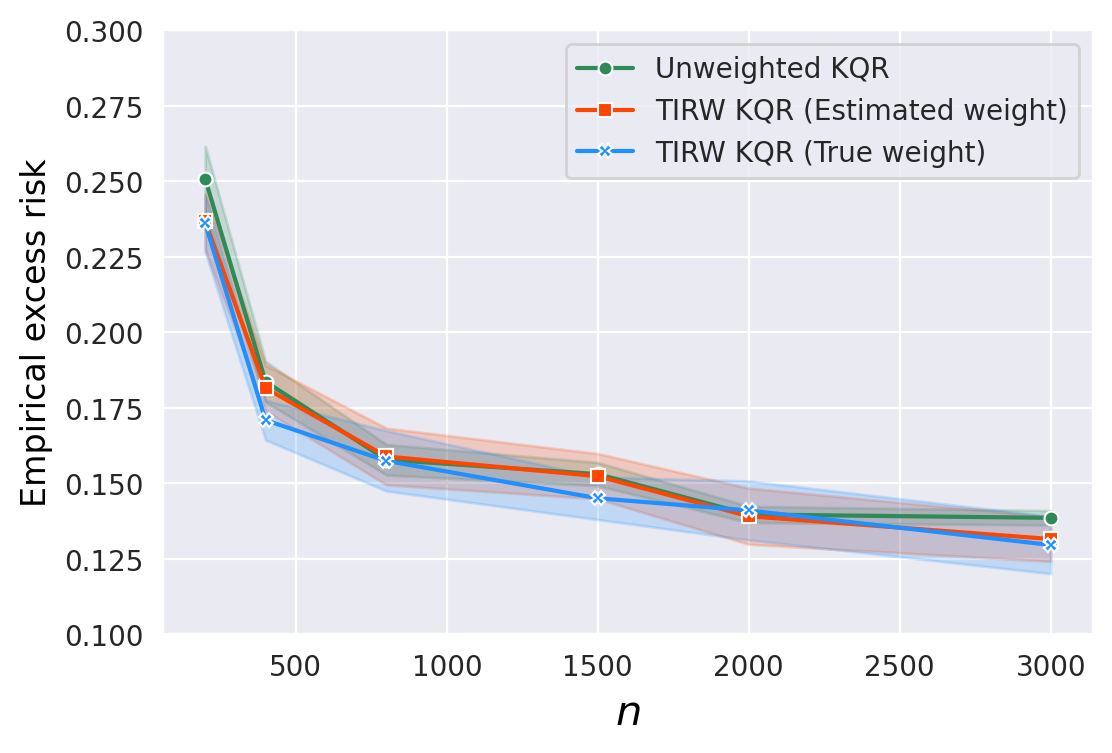}
\label{KLR_dimension3_uniformbounded_Empirical_2}}
   \subfigure[]{
    \includegraphics[width=1.6in,height=1.09in]{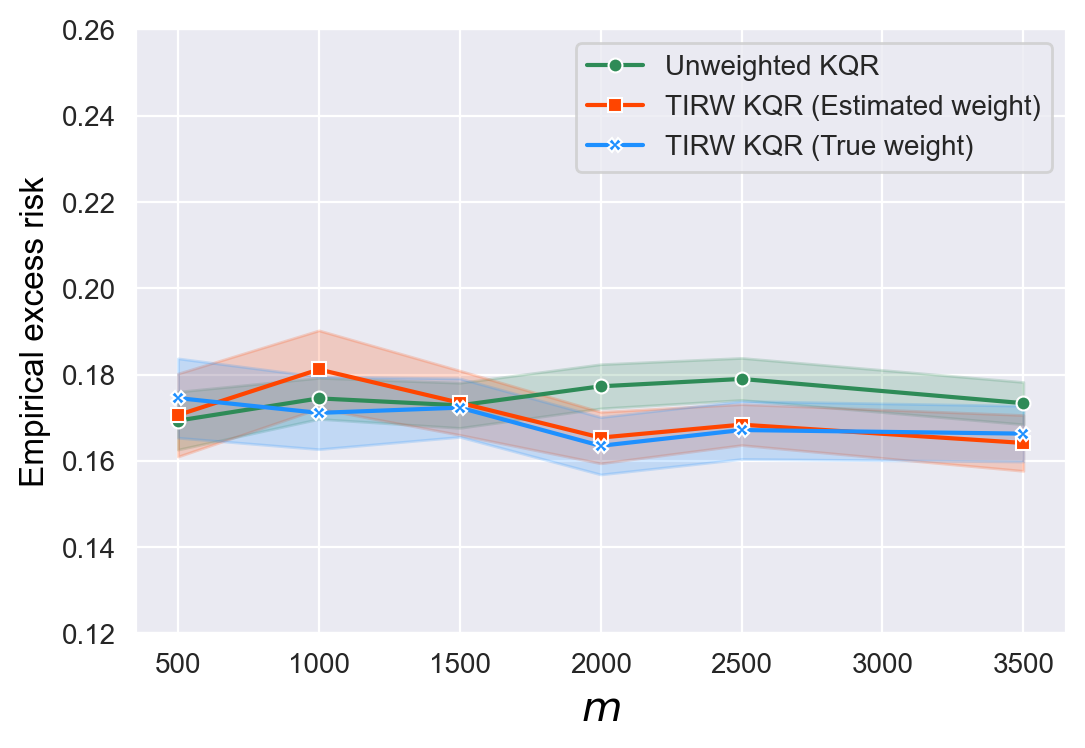}\label{KLR_dimension3_uniformbounded_Empirical_3}}
\caption{\footnotesize{Average misclassification rate  and empirical excess risk for unweighted KLR, TIRW KLR with true weight and estimated weight, respectively (in (a) and (d), the curves are plotted with respect to $\log_{10} \lambda$ with $n=500, m=1000$; in (b) and (e) the curves are plotted with respect to $n$ with fixed $m = 1000,\lambda=5\times10^{-5}$; in (c) and (f), the curves are plotted with respect to $m$ with fixed $n=500,\lambda=5\times 10^{-5}$)}}
\label{KLR_dimension3_bounded}
\end{figure}

\begin{figure}[H]
\graphicspath{{unbounded_KLR_3_d/}}
    \centering
    \subfigure[]{
    \includegraphics[width=1.6in,  height=1.09in]{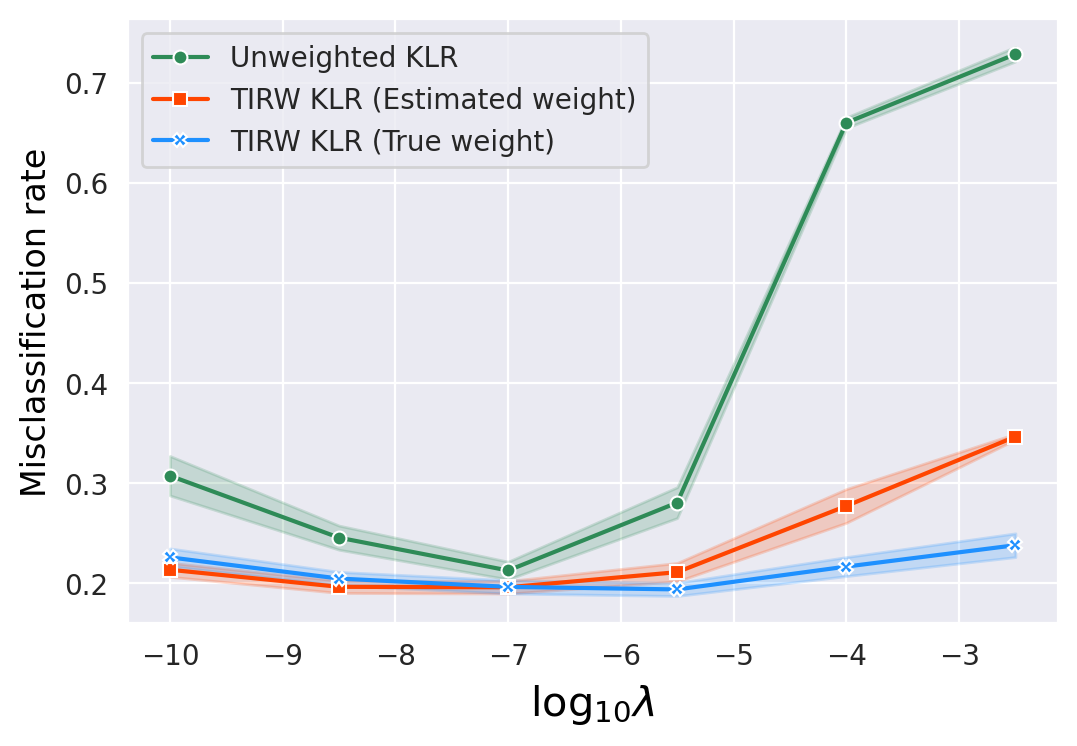}\label{KLR_dimension3_Misclassification_1}}
    \subfigure[]{
	\includegraphics[width=1.6in,  height=1.09in]{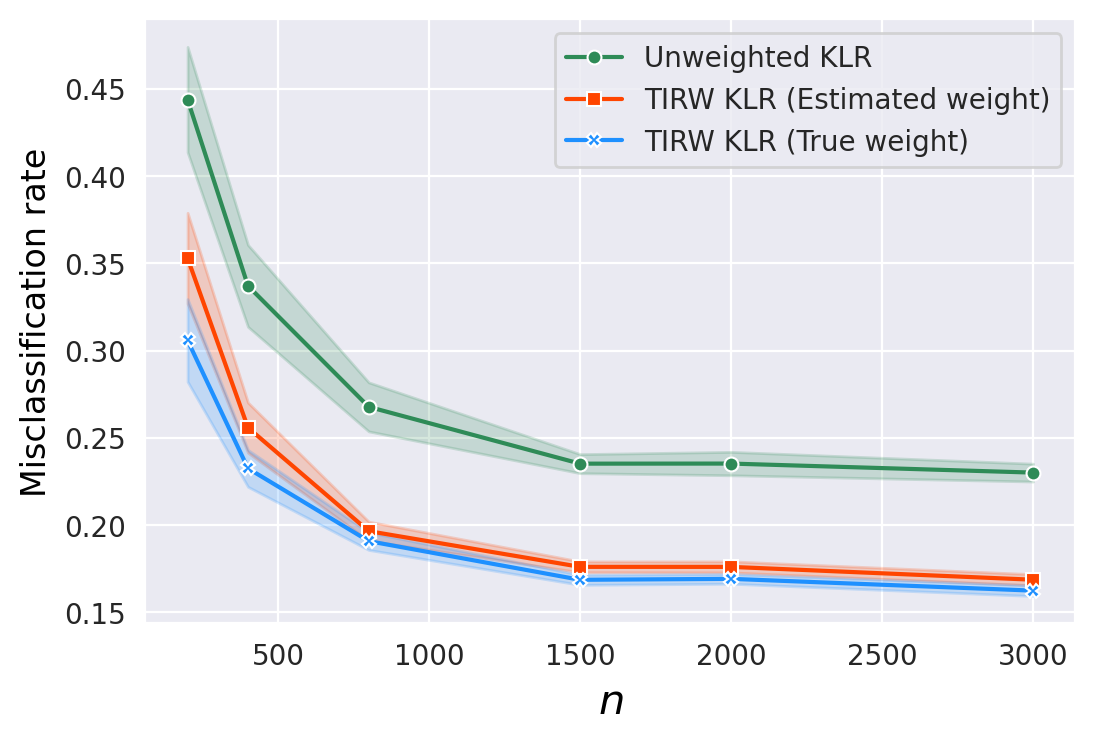}
\label{KLR_dimension3_Misclassification_2}}
   \subfigure[]{
\includegraphics[width=1.6in,height=1.09in]{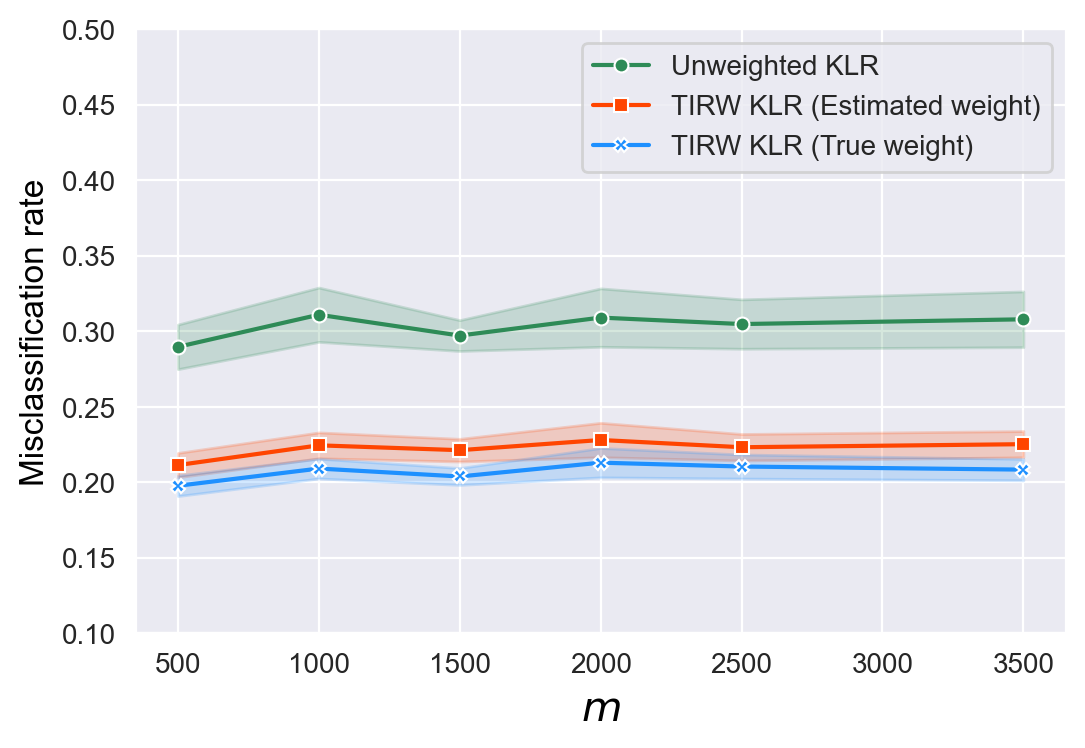}\label{KLR_dimension3_Misclassification_3}}
  \subfigure[]{  \includegraphics[width=1.6in,  height=1.09in]{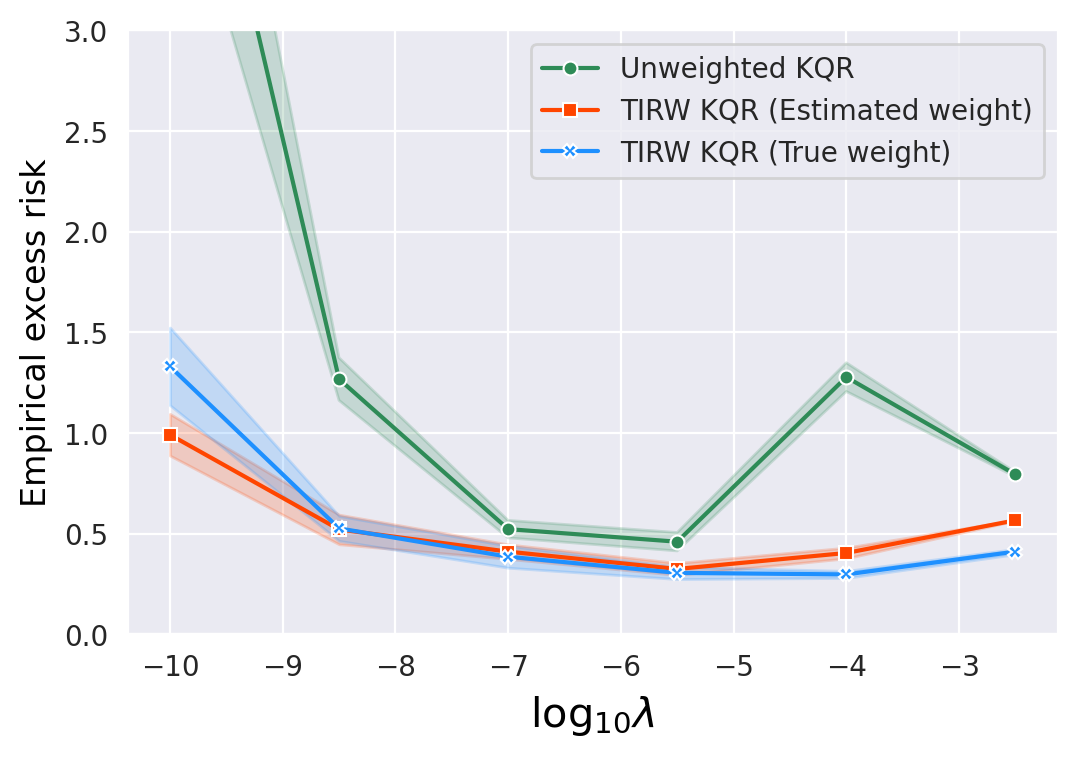}\label{KLR_dimension3_Empirical_1}}
    \subfigure[]{
	\includegraphics[width=1.6in,  height=1.09in]{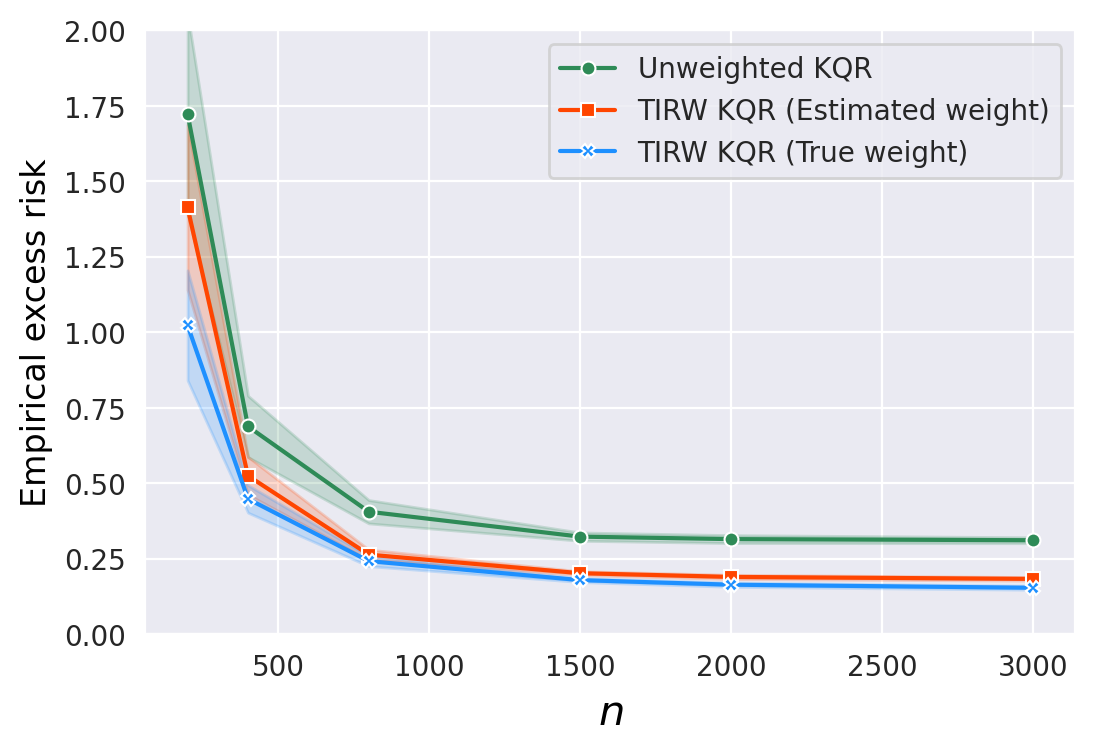}
\label{KLR_dimension3_Empirical_2}}
   \subfigure[]{
\includegraphics[width=1.6in,height=1.09in]{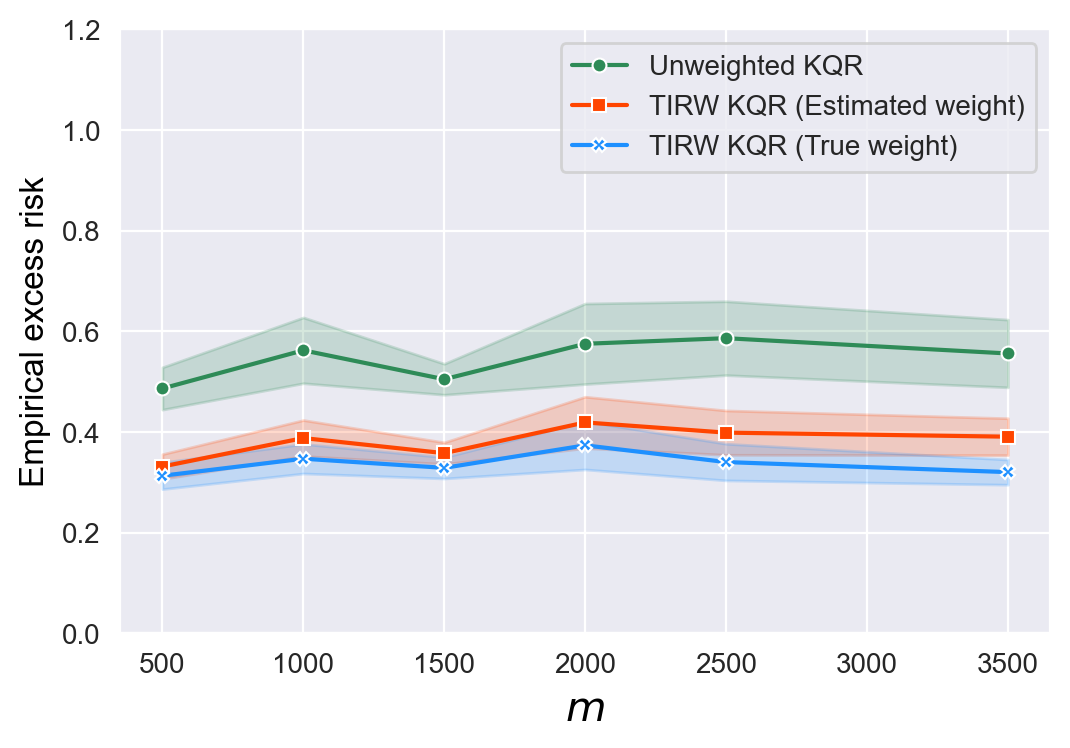}\label{KLR_dimension3_Empirical_3}}
\caption{\footnotesize{Average misclassification rate  and empirical excess risk for unweighted KLR, TIRW KLR with true weight and estimated weight, respectively (in (a) and (d), the curves are plotted with respect to $\log_{10} \lambda$ with $n=500, m=1000$; in (b) and (e) the curves are plotted with respect to $n$ with fixed $m = 1000,\lambda=5\times10^{-5}$; in (c) and (f), the curves are plotted with respect to $m$ with fixed $n=500,\lambda=5\times10^{-5}$)}}
\label{KLR_dimension3_unbounded}
\end{figure}

For the logistic loss function, we consider the following example.\\
\noindent{\bf Example S5}: The response $y$ is generated by 
$P(y=1| \bx)=\frac{1}{1+\exp(-f(\bx))}$ where $ {\bx}=(x_0,x_1,x_2)^\top\in \mathcal{R}^3$ and  $f({\bx})=-x_0^2+3\sin(3\pi x_0)+e^{x_1^2-x_2^2}$. 
We consider  $\rho^S_{\bx}({\bx})=g(x_0; \alpha_s,\beta_s)$ and  $\rho^T_{\bx}({\bx})=g(x_0; \alpha_t,\beta_t)$ with   $\alpha_s=2.5,\beta_s=2,\alpha_t=3,\beta_t=4$ for the uniformly bounded case and $\alpha_s=4,\beta_s=1,\alpha_t=3,\beta_t=6$ for the moment bounded case, respectively. Note that for any learned $\widehat{f}$, the classification rule is specified as $\sign(\widehat{f}(\bx))$. Figure \ref{KLR_dimension3_bounded} presents the results for the uniformly bounded case, and the results for the moment bounded case are presented in Figure \ref{KLR_dimension3_unbounded}. 

\setcounter{subfigure}{0}
\renewcommand{\thesubfigure}{(\alph{subfigure})}

Clearly, Figure \ref{KLR_dimension3_bounded} demonstrates that the difference between the performance of the TIRW estimator and the unweighted estimator is very close under the uniformly bounded case, which confirms our theoretical findings. In addition, as we found in the other examples, we can conclude from (a) and (d) in Figure \ref{KLR_dimension3_unbounded} that the performance of the TIRW estimator, in terms of both misclassification rate and empirical excess risk, maintains relative stability for different choices of $\lambda$. But if we don't select $\lambda$ carefully, the unweighted estimator performs extremely poor. Other cases in Figure \ref{KLR_dimension3_unbounded} confirm that the TIRW estimator takes a remarkable benefit.
\subsection{Real applications for multi-source datasets} \label{sec A.4}

In this section, we apply  KSVM to a wide range of real datasets that are available in the  UCI archive \url{https://archive.ics.uci.edu/ml/datasets.php}, including the ionosphere dataset, the dry bean dataset, the magic04 dataset, and the banknote authentication dataset. Specifically, the ionosphere dataset contains 350 instances and 34 covariates, and we select all the instances with $3$-$7$-th covariates into the model.  The dry bean dataset contains 13611 instances and 16 covariates, and we randomly select 30\% of this dataset with the $1$-$8$-th covariates.  The   magic04 dataset contains 19019 instances and 10 covariates, and we randomly select 10\% of this dataset with the $2$-$10$-th covariates. The banknote authentication dataset contains 1372 instances and 4 covariates. The numerical performance is summarized in the following table. 

\begin{table}[H]
\centering
\caption{Classification performance on multi-source datasets.}
\label{table2}
\begin{threeparttable}
\begin{footnotesize}
\scalebox{0.9}{
\setlength{\tabcolsep}{1.3mm}{
\begin{tabular}{ l  c   c  c c c c} 
\hline
\hline
Dataset & Estimator    &   $C=0.01$    & $C=0.1$    & $C=1$ & $C=10$ & $C=100$  \\
\hline
\multirow{2}*{Ionosphere} &Unweighted &  $0.260\pm 0.007$    & $0.266\pm0.013$   &$0.602\pm0.098$&$\mathbf{0.637\pm0.104}$&$0.625\pm0.102$\\
&TIRW & $0.740\pm0.007$  & $\mathbf{0.749\pm0.015}$ & $ 0.735\pm0.108$ &$0.665\pm0.108$&$0.643\pm0.103$ \\
\hline
\multirow{2}*{Dry Bean} &Unweighted &  $0.266\pm 0.008$    & $0.616\pm0.012$   &$\mathbf{0.748\pm0.014}$&$0.719\pm0.021$&$0.668\pm0.020$\\
&TIRW & $\mathbf{0.824\pm0.022}$  & $0.773\pm0.013$ & $ 0.764\pm0.014$ &$0.713\pm0.022$&$0.696\pm0.028$ \\
\hline
\multirow{2}*{Magic04} &Unweighted &  $0.621\pm 0.006$    & $0.625\pm0.007$   &$\mathbf{0.779\pm0.011}$&$0.752\pm0.012$&$0.744\pm0.012$\\
&TIRW & $0.621\pm0.006$  & $0.624\pm0.007$ & $\mathbf{0.804\pm0.041}$ &$0.768\pm0.011$&$0.748\pm0.013$ \\
\hline
\multirow{2}*{Authentication} &Unweighted &  $0.237\pm 0.007$    & $0.718\pm0.031$  &$\mathbf{0.940\pm0.033}$&$0.920\pm0.036$&$0.919\pm0.036$\\
&TIRW & $0.811\pm 0.037$  & $\mathbf{0.986 \pm0.015}$ & $ 0.973\pm0.023$ &$0.928\pm0.032$&$0.919\pm0.036$ \\
\hline
\hline
\end{tabular}}
}
\end{footnotesize}  
 \end{threeparttable}
\end{table}

As shown in Table \ref{table2}, the TIRW  estimator outperforms the unweighted estimator on each dataset for almost all the choices of $C_\lambda$. We also observe that the unweighted estimator has a much less satisfying accuracy of prediction for small choices of $C_\lambda$. Nevertheless, with importance ratio correction, the accuracy rate has been significantly improvded for small choices of $C_\lambda$, even attaining nearly optimal for the first two datasets. For a large choice of $C_\lambda$, these two estimators have a negligible gap in accuracy rate.

 \subsection{Kullback-Leibler importance estimation procedure}\label{sec A.7}
In this section, we introduce the importance ratio estimation procedure based on Kullback-Leibler divergence \citep{sugiyama2007direct}. Recall that we have the source input data ${\bx}_1^S,...,{\bx}_n^S$  generated from $\rho_{\bx}^S$ and the target input data ${\bx}_1^T,...,{\bx}_m^T$ generated from $\rho_{\bx}^T$ and our goal is to estimate the ratio $\phi(\bx)$. Since $\rho_{\bx}^T(\bx)=\phi(\bx)\rho_{\bx}^S(\bx)$, the true ratio $\phi(\bx)$ can be correctly identified by solving the population version of the optimization task that
\begin{align}\label{KLIEP_1}
  \underset{g(\bx)}{\text{minimize}}\quad \text{KL}\big(\rho_{\bx}^T(\bx)\| g(\bx)\rho_{\bx}^S(\bx)\big),
\end{align}
where $\text{KL}(p\| q)=\int p(\bx)\log\frac{p(\bx)}{q(\bx)}d\bx$ denotes Kullback–Leibler divergence between $P$ and $Q$ with probability densities $p$ and $q$ respectively. Note that either the bounded case or second moment bounded case considered in the main text,   $\rho_{\bx}^T(\bx)$ is absolutely continuous with respect to $\rho_{\bx}^S(\bx)$, which ensures the optimization problem \eqref{KLIEP_1} is well defined. Since 
\begin{align*}
  \text{KL}(\rho_{\bx}^T(\bx)\| g(\bx)\rho_{\bx}^S(\bx))=&\int \rho_{\bx}^T(\bx)\log\left(\frac{\rho_{\bx}^T(\bx)}{g(\bx)\rho_{\bx}^S(\bx)}\right)d\bx\\
  =&\int \rho_{\bx}^T(\bx)\log\left(\frac{\rho_{\bx}^T(\bx)}{\rho_{\bx}^S(\bx)}\right)d\bx-\int \rho_{\bx}^T(\bx)\log g(\bx)d\bx,
\end{align*}
we can rewrite the objective function as $-\int \rho_{\bx}^T(\bx)\log(g(\bx))d\bx$ by ignoring the constant term. It can be approximated by its empirical version that $-\frac{1}{m}\sum_{i=1}^{m}\log g({\bx}_i^T)$ with $g(\bx)=\sum_{k=1}^b\alpha_k K({\bx}_k,\bx)$ and $\{{\bx}_1,...,{\bx}_b\}$ denoting a subset of target input data with  $b$ as a pre-fixed number and $K(\cdot,\cdot)$ denoting some certain kernel function. Since the true ratio $\phi(\bx)$ is non-negative and satisfies $\frac{1}{n}\sum_{i=1}^n\phi({\bx}_i^S)\approx \int \phi(\bx)\rho_{\bx}^S(\bx)d\bx=1$. We add some constraints that $\alpha_k\ge 0$ and $\frac{1}{m}\sum_{i=1}^{n} \sum_{k=1}^b\alpha_k K({\bx}_k,{\bx}_i^S)=1$. 
This leads to the optimization problem:  \begin{align*}
\underset{\bfalpha\in \mathcal{R}^b}{\text{maximum}} \quad&  \frac{1}{m}\sum_{i=1}^{m}\log \left(\sum_{k=1}^b\alpha_k K({\bx}_k,{\bx}_i^T)\right),\\
\text{s.t.}\quad & \alpha_k\ge 0, \ k=1,...,b,\\
& \frac{1}{m}\sum_{i=1}^{n} \sum_{k=1}^b\alpha_k K({\bx}_k,{\bx}_i^S)=1.
\end{align*}
\subsection{Importance weighted cross validation}\label{sec A.8}
\cite{sugiyama2007covariate} points out that cross-validation (CV) on the unweighted training data introduces an additional source of bias in making predictions on test data due to covariate shift. They propose a method called importance weighted cross validation (IWCV) according to important ratio to compensate for the effect of covariate shift. First, one can randomly divide the training set $\{(\bx_i,y_i)\}_{i=1}^n$
into $b$ disjoint non-empty subsets $\{\mathcal{T}_i\}^{b}_{k=1}$. Then, denoting the learned function by using dataset $\{\mathcal{T}_k\}_{k\not= j}$ as $\widehat{f}_j$. Instead of the classical CV procedure, the IWCV aims to minimize
\begin{align*}
  \widehat{R}_{IWCV}=\frac{1}{b}\sum_{k=1}^b \frac{1}{\left|\mathcal{T}_k\right|} \sum_{(\bx_i, y_i) \in \mathcal{T}_k} \phi({\bx}_i) L\big (y_i, \widehat{f}_{k}({\bx}_i)\big ).   
\end{align*}

\subsection{Discussion on Assumption 2}\label{sec A.9}
Assumption 2 in the main text is a local $c_0$-strongly convexity condition on the expected loss function with respect to $\mathcal{L}^2(\mathcal{X},P^S_{\mathbf{x}})$ and $\mathcal{L}^2(\mathcal{X},P^T_{\mathbf{x}})$ at $f^*$. So verifying Assumption 2 is equivalent to verifying the local $c_0$ strongly convexity of the loss function. Here are some examples: 

\begin{itemize}
    \item For the squared loss $L(y,f(\mathbf{x}))=(y-f(\mathbf{x}))^2$, note that for any $y \in \mathcal{R}$, the function $z \rightarrow (y-z)^2$ is strongly convex with parameter $c_0=1$, so $f \rightarrow L(y,f(\mathbf{x}))$ satisfies the condition in Assumption 2 with $c_0=1$.
    \item For the Huber loss ${L}(y, f(\mathbf{x}))=(y-f(\mathbf{x}))^2$, if $|y-f(\mathbf{x})| \leq \delta; \delta|y-f(\mathbf{x})|-\frac{1}{2} \delta^2$, otherwise, since this loss function is locally equivalent to the squared loss, so it is locally strongly convex under mild tail condition on $y-f^*(\mathbf{x})$.
    \item For the check loss ${L}(y, f(\mathbf{x}))=(y-f(\mathbf{x}))\left(\tau-I_{\{y\leq f(\mathbf{x})\}}\right)$, the local strong convexity holds if the conditional density of $y-f^*(\mathbf{x})$ given $\mathbf{x}$ is bounded away from $c_0$ uniformly \citep{lian2022distributed}.
\end{itemize}
For the other loss functions, including logistic loss and hinge loss,  more detailed discussions and verifications can be found on Pages 470-472 in \cite{wainwright2019high}. Interested readers are referred to it for more details.

\section{Algorithm details}\label{B}
In this section, we provide the computing details for different loss functions considered in our experiments.  

\textbf{Kernel ridge regression.} For the squared loss, the minimizer $\widehat{f}$ takes the form of $\widehat{f}(\bx)=\sum_{i=1}^n\widehat{\alpha}_iK({\bx}_i,\bx), $ due to the representer theorem \citep{Smale2007}. Let $\boldsymbol{\widehat{\alpha}}=(\widehat{\alpha}_1,...,\widehat{\alpha}_n)^\top\in\mathcal{R}^n$, then the solution is given by
$\boldsymbol{\widehat{\alpha}}=(\mathbf{K} W\mathbf{K}+n\lambda \mathbf{K})^{-1}\mathbf{K} \bW \by=(\bW\mathbf{K}+n\lambda I)^{-1} \bW\by$. Here  $\mathbf{K}$ is $n\times n$ invertible matrix with elements $K({\bx}_i, {\bx}_j)$, $\bW$ is a diagonal matrix with $W_{ii}=\widehat{\phi}_n(\mathbf{x}_i)$ and $\by=(y_1,...,y_n)^\top$, where $\widehat{\phi}_n(\bx)$ is the truncation version of the KLIEP estimator $\widehat{\phi}(\bx)$.

\textbf{Kernel quantile regression.} For the check loss, there is no explicit form for the solution. Then, we attempt to search the optimal solution based on the solution of the dual problem \citep{takeuchi2006nonparametric}.  In the presence of covariate shift, we derive the dual problem to be the following convex optimization task \citep{boyd2004convex} that
\begin{align*}
    \text{minimize}\quad &\frac{1}{2}\boldsymbol{\alpha}^\top \mathbf{K}\boldsymbol{\alpha}-{\by}^\top\boldsymbol{\alpha},\\
    \text{s.t.}\quad &C_\lambda (\tau-1) \widehat{\phi}_n({\bx}_i)\leq\alpha_i\leq C_\lambda \tau \widehat{\phi}_n({\bx}_i),\ \text{for}\  1\leq i\leq n,\\
    & \Vec{1}^\top\boldsymbol{\alpha}=0,
\end{align*}
where $\Vec{1}$ denotes the vector whose elements are 1, $\by=(y_1,...,y_n)^\top$ and $C_\lambda=1/(n\lambda)$. $b$ is the dual variable to the constraint $\Vec{1}^\top\boldsymbol{\alpha}=0$.

\textbf{Kernel support vector machine.} For the hinge loss,  we also  solve the duality problem \citep{scholkopf2002learning}. Specifically, in the presence of covariate shift, we solve the following convex optimization task that
\begin{align*}
\text { minimize} \quad &\frac{1}{2}\boldsymbol{\eta}^\top \widetilde{\mathbf{K}}\boldsymbol{\eta}-\vec{1}^{\top} \boldsymbol{\eta}, \\
\text { s.t. } \quad & {\by}^\top \boldsymbol{\eta}=0, \\& 0\leq\eta_i\leq C_\lambda\widehat{\phi}_n({\bx}_i), \quad i=1, \cdots, n,
\end{align*}
where $\widetilde{\mathbf{K}}$ denotes the $n\times n$ matrix with entries $K(x_i,x_j)y_iy_j$ and $\boldsymbol{\alpha}=(\eta_1y_1,...,\eta_ny_n)^\top$.  $b$ is the dual variable to the constraint ${\by}^\top \boldsymbol{\eta}=0$.

\textbf{Kernel logistic regression.} For the logistic loss, we use the Newton–Raphson algorithm to solve the optimization task \citep{keerthi2005fast}. Let $\mathbf{K}_{i}$ correspond  the $i$-th row of $\mathbf{K}$ and  $\boldsymbol{\widehat{\phi}}=(\widehat{\phi}({\bx}_1),...,\widehat{\phi}({\bx}_n))^\top$, $\boldsymbol{p}=(p_1,...,p_n)^\top$ with $p_i=\frac{\exp (\mathbf{K}_{i}{\boldsymbol{\alpha}})}{1+\exp (\mathbf{K}_{i}{\boldsymbol{\alpha}})}$. We conduct the iterative algorithm  
\begin{align*}
   \boldsymbol{\alpha}_{k+1}=\boldsymbol{\alpha}_{k}-J(\boldsymbol{\alpha}_{k})^{-1}F(\boldsymbol{\alpha}_{k})
   =& (\mathbf{K} \bW\mathbf{K}+\lambda\mathbf{K})^{-1}(\mathbf{K} \bW\mathbf{K}\boldsymbol{\alpha}_k+ \mathbf{K}\by\odot(1-\boldsymbol{p})\odot \boldsymbol{\widehat{\phi}})\\
   =& ( \bW\mathbf{K}+\lambda I)^{-1}(\bW\mathbf{K}\boldsymbol{\alpha}_k+ \by\odot(1-\boldsymbol{p})\odot \boldsymbol{\widehat{\phi}}),
\end{align*}
where  $W$ denotes the diagonal matrix with $W_{ii}=\widehat{\phi}_n({\bx}_i) p_i(1-p_i)$ and  $\ba\odot \bc=(a_1c_1,...,a_nc_n)^\top$ for two vectors $\ba$ and $\bc$.

\section{Technical proofs}\label{C}
This part provides the proofs of all the theorems and corollaries in the main text. Note that our theoretical analysis mainly employs the symmetrization technique and concentration inequality in learning theory. For the second moment bounded case, Section \ref{C.1} is devoted to the proof of Theorem \ref{thm2}  that indicates the TIRW estimator achieves optimal rate, and Section \ref{C.2} gives the proof of Theorem \ref{thm3}  showing the sub-optimal rate for the unweighted estimator. For the uniformly bounded case,  Section \ref{C.3} presents the proof of Theorem \ref{thm1}   to show the optimal rate for the unweighted estimator, which follows a similar argument as in Section \ref{C.2}.  Section \ref{C.4} gives the detailed derivation of all the corollaries. Section \ref{C.5} discusses the results of the minimax lower bound when some specified loss functions are used. In Section \ref{C.6}, we further discuss the theoretical gap with the importance ratio replaced by its plugin
estimator and potential future direction. For ease of notation, we discard the superscripts of ${\bx}_i^S$ and $y_i^S$ to ${\bx}_i$ and $y_i$ in our proofs, that is, $\{{\bx}_i, y_i\}_{i=1}^n$ is driven from the source model.
Additionally,  we define {$P\varphi:=E_{S}[\varphi(\mathbf{x},y)]$  and ${P_n}\varphi:=(1/n)\sum_{i=1}^n \varphi(\mathbf{x}_i,y_i)$} for a measurable function $\varphi(\mathbf{x},y)$, and clarify that the expectation $E[ \cdot ]$ in our proof is taking with respect to all random variables contained in it. Note that we remain $\|f^*\|_K$ in our proof and the theoretical results in the  main text can be obtained by letting $\|f^*\|_K=1$.

\subsection{Proof of Theorem \ref{thm2}} \label{C.1}
The following Lemma states Talagrand's concentration inequality for random elements taking
values in some space $\mathcal{Z}$. One can refer to \cite{bousquet2002bennett} for detailed proof. 
\begin{lem}\label{lem1}
Let $Z_1, \ldots, Z_n$ be independent random elements taking values in some space $\mathcal{Z}$ and let $\Xi$ be a class of real-valued functions on $\mathcal{Z}$, if we have 
\begin{align*}
\|\xi\| \leq \eta_n \quad \text { and } \quad \frac{1}{n} \sum_{i=1}^n \Var \left(\xi\left(Z_i\right)\right) \leq \zeta_n^2, \quad \forall \xi \in \Xi.
\end{align*}
Define $\boldsymbol{Z}:=\sup _{\xi \in \Xi}\left|\frac{1}{n} \sum_{i=1}^n\left(\xi\left(Z_i\right)-{E} \xi\left(Z_i\right)\right)\right|$. Then for $t>0$
\begin{align*}
P\left(\boldsymbol{Z} \geq E(\boldsymbol{Z})+t \sqrt{2\left(\zeta_n^2+2 \eta_n E(\boldsymbol{Z})\right)}+\frac{2 \eta_n t^2}{3}\right) \leq \exp \left(-n t^2\right) .
\end{align*}
\end{lem}

The following Lemma is the core of our proofs. It bounds the supremum of the difference between the empirical average dependent on the source data and the target expectation within a local ball using the   Rademacher complexity function and Lemma \ref{lem1}.
 
 \vspace{0.5em}
 
\begin{lem}\label{lem2}
For any radii $\delta>0$, we define event $\mathcal{K}(\delta)$ as 
$$
\sup_{f \in \Theta(\delta)}\left|\frac{1}{n}\sum_{i=1}^n\phi_n({\bx}_i)\big(L(y_i, f({\bx}_i))-L(y_i, f^*({\bx}_i))\big)-E_{T}\left[L(y, f({\bx}))-L(y, f^*({\bx}))\right]\right|\leq \mathcal{M}(\delta),   
$$
where $\Theta(\delta):=\{f \in \mathcal{H}_K \mid \|f-f^*\|_T \leq \delta, \text{and}\  \|f-f^*\|_K \leq 3\|f^*\|_K\}$ and $\mathcal{M}(\delta)=C\sqrt{\beta^2}\log nR(\delta)$, then  $\mathcal{K}(\delta)$ holds with probability at least $1-n^{-c_3}$.
\end{lem}

\vspace{0.5em}

Before providing the detailed proof of  Lemma \ref{lem2}, we give some discussions to illustrate the motivation of the proof.  Specifically,
Lemma \ref{lem2} states a general uniform law for the Lipschitz loss functions under covariate shift. Note that by the empirical process theory, the empirical average $\frac{1}{n}\sum_{i=1}^n\phi_n({\bx}_i)\big(L(y_i, f({\bx}_i))-L(y_i, f^*({\bx}_i)))$ approximates its population counterpart $E_S[\phi_n({\bx})(L(y, f({\bx}))-L(y, f^*({\bx})))]$ uniformly on some function class. When the truncation $\gamma_n$ diverges fast enough as $n$ grows, the quantity $| E_S[\phi_n({\bx})(L(y, f({\bx}))-L(y, f^*({\bx})))]-E_T[L(y, f({\bx}))-L(y, f^*({\bx}))]|$  is negligible. Therefore, it is expected  that $\frac{1}{n}\sum_{i=1}^n\phi_n({\bx}_i)\big(L(y_i, f({\bx}_i))-L(y_i, f^*({\bx}_i)))$ is close to $E_T[L(y, f({\bx}))-L(y, f^*({\bx}))]$ uniformly. Our proof is precisely motivated by this intuition. Moreover, in the proof, we decompose the total error into the empirical error $\sup_{f \in \Theta(\delta)}|\frac{1}{n}\sum_{i=1}^n\phi_n({\bx}_i)\big(L(y_i, f({\bx}_i))-L(y_i, f^*({\bx}_i)))-E_S[\phi_n({\bx})(L(y, f({\bx}))-L(y, f^*({\bx})))]|$ and the approximation error $\sup_{f \in \Theta(\delta)}|E_S[\phi_n({\bx})(L(y, f({\bx}))-L(y, f^*({\bx})))]-E_T[L(y, f({\bx}))-L(y, f^*({\bx}))]|$. Recall that $\phi_n(\bx)=\min\{\phi(\bx),\gamma_n\}$, and here, the truncation parameter $\gamma_n$ plays a key role in balancing such two errors. For example, a fast diverging $\gamma_n$ may reduce the approximation error but compromise the empirical error.  Therefore, an optimal $\gamma_n$ is the one that diverges at a certain rate (i.e., $\gamma_n=O(\sqrt{n})$) to achieve the optimal tradeoff between empirical and approximation errors.

\vspace{0.5em}

\noindent \textbf{Proof of Lemma \ref{lem2}.} We first make the following decomposition that
   \begin{align*}
&\sup_{f \in \Theta(\delta)}\left|\frac{1}{n}\sum_{i=1}^n\phi_n({\bx}_i)\big(L(y_i, f({\bx}_i))-L(y_i, f^*({\bx}_i))\big)-E_{T}\left[L(y, f({\bx}))-L(y, f^*({\bx}))\right]\right|\\
&\leq \underbrace{\sup_{f \in \Theta(\delta)}\left|\frac{1}{n}\sum_{i=1}^n\phi_n({\bx}_i)\big(L(y_i, f({\bx}_i))-L(y_i, f^*({\bx}_i))\big)-E_{S}\left[\phi_n(\bx)\left(L(y, f({\bx}))-L(y, f^*({\bx}))\right)\right]\right|}_{\text{Empirical error} \ D_1}\\
&+\underbrace{\sup_{f \in \Theta(\delta)}\left|E_{S}\left[\phi_n(\bx)\left(L(y, f({\bx}))-L(y, f^*({\bx}))\right)\right]-E_{T}\left[L(y, f({\bx}))-L(y, f^*({\bx}))\right]\right|}_{\text{Approximation error} \ D_2}.
\end{align*} 
Then, we only need to bound $D_1$ and $D_2$ separately. To bound $D_1$, we firstly use the standard symmetrization technique in empirical process \citep{pollard2012convergence, wainwright2019high} to bound $E[D_1]$ that
\begin{equation}
\begin{aligned}\label{A.1}
E[D_1]&=E\Big [\sup_{f \in \Theta(\delta)}\big |(P_n-P)\phi_n(\bx)\left(L(y, f({\bx}))-L(y, f^*({\bx}))\right)\big| \Big]  \\
&\overset{(\romannumeral1)}{\leq} \frac{2}{n}E\Big[\sup_{f \in \Theta(\delta)}\big |\sum_{i=1}^n\sigma_i\phi_n({\bx}_i)\big (L(y_i, f({\bx}_i))-L(y_i, f^*({\bx}_i))\big)\big | \Big]\\
& \overset{(\romannumeral2)}{\leq} \frac{4c_{L}}{n}E\Big [\sup_{f \in \Theta(\delta)}\big|\sum_{i=1}^n\sigma_i\phi_n({\bx}_i)(f({\bx}_i)-f^*({\bx}_i))\big| \Big],
\end{aligned}   
\end{equation}
where $\{\sigma_i\}'s$ denote the Rademacher variables taking values in $\{-1,1\}$ with equal probability, the inequality (\romannumeral1) is from the symmetrization technique that for any class of measurable function $\mathcal{F}$, we have $E[\sup_{\varphi\in \mathcal{F}} (P_n-P)\varphi]\le 2E[\sup_{\varphi\in \mathcal{F}} (1/n) \sum_{i=1}^n \sigma_i \varphi({\mathbf{x}}_i,y_i)  ]$. The inequality (\romannumeral2) follows from the fact that  the loss function is $c_{L}$-Lipschitz continuous and the Ledoux–Talagrand contraction
inequality \citep{wainwright2019high}. 

For any $f \in \Theta(\delta)$, we denote $g=f-f^* \in {\cal H}_K$ and then, there holds $g=\sum_{j=1}^{\infty}g_j\psi_j$ with $g_j=\int_{\mathcal{X}}f(\bx)\psi_j(\bx)\rho_{\bx}^T(\bx)d\bx$. Clearly, we have $\|g\|_T\leq \delta$ and $\|g\|_K\leq 3\|f^*\|_K$, which implies that $\sum_{j=1}^{\infty}g_j^2\leq \delta^2$ and $\sum_{j=1}^{\infty}g_j^2/\mu_j\leq 9\|f^*\|_K^2$. Combining these two results, there holds
\begin{align}\label{A.2}
\sum_{j=1}^{\infty}\frac{g_j^2}{\min(\delta^2,\mu_j\|f^*\|_K^2)}\leq 10.   
\end{align}
Then, we have
\begin{equation}
\begin{aligned}\label{A.3}
&\Big |\sum_{i=1}^n\sigma_i\phi_n({\bx}_i)(f({\bx}_i)-f^*({\bx}_i))\Big|=   \Big|\sum_{i=1}^n\sigma_i\phi_n({\bx}_i)\sum_{j=1}^{\infty}g_j\psi_j({\bx}_i)\Big| \\
=& \Big|\sum_{j=1}^{\infty}\frac{g_j}{\sqrt{\min(\delta^2,\mu_j\|f^*\|_K^2)}}\sqrt{\min(\delta^2,\mu_j\|f^*\|_K^2)}\sum_{i=1}^n\sigma_i\phi_n({\bx}_i)\psi_j({\bx}_i)\Big|\\
\overset{(\romannumeral1)}{\leq}& \sqrt{10}\left\{\sum_{j=1}^{\infty}\min(\delta^2,\mu_j\|f^*\|_K^2)\left(\sum_{i=1}^n\sigma_i\phi_n({\bx}_i)\psi_j({\bx}_i)\right)^2\right\}^{1/2},
\end{aligned}    
\end{equation}
{where the inequality (\romannumeral1) follows from Cauthy-Schwarz
inequality and the fact  \eqref{A.2}}.  Moreover, by plugging \eqref{A.3} into \eqref{A.1}, we have 
\begin{align*}
E[D_1]&\leq \frac{4\sqrt{10}c_{L}}{n}E \Big[\sum_{j=1}^{\infty}\min(\delta^2,\mu_j\|f^*\|_K^2)\left(\sum_{i=1}^n\sigma_i\phi_n({\bx}_i)\psi_j({\bx}_i)\right)^2\Big]^{1/2}   \\
& \overset{(\romannumeral1)}{\leq} \frac{4\sqrt{10}c_{L}}{n}\left\{\sum_{j=1}^{\infty}\min(\delta^2,\mu_j\|f^*\|_K^2)E_{\bx,\sigma}\left[\sum_{i=1}^n\sigma_i\phi_n({\bx}_i)\psi_j({\bx}_i)\right]^2\right\}^{1/2} \\
& \overset{(\romannumeral2)}{=}\frac{4\sqrt{10}c_{L}}{n}\left\{\sum_{j=1}^{\infty}\min(\delta^2,\mu_j\|f^*\|_K^2)\sum_{i=1}^nE_{\bx,\sigma}\left[\sigma_i^2\phi_n^2({\bx}_i)\psi_j^2({\bx}_i)\right]\right\}^{1/2}\\
&\overset{(\romannumeral3)}{\leq} \frac{4\sqrt{10}c_{L}}{n}\left\{\sum_{j=1}^{\infty}\min(\delta^2,\mu_j\|f^*\|_K^2)\sum_{i=1}^nE[\phi^2({\bx}_i)]\right\}^{1/2},
\end{align*}
where the first inequality (\romannumeral1) follows from Jensen's inequality, the second inequality (\romannumeral2) follows from the fact that $E_{\bx,\sigma}[\sigma_i\phi_n({\bx}_i)\psi_j({\bx}_i)]=0$ for each $i$, and the last inequality (\romannumeral3) follows from the assumption that$\|\psi_j\|_{\infty}\leq 1$ for all $j \geq 1$ and the fact that $\phi_n({\bx}_i) \leq \phi({\bx}_i)$. Note that $E[\phi^2({\bx}_i)]\leq \beta^2$, and thus we have 
\begin{align}\label{A.4}
E(D_1)\leq 4\sqrt{10}\sqrt{\frac{\beta^2c_{L}^2}{n}\sum_{j=1}^{\infty}\min(\delta^2,\mu_j\|f^*\|_K^2)}=4\sqrt{10\beta^2}c_{L}R(\delta) .   
\end{align}

Next, we turn to bound $D_1-E(D_1)$. Recall that $\sum_{j=1}^{\infty}\frac{g_j^2}{\min(\delta^2,\mu_j\|f^*\|_K^2)}\leq 10$ and $\|\psi_j\|_{\infty}\leq 1$, and then there holds
\begin{align*}
|g(\bx)|=\left|\sum_{j=1}^{\infty}g_j\psi_j(\bx)\right|&\overset{(\romannumeral1)}{\leq} \sqrt{\sum_{j=1}^{\infty}\frac{g_j^2}{\min(\delta^2,\mu_j\|f^*\|_K^2)}} \sqrt{\sum_{j=1}^{\infty}\min(\delta^2,\mu_j\|f^*\|_K^2)\psi_j^2(\bx)}\\
&\leq \sqrt{10\sum_{j=1}^{\infty}\min(\delta^2,\mu_j\|f^*\|_K^2)}=\sqrt{10n}R(\delta),
\end{align*}
where the first inequality (\romannumeral1) follows from Cauthy-Schwarz
inequality. Consequently, we have 
\begin{align*}
\left|\phi_n({\bx}_i)\left(L(y_i, f({\bx}_i))-L(y_i, f^*({\bx}_i))\right)\right|&\leq\gamma_n\left|\left(L(y_i, f({\bx}_i))-L(y_i, f^*({\bx}_i))\right)\right |\\
&\leq\gamma_nc_{L}\left|g({\bx}_i)\right | \leq\gamma_nc_{L}\sqrt{10n}R(\delta)=\sqrt{10\beta^2}nc_{L}R(\delta),
\end{align*}
where we use $\gamma_n = \sqrt{n\beta^2}$.  Furthermore, we have 
\begin{align*}
E\Big[\big\{\phi_n({\bx}_i)\left(L(y_i, f({\bx}_i))-L(y_i, f^*({\bx}_i))\right)\big\}^2\Big]&\leq  c_{L}^2E\left[\phi_n^2({\bx}_i)g^2({\bx}_i)\right]\leq 10n\beta^2c_{L}^2R^2(\delta),
\end{align*}
where we use the fact that $E[\phi_n^2({\bx}_i)]\leq E[\phi^2({\bx}_i)]\leq\beta^2$. Clearly, all the required  conditions in Lemma \ref{lem1} are satisfied by taking $\eta_n=\sqrt{10\beta^2}nc_{L}R(\delta)$ and $\zeta_n^2=10n\beta^2c_{L}^2R^2(\delta)$. Let $t=\sqrt{\frac{c_3\log n}{n}}$, with probability at least $1-n^{-c_3}$, there holds that 
\begin{equation}\label{A.5}
\begin{aligned}
D_1-E[D_1]\leq &  \sqrt{\frac{c_3\log n}{n}\left(20n\beta^2c_{L}^2R^2(\delta)+4 nc_{L}\sqrt{10\beta^2}R(\delta) E[D_1]\right)}+\frac{2\sqrt{10\beta^2}c_3 c_{L}}{3}\log nR(\delta)\\
\overset{(\romannumeral1)}{\leq} & 3c_{L}\sqrt{20c_3\beta^2\log n}R(\delta)+\frac{2\sqrt{10\beta^2}c_3 c_{L}}{3}\log nR(\delta) \\
\leq&\left(3c_{L}\sqrt{20c_3}+\frac{2\sqrt{10}c_3 c_{L}}{3} \right)\sqrt{\beta^2}\log nR(\delta),
\end{aligned}
\end{equation}
where the first inequality (\romannumeral1) follows from \eqref{A.4}.

Then, combining \eqref{A.4} and \eqref{A.5}, with probability at least $1-n^{-c_3}$, we have 
\begin{align}\label{A.6}
D_1\leq C\sqrt{\beta^2}\log nR(\delta), 
\end{align}
where $C=4\sqrt{10}c_{L}+3c_{L}\sqrt{20c_3}+\frac{2\sqrt{10\beta^2}c_3 c_{L}}{3}$.

Now we turn to bound $D_2$. Note that 
\begin{equation}\label{A.7}
\begin{aligned}
D_2 &\leq \sup_{f \in \Theta(\delta)}\left|E_{T}\left[L(y, f({\bx}))-L(y, f^*({\bx}))\right]-E_{T}\left[I_{\{\phi(\bx)\leq \gamma_n\}}\left(L(y, f({\bx}))-L(y, f^*({\bx}))\right)\right]\right| +\\
&\hspace{3cm} \gamma_n\sup_{f \in \Theta(\delta)}\left|E_{S}\left[I_{\{\phi(\bx)> \gamma_n\}}\left(L(y, f({\bx}))-L(y, f^*({\bx}))\right)\right]\right|\\
&= \sup_{f \in \Theta(\delta)}\left|E_{T}\left[I_{\{\phi(\bx)> \gamma_n\}}\left(L(y, f({\bx}))-L(y, f^*({\bx}))\right)\right]\right| + \\
&\hspace{3cm} \gamma_n\sup_{f \in \Theta(\delta)}\left|E_{S}\left[I_{\{\phi(\bx)> \gamma_n\}}\left(L(y, f({\bx}))-L(y, f^*({\bx}))\right)\right]\right|\\
& \leq E_{T}\left[I_{\{\phi(\bx)> \gamma_n\}}\sup_{f \in \Theta(\delta)}\left|L(y, f({\bx}))-L(y, f^*({\bx}))\right|\right] +\\
&\hspace{3cm} \gamma_nE_{S}\left[I_{\{(\phi(\bx)> \gamma_n\}}\sup_{f \in \Theta(\delta)}\left|L(y, f({\bx}))-L(y, f^*({\bx}))\right|\right]\\
&\leq c_{L}E_{T}\left[I_{\{\phi(\bx)> \gamma_n\}}\right] \sup_g\left\|g\right\|_{\infty}+\gamma_nc_{L}E_{S}\left[I_{\{\phi(\bx)> \gamma_n\}}\right]\sup_g\left\|g\right\|_{\infty}\\
& \overset{(\romannumeral1)}{\leq}\frac{\beta^2c_{L}}{\gamma_n}\sqrt{10n}R(\delta)+\gamma_n\frac{\beta^2c_{L}}{\gamma_n^2}\sqrt{10n}R(\delta)\leq 2\sqrt{10\beta^2}c_{L}R(\delta) ,
\end{aligned}    
\end{equation}
where the inequality (\romannumeral1) follows from Markov inequality.

Combining \eqref{A.6} and \eqref{A.7}, with probability at least $1-n^{-c_3}$, there holds
\begin{align*}
D_1+D_2 \leq C\sqrt{\beta^2}\log nR(\delta) =\mathcal{M}(\delta),
\end{align*}
where $C=4\sqrt{10}c_{L}+3c_{L}\sqrt{20c_3}+\frac{2\sqrt{10\beta^2}c_3 c_{L}}{3}+2\sqrt{10}c_{L}$.
This completes the proof.
\hfill$\blacksquare$

\vspace{0.5em}

Following the Lemma \ref{lem2}, we give the proof of Theorem \ref{thm2} .

\noindent \textbf{Proof of Theorem \ref{thm2}.} 
Denote $\delta_{\lambda}=\sqrt{\delta_n^{2}+2c_0^{-1}\lambda\|f^*\|_K^2}$ and note that $\mathcal{M}(\delta)/\delta$ is non-increasing in $\delta$, then
\begin{align*}
\frac{\mathcal{M}(\delta_{\lambda})}{\delta_{\lambda}}\leq  \frac{\mathcal{M}(\delta_{n})}{\delta_{n}}\leq \frac{1}{2}c_0\delta_n\leq \frac{1}{2}c_0\delta_{\lambda},
\end{align*}
where the second inequality follows from the definition of $\delta_n$. Then,  we have $\mathcal{M}(\delta_{\lambda})\leq c_0\delta_{\lambda}^2/2$. In the following, we first establish the upper bound on ${\cal L}^2( P_{\bx}^T)$-error by showing the following inequality holds conditioning on the event $\mathcal{K}(\delta_{\lambda})$
\begin{align}\label{A.8}
\inf_{f \in \mathcal{H}_K, f \notin \Theta(\delta_{\lambda})} \frac{1}{n}\sum_{i=1}^n\phi_n({\bx}_i)\left\{L(y_i, f({\bx}_i))-L(y_i, f^*({\bx}_i))\right\}+\lambda\|f\|_K^2-\lambda\|f^*\|_K^2 > 0,
\end{align}
where the definitions of $\mathcal{K}(\delta_{\lambda})$ and $\Theta(\delta_{\lambda})$ are provided in Lemma \ref{lem2}. 
Note that it suffices to prove that \eqref{A.8} holds on the boundary of $\Theta(\delta_{\lambda})$, denoted by $B(\Theta(\delta_{\lambda}))$. To see this, for any $f\in \mathcal{H}_K$ and $f \notin \Theta(\delta_{\lambda})$, by the convexity of the two sets $\mathcal{H}_K$ and $\Theta(\delta_{\lambda})$, there exists  $0< \alpha\le 1$ such that $\Tilde{f}=\alpha f+(1-\alpha)f^*\in B(\Theta(\delta_{\lambda}))$. Applying Jensen's inequality yields
\begin{align*}
  &\phi_n({\bx}_i)\left\{L(y_i, \Tilde{f}({\bx}_i))-L(y_i, f^*({\bx}_i))\right\}+\lambda\|\Tilde{f}\|_K^2-\lambda\|f^*\|_K^2\\
 & \hspace{3cm} \le  \alpha \left\{\phi_n({\bx}_i)\left\{L(y_i, f({\bx}_i))-L(y_i, f^*({\bx}_i))\right\}+\lambda\|f\|_K^2-\lambda\|f^*\|_K^2\right\}.
\end{align*}
Therefore, we only need to show 
\begin{align*}
\frac{1}{n}\sum_{i=1}^n\phi_n({\bx}_i)\left\{L(y_i, f^*({\bx}_i)))-L(y_i, \Tilde{f}({\bx}_i)\right\}+\lambda\|f^*\|_K^2-\lambda\|\Tilde{f}\|_K^2   <0.
\end{align*}
For $\Tilde{f}\in B(\Theta(\delta_{\lambda}))$, we consider the following two cases: (\romannumeral 1) If $\|\Tilde{f}-f^*\|_T=\delta_{\lambda}$ and $\|\Tilde{f}-f^*\|_K \leq 3\|f^*\|_K$, we have 
\begin{align*}
&\frac{1}{n}\sum_{i=1}^n\phi_n({\bx}_i)\left\{L(y_i, f^*({\bx}_i)))-L(y_i, \Tilde{f}({\bx}_i)\right\}+\lambda\|f^*\|_K^2-\lambda\|\Tilde{f}\|_K^2\\ 
\overset{(\romannumeral1)}{\leq}& \mathcal{M}(\delta_{\lambda})-E_{T}\left[L(y, \Tilde{f}({\bx}))-L(y, f^*({\bx}))\right]+\lambda\|f^*\|_K^2-\lambda\|\Tilde{f}\|_K^2\\
\overset{(\romannumeral2)}{\leq} & \mathcal{M}(\delta_{\lambda})-c_0\|\Tilde{f}-f^*\|_T^2+\lambda\|f^*\|_K^2\leq-\frac{c_0}{2}\delta^2_{\lambda}+\lambda\|f^*\|_K^2=-\frac{c_0}{2}\delta^{2}_{n} < 0,
\end{align*}
where the inequality (\romannumeral1) follows from Lemma \ref{lem2}, (\romannumeral2) is from Assumption 2.
If $\|\Tilde{f}-f^*\|_T\leq \delta_{\lambda}$ and $\|\Tilde{f}-f^*\|_K = 3\|f^*\|_K$, we have 
\begin{align*}
&\frac{1}{n}\sum_{i=1}^n\phi_n({\bx}_i)\left\{L(y_i, f^*({\bx}_i)))-L(y_i, \Tilde{f}({\bx}_i)\right\}+\lambda\|f^*\|_K^2-\lambda\|\Tilde{f}\|_K^2\\    
\leq& \mathcal{M}(\delta_{\lambda})-E_{T}\left[L(y, \Tilde{f}({\bx}))-L(y, f^*({\bx}))\right]+\lambda\|f^*\|_K^2-\lambda\|\Tilde{f}\|_K^2\\ 
\leq& \mathcal{M}(\delta_{\lambda})+\lambda\|f^*\|_K^2-\lambda\|\Tilde{f}\|_K^2
\overset{(\romannumeral1)}{\leq}\frac{c_0}{2}\delta^2_{\lambda}-3\lambda\|f^*\|_K^2 =\frac{c_0}{2}\delta^{2}_{n}-2\lambda\|f^*\|_K^2\overset{(\romannumeral2)}{<} 0,
\end{align*}
where the inequality (\romannumeral1) follows from the fact that $\|\Tilde{f}\|_K\geq 2\|f^*\|_K$ by triangle inequality, and the inequality (\romannumeral2) follows from the definition of $\lambda$.

Combining the above results, the inequality \eqref{A.8} holds. Then, we can conclude that
\begin{align*}
\|\widehat{f}^{\phi}-f^*\|_T^2\leq \delta_{\lambda}^2=\delta_n^2+2c_0^{-1}\lambda\|f^*\|_K^2 \end{align*}
holds by the definition of $\widehat{f}^{\phi}$. 

For the bound of excess risk, note that 
\begin{align*}
&\mathcal{E}^{L}_T(\widehat{f}^{\phi})-\mathcal{E}^{L}_T(f^*)   \\
 \overset{(\romannumeral1)}{\leq}& \mathcal{E}^{L}_T(\widehat{f}^{\phi})-\mathcal{E}^{L}_T(f^*)-\frac{1}{n}\sum_{i=1}^n\phi_n({\bx}_i)\big(L(y_i, \widehat{f}^{\phi}({\bx}_i))-L(y_i, f^*({\bx}_i))\big)+\lambda\|f^*\|_K^2-\lambda\|\widehat{f}^{\phi}\|_K^2\\
\overset{(\romannumeral2)}{\leq}& \mathcal{M}(\delta_{\lambda})+\lambda\|f^*\|_K^2-\lambda\|\widehat{f}^{\phi}\|_K^2 =\mathcal{M}(\delta_{\lambda})-2\lambda\langle f^*,\widehat{f}^{\phi}-f^*\rangle_K-\lambda\|\widehat{f}^{\phi}-f^*\|_K^2\\
\leq& \frac{c_0\delta_{\lambda}^2}{2}+2\lambda\|f^*\|_K\|\widehat{f}^{\phi}-f^*\|_K-\lambda\|\widehat{f}^{\phi}-f^*\|_K^2\\
\overset{(\romannumeral3)}{\leq}& \frac{c_0\delta_{\lambda}^2}{2}+\lambda\|f^*\|_K^2+\lambda\|\widehat{f}^{\phi}-f^*\|_K^2-\lambda\|\widehat{f}^{\phi}-f^*\|_K^2\\
=&\frac{c_0\delta_{\lambda}^2}{2}+\lambda\|f^*\|_K^2=\frac{1}{2}c_0\delta_{n}^{2}+2\lambda\|f^*\|_K^2,
\end{align*}
where the inequality (\romannumeral1) follows from the definition of $\widehat{f}^{\phi}$, (\romannumeral2) is from Lemma \ref{lem2}, and (\romannumeral3) is from the basic inequality. This completes the proof.
\hfill$\blacksquare$

\subsection{Proof of Theorem \ref{thm3} }\label{C.2}
To prove Theorem \ref{thm3} , we first provide a Lemma which is similar to Lemma \ref{lem2} and bounds the regular supremum of empirical process within a local ball.
\begin{lem}\label{lem3}
For any radii $\nu>0$, define the event $\mathcal{K}^\prime(\nu)$ as
\begin{align*}
\sup_{f \in \Theta^\prime(\nu)}\left|\frac{1}{n}\sum_{i=1}^n\left(L(y_i, f({\bx}_i))-L(y_i, f^*({\bx}_i))\right)-E_{S}\left[L(y, f({\bx})-L(y, f^*({\bx}))\right]\right|\leq \mathcal{M}^\prime(\nu),
\end{align*}

where $\Theta^\prime(\nu):= \Big \{f \in \mathcal{H}_K\mid  \|f-f^*\|_S\leq c_0\nu^2/(c_{L}\sqrt{\beta^2})\ and \   \|f-f^*\|_K \leq 3\|f^*\|_K\} \Big \}$, then  $\mathcal{K}^\prime(\nu)$ holds with probability at least $1-n^{-c_2}$.
\end{lem}

\vspace{0.5em}
Note that
\begin{equation}\label{A.10}
\begin{aligned}
c_0\|f-f^*\|_T^2\overset{(\romannumeral1)}{\leq} & E_{T}\left[L(y, f({\bx}))-L(y, f^*({\bx}))\right]
=E_{S}\left[\phi(\bx)\left(L(y, f({\bx}))-L(y, f^*({\bx}))\right)\right]\\
\overset{(\romannumeral2)}{\leq}& c_{L}\sqrt{\beta^2}\|f-f^*\|_S, 
\end{aligned}    
\end{equation}
where the inequality (\romannumeral1) follows from Assumption 2, the inequality (\romannumeral2) follows from Cauchy-Schwarz inequality and the fact that the loss function is $c_{L}$-Lipschitz continuous.

Denote
\begin{align*}
D=\sup_{f \in \Theta^\prime(\nu)}\left|\frac{1}{n}\sum_{i=1}^n\big(L(y_i, f({\bx}_i))-L(y_i, f^*({\bx}_i))\big)-E_{S}\left[L(y, f({\bx}))-L(y, f^*({\bx}))\right]\right|.
\end{align*}
To bound $E[D]$, by following the similar argument as Lemma \ref{lem2}, we have
\begin{equation}\label{A.11}
E[D]\leq\frac{4c_{L}}{n}E\Big [ \sup_{f \in \Theta^\prime(\nu)}\left|\sum_{i=1}^n\sigma_i(f({\bx}_i)-f^*({\bx}_i))\right|\Big ].  
\end{equation}
Denote $g=f-f^* \in \mathcal{H}_K$ and then, we have $g=\sum_{j=1}^{\infty}g_j\psi_j$ with $g_j=\int_{\mathcal{X}}f(\bx)\psi_j(\bx)\rho_{\bx}^T(\bx)d\bx$ and 
\begin{equation}\label{A.12}
\begin{aligned}
\left|\sum_{i=1}^n\sigma_i(f({\bx}_i)-f^*({\bx}_i))\right|&=   \left|\sum_{i=1}^n\sigma_i\sum_{j=1}^{\infty}g_j\psi_j({\bx}_i)\right| \\
& = \left|\sum_{j=1}^{\infty}\frac{g_j}{\sqrt{\min(\nu^2,\mu_j\|f^*\|_K^2)}}\sqrt{\min(\nu^2,\mu_j\|f^*\|_K^2)}\sum_{i=1}^n\sigma_i\psi_j({\bx}_i)\right|\\
&\overset{(\romannumeral1)}{\leq}  \sqrt{10}\left\{\sum_{j=1}^{\infty}\min(\nu^2,\mu_j\|f^*\|_K^2)\left(\sum_{i=1}^n\sigma_i\psi_j({\bx}_i)\right)^2\right\}^{1/2},
\end{aligned}
\end{equation}
where the inequality (\romannumeral1) follows from Cauthy-Schwarz
inequality and the fact that $\sum_{j=1}^{\infty}\frac{g_j^2}{\min(\nu^2,\mu_j\|f^*\|_K^2)}\leq 10$ by using $\|g\|_T\le \nu$ from \eqref{A.10} and $\|g\|_K\le 3\|f^*\|_K$.

Plugging  \eqref{A.12} into \eqref{A.11}, we have 
\begin{equation}\label{A.13}
  \begin{aligned}
E[D]&\leq \frac{4\sqrt{10}c_{L}}{n}E\Big [\sum_{j=1}^{\infty}\min(\nu^2,\mu_j\|f^*\|_K^2)\left(\sum_{i=1}^n\sigma_i\psi_j({\bx}_i)\right)^2\Big ]^{1/2}\\
&\overset{(\romannumeral1)}{\leq} \frac{4\sqrt{10}c_{L}}{n}\left\{\sum_{j=1}^{\infty}\min(\nu^2,\mu_j\|f^*\|_K^2)E_{\bx,\sigma}\big[\sum_{i=1}^n\sigma_i\psi_j({\bx}_i)\big ]^2\right\}^{1/2} \\
&\overset{(\romannumeral2)}{\leq}  4\sqrt{10}c_{L}R(\nu),
\end{aligned}  
\end{equation}
where the inequality (\romannumeral1) follows from Jensen's inequality, the inequality (\romannumeral2) is from the fact that $E_{\bx,\sigma}[\sigma_i\psi_j({\bx}_i)]=0$, for each $i$ and the the assumption that $\|\psi_j\|_{\infty}\leq 1$, for all $j \geq 1$. 

Next, we turn to bound the term $D-E[D]$ and  following the similar argument  as \eqref{A.12} yields that
\begin{align*}
|g(\bx)|\leq \sqrt{\sum_{j=1}^{\infty}\frac{g_j^2}{\min(\nu^2,\mu_j\|f^*\|_K^2)}} \sqrt{\sum_{j=1}^{\infty}\min(\nu^2,\mu_j\|f^*\|_K^2)\psi_j^2(\bx)}\leq \sqrt{10n}R(\nu).
\end{align*}
Consequently, we have 
\begin{align*}
\left|L(y_i, f({\bx}_i))-L(y_i, f^*({\bx}_i))\right|\leq c_{L} |g({\bx}_i)|\leq \sqrt{10n}c_{L}R(\nu),
\end{align*}
and 
\begin{align*}
E\left[(L(y_i, f({\bx}_i))-L(y_i, f^*({\bx}_i)))\right]^2\leq  c_{L}^2E\left[g^2({\bx}_i)\right]
\leq 10nc_{L}^2R^2(\nu).
\end{align*}
Then conditions in Lemma \ref{lem1} are satisfied with $\eta_n=\sqrt{10n}c_{L}R(\nu)$ and $\zeta_n^2=10nc_{L}^2R^2(\nu)$. Let $t=\sqrt{\frac{c_2\log n}{n}}$, with probability at least $1-n^{-c_2}$, there holds that 
\begin{equation}\label{A.14}
\begin{aligned}
D-E[D]\leq &  \sqrt{\frac{c_2\log n}{n}\left(20nc_{L}^2R^2(\nu)+4 c_{L}\sqrt{10n}R(\nu) E[D]\right)}+\frac{2\sqrt{10}c_2 c_{L}}{3}\frac{ \log n}{\sqrt{n}}R(\nu)\\
\overset{(\romannumeral1)}{\leq} & c_{L}\sqrt{\frac{20c_2\log n}{n}(n+8\sqrt{n})}R(\nu)+\frac{2\sqrt{10}c_2 c_{L}}{3}\frac{ \log n}{\sqrt{n}}R(\nu) \\
\overset{(\romannumeral2)}{\leq}&\left(3c_{L}\sqrt{20c_2}+\frac{\sqrt{20}c_2 c_{L}}{3} \right)\sqrt{\log n}R(\nu) 
\end{aligned}
\end{equation}
where the inequality (\romannumeral1) follows from \eqref{A.13}, and the inequality (\romannumeral2) follows from the fact that $\frac{\log n}{n} <1/2$, for $n\geq 2$. 

Combining \eqref{A.13} and \eqref{A.14}, with probability at least $1-n^{-c_2}$, we have 
\begin{align}\label{A.15}
D\leq C \sqrt{\log n} R(\nu),
\end{align}
where $C=4\sqrt{10}c_{L}/\sqrt{\log 2}+3c_{L}\sqrt{20c_2}+\sqrt{20}c_2 c_{L}/3$. Thus we complete the proof by taking $\mathcal{M}^\prime(\nu)= C \sqrt{\log n} R(\nu)$ .
\hfill$\blacksquare$

\vspace{0.5em}

\noindent \textbf{Proof of Theorem \ref{thm3}.} 
Let $\delta=c_0\nu^2/(c_{L}\sqrt{\beta^2})$, $\mathcal{Q}(\delta)=\mathcal{M}^\prime\left((c_0^{-1}c_{L}\sqrt{\beta^2}\delta)^{1/2}\right)$, the function class $\mathcal{G}(\delta)=\Theta^\prime\left((c_0^{-1}c_{L}\sqrt{\beta^2}\delta)^{1/2}\right)$ and the event $\mathcal{P}(\delta)=\mathcal{K}^\prime\left((c_0^{-1}c_{L}\sqrt{\beta^2}\delta)^{1/2}\right)$.  
Denote $\delta_{\lambda}=\sqrt{\delta_n^{2}+2c_0^{-1}\lambda\|f^*\|_K^2}$ with $\mathcal{Q}(\delta)\le c_0\delta^2/2$.  Since $\mathcal{M}^\prime(\nu)/\nu$ is non-increasing in $\nu$, then it is easy to check $\mathcal{Q}(\delta)/\delta$ is non-increasing in $\delta$ by  
\begin{align*}
\frac{\mathcal{Q}(\delta)}{\delta}=\frac{\mathcal{M}^\prime\left((c_0^{-1}c_{L}\sqrt{\beta^2}\delta)^{1/2}\right)}{(c_0^{-1}c_{L}\sqrt{\beta^2}\delta)^{1/2}}\left\{c_0\delta/(c_{L}\sqrt{\beta^2})\right\}^{-1/2},  
\end{align*}
hence we also have $\mathcal{Q}(\delta_{\lambda})\leq c_0\delta_{\lambda}^2/2$. Following a similar treatment as that in the proof of Theorem \ref{thm2} , we can show that 
\begin{align}\label{A.16}
 \inf_{f \in \mathcal{H}_K, f \notin \mathcal{G}(\delta_{\lambda})}\frac{1}{n}\sum_{i=1}^n\left\{L(y_i, f({\bx}_i))-L(y_i, f^*({\bx}_i))\right\}+\lambda\|f\|_K^2-\lambda\|f^*\|_K^2 > 0.
\end{align}

It implies by the definition of $\widehat{f}$ that  
$
\|\widehat{f}-f^*\|_S\leq \delta_{\lambda}$
with probability at least $1-n^{-c_2}$. By \eqref{A.10}, we have 
\begin{align*}
\|\widehat{f}-f^*\|_T^2\leq   c_0^{-1}c_{L}\sqrt{\beta^2}\delta_{\lambda}= c_0^{-1}c_{L}\sqrt{\beta^2}\sqrt{\delta_n^{2}+2c_0^{-1}\lambda\|f^*\|_K^2},
\end{align*}
and 
\begin{align*}
\mathcal{E}^{L}_T(\widehat{f})-\mathcal{E}^{L}_T(f^*) =E_{T}\left[L(y, f({\bx}))-L(y, f^*({\bx}))\right] \leq    c_{L}\sqrt{\beta^2}\sqrt{\delta_n^{2}+2c_0^{-1}\lambda\|f^*\|_K^2}
\end{align*}
with probability at least $1-n^{-c_2}$.
\hfill$\blacksquare$

\subsection{Proof of Theorem \ref{thm1} }\label{C.3}
 Note that the density ratio is bounded that  $\sup_{\bx \in \mathcal{X}}\phi(\bx)\leq \alpha$, which implies 
\begin{align}\label{A.17}
    \|f-f^*\|_T\le\sqrt{\alpha}\|f-f^*\|_S.
\end{align}

We can establish the similar result as that in Lemma \eqref{lem2} by taking $\Theta^\prime(\nu):=\{f \in \mathcal{H}_K\mid  \|f-f^*\|_S\le \nu/\sqrt{\alpha} \ \text{and} \   \|f-f^*\|_K \leq 3\|f^*\|_K\}$.  Then,  by choosing $\delta=\nu/\sqrt{\alpha}$ and following the similar treatment 
 as that in the proof of Theorem \ref{thm3} , we have 
\begin{align*}
\|\widehat{f}-f^*\|_S^2 \leq \delta_n^2+2c_0^{-1}\lambda\|f^*\|_K^2.  
\end{align*}
with probability at least $1-n^{-c_1}$, where $\delta_n$ satisfies $\mathcal{M}^\prime(\sqrt{\alpha}\delta)=C\sqrt{\log n} R(\sqrt{\alpha}\delta)\le \frac{c_0\delta^2}{2}$. 
Together with \eqref{A.17}, we have 
\begin{align*}
\|\widehat{f}-f^*\|_T^2 \leq \alpha\left(\delta_n^2+2c_0^{-1}\lambda\|f^*\|_K^2\right)   
\end{align*}
with probability at least $1-n^{-c_1}$. On the other hand, by \eqref{A.10}, there holds
\begin{align*}
\mathcal{E}^{L}_T(\widehat{f})-\mathcal{E}^{L}_T(f^*) =E_{T}\left[L(y, f({\bx}))-L(y, f^*({\bx}))\right] \leq    c_{L}\alpha\sqrt{(\delta_n^{2}+2c_0^{-1}\lambda\|f^*\|_K^2)}
\end{align*}
with probability at least $1-n^{-c_1}$.
Thus we complete the proof.
\hfill$\blacksquare$

\vspace{0.5em}

\begin{rem}
One can combine the proofs of Theorem \ref{thm1}  and Theorem \ref{thm3}  to find out why the unweighted estimator for the bounded case achieves the optimal rate in terms of the  ${\cal L}^2( P_{\bx}^T)$-error, rather than only attaining sub-optimal for the moment bounded case.  Both the two proofs first bound the supremum of empirical process under the classic regime, that is, without covariate shift. Then the fundamental distinction lies in that the inequalities \eqref{A.10} and \eqref{A.17} give two different convergence rates compared to $\|\widehat{f}-f^*\|_S$.
\end{rem}

\subsection{Proof of the corollaries} \label{C.4}
\noindent \textbf{Proof of Corollary 1.} From the definition that $d(\delta)=\min\{j \geq 1 | \mu_j \leq \delta^2\}$ and the assumption that  $\|f^*\|_K^2=1$, we have
\begin{align*}
\sum_{j=1}^{\infty}\min(\delta^2,\mu_j\|f^*\|_K^2)=\sum_{j=1}^{d(\delta)}\min(\delta^2,\mu_j)+\sum_{j=d(\delta)+1}^{\infty}\min(\delta^2,\mu_j) \leq d(\delta)\delta^2+ C d(\delta)\delta^2\asymp d(\delta)\delta^2,
\end{align*}
where we use the definition of the regular kernel. So the inequality $C\sqrt{\log n} R(\sqrt{\alpha}\delta)\le \frac{c_0\delta^2}{2}$ can be simplified to 
\begin{align*}
\sqrt{\frac{\alpha\log n}{n}d(\sqrt{\alpha}\delta)}\leq C \delta.
\end{align*}
This proves the inequality (7) in the main text. For the finite-rank $D$ case, $\sum_{j=1}^{\infty}\min(\delta^2,\mu_j) \leq D\delta^2$,
which implies
\begin{align*}
\delta_{n}^{ 2}\leq C\frac{D\alpha\log n}{n}.
\end{align*}
Combining the choice of $\lambda$ and Theorem \ref{thm1}  gives 
\begin{align*}
\|\widehat{f}-f^*\|_T^2  \asymp \mathcal{E}^{L}_T(\widehat{f})-\mathcal{E}^{L}_T(f^*)  \leq C\frac{D\alpha^2\log n}{n}.
\end{align*}
For the eigenvalues with polynomial decay, such that $\mu_j \leq Cj^{-2r}$, so we have $d(\delta)\leq C(1/\delta)^{1/r}$, which implies that $\delta_n$ satisfies
\begin{align*}
\left(\frac{\log n}{n} \alpha^{\frac{2r-1}{2r}}\right)^{\frac{2r}{2r+1}}\le C\delta^2.
\end{align*}
The simple derivation leads to the desired result. Thus we complete the proof.
\hfill$\blacksquare$

\vspace{0.5em}

\noindent \textbf{Proof of Corollary 2.} From the definition that $d(\delta)=\min\{j \geq 1 | \mu_j \leq \delta^2\}$ and the assumption that  $\|f^*\|_K^2=1$, we have
\begin{align*}
\sum_{j=1}^{\infty}\min(\delta^2,\mu_j)=\sum_{j=1}^{d(\delta)}\min(\delta^2,\mu_j)+\sum_{j=d(\delta)+1}^{\infty}\min(\delta^2,\mu_j) \leq d(\delta)\delta^2+ C d(\delta)\delta^2 ,
\end{align*}
where we use the definition of the regular kernel. So the inequality $\mathcal{M}(\delta)\leq \delta^2/2$ can be simplified to 
\begin{align*}
\sqrt{\frac{\beta^2\log^2 n}{n}d(\delta)}\leq C \delta.
\end{align*}
This also proves the inequality (16) in the main text. For the finite-rank $D$ case, $\sum_{j=1}^{\infty}\min(\delta^2,\mu_j) \leq D\delta^2$,
which implies
\begin{align}\label{p_cor_1}
\delta_{n}^{ 2}\leq C\frac{D\beta^2\log^2 n}{n}.
\end{align}
Combine \eqref{p_cor_1}, the choice of $\lambda$ and Theorem \ref{thm2} , we have 
\begin{align*}
\|\widehat{f}^{\phi}-f^*\|_T^2  \asymp \mathcal{E}^{L}_T(\widehat{f}^{\phi})-\mathcal{E}^{L}_T(f^*) \leq C\frac{D\beta^2\log^2 n}{n}.
\end{align*}
For the eigenvalues with polynomial decay, such that $\mu_j \leq Cj^{-2r}$, so we have $d(\delta)\leq C(1/\delta)^{1/r}$. According to inequality (16) in the main text, we have 
\begin{align*}
\delta^2+\frac{\beta^2\log^2 n}{n}d(\delta)\leq \delta^2+C\frac{\beta^2\log^2 n}{n}(1/\delta)^{1/r},
\end{align*}
which leads to an optimal choice $\delta^2=C(\frac{\beta^2\log^2 n}{n})^{\frac{2r}{2r+1}}$. 
\hfill$\blacksquare$

\noindent \textbf{Proof of the sub-optimality of $\widehat{f}$ in moment bounded case.} Now, we come to verify the result in Table 1 for the unweighted estimator in the moment bounded case. For the kernel with finite $R$,  the inequality $C\sqrt{\log n}R((c_{L}\sqrt{\beta^2}\delta/c_0)^{1/2})\leq \frac{c_0\delta^2}{2}$ can be simplified to 
\begin{align*}
\sqrt{\frac{\log n}{n}D\sqrt{\beta^2}\delta}\le C\delta^2.    
\end{align*}
Simple derivation yields $\delta_n\le (\frac{\log n}{n}D\sqrt{\beta^2})^{1/3}$. The desired convergence rate follows by setting $\lambda\asymp (\frac{D\sqrt{\beta^2}\log n}{n} )^{2/3}$. 
 For the eigenvalues with polynomial decay, by the argument as before, the  inequality $C\sqrt{\log n}R((c_{L}\sqrt{\beta^2}\delta/c_0)^{1/2})\leq \frac{c_0\delta^2}{2}$ can be simplified to 
\begin{align*}
 \sqrt{\frac{\log n}{n}(\sqrt{\beta^2}\delta)^{\frac{2r-1}{2r}}}\le C\delta^2,     
\end{align*}
which leads to $\delta_n^2\leq C(\frac{\log n}{n}(\beta^2)^{\frac{2r-1}{4r}})^{\frac{4r}{6r+1}}$.  
Thus we complete the proof by applying Theorem \ref{thm3}  with $\lambda\asymp (\frac{\log n}{n}(\beta^2)^{\frac{2r-1}{4r}})^{\frac{4r}{6r+1}}$. \hfill$\blacksquare$

\noindent \textbf{Proof of the convergence rates with Gaussian kernel.} At last, for Gaussian kernel, the eigenvalues $\mu_j$ decay exponentially, that is $\mu_j\asymp e^{-Cj\log j}$ \cite{bach2002kernel}. By the definition of $d(\delta)$, we have $d(\delta)\le -C\log\delta^2$ for $0<\delta<1$. We first consider the moment bounded case. For the TIRW estimator, by applying $\sum_{j=1}^{\infty}\min(\delta^2,\mu_j)\le Cd(\delta)\delta^2$, the inequality $C\sqrt{\beta^2}\log nR(\delta)\le \frac{c_0\delta^2}{2}$ can be simplified to
\begin{align*}
    C\beta^2 \frac{(\log n)^2}{n}\log(1/\delta^2)\le \delta^2,
\end{align*}
which yields $\delta^2_n\le C \beta^2 \frac{(\log n)^3}{n}$. With $\lambda\asymp \beta^2 \frac{(\log n)^3}{n}$, Theorem \ref{thm2}  implies 
\begin{align*}
 \|\widehat{f}^\phi-f^*\|_T^2  \asymp \mathcal{E}^{L}_T(\widehat{f})-\mathcal{E}^{L}_T(f^*)  \leq C\beta^2 \frac{(\log n)^3}{n}.
\end{align*}
For the unweighted estimator,  $C\sqrt{\log n}R((c_{L}\sqrt{\beta^2}\delta/c_0)^{1/2})\leq \frac{c_0\delta^2}{2}$ can be simplified to 
\begin{align*}
    C\sqrt{\frac{\log n}{n}\log\left(\frac{1}{\sqrt{\beta^2}\delta}\right)\sqrt{\beta^2}\delta}\le \delta^2,
\end{align*}
which yields $\delta_n\le C (\sqrt{\beta^2}\frac{(\log n)^2}{n})^{1/3}$. With $\lambda\asymp (\sqrt{\beta^2}\frac{(\log n)^2}{n})^{2/3}$, Theorem \ref{thm3}  implies 
\begin{align*}
 \|\widehat{f}-f^*\|_T^2  \asymp \mathcal{E}^{L}_T(\widehat{f})-\mathcal{E}^{L}_T(f^*)  \leq C(\beta^4\frac{(\log n)^2}{n})^{1/3}.
\end{align*}
We next consider the uniformly bounded case. For the TIRW estimator, it is straightforward to obtain that
\begin{align*}
 \|\widehat{f}^\phi-f^*\|_T^2  \asymp \mathcal{E}^{L}_T(\widehat{f})-\mathcal{E}^{L}_T(f^*)  \leq C\alpha \frac{(\log n)^3}{n}.
\end{align*}
For the unweighted estimator, the inequality $C\sqrt{\log n} R(\sqrt{\alpha}\delta)\le \frac{c_0\delta^2}{2}$ can be simplified to 
\begin{align*}
 C\sqrt{\frac{\log n}{n}\log\left(\frac{1}{\alpha \delta^2}\right)\alpha \delta^2} \le \delta^2,
\end{align*} 
which yields $\delta_n^2\le C \alpha \frac{(\log n)^2}{n}$. With $\lambda\asymp  C \alpha \frac{(\log n)^2}{n}$, Theorem \ref{thm1}  implies 
\begin{align*}
 \|\widehat{f}-f^*\|_T^2  \asymp \mathcal{E}^{L}_T(\widehat{f})-\mathcal{E}^{L}_T(f^*)  \leq C \alpha^2 \frac{(\log n)^2}{n}.
\end{align*}

\subsection{Discussion on the minimax lower bound} \label{C.5}
Based on a standard application of Fano's inequality, \cite{ma2022optimally} establish a minimax lower bound for the regular kernel class by using the squared loss. For completeness of our paper, we present the relevant result below, which gives a relatively conservative lower bound by only taking the regression-based problems into consideration, which covers the squared loss and check loss. To be specific, we suppose that the conditional density of $\varepsilon:=y-f^*(\bx)$ given $\bx$ follows the normal distribution with mean zero and variance $\sigma^2$.
\begin{thm}
For any $\alpha>0$, there exists a pair of marginal distributions  $(P_{\bx}^S,P_{\bx}^T)$ with  $\alpha$-uniformly bounded importance ratio  and an orthonormal  basis $\{\psi_j\}_{j\ge1}$ of $\mathcal{L}^2(\mathcal{X},P_{\bx}^T)$ such that for any regular  kernel class with eigenvalues $\{\mu_j\}_{j\ge 1}$, we have 
\begin{equation} \label{minimax-bound}
\inf_{\widehat{f}}\sup_{f^*\in \mathcal{B}_{\mathcal{H}}(1)}E\left[\|\widehat{f}-f^*\|_T^2\right]\ge C\inf_{\delta>0}\left\{\delta^2+\sigma^2\alpha \frac{d(\delta)}{n}\right\}, 
\end{equation}
where $\mathcal{B}_{\mathcal{H}}(1)=\{f \in {\cal H}_K \mid  \|f\|_K\le 1\}$ represents the unit Hilbert ball. 
\end{thm}
By simply comparing the lower bound in \eqref{minimax-bound} to the upper bound in (7) and (14) in the main text, we can see that this lower bound is sharp since it is achieved by both the unweighted estimator and the TIRW estimator up to a logarithmic factor. And hence, in the uniformly bounded case, the unweighted estimator achieves minimax optimality, which indicates the TIRW estimator may not be necessary. In the moment bounded case, the upper bound in Theorem \ref{thm2}  also attains the lower bound in \eqref{minimax-bound}  up to logarithmic factors. For the reason that the second moment bounded class contains the uniformly bounded class, we can conclude the TIRW estimator is still preserving minimax optimality, whereas the unweighted estimator is far from optimal compared to the minimax lower bound.

\subsection{Remark about  the importance ratio} \label{C.6}
It is worthy pointing out that in practice, it is unrealistic to obtain the true importance ratio $\phi(\bx)$, and it should be estimated  from data, where we denote the estimator of $\phi(\bx)$ by $\widehat{\phi}(\bx)$. As illustrated in Section 4 of  the main text, we adopted the KLIEP algorithm \citep{sugiyama2007direct} to obtain  $\widehat{\phi}(\bx)$ in all the numerical examples.  While the   theoretical results  are established under the case that$\phi(\bx)$ is known. We want to emphasize that to the best of our knowledge, such a gap commonly appears in the existing literature, possibly due to inherent theoretical challenges.  We decide to leave such a promising topic as potential future work, and we add some detailed discussions on the possible route for establishing the theoretical results.  Specifically, the key step is that we need to bound the term 
\begin{align*}
\sup_{f \in \Theta(\delta)}\Big|(1/n)\sum_{i=1}^n(\widehat{\phi}_n({\mathbf{x}}_i)-{\phi}_n({\mathbf{x}}_i))(L(y_i, f({\mathbf{x}}_i))-L(y_i, f^*({\mathbf{x}}_i))) \Big|.   
\end{align*}
Thus the strong convergence rate of $\widehat{\phi}_n-{\phi}_n$ is required. It's important to note that the components within this term are not independent, as the estimated importance ratio $\widehat{\phi}$ relies on the source data. To address these intricacies, advanced technical tools are essential.  Once we successfully bound this term, we can establish results similar to those presented in Theorem \ref{thm2}  by leveraging existing proof techniques with slight modification.

\end{document}